\documentclass[10pt]{article}

\usepackage{times}
\usepackage[a4paper, total={6.5in, 8.6in}]{geometry}
\usepackage{authblk}
\usepackage[table]{xcolor}
\usepackage{url}
\usepackage{todonotes}
\usepackage{csquotes}
\usepackage{multirow, bigdelim}
\usepackage{tabularx}
\usepackage{pgfplots}
\usepgfplotslibrary{groupplots}
\usepackage{amsthm}
\usepackage{thmtools} 
\usepackage{longtable}
\usepackage{subcaption}  
\usepackage{tikz} 
\usetikzlibrary{shapes,arrows,calc,matrix,trees,hobby,shapes.misc} 
\usepackage{bm}
\usepackage{paralist}
\usepackage{amssymb}
\usepackage{amsmath} 
\usepackage{pgfplotstable}
\usepgfplotslibrary{statistics} 
\pgfplotsset{compat=1.14}
\usetikzlibrary{snakes}
\usetikzlibrary{patterns}
\usepackage{tabulary}

\definecolor{ggray}{gray}{0.50}

\tikzset{ 
    mega thick/.style={line width=2.2pt}
}

\definecolor{darkgreen}{rgb}{0.09, 0.45, 0.27}
\newcommand{\cellg}[1]{{\cellcolor{gray!35} #1}}

\usepackage{lineno,hyperref}
\modulolinenumbers[5]

\tikzset{snake it/.style={decorate, decoration=snake}}

\newcommand{\fdiff}{{\sf diff}}

\newcommand{\prob}{P}
\newcommand{\allprob}{\mathcal{P}}
\mathchardef\mhyphen="2D 

\newcommand{\lab}{\mathcal{L}}

\newcommand{\ext}{\mathcal{E}}

\newcommand{\lin}{{\sf in}}
\newcommand{\lout}{{\sf out}}
\newcommand{\lundec}{{\sf und}}

\newcolumntype{P}[1]{>{\centering\arraybackslash}m{#1}}
\newcommand*\rot{\rotatebox{90}}

\newcommand*\cm{\checkmark}
\newcommand*\xm{$\times$}

\newcommand{\AF}{F} 
\newcommand{\BAF}{BF}
\newcommand{\TAF}{TF}
\newcommand{\subs}{{\sf Sub}}

\newcommand{\tframec}[1]{\multicolumn{2}{P{2.3cm}|}{#1}}

\newcommand{\args}{\mathcal{A}}
\newcommand{\atts}{\mathcal{R}}
\newcommand{\sups}{\mathcal{S}}
\newcommand{\deps}{\mathcal{D}}

\newcommand{\cf}{{\sf cf}}
\newcommand{\ad}{{\sf ad}}
\newcommand{\co}{{\sf co}}
\newcommand{\pr}{{\sf pr}}
\newcommand{\gr}{{\sf gr}}
\newcommand{\st}{{\sf st}}
\newcommand{\semantics}{\ad, \co, \pr, \gr, \st}

\theoremstyle{plain} 
\newtheorem{theorem}{Theorem}[section]

\theoremstyle{definition}
\newtheorem{definition}[theorem]{Definition}
\declaretheorem[style=definition]{example}
\renewcommand\thmcontinues[1]{Continued}

\makeatletter

\newcommand\frontmatter{%
    \cleardoublepage
  \pagenumbering{roman}
}

\newcommand\mainmatter{%
    \cleardoublepage
  \pagenumbering{arabic}}

\newcommand\backmatter{%
  \if@openright
    \cleardoublepage
  \else
    \clearpage
  \fi
   }

\makeatother

\providecommand{\keywords}[1]{\textbf{\textit{Keywords---}} #1}
 
\begin{document}
\frontmatter

\title{Empirical Evaluation of Abstract Argumentation: Supporting the Need for Bipolar and Probabilistic Approaches}
 \date{}

\author{Sylwia Polberg} 

\author{Anthony Hunter\footnote{This research is funded by EPSRC Project EP/N008294/1 \enquote{Framework for Computational Persuasion}.We thank the reviewers for their valuable comments that helped us to improve this paper.}}

\affil{University College London, Department of Computer Science \\ 66-72 Gower Street, London WC1E, United Kingdom} 
 
 \maketitle

\begin{abstract}
In dialogical argumentation, it is often assumed that the involved parties will always correctly identify  
the intended statements posited by each other and realize all of the associated relations, 
conform to the three acceptability states (accepted, rejected, undecided), 
adjust their views whenever new and correct information comes in, and that a framework handling only attack relations 
is sufficient to represent their opinions. 
Although it is natural to make these assumptions as a starting point for further research, dropping some of them has 
become quite challenging.  

Probabilistic argumentation is one of the approaches that can be harnessed for more accurate user modelling. 
The epistemic approach allows us to represent 
how much a given argument is believed or disbelieved by a given person, offering us the possibility to express more than just three agreement states. 
It comes equipped with a wide range of postulates, including those that do not make any restrictions concerning how initial arguments should 
be viewed. Thus, this approach is potentially more suitable for handling beliefs of the people that have not fully disclosed their opinions or counterarguments 
with respect to standard Dung's semantics. The constellation approach can be used to represent the views of different people concerning
the structure of the framework we are dealing with, including situations in which not all relations are acknowledged or when they are seen differently
than intended. Finally, bipolar argumentation frameworks can be used to express both positive and negative relations between arguments. 

In this paper we will describe the results of an experiment in which participants were asked to judge dialogues in terms of agreement and structure. 
We will compare our findings with the aforementioned assumptions as well as with the constellation and epistemic approaches to probabilistic argumentation and 
bipolar argumentation.  
\end{abstract} 


\mainmatter
\newpage 

\tableofcontents

\newpage
\section{Introduction} 

At the heart of abstract argumentation lies Dung's framework developed in \cite{Dung:1995}, which treats arguments as abstract 
 atomic entities that can be connected through an attack relation. 
Since its introduction, this framework has been endowed with numerous new semantics as well as 
generalized in various ways. These generalizations include structures that can represent new types of relations, such as support, 
as well as those that handle properties such as preferences 
or probabilities of arguments or relations \cite{general}. 
Despite the available approaches, dialogical argumentation appears to rely heavily on frameworks 
handling only attacks between arguments, 
such as Dung's framework. It also makes certain assumptions that are supported by the argumentation theory, 
but not always by empirical results. 
Argumentation, in many ways, simplifies human reasoning and appears to consider people to be in principle rational, good reasoners, 
that may be simply uninformed. Unfortunately, this approach might not always be adequate. 

One of the core concepts of defeasible reasoning and therefore 
abstract argumentation is the fallibility of human perception. Thus, we need to be able to reason even with incomplete information  
and be prepared to retract our conclusions in the face of new data. 
From a certain perspective, most of the abstract argumentation 
approaches can be quite conservative in their handling of these issues. Although the defeasibility of arguments and notions such as undercutting attack 
are widely acknowledged, at the same time there is an assumption that a universal attack relation exists. 
In other words, it is believed that every person participating in a given dialogue 
will interpret arguments and the relations between them in exactly the same way, and every time new information is presented, 
it will be understood and linked to the existing arguments in the \enquote{intended and correct} way. 
However, this does not seem realistic in various scenarios. For example, the transcript of a TV debate \enquote{as it is} can be perceived differently from 
its processed version, i.e. one in which arguments are identified, organized and clearly presented w.r.t. the chosen argument structure. 
Not all relevant pieces of information have to be explicitly stated and it can happen that people taking part in the debate may be purposefully ambiguous. 
Consequently, we may have to deal with arguments that have implicit premises or conclusions and 
thus run the risk of participants interpreting them differently. 
Therefore, we need to acknowledge that people can view the nature of the framework associated with a given dialogue differently 
and that we should be able to represent such differences and uncertainties. Furthermore, these issues can themselves 
become a part of the discussion, which due to the aforementioned assumptions is rarely considered.

The fact that dialogical argumentation relies so heavily on attack--based frameworks for modelling purposes has also 
affected the way we perceive dialogues on a higher level. 
This approach became rather heavily conflict--centered, by which we understand that people participating in a dialogue 
view each other as opponents whose arguments should be defeated and knowledge \enquote{rectified}. 
Although this view may very well work 
in a court room, it has negative effects in the context of, for example, physical or mental health, as seen in \cite{masthoff08}. For these applications, it may be more fruitful to consider dialogue parties as partners rather than opponents 
and use a more support--oriented approach in the discussion. 
Thus, rather than giving arguments against the current views of 
our dialogue partners, we try to motivate them or give them arguments for changing their behaviour and opinions.  A given 
piece of information can also be framed in a positive or negative way, which affects the way people react to it \cite{Tversky1985,rothman1997}. 
Hence, not only negative, but also 
positive relations between arguments can play a role in dialogical argumentation, and the use of various
types of support should be considered. 

Finally, the applicability of argumentation semantics in dialogical argumentation is not sufficiently verified \cite{Cerutti2014}. 
The use of two or three values, as promoted 
by Dung's semantics, oversimplifies the varying degrees to which we may agree or disagree with a given argument. 
For example, while a mother of an 8 year old girl might disagree with her staying the night at her best friend's house, 
this is probably nowhere near  
to her disagreement with the child going alone to a heavy metal concert. 
Thus, there is a need for expressing varying degrees of agreement with a given argument. 
Another issue lies in the nature of the semantics themselves. 
For example, successful persuasion is often seen as synonymous with winning a dialogue game 
w.r.t. a given semantics, such as grounded or admissible. 
However, it is easy to see that in the cases such as a doctor persuading a patient to stop smoking or to go on a diet, 
holding a dialectically winning position and actually convincing a person to do or not to do something can be two 
completely different things. A person taking part in a dialogue might also exhibit a number of perception biases or reasoning paradoxes 
and, due to inability or lack of cooperation, withhold information from the doctor. 
Hence, he or she does not have to reason in a way that adheres to the current 
argumentation semantics \cite{Olson92,ogden2012health}. 

To summarize, there is a need for empirical evaluation of argumentation approaches, which so far have received limited 
attention \cite{Rahwan2011,Cerutti2014,Rosenfeld16}. 
Although this task can be quite challenging, 
as what we are dealing with is human perception and judgment which are not always perfect, it is necessary. 
One has to realize that 
due to the nature of abstract argumentation, the validity of its approaches is typically inherited from other methods or obtained through novelty and technical correctness. 
For example, in order to defend the introduction of a new framework, we can explain how it can be instantiated with 
a given formalism, and show that the answers the framework produces exhibit certain desirable properties connected to the 
formalism we instantiated it with. The validity of the new framework thus depends on the validity of the formalisms it is linked to. 
Another possible way to argue in favour of a new approach is through the abstract argumentation itself, where the new 
semantics or framework is shown to be considerably different or more expressive than the existing ones. 
This method is often paired with 
presenting a realistic motivating example that argues in favor of the new framework or semantics. 
Although both of these strategies  promote 
new ideas and innovation, there comes a time when they need to be verified in real life. 
Without empirical evidence, we can accidentally increase a gap between applying argumentation and \textit{successfully} applying argumentation in real life situations. 

In this paper we describe the results 
of an empirical study in which participants are presented with dialogues separated into five stages. At every stage, they are 
asked to declare why and how much they agree or disagree with the presented statements and 
how they view the relations between them. 
The purpose of this study was to investigate certain aspects of abstract argumentation as used by laypeople, 
rather than to make a number of initial assumptions that we prove or disprove based purely on the behaviour of experts. 
We have been able to draw a number of important observations and gained some evidence concerning various 
formalisms available in abstract argumentation. However, we would like to note that due to the exploratory nature of 
our experiments, our results should be treated as indicative and as a basis for further studies, rather than as an 
indisputable proof for or against a given argumentation approach. 

\begin{itemize}
\item[\textbf{Observation 1}] The data supports the use of the \textbf{constellation approach} to probabilistic argumentation  -- people may interpret 
statements and relations between them differently, and not necessarily in the intended manner. The constellation approach can represent 
our uncertainty about the argument graphs describing our opponents views.
 
\item[\textbf{Observation 2}] People may explicitly declare
that two statements are connected, however, they might \textbf{not be sure of the exact nature of the relation} between them. We therefore also need 
to express the uncertainty that a person has about his or her own views, 
which can potentially be addressed with the constellation approach 
or with the introduction of a suitable framework.

\item[\textbf{Observation 3}] The data supports the use of \textbf{epistemic approach} to probabilistic argumentation: 
\begin{itemize}
\item people may assign levels of agreement to statements going beyond the 3--valued Dung's approach,
\item the epistemic postulates, in contrast to the standard semantics, can be highly adhered to and due to their nature, allow us to analyze why 
classical semantics may fail to explain the participants behaviour,
\item the extended epistemic postulates allow us to model situations where the perceived \enquote{strength} of a relation might not necessarily be tightly connected with the level of agreement assigned to its source.
\end{itemize}

\item[\textbf{Observation 4}] The data supports the use of \textbf{bipolar argumentation frameworks} -- the notion of defence does not necessarily account for all of the positive relations between the statements 
viewed by the participants.

\item[\textbf{Observation 5}] The data supports the use of \textbf{bipolar argumentation in combination with the prudent/careful approaches} --  many additional attacks perceived by the participants can be explained by the 
existing notions of indirect conflicts in these settings.

\item[\textbf{Observation 6}] The data shows that \textbf{people use their own personal knowledge} in order to make judgments and might not necessarily disclose it.
 
\item[\textbf{Observation 7}] The data shows that presenting a new and correct piece of information that a given person was not aware of \textbf{does not 
necessarily lead to changing that person's beliefs}.
\end{itemize}

This paper is organized as follows. In Sections \ref{sec:dung} to \ref{sec:probarg} we review the necessary background on Dung's argumentation 
frameworks, bipolar argumentation frameworks and probabilistic argumentation. 
In Section \ref{sec:experiment} we explain the set up of our experiment and analyze its results in Section \ref{sec:results}. This includes the 
analysis of
\begin{inparaenum}[\itshape 1\upshape)]
\item the argument graphs created for the experiment as well as those provided by the participants, 
\item the satisfaction rates of various epistemic 
postulates on the provided frameworks, 
\item the connection between the level of agreement assigned to a given statement 
and the way the relations it carries out are perceived, and 
\item changes in participants' opinions throughout the experiment. 
\end{inparaenum}
We close this paper with the discussion on the works related to our study and pointers for future work.

\section{Dung's Argumentation Framework}
\label{sec:dung} 

We start with a review of abstract argumentation as proposed by Dung in \cite{Dung:1995}. He introduced the following, straightforward framework, 
that can be easily depicted using a graph where nodes play the role of arguments and edges represent conflicts: 

\begin{definition}
A \textbf{Dung's abstract argumentation framework} (AF for short) is a pair $\AF=(\args, \atts)$, where $\args$ is a set of \textbf{arguments} and 
$\atts \subseteq \args \times \args$ represents an \textbf{attack} relation.
\end{definition} 

An argument $A \in \args$ is an {\bf attacker} of $B \in \args$ iff $(A,B) \in \atts$.  
By abuse of notation, we say that a set of elements attacks another element if it contains an appropriate attacker. 
The way we decide which arguments can be accepted 
or rejected (or neither) is called a semantics. Depending on whether a set of arguments or a labeling is returned, we deal with the 
extension--based and labeling--based semantics  \cite{Caminada:2009,Baroni:2011}.
 
An {\bf extension} is a set of arguments $\ext \subseteq \args$ that satisfies the requirements imposed by a given semantics. 
The classical semantics \cite{Dung:1995} are built on the notion of defence:

\begin{definition}\label{def:semantics}
Let $\AF = (\args, \atts)$ be a Dung's framework. An argument $A \in \args$ is \textbf{defended} 
by a set $\ext \subseteq \args$ in $\AF$\footnote{Defence is often
also referred to as acceptability: we say that $A$ is acceptable w.r.t. $\ext$ iff $\ext$ defends $A$.}
if for every $B \in \args$ s.t. $B$ attacks $A$, there exists $C \in \ext$ that attacks $B$. A set $\ext \subseteq \args$ is:
\begin{itemize}
\item[($\cf$)] \textbf{conflict--free} in $\AF$ iff for no $A, B \in \ext$, $A$ is an attacker of $B$.
\item[($\ad$)] \textbf{admissible} in $\AF$ iff it is conflict--free in $\AF$ and defends all of its members.
\item[($\pr$)]  \textbf{preferred} in $\AF$ iff it is maximal w.r.t. set inclusion admissible in $\AF$.
\item[($\co$)]  \textbf{complete} in $\AF$ iff it is admissible in $\AF$ and all arguments defended by $\ext$ are contained in $\ext$.
\item[($\st$)]  \textbf{stable} in $\AF$ iff it is conflict--free in $\AF$ and for every $A\in \args \setminus \ext$ there exists an argument $B \in \ext$ 
that attacks $A$.
\item[($\gr$)] \textbf{grounded} in $\AF$ iff it is the least complete extension of $\AF$.
\end{itemize}
\end{definition}
%

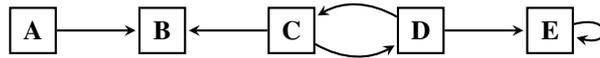
\begin{figure}[!ht]
\centering
  \begin{tikzpicture}
[->,>=stealth,shorten >=1pt,auto,node distance=1.7cm,
  thick,main node/.style={rectangle,fill=none,draw,minimum size = 0.6cm,font=\normalsize\bfseries}]

\node[main node] (a) {A};
\node[main node] (b) [right of=a] {B};
\node[main node] (c) [right of=b] {C};
\node[main node] (d) [right of=c] {D};
\node[main node] (e) [right of=d] {E};  
 
 \path
	(a) edge node {} (b)
	(c) edge node {} (b)
	(d) edge node {} (e)
    	(e) edge [loop right] node {} (e)
 	 (c) edge [bend right] node {} (d)
    (d) edge [bend right] node {} (c);
\end{tikzpicture}
\caption{Sample argument graph}
\label{fig:af}
\end{figure}

\begin{example}[label=ex:af] 
Consider the Dung's framework $\AF = (\args, \atts)$ 
with $\args = \{A$, $B$, $C$, $D$, $E\}$ and the attack relation $\atts= \{(A,B)$, $(C,B)$, $(C,D)$, $(D,C)$, $(D,E)$, $(E,E)\}$, as
depicted in Figure~\ref{fig:af}. It has eight conflict--free extensions in total: 
$\{A, C\}$,$\{A, D\}$, $\{B, D\}$, $\{A\}$, $\{B\}$, $\{C\}$, $\{D\}$ and $\emptyset$. As $B$ is attacked by an unattacked argument, it
cannot be defended against it. Consequently, $\{A, C\}$,$\{A, D\}$, $\{A\}$, $\{C\}$, $\{D\}$ and $\emptyset$ are our admissible 
sets.  
From this $\{A,C\}$, $\{A,D\}$ and $\{A\}$ are complete. 
We end up with two preferred extensions, 
$\{A,C\}$ and $\{A,D\}$. However, only $\{A,D\}$ is stable. Finally, 
$\{A\}$ is the 
grounded extension.  
\end{example}

More types of extension--based semantics have been proposed in the recent years \cite{Baroni:2011} and even though we will not consider all of them here, 
we would like to recall certain notions that will be useful in the next sections. As observed in the \textbf{prudent and careful semantics} \cite{inproc:careful,inproc:prudent}, 
based on the interplay of attack and defence, 
there might be additional positive and negative indirect interactions between the arguments that go beyond direct attack and defence:

\begin{definition} 
\label{def:prudent}
Let $\AF = (\args, \atts)$ be a Dung's framework and let $A, B \in  \args$ be two arguments:
\begin{itemize}
\item $A$ \textbf{indirectly attacks} $B$ iff there exists an odd--length path from $A$ to $B$ in $\AF$.
\item $A$ \textbf{indirectly defends} $B$ iff there exists an even--length path from $A$ to $B$ in $\AF$. The length of this path is not zero.
\item $A$ is \textbf{controversial} w.r.t. $B$ iff $A$ indirectly attacks and indirectly defends $B$.
\end{itemize}
Additionally, for arguments $A, B, C \in \args$, the pair $(A,B)$ is \textbf{super--controversial} w.r.t. $C$ iff $A$ indirectly attacks $C$
and $B$ indirectly defends $C$.
\end{definition} 
 
Originally, these notions were used to impose stronger restrictions on the conflict--free sets. 
Although we will not recall the semantics themselves, the aforementioned notions will be useful to us in the context of this work. 
In particular, we will test whether certain elements of the argument graphs sourced from the participants of our experiments 
could potentially be the manifestations of the above definitions.

\begin{example}[continues=ex:af] 
In the considered framework, we can observe that each of $C$, $D$ and $E$ function as indirect attackers and defenders of $E$. 
In addition to that, we can consider $C$ and $D$ as indirectly attacking each other. They also serve as their own defenders. 
We can observe that $C$ and $D$ also  
respectively indirectly attack and defend $B$.
\end{example}

Let us now focus on the labeling--based semantics, in particular those corresponding to the classical extension--based ones. 
An {\bf argument labeling} is a total function $\lab: \args \rightarrow \{ \lin, \lout, \lundec\}$ \cite{Caminada:2009,Baroni:2011}. 
By $\lin(\lab), \lout(\lab)$ and $\lundec(\lab)$ we denote the arguments mapped respectively to $\lin, \lout$ and $\lundec$(ecided) by 
$\lab$. We will often write a labeling
as a triple $(I,O,U)$, where $I$, $O$ and $U$ are sets of arguments mapped to $\lin$, $\lout$ and $\lundec$. 
We can now introduce the notion 
of legality, on which many semantics are based.

\begin{definition}
Let $\AF = (\args, \atts)$ be a Dung's framework and let $\lab: \args \rightarrow \{\lin, \lout, \lundec\}$ be a labeling:
\begin{itemize} 
\item $X \in \lin(\lab)$ is {\bf legally $\bm{\lin}$} iff all its attackers are in $\lout(\lab)$.
\item $X \in \lout(\lab)$ is {\bf legally $\bm{\lout}$} iff it has an attacker in $\lin(\lab)$.
\item $X \in \lundec(\lab)$ is {\bf legally $\bm{\lundec}$}
iff not all of its attackers are in $\lout(\lab)$ and it does not have an attacker in $\lin(\lab)$. 
\end{itemize}
\end{definition}


\begin{definition}\label{def:labsemantics}
Let $\lab: \args \rightarrow \{\lin, \lout, \lundec\}$ be a labeling:
\begin{itemize}
\item[(${\sf lab}{\mhyphen}\cf$)] $\lab$ is {\bf conflict--free} iff every $A \in \lout(\lab)$ is legally ${\lout}$ 
and there are no arguments $A, B \in \lin(\lab)$ s.t. $A$ is an attacker of $B$.
\item[(${\sf lab}{\mhyphen}\ad$)] $\lab$ is {\bf admissible} iff every $A \in \lin(\lab)$ is legally $\lin$ and every $A \in \lout(\lab)$ is legally $\lout$.
\item[(${\sf lab}{\mhyphen}\co$)] $\lab$ is {\bf complete} iff it is admissible and every $A \in \lundec(\lab)$ is legally $\lundec$.
\end{itemize}
\end{definition} 
The preferred, stable and grounded labelings are obtained from the complete ones by using the constraints from Table \ref{tab:dung}.

\begin{table}[!ht]
\centering 
  \begin{tabular}{ |c | c| }
\hline
Restriction on a complete labeling $\lab$ & Semantics \\
    \hline\hline 
No argument $A \in \args$ s.t. $\lab(A)= \lundec $ & stable {\bf (ST)}\\
Maximal no. of $A \in \args$ s.t. $\lab(A)= \lin$ & preferred {\bf (PR)}\\
Maximal no. of $A \in \args$ s.t. $\lab(A) = \lout$ & preferred {\bf (PR)}\\
Maximal no. of $A \in \args$ s.t. $\lab(A)= \lundec$ & grounded {\bf (GR)}\\
Minimal no. of $A \in \args$ s.t. $\lab(A) = \lin$ & grounded {\bf (GR)}\\
Minimal no. of $A \in \args$ s.t. $\lab(A) = \lout$ & grounded {\bf (GR)}\\ 
   \hline
  \end{tabular}
\caption{Relation between different labelings}
\label{tab:dung}
 \end{table}
 
The properties of the labeling--based semantics and their correspondence to the classical extension--based family have 
already been studied in \cite{Caminada:2009,Baroni:2011}: 

\begin{theorem}[\cite{Baroni:2011}]
\label{thm:extlabaf}
Let $\AF = (A,R)$ be a Dung's framework and $\ext \subseteq A$ be a $\sigma$--extension of $\AF$, 
where $\sigma \in \{\cf,\ad, \co, \gr, \pr, \st\}$. Let $\ext^+ = \{B \mid$ there exists $A \in \ext$ s.t. $A$ is an attacker of $B\}$. 
Then $(\ext, \ext^+, A\setminus(\ext \cup \ext^+))$ is a $\sigma$--labeling of $\AF$.
Let $\lab$ be a $\sigma$--labeling of $\AF$, where $\sigma \in \{\cf,\ad, \co, \gr, \pr, \st\}$.
Then $\lin(\lab)$ is a $\sigma$--extension of $\AF$.
\end{theorem}

\begin{example}[continues = ex:af]
Let us come back to the previously analyzed framework. Its admissible labelings are visible in Table \ref{tab:admtab}. 
We can observe that one admissible extension
can be associated with more than one labeling. However, out of the possible interpretations, only 
$\lab_3$, $\lab_9$ and $\lab_{13}$ are complete. They are now also in one--to--one relation with the complete
extensions. $\lab_3$ is the grounded labeling, while $\lab_9$ and $\lab_{13}$ are preferred. Only $\lab_{13}$ is stable; we can observe
that this is the only labeling in which no argument is assigned $\lundec$. 

\begin{table}[!ht]
\centering
\resizebox{\textwidth}{!}{
\begin{tabular}{|c|c|c|c|c|c|c|c|c|c|c|c|c|c|c|}
%
%
%
%
\hline
\multicolumn{2}{|c|}{Extension}			&	$\emptyset$ 	&	$\{A\}$ 	&	$\{A\}$ 	&	$\{C\}$ 	&	$\{C\}$ 	&	$\{D\}$ 	&	$\{D\}$ 	&	$\{A,C\}$ 	&	$\{A,C\}$ 	&	$\{A,D\}$ 	&	$\{A,D\}$ 	&	$\{A,D\}$ 	&	$\{A,D\}$ 	\\
\hline
\multirow{6}{*}{\rot{Labeling}}	&	\#	&$\lab_1$ 	&$\lab_2$ 	&$\lab_3$ 	&$\lab_4$ 	&$\lab_5$ 	&$\lab_6$ 	&$\lab_7$ 	&$\lab_8$ 	&$\lab_9$ 	&$\lab_{10}$ 	&$\lab_{11}$ 	&$\lab_{12}$ 	&$\lab_{13}$ 	\\
\hline
	&A 	&\lundec 	&\lin 	&\lin 	&\lundec 	&\lundec 	&\lundec 	&\lundec 	&\lin 	&\lin 	&\lin 	&\lin 	&\lin 	&\lin 	\\
	&B 	&\lundec 	&\lundec 	&\lout 	&\lout 	&\lundec 	&\lundec 	&\lundec 	&\lundec 	&\lout 	&\lundec 	&\lundec 	&\lout 	&\lout 	\\
	&C 	&\lundec 	&\lundec 	&\lundec 	&\lin 	&\lin 	&\lout 	&\lout 	&\lin 	&\lin 	&\lout 	&\lout 	&\lout 	&\lout 	\\
	&D 	&	\lundec 	&	\lundec 	&	\lundec 	&\lout 	&\lout 	&\lin 	&\lin 	&\lout 	&\lout 	&\lin 	&\lin 	&\lin 	&\lin 	\\
	&E 	&\lundec 	&\lundec 	&\lundec 	&\lundec 	&\lundec 	&\lundec 	&\lout 	&\lundec 	&\lundec 	&\lundec 	&\lout 	&\lundec 	&\lout 	\\
\hline
\end{tabular}
}
\caption{Admissible labelings of the framework $(\{A,B,C,D,E\}$, $\{(A,B)$, $(C,B)$, $(C,D)$, $(D,C)$, $(D,E)$, $(E,E)\})$}
\label{tab:admtab}
\end{table}
\end{example}

\section{Bipolar Argumentation}
\label{sec:bipolar}

In Dung's framework, from direct attacks we can derive defence, which can be seen as a type of a positive indirect relation
between arguments. However, defence does not account for all the possible forms of support between arguments, and a structure 
going beyond attack was required. Consequently, the notion of abstract 
support and the bipolar argumentation framework \cite{incoll:bipolar,article:coalitions,article:newbaf} were introduced, 
followed by the deductive \cite{inproc:support}, necessary \cite{incoll:afn,incoll:newafn} and evidential supports \cite{inproc:eas,inproc:moving,inproc:easafn}, 
with the latter two developed in their own dedicated frameworks. Although there are significant differences between these supports and the 
way that their dedicated frameworks model them, particularly in the context of support cycles, they have also been partially 
recreated in bipolar argumentation frameworks in order to perform a comparative study \cite{article:newbaf}. Further results on this topic can be found 
in \cite{inproc:easafn,report:trans}. 
The bipolar argumentation framework itself is defined as follows: 

\begin{definition}
The \textbf{bipolar argumentation framework} (BAF for short) is a tuple $(\args, \atts, \sups)$, where $\args$ is a set of \textbf{arguments}, 
$\atts \subseteq \args \times \args$ represents the \textbf{attack} relation and $\sups \subseteq \args \times \args$ the \textbf{support}. 
\end{definition}

We say that there is a sequence of supports between arguments $A, B \in \args$ if there is a sequence of arguments $(A, C_1,...C_n, B)$ s.t. 
$A \sups C_1$, $C_1 \sups C_2$, \ldots, $C_n \sups B$.

One of the ways bipolar argumentation frameworks model support on the semantics level is by transforming 
it into attack \cite{article:newbaf,axiomatic}. The base 
framework is extended with indirect conflicts, stemming from the interplay of existing attacks and supports. 
The resulting structure can then be evaluated like 
a Dung's framework, particularly when semantics that are at least complete are concerned. 
The type of support we try to model affects what sort of additional indirect 
attacks are created. Although this approach does not account for all possible support semantics \cite{report:trans}, it 
is sufficient for the purpose of this work. 
Typically, the following indirect conflicts are distinguished:

\begin{definition}
\label{def:indiratts}
Let $\BAF = (\args,\atts, \sups)$ be a BAF. The indirect attacks of $\BAF$ are as follows:
\begin{itemize}  
\item there is a \textbf{supported attack} from $A$ to $B$ iff there exists an argument $C$ s.t. there is a sequence of supports from $A$ to $C$
and $(C,B) \in \atts$.
\item there is a \textbf{secondary attack}\footnote{This attack was also referred to as diverted in \cite{article:coalitions}.} 
from $A$ to $B$ iff there exists an argument $C$ s.t. there is a sequence of supports from $C$ to $B$
and $(A,C) \in \atts$.
\item there is an \textbf{extended attack}\footnote{We recall only one form of the extended
attack, as other ones are already subsumed by the direct and secondary attacks.} 
from $A$ to $B$ iff there exists an argument $C$ s.t. there is a sequence of supports from $C$ to $A$
and $(C,B) \in \atts$.
\item there is a \textbf{mediated attack} from $A$ to $B$ iff there exists an argument $C$ s.t. there is a sequence of supports from $B$ to $C$
and $(A,C) \in \atts$.
\item there is a \textbf{super--mediated attack} from $A$ to $B$ iff there exists an argument $C$ s.t. there is a sequence of supports from $B$ to $C$
and a direct or supported attack from $A$ to $C$. 
\item there is a \textbf{super--extended attack}\footnote{In \cite{axiomatic} this attack is referred as an n+-attack.} from $A$ to $B$ iff there exists an argument $C$ s.t. there is a sequence of supports from $C$ to $A$
and a direct or secondary attack from $C$ to $B$.
\end{itemize}  
\end{definition}

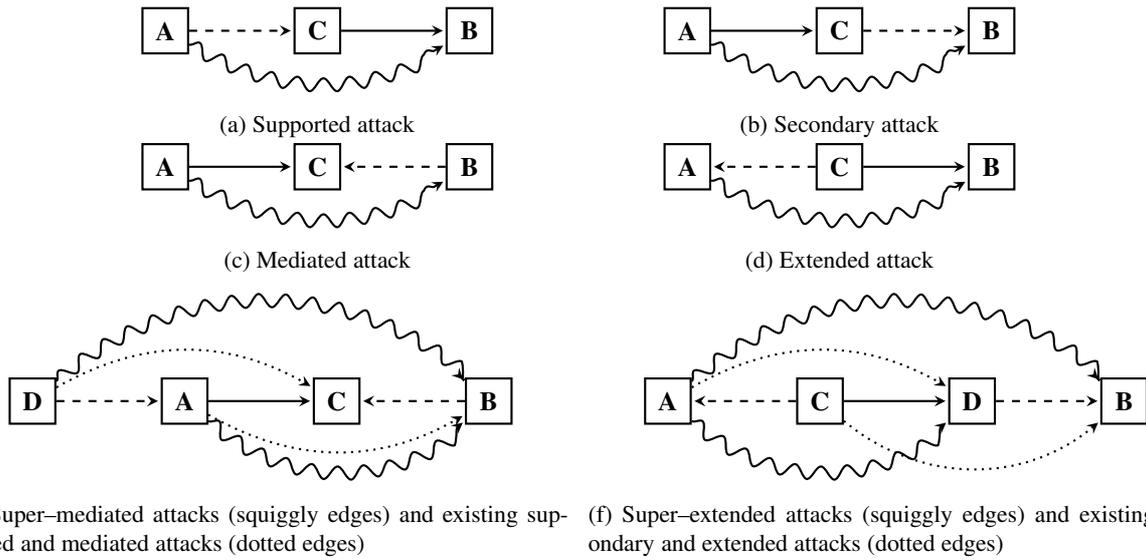
\begin{figure}[t!]
    \centering
    \begin{subfigure}[t]{0.40\textwidth}
        \centering
       
\begin{tikzpicture}
[->,>=stealth,shorten >=1pt,auto,node distance=2cm,
  thick,main node/.style={rectangle,fill=none,draw,minimum size =0.6cm,font=\normalsize\bfseries}]

\node[main node] (a) {A};
\node[main node] (c) [right of = a] {C}; 
\node[main node] (b) [right of=c] {B};

 \path
	(a) edge [color=black, bend right,snake it] node  {} (b)
	(a) edge [dashed] node {} (c)
	(c) edge node {} (b);
\end{tikzpicture}
        \caption{Supported attack}
    \end{subfigure} 
    ~ 
    \begin{subfigure}[t]{0.40\textwidth}
        \centering
    	
\begin{tikzpicture}
[->,>=stealth,shorten >=1pt,auto,node distance=2cm,
  thick,main node/.style={rectangle,fill=none,draw,minimum size =0.6cm,font=\normalsize\bfseries}]

\node[main node] (a) {A};
\node[main node] (c) [right of = a] {C}; 
\node[main node] (b) [right of=c] {B};

 \path
	(a) edge [color=black, bend right,snake it] node  {} (b)
	(a) edge node {} (c)
	(c) edge [dashed] node {} (b);
\end{tikzpicture}
        \caption{Secondary attack}
    \end{subfigure}
    
    \begin{subfigure}[t]{0.40\textwidth}
        \centering
\begin{tikzpicture}
[->,>=stealth,shorten >=1pt,auto,node distance=2cm,
  thick,main node/.style={rectangle,fill=none,draw,minimum size =0.6cm,font=\normalsize\bfseries}]

\node[main node] (a) {A};
\node[main node] (c) [right of = a] {C}; 
\node[main node] (b) [right of=c] {B};

 \path
	(a) edge [color=black, bend right,snake it] node  {} (b)
	(a) edge  node {} (c)
	(b) edge [dashed] node {} (c);
\end{tikzpicture}
     \caption{Mediated attack}
    \end{subfigure}
~
  \begin{subfigure}[t]{0.40\textwidth}
        \centering
\begin{tikzpicture}
[->,>=stealth,shorten >=1pt,auto,node distance=2cm,
  thick,main node/.style={rectangle,fill=none,draw,minimum size =0.6cm,font=\normalsize\bfseries}]

\node[main node] (a) {A};
\node[main node] (c) [right of = a] {C}; 
\node[main node] (b) [right of=c] {B};

 \path
	(c) edge [dashed] node {} (a)
	(c) edge node {} (b);
 
\path
	(a) edge [color=black, bend right,snake it] node {} (b);

\end{tikzpicture}
     \caption{Extended attack}
    \end{subfigure}

\begin{subfigure}[t]{0.49\textwidth}
    \centering 
\begin{tikzpicture}
[->,>=stealth,shorten >=1pt,auto,node distance=2cm,
  thick,main node/.style={rectangle,fill=none,draw,minimum size =0.6cm,font=\normalsize\bfseries}]

\node[main node] (a) {A};
\node[main node] (c) [right of = a] {C}; 
\node[main node] (b) [right of=c] {B};
\node[main node] (d) [left of = a] {D};

 \path
	(d) edge [snake it, bend left = 40] node {} (b)
	(a) edge [snake it, bend right = 40] node  {} (b)
	(d) edge [dashed] node {} (a)
	(d) edge [dotted, bend left] node {} (c)
	(a) edge [dotted, bend right] node  {} (b)
	(a) edge  node {} (c)
	(b) edge [dashed] node {} (c);
\end{tikzpicture}
        \caption{Super--mediated attacks (squiggly edges) and existing supported and mediated attacks (dotted edges)} 
\end{subfigure}
~
\begin{subfigure}[t]{0.49\textwidth}
    \centering 
\begin{tikzpicture}
[->,>=stealth,shorten >=1pt,auto,node distance=2cm,
  thick,main node/.style={rectangle,fill=none,draw,minimum size =0.6cm,font=\normalsize\bfseries}]

\node[main node] (c) {C}; 
\node[main node] (a) [left of = c] {A};
\node[main node] (d) [right of = c] {D}; 
\node[main node] (b) [right of = d] {B}; 

 \path
(c) edge  node {} (d)
(d) edge [dashed]  node {} (b)
(c) edge [dashed] node {} (a) 

(a) edge [color=black, bend left=40,snake it] node {} (b)
(a) edge [color=black, bend right=40,snake it] node {} (d)
(a) edge [color=black, bend left,dotted] node {} (d)
(c) edge [color=black, bend right=40,dotted] node {} (b)
;
\end{tikzpicture}
        \caption{Super--extended attacks (squiggly edges) and existing secondary and extended attacks (dotted edges)} 
\end{subfigure}
\caption{Indirect attacks in BAFs. Solid edges represent direct attacks, dashed edges represent support, and squiggly or dotted 
edges represent indirect attacks.}
\label{fig:indiratts}
\end{figure}

Additionally, the interactions between these auxiliary conflicts can also lead to the creation of new attacks. 
In particular, the super--mediated attack 
is a mixture of supported and mediated attacks (more details can be found in \cite{report:trans}). 
Nevertheless, the above notions are sufficient for the remainder of this report.  We can see them exemplified in Figure \ref{fig:indiratts}.

We would like to stress that even 
though there are many types of conflicts available, it does not mean that all of them need to be used -- the choice 
depends on what we intend
to use a given BAF for. Usually, only some of the conflicts are studied at a time, particularly if we consider specialized forms 
of support rather than just the abstract type.  
What needs to be stated explicitly is that BAFs were meant as research frameworks for analyzing the different
types of support and the consequences of their interplay with attack. Therefore, there is no \enquote{absolute} way to choose what sort of indirect attacks
need to be taken into account and different interpretations of support might call for different attacks. 
In our experiment, we will focus on finding the existing notions 
that would allow us to reproduce the relations identified by the participants, rather than on fixing the interpretation of the used supports 
and stating that given indirect conflicts should have been used.

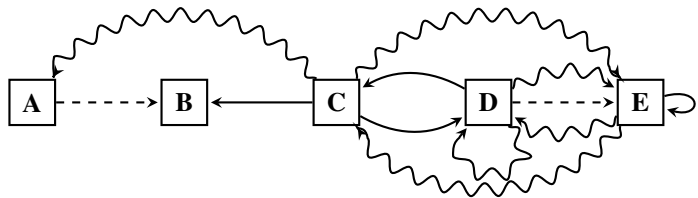
\begin{figure}[!ht]
\centering
  \begin{tikzpicture}
[->,>=stealth,shorten >=1pt,auto,node distance=2cm,
  thick,main node/.style={rectangle,fill=none,draw,minimum size = 0.6cm,font=\normalsize\bfseries}]

\node[main node] (a) {A};
\node[main node] (b) [right of=a] {B};
\node[main node] (c) [right of=b] {C};
\node[main node] (d) [right of=c] {D};
\node[main node] (e) [right of=d] {E};  
 
 \path
	(d) edge [bend left, snake it] node {} (e)

	(e) edge [bend left, snake it] node {} (d)
	
	(d) edge [loop below, looseness=5, in=230, out = 310, snake it] node {} (d)
	
	(c) edge [bend left=50, snake it] node {} (e)
	(e) edge [bend left=50, snake it] node {} (c)

	(c) edge [bend right=50, snake it] node {} (a)

	(a) edge [dashed] node {} (b)
	(c) edge node {} (b)
	(d) edge [dashed] node {} (e)
    	(e) edge [loop right] node {} (e)
 	 (c) edge [bend right] node {} (d)
    (d) edge [bend right] node {} (c);
\end{tikzpicture}
\caption{Sample bipolar argument graph. Solid edges represent attack, dashed edges represent support and squiggly edges represent indirect attacks.}
\label{fig:baf}
\end{figure}

\begin{example}[label=ex:baf] 
Consider the bipolar argumentation framework $\BAF = (\args, \atts, \sups)$ 
with $\args = \{A$, $B$, $C$, $D$, $E\}$,  $\atts= \{(C,B)$, $(C, D)$, $(D,C)$, $(E,E)\}$ and $\sups = \{(A, B)$, $(D, E)\}$, 
depicted in Figure~\ref{fig:baf}. 
We can create the following indirect conflicts for this framework. Since $D$ supports $E$, which is a self--attacker, we can create a supported 
attack $(D, E)$ and a (super) mediated one $(E, D)$. These conflicts also lead to the super--mediated attack $(D, D)$. 
Due to the same support, we can create a secondary attack $(C, E)$ and a (super) extended one $(E, C)$. 
As $C$ attacks $B$, which is supported by $A$, we can create a (super) mediated attack $(C, A)$.
 
If we decide to use all of the aforementioned attacks, then the Dung's framework associated with $\BAF$ is $\AF = (\args, \atts')$, 
where $\atts' = \atts \cup \{(C, A)$, $(C, E)$, $(D, D)$, $(D, E)$, $(E, C)$, $(E, D)\}$. The sets $\emptyset$ and $\{C\}$ are its admissible and complete extensions. $\{C\}$ is the stable and preferred extension, while $\emptyset$ is grounded.  
Following the approach from \cite{article:newbaf}, we can treat these extensions as the extensions of $\BAF$. 

If we consider only the secondary and (super) extended attacks, we obtain a Dung's graph with the set 
of attacks $\atts'' = \atts \cup \{(C,E)$, $(E,C)\}$. The admissible extensions of this framework are
$\emptyset$, $\{A\}$, $\{C\}$, $\{D\}$, $\{A, C\}$, $\{A, D\}$, $\{B, D\}$ and $\{A, B, D\}$. 
Out of this, $\{A\}$, $\{A,C\}$ and $\{A,B,D\}$ are complete, with the first set being grounded and the other two preferred. Only
$\{A,C\}$ is stable. 
\end{example}

\section{Probabilistic Argumentation}
\label{sec:probarg}

The proposals for probabilistic argumentation extend Dung's framework to address various aspects of uncertainty arising in argumentation. 
The two main approaches to probabilistic argumentation are the constellations and the epistemic approaches \cite{Hunter:2013}. 
In the {\bf constellations approach}, the uncertainty is in the topology of the graph. This approach is useful when one agent is not sure 
what arguments and attacks another agent is aware of. This can be captured by a probability distribution over the space of possible 
argument graphs, where each graph has a chance of being the real model of the agent. In the {\bf epistemic approach}, the topology of the argument graph is fixed, but there is uncertainty as to the degree 
to which each argument is believed. It thus provides us with a more fine--grained approach towards modelling the acceptability of arguments. 
In this section we provide a brief revision of these two probabilistic formalisms and for further reading concerning 
the differences between them we refer the readers to \cite{Hunter:2013,probattack}.

\subsection{Epistemic Probabilistic Argumentation}
\label{sec:epistemic} 

We now turn to the epistemic approach to probabilistic argumentation \cite{Thimm:2012,Hunter:2013,HT14ecai,BGV14,HunterThimm17}. 
In this section we review and extend the results from \cite{HT14arxiv}. 

\begin{definition}
A {\bf mass distribution} over arguments $\args$ is a function $\prob:2^\args \rightarrow \lbrack 0, 1 \rbrack$ s.t. $\sum_{\ext \subseteq \args} \prob(\ext) = 1$.
The {\bf probability of an argument} $A \in \args$ 
is  $\prob(A)  = \sum_{\ext \subseteq \args \mbox{ s.t. } A \in \ext} \prob(\ext)$.
\end{definition}
 
The probability of a single argument is understood as the belief that an agent has in it, i.e. we 
say that an agent believes an argument $A$ to some degree when $\prob(A) > 0.5$,
disbelieves an argument to some degree when $\prob(A) < 0.5$,
and neither believes nor disbelieves an argument when $\prob(A) = 0.5$. This belief can be interpreted in various ways 
\cite{Hunter:2013}, for 
example, if we assume that an argument has a standard premise--conclusion structure, then the degree of belief in an 
argument can be seen as the degree to which the 
agent believes the premises and the conclusion drawn from those premises.  

Just like argument graphs come equipped with different types of semantics that are meant to capture various intuitions as to what
is a \enquote{good} collection of arguments (and/or attacks), the probabilistic frameworks are accompanied by a number of postulates
that capture the properties of a probability distribution. While classical semantics tend to represent a number of 
properties at the same time, a single postulate focuses on a single aspect at a time. Thus, while there are many epistemic 
postulates, they 
tend to be quite simple. They also allow 
a more detailed view on the participant behaviour and can allow us to analyze the cases in which classic semantics may fail to explain it. 
Consequently, they can 
provide more feedback to argumentation systems, such as for computational persuasion, than normal semantics do. 
For example, the low performance of the argumentation semantics such as complete does not really inform the system what aspect of the participant 
reasoning does not meet the semantics requirements. In contrast, through the use of the postulates we could analyze whether 
the issue lies within conflict--freeness, defense or not accepting/rejecting arguments that should be accepted/rejected. 

We will recall some of the existing postulates for the epistemic semantics and introduce certain argument--centered alternatives of the 
properties from \cite{probattack}, namely the preferential postulate PRE, strict STC, 
protective PRO, restrained RES, discharging DIS, guarded GRD, trusting TRU, anticipating ANT, demanding DEM, binary BIN and n--valued VAL$^n$. 
The available postulates can be grouped in various ways. In what follows, we separate them into the following, not necessarily disjoint, types: 
preferential (rational), explanatory and value families.

\subsubsection{Preferential Postulates}

The first family of postulates focuses on resolving a conflict between two arguments and deciding how 
believed or disbelieved the attacker and attackee should be. They are primarily of the form \enquote{\textit{for every $A, B \in \args$, 
if $(A, B) \in \atts$, then $X$ holds}}, where $X$ specifies the conditions on the beliefs assigned to the attacker and the attackee. 
The intuitions behind the introduced postulates are as follows. 
Let us assume that we believe an argument attacking another argument. The question now is what should be our belief in the attackee. The least restrictive, PRE postulate, allows 
us to believe the attackee 
as long as it is believed more than the attacker. If we follow RAT, then we do not believe the attackee, and if STC, 
then we disbelieve it. In a dual manner, the PRO postulate ensures that if we believe the attackee, we disbelieve its attackers. 
The RES postulate strengthens PRO and STC and is equivalent to saying that if two arguments are in conflict, 
at least one of them should be disbelieved. 
Hence, we no longer have the option to be undecided about an argument. 

\begin{definition}
\label{def:rational}
A probability mass distribution $\prob$ for $\AF$ is:   
\begin{itemize}
\item[{\bf (PRE)}] {\bf preferential} 
if for every $A,B \in \args$ s.t. $(A,B) \in \atts$, if $\prob(A) >0.5$ and $\prob(B)>0.5$, then $\prob(A) < \prob(B)$.

\item[{\bf (RAT)}] {\bf rational} 
 if for every $A, B \in \args$ s.t. $(A, B) \in \atts$, $\prob(A)  > 0.5$ implies $\prob(B) \leq 0.5$.

\item[{\bf (STC)}] {\bf strict} 
 if for every $A, B \in \args$ s.t. $(A, B) \in \atts$, $\prob(A)  > 0.5$ implies $\prob(B) < 0.5$.

\item[{\bf (PRO)}] {\bf protective} 
if for every $A, B \in \args$ s.t. $(A, B) \in \atts$, $\prob(B)  > 0.5$ implies $\prob(A) < 0.5$.

\item[{\bf (RES)}] {\bf restrained} 
if for every $A,B \in \args$, if $(A,B) \in \atts$ then $\prob(A) \geq 0.5$ implies $\prob(B) < 0.5$.

%

\end{itemize}
\end{definition}

\subsubsection{Explanatory Postulates}

We now come to the explanatory family. The purpose of the postulates of this type is to demand that the degree of belief assigned to an argument 
is justified by the degrees of belief associated with the arguments related to it. They are roughly of the form \enquote{\textit{for every 
argument $B \in \args$, if $P(B)$ meets a condition $X$ then it has an attacker $A$ s.t. $P(A)$ meets condition $Y$}}
or \enquote{\textit{for every 
argument $B \in \args$, if for all of its attackers $A$, $P(A)$ meets condition $Y$, then $P(B)$ meets a condition $X$}}.
For instance, DIS demands that if an argument is disbelieved, then it possesses a believed attacker. GRD postulate relaxes this requirement by allowing the use of undecided attackers as well. 
On the other hand, the TRU property requires us to believe an argument when we have no reason against it, i.e. when all of its attackers are 
disbelieved. ANT modifies TRU by saying that lack of believed attackers is a good reason to believe the attackee. Finally, the DEM property 
requires that a completely disbelieved argument has to be paired with a completely believed attacker, and a completely believed argument can 
only be attacked by completely disbelieved arguments. 

\begin{definition}
\label{def:explanatory}
A probability mass distribution $\prob$ for $\AF$ is:   
\begin{itemize}
\item[{\bf (DIS)}] {\bf discharging} 
if for every $B \in  \args$, 
if $\prob(B) < 0.5$ then there exists an argument $A \in  \args$ s.t. 
$(A, B) \in \atts$ and $\prob(A) > 0.5$. 

\item[{\bf (GRD)}] {\bf guarded} 
if for every $B \in  \args$, 
if $\prob(B) < 0.5$ then there exists an argument $A \in  \args$ s.t. 
$(A, B) \in\atts$ and $\prob(A) \geq 0.5$. 

\item[{\bf (TRU)}] {\bf trusting} 
if for every $B \in \args$, 
if $\prob(A) <0.5$ for all $A \in \args$ s.t. $(A, B) \in \atts$, then $\prob(B) >0.5$. 
 
\item[{\bf (ANT)}] {\bf anticipating} 
if for every $B \in \args$, 
if $\prob(A) \leq 0.5$ for all $A \in \args$ s.t. $(A, B) \in\atts$, then $\prob(B) >0.5$. 

\item[{\bf (DEM)}] {\bf demanding} 
if for every $A \in \args$, if $\prob(A) = 1$, then for every $B \in \args$ s.t. $(B,A) \in \atts$, $\prob(B) = 0$,
and if $\prob(A) = 0$, then $\exists B \in \args$ s.t. $(B, A) \in \atts$ and $\prob(B) = 1$.
\end{itemize}
\end{definition}

\subsubsection{Value Postulates} 
 
All of the previously listed postulates are relatively general and tell us whether to believe
or disbelieve an argument, but not to what degree. This is where the value family of postulates comes in. 
The FOU and SFOU postulates tell us how much we should believe initial arguments. 
OPT and SOPT provide us with lower bounds for the degrees of belief we have in an element based on the beliefs we have in its attackers. 
The BIN postulate prohibits any indecisiveness we may have about the arguments. MIN, NEU, MAX permit the use of only $0$, $0.5$ and $1$ 
degrees of belief respectively, while TER allows for all three. Finally, with VAL$^n$ postulate we can distinguish those distributions that assign to their 
arguments no more than $n$ distinct values in total. 

\begin{definition}
\label{def:value}
A probability mass distribution $\prob$ for $\AF$ is:   
\begin{itemize}
\item[{\bf (SFOU)}] {\bf semi--founded} 
if $\prob(A) \geq 0.5$ for every initial $A \in \args$. 

\item[{\bf (FOU)}] {\bf founded} 
if $\prob(A) = 1$ for every initial $A \in  \args$.
 
\item[{\bf (SOPT)}] {\bf semi--optimistic} 
if 
$\prob(A) \geq 1 - \sum_{B \in \{A\}^-}\prob(B)$ for every $A \in  \args$ that is not initial. 

\item[{\bf (OPT)}] {\bf optimistic} 
if  $\prob(A) \geq 1 - \sum_{B\mid (B,A) \in \atts}\prob(B)$ for every $A \in  \args$.

\item[{\bf (BIN)}] {\bf binary} 
 if for no $A \in \args$, $\prob(A) = 0.5$.

\item[{\bf (TER)}] {\bf ternary} 
 if $\prob(A) \in \{0, 0.5, 1\}$ for every $A \in  \args$.

\item[{\bf (NEU)}] {\bf neutral} 
if $\prob(A) = 0.5$ for every $A \in \args$.

\item[{\bf (MAX)}] {\bf maximal} 
if $\prob(A) = 1$ for every $A \in \args$. 

\item[{\bf (MIN)}] {\bf minimal} 
if $\prob(A) = 0$ for every $A \in \args$.

\item[{\bf (VAL$^n$)}] {\bf n--valued} 
if $\lvert \{ x \mid \exists A \in \args$ s.t. $\prob(A) = x\}\rvert \leq n$.
\end{itemize}
\end{definition} 

\subsubsection{Multi--type Postulates}

Finally, we have the postulates that can be seen as shared between the families. 
In particular, on the intersection of the rational (preferential) and value types 
are the COH, INV and JUS postulates. COH gives us the upper bound of the belief we might have in an argument based on
 its (strongest) attacker. By combining this upper bound with the lower bound from the OPT postulate, 
we receive the JUS property. The INV postulate 
requires that the belief in the attackee is dual to the belief in the attacker. 
Please note that the DEM, SFOU and FOU postulates can be seen as shared between the value and explanatory families, rather than belonging 
to just one of them. We leave it to the reader to classify them as he or she sees fit.  

\begin{definition}
\label{def:shared}
A probability mass distribution $\prob$ for $\AF$ is:   
\begin{itemize} 
\item[{\bf (COH)}] {\bf coherent} 
 if for every $A, B \in  \args$ s.t.  $(A, B) \in \atts$, $\prob(A) \leq 1 - \prob(B)$. 

\item[{\bf (INV)}] {\bf involutary} 
if for every $A,B \in \args$, if $(A, B) \in \atts$ then $\prob(A) = 1 - \prob(B)$. 

\item[{\bf (JUS)}] {\bf justifiable} 
if it is coherent and optimistic.
\end{itemize}
\end{definition}

\subsubsection{Properties of Epistemic Postulates}

\begin{figure}[!h]
\centering
\begin{tikzpicture}[->,>=latex,thick,,scale=0.9,main node/.style={draw=none,fill=none,font=\small\bfseries}]
\node[main node] (ALL) at (6,7.5) {${\cal P}$};

\node[main node] (PRE) at (11,7.5) {${\cal P}_{PRE}$};

\node[main node] (RAT) at (11,6) {${\cal P}_{RAT}$}; 

\node[main node] (SFOU) at (2,6) {${\cal P}_{SFOU}$};

\node[main node] (STC) at (10,4.5) {${\cal P}_{STC}$};
\node[main node] (PRO) at (12,4.5) {${\cal P}_{PRO}$};

\node[main node] (TRU) at (0,4.5) {${\cal P}_{TRU}$}; 
\node[main node] (GRD) at (2,4.5) {${\cal P}_{GRD}$};  
\node[main node] (FOU) at (4,4.5) {${\cal P}_{FOU}$};  
\node[main node] (SOPT) at (6,4.5) {${\cal P}_{SOPT}$};

\node[main node] (COH) at (10,3) {${\cal P}_{COH}$};

\node[main node] (ANT) at (0,3) {${\cal P}_{ANT}$}; 
\node[main node] (DIS) at (2,3) {${\cal P}_{DIS}$}; 
\node[main node] (OPT) at (4,3) {${\cal P}_{OPT}$}; 
\node[main node] (INV) at (6,3) {${\cal P}_{INV}$};

\node[main node] (MAX) at (1,1.5) {${\cal P}_{MAX}$};  
\node[main node] (NEU) at (3,1.5) {${\cal P}_{NEU}$};
\node[main node] (JUS) at (5,1.5) {${\cal P}_{JUS}$};
\node[main node] (MIN) at (10,1.5) {${\cal P}_{MIN}$};
\node[main node] (RES) at (12,3) {${\cal P}_{RES}$};

\node[main node] (EMP) at (6,0) {$\emptyset$};

\draw[] (PRE)--(ALL)  ;   
\draw[] (SOPT)--(ALL)  ;   
\draw[] (SFOU)|-(ALL)  ;   		

\path (RAT) edge (PRE);			
\path (PRO) edge (RAT);
\path (COH) edge (PRO);			
\path (JUS) edge (COH);
\path (STC) edge (RAT);
\path (COH) edge (STC); 	
\draw[] (RES)--(PRO)  ; 
\draw[] (RES)--(STC)  ; 

\draw[] (OPT)--(SOPT)  ;   		
\draw[] (JUS)|-(OPT)  ; 

\draw[] (EMP)-|(JUS)  ;	
\draw[] (EMP)-|(RES)  ;
\draw[] (EMP)-|(MIN)  ;
\draw[] (EMP)-|(NEU)  ;
\draw[] (EMP)-|(MAX)  ;

\draw[] (FOU)--(SFOU)  ; 
\draw[] (OPT)--(FOU)  ;   
\draw[] (TRU)|-(SFOU)  ; 
\draw[] (GRD)--(SFOU)  ; 
\draw[] (ANT)--(TRU)  ; 
\draw[] (DIS)--(GRD)  ;

\draw[] (NEU)--(INV)  ;  
\draw[] (NEU)|-(DIS)  ;
\draw[] (INV)--(SOPT);
\draw[] (INV)--(COH)  ;
\draw[] (MIN)--(COH)  ;
\draw[] (MAX)--(OPT)  ;
\draw[] (MAX)-|(ANT)  ;
\draw[] (MAX)|-(DIS)  ;

\end{tikzpicture}
\begin{tikzpicture}[->,>=latex,thick,,scale=0.9,main node/.style={draw=none,fill=none,font=\small\bfseries}]
\node[main node] (ALL) at (7,0) {${\cal P}$};  
\node[main node] (EMP) at (0,0) {$\emptyset$};
\node[main node] (TER) at (5,0) {${\cal P}_{TER}$}; 
\node[main node] (BIN) at (5,-1.5) {${\cal P}_{BIN}$}; 
 
\node[main node] (NEU) at (2.5,1.5) {${\cal P}_{NEU}$}; 
\node[main node] (MIN) at (2.5,0) {${\cal P}_{MIN}$};
\node[main node] (MAX) at (2.5,-1.5) {${\cal P}_{MAX}$};   
 \node[main node] (DEM) at (5,1.5) {${\cal P}_{DEM}$};

\draw[] (TER)--(ALL)  ;   		
\draw[] (BIN)-|(ALL)  ;   		
\draw[] (DEM)-|(ALL)  ;
\draw[] (NEU)--(DEM)  ;
\draw[] (EMP)|-(MIN)  ;
\draw[] (EMP)|-(NEU)  ; 
\draw[] (EMP)|-(MAX)  ; 
\draw[] (NEU)--(TER)  ;
\draw[] (MIN)--(TER)  ;
\draw[] (MAX)--(TER)  ;
\draw[] (MIN)--(BIN)  ;
\draw[] (MAX)--(BIN)  ; 
\end{tikzpicture}
\caption{Classes of probability functions 
where ${\cal P}_{\mu_1} \rightarrow {\cal P}_{\mu_2}$ denotes ${\cal P}_{\mu_1} \subseteq {\cal P}_{\mu_2}$}
\label{fig:hasseepiarg}
\end{figure}
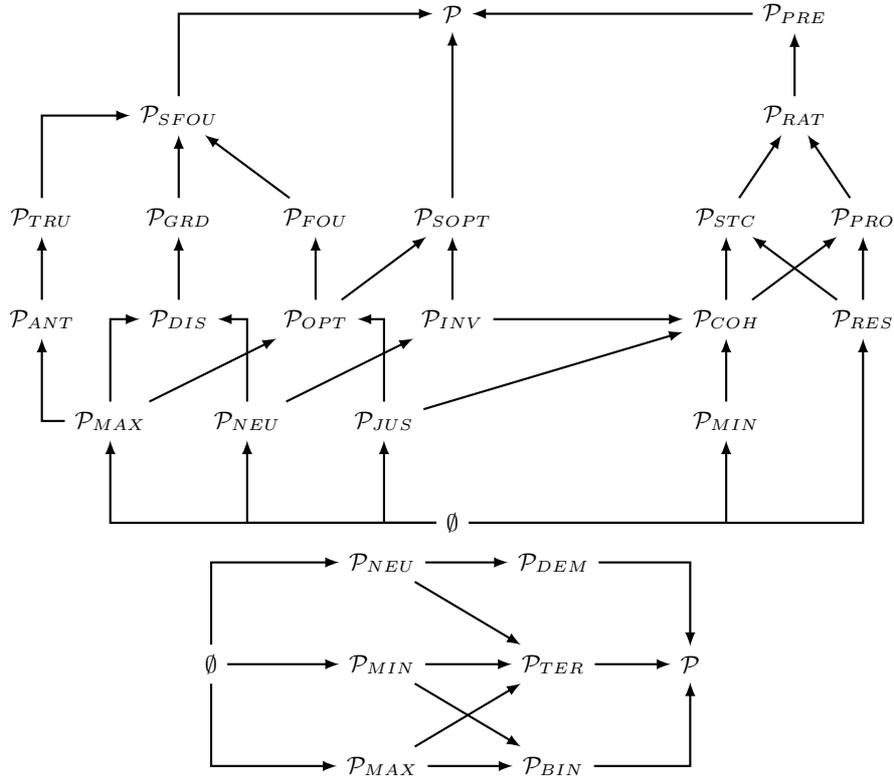

In Proposition \ref{prop:eproperties} and Figure \ref{fig:hasseepiarg} we show some of the relations between the presented postulates, extending 
the results available in \cite{Hunter:2016a}. 
These properties make it more explicit which postulates are more restrictive than the others and show additional 
connections between the different 
families of postulates we have presented. 

\begin{restatable}{proposition}{eproperties}
\label{prop:eproperties}
Let $\AF$ be an argument graph and let $\allprob_X$ be the collection of all distributions on $\AF$ satisfying postulate $X$. 
The following holds:
\begin{center}
\renewcommand*{\arraystretch}{1.3}
\begin{tabular}{c c} 

$\allprob_{OPT}(\AF) = \allprob_{SOPT}(\AF) \cap \allprob_{FOU}(\AF)$ 
&$\allprob_{TER}(\AF) \cap \allprob_{TRU}(\AF) \subseteq \allprob_{FOU}(\AF)$ \\

\multicolumn{2}{c}{$(\allprob_{COH}(\AF) \cap \allprob_{TER}(\AF))  = (\allprob_{PRO}(\AF) \cap \allprob_{STC}(\AF)  \cap \allprob_{TER}(\AF))$} \\

\multicolumn{2}{c}{$\allprob_{INV}(\AF) \cap \allprob_{FOU} \subseteq \allprob_{DEM}(\AF)$} \\
\end{tabular}
\end{center}
\end{restatable}

\begin{example}[continues=ex:af] 
\label{ex:postulate1}
Let us come back to the argument graph $\AF = (\{A,B,C,D,E\}$, $\{(A,B)$, $(C,B)$, $(C,D)$, $(D,C)$, $(D,E)$, $(E,E) \})$ from Figure \ref{fig:af}. 
In Table \ref{tab:post} we have listed a number of probability distributions and marked which postulates are or are not satisfied 
by them under $\AF$. For the sake of simplicity, 
we have only focused on the resulting argument probabilities, not on the full description of the distributions.   
\end{example}

\begin{table}[!ht]
\centering
\resizebox{\textwidth}{!}{
\renewcommand*{\arraystretch}{1.2}
\begin{tabular}{|c|c|c|c|c|c|c|c|c|c|c|c|c|c|c|c|c|c|c|c|c|c|c|c|}
\hline  
\#	&	 A 	&	 B 	&	 C 	&	 D 	&	 E 	&	\rot{PRE} 	&	\rot{RAT} 	&	\rot{STC} 	&	\rot{PRO} 	&	\rot{RES} 	&	\rot{COH} 	&	 \rot{JUS}	&	 \rot{INV}	&	\rot{DIS} 	&	\rot{GRD}	&	\rot{TRU} 	&	\rot{ANT}	&	\rot{DEM}	&	\rot{SFOU} 	&	\rot{FOU} 	&	\rot{SOPT} 	&	\rot{OPT}	\\
\hline																																													
$\prob_1$ 	&	0.6	&	0.5	&	0.2	&	0.4	&	0.8	&	\xm	&	\xm 	&	\xm 	&	\xm 	&	\xm	&	\xm 	&	\xm	&	\xm	&	\xm 	&	\xm	&	\xm 	&	\xm	&	\cm	&	\cm 	&	\xm 	&	\xm 	&	\xm 	\\
\hline																																													
$\prob_2$ 	&	0.3	&	0.5	&	0.6	&	0.2	&	0.5	&	\cm	&	\cm 	&	\xm 	&	\cm 	&	\xm	&	\xm 	&	\xm	&	\xm	&	\xm 	&	\xm	&	\xm 	&	\xm	&	\cm	&	\xm 	&	\xm 	&	\xm 	&	\xm 	\\
\hline																																													
$\prob_3$ 	&	0.7	&	0.4	&	0.6	&	0.4	&	0.5	&	\cm	&	\cm 	&	\cm 	&	\cm 	&	\xm	&	\xm 	&	\xm	&	\xm	&	\cm 	&	\cm	&	\cm 	&	\xm	&	\cm	&	\cm 	&	\xm 	&	\cm 	&	\xm 	\\
\hline																																													
$\prob_4$ 	&	1	&	0.5	&	1	&	0	&	0.4	&	\cm	&	\cm 	&	\xm	&	\cm 	&	\cm	&	\xm 	&	\xm	&	\xm	&	\xm 	&	\xm	&	\xm 	&	\xm	&	\cm	&	\cm 	&	\cm 	&	\xm 	&	\xm	\\
\hline																																													
$\prob_5$ 	&	1	&	0	&	1	&	0	&	0.5	&	\cm	&	\cm 	&	\cm 	&	\cm 	&	\xm	&	\cm 	&	\cm	&	\xm	&	\cm 	&	\cm	&	\cm 	&	\xm	&	\cm	&	\cm 	&	\cm 	&	\cm 	&	\cm 	\\
\hline																																													
$\prob_6$ 	&	1	&	0	&	0.5	&	0.4	&	0.5	&	\cm	&	\cm 	&	\cm 	&	\cm 	&	\xm	&	\cm 	&	\xm	&	\xm	&	\xm 	&	\cm	&	\xm 	&	\xm	&	\cm	&	\cm 	&	\cm 	&	\xm 	&	\xm 	\\
\hline																																													
$\prob_7$ 	&	0.5	&	0	&	1	&	0	&	0.5	&	\cm	&	\cm 	&	\cm 	&	\cm 	&	\xm	&	\cm 	&	\xm	&	\xm	&	\cm 	&	\cm	&	\xm 	&	\xm	&	\cm	&	\cm 	&	\xm 	&	\cm 	&	\xm 	\\
\hline																																													
$\prob_8$ 	&	0.5	&	1	&	0.5	&	0.5	&	0.5	&	\cm	&	\cm 	&	\cm 	&	\xm 	&	\xm	&	\xm 	&	\xm	&	\xm	&	\cm 	&	\cm	&	\xm 	&	\xm	&	\xm	&	\cm 	&	\xm 	&	\cm 	&	\xm 	\\
\hline																																													
$\prob_9$ 	&	0.6	&	1	&	0.4	&	0.7	&	0.5	&	\cm	&	\xm 	&	\xm	&	\xm 	&	\xm	&	\xm 	&	\xm	&	\xm	&	\cm 	&	\cm	&	\cm	&	\cm	&	\xm	&	\cm 	&	\xm 	&	\cm 	&	\xm	\\
\hline						
$\prob_{10}$ 	&	0.5	&	0.5	&	0.5	&	0.5	&	0.5	&	\cm	&	\cm 	&	\cm 	&	\cm 	&	\xm	&	\cm 	&	\xm	&	\cm	&	\cm 	&	\cm	&	\xm	&	\xm	&	\cm	&	\cm 	&	\xm 	&	\cm 	&	\xm	\\
\hline							
\end{tabular}
}
\caption{Satisfaction of epistemic postulates on Dung's framework from Figure \ref{fig:af}.}
\label{tab:post}
\end{table}

\subsubsection{Relation to Labeling--based Semantics}

In the previous sections we could have observed that the epistemic approach offers a wide variety of postulates describing certain 
properties that a given probability function may or may not possess. In this section, we would like to show how some of the combinations of these
postulates can capture the intuitions behind the standard labeling based semantics. This information will become useful once we analyze 
the postulate adherence in our experiment. 

\begin{definition}
Let $\AF = (\args, \atts)$ be an argument graph and let  $\prob$ be a probability distribution. 
The {\bf epistemic labeling} associated with $\prob$ 
is  
$\lab_{\prob} = (I, O, U)$, where 
$I = \{ A \in \args \mid \prob(A) > 0.5 \}$, 
$O = \{ A \in \args \mid \prob(A) < 0.5 \}$, and 
$U = \{ A \in \args \mid \prob(A) = 0.5 \}$. 
A labeling
$\lab$ and a probability function $\prob$ are \textit{congruent}, denoted $\lab \sim \prob$, if for all $A \in \args$ we have that 
$\lab(A) = \lin \Leftrightarrow \prob(A) = 1$, $\lab(A) = \lout \Leftrightarrow \prob(A) = 0$, and 
$\lab(A) = \lundec \Leftrightarrow \prob(A) = 0.5$. 
\end{definition} 

Please note that if $\lab \sim \prob$, then $\lab = \lab_{\prob}$, i.e. if a labeling and a probability function are \textbf{congruent} then 
this labeling is also an epistemic labeling of this function. 

\begin{restatable}{proposition}{propepiclassem}
\label{prop:epiclassem}
Let $\prob$ be an epistemic distribution and let $\AF= (\args,  \atts)$ be a Dung's framework. 
\begin{itemize} 
\item $\prob \in \allprob_{RAT}(\AF) \cap \allprob_{DIS}(\AF)$ iff $\lab_{\prob} \in \cf(\AF)$.
\item $\prob\in \allprob_{PRO}(\AF) \cap \allprob_{DIS}(\AF)$ iff $\lab_{\prob} \in \ad(\AF)$.
\item $\prob \in \allprob_{PRO}(\AF)  \cap \allprob_{STC}(\AF) \cap \allprob_{DIS}(\AF) \cap \allprob_{TRU}(\AF)$.  
iff $\lab_{\prob} \in \co(\AF)$.
\end{itemize}   
\end{restatable} 

If we were to consider retrieving the extension--based semantics through epistemic postulates, the above results should be 
rephrased in the following 
manner: if a probability distribution satisfies the given postulates, then the set of $\lin$ arguments of the associated labeling is 
an extension 
of a given type, and if a set of arguments is an extension of a given type, then there exists a probability distribution satisfying the 
given postulates
s.t. the set of believed arguments coincides with the extension. The only relaxation we are aware of is in the case of conflict--free semantics, 
where a set of arguments is conflict--free iff the associated probability distribution is rational. 

\begin{example}[continues=ex:af]
For every labeling listed in Table \ref{tab:admtab}, we can create a congruent probability function and show that it satisfies the required postulates. 
For example, we can consider the admissible labeling $\lab_{6}$ and its congruent distribution s.t. $\prob(A) = \prob(B) = \prob(E) = 0.5$, 
$\prob(C) = 0$ and $\prob(D) = 1$. The protective and discharging postulates are easily satisfied, however, the trusting and strict ones are 
not -- in particular, $A$ should have been believed and $B$, $E$ disbelieved. Addressing these issues would produce a distribution that  is associated with 
the complete labeling $\lab_{13}$.
 
Let us now look at the distributions in Table \ref{tab:post}. All $\prob_3$, $\prob_5$, $\prob_7$, $\prob_8$ and $\prob_{10}$ are both rational 
and discharging. It is easy to verify that the associated labelings 
$\lab_{\prob_3} =\lab_{\prob_5} =  \{A:\lin, B:\lout, C:\lin, D:\lout, E:\lundec\}$,
$\lab_{\prob_7} = \{A:\lundec, B:\lout, C:\lin, D:\lout, E:\lundec\}$, 
$\lab_{\prob_8} = \{A:\lundec, B:\lin, C:\lundec, D:\lundec, E:\lundec\}$ and 
$\lab_{\prob_{10}} = \{A:\lundec, B:\lundec, C:\lundec, D:\lundec, E:\lundec\}$ are conflict--free. All of them, with the exception of 
$\lab_{\prob_8}$, are also admissible -- at the same time, we can observe that $\prob_8$ is not protective. Finally, only $\lab_{\prob_3}$ and 
$\lab_{\prob_5}$ are complete, which is reflected by $\prob_3$ and $\prob_5$ being the only of the listed functions that are additionally strict and trusting.
\end{example}

Therefore, as we can observe, epistemic probability, understood as the degree of belief, can also express various levels of agreement in the Dung's sense, 
i.e. accepting, rejecting and being undecided about an argument.

\subsection{Constellation Probabilistic Argumentation}
\label{sec:constellation}

Here we review the constellation approach to probabilistic argumentation from \cite{Hunter:2012}, 
which extends the methods from \cite{Dung:2010} and \cite{Li:2011}. 
The constellation approach allows us to represent the uncertainty over the topology of the graph. Each subgraph of the original graph is assigned a probability which is understood as the chances of it being the actual argument graph of the agent.
It can be used to model what arguments and attacks an agent is aware of.  If our uncertainty is about which arguments appear in the graph, then only the full (induced) subgraphs of the argument graph have a non--zero probability.
If we are only uncertain about which attacks appear, then it is the spanning subgraphs of the argument graph that can have 
a non--zero probability.
\begin{definition}
Let $\AF = (\args, \atts)$ and $\AF' = (\args', \atts')$ be two argument graphs. 
$\AF'$ is a subgraph of $\AF$, denoted $\AF' \sqsubseteq \AF$, iff
$\args' \subseteq \args$ and $\atts' \subseteq (\args' \times \args') \cap \atts$.
The set of subgraphs of $\AF$ is $\subs(\AF) = \{\AF' \mid \AF' \sqsubseteq \AF \}$. 
A subgraph $(\args',\atts')$ is {\bf full} iff $\args' \subseteq \args$
and $\atts' = (\args' \times \args') \cap \atts$.
A subgraph $(\args',\atts')$ is {\bf spanning} iff $\args' = \args$
and $\atts' \subseteq \atts$.
\end{definition}

\begin{definition}
A {\bf subgraph distribution} is a function $\prob^c: \subs(\AF) \rightarrow \lbrack 0, 1 \rbrack$ with
$\sum_{\AF' \in \subs(\AF)} \prob^c(\AF') = 1$.
A  subgraph distribution $\prob^c$ is
a {\bf full subgraph distribution}
iff if $(\args',\atts')$ is not a full subgraph,
then
$\prob^c((\args',\atts')) = 0$.
A  subgraph distribution $\prob^c$ is
a {\bf spanning subgraph distribution}
iff if $(\args',\atts')$ is not a spanning subgraph,
$\prob^c((\args',\atts')) = 0$.
\end{definition}

The constellation semantics can be seen as a two--level construct. Each graph in the subgraph distribution is evaluated with 
a given base semantics, the results of which are later paired with the probabilities of the frameworks producing them, thus 
leading to constellation semantics. Although we will not be using this particular method for evaluating argument graphs 
in this paper, we briefly recall and exemplify it so that it is more clear that it is different from the epistemic approach. 
Further details can be found in \cite{Hunter:2013,probattack}.

Determining the probability that a set of arguments or a labeling follows the semantics 
of a particular type (e.\,g.\ grounded, preferred, etc.) 
is done by collecting the probabilities of the subgraphs
producing the desired extensions or labelings. 
In a similar fashion, we can derive the probability of an argument being accepted in a labeling of a given type.

\begin{definition}
For $\ext \subseteq A$ and $\sigma \in \{\cf, \semantics\}$,
the probability 
that 
$\lab: \ext \rightarrow \{\lin, \lout,\lundec\}$ is a $\sigma$--labeling  is:
\[
\prob_{\sigma}(\lab) = \sum_{\AF' \in \subs(\AF) \mbox{ s.t. } \lab \in \sigma(\AF')} \prob^c(\AF')
\]
\end{definition}

\begin{definition}
Given a semantics $\sigma \in \{\semantics\}$, 
the probability that $A \in \args$ is assigned an $\lin$ status in a $\sigma$--labeling is 
\[
\prob_{\sigma}(A) = \sum_{\AF' \in \subs(\AF) \mbox{ s.t. } \lab \in \sigma(\AF') \mbox{ and } A \in \lin(\lab)} \prob(\AF')
\]
\end{definition}

\begin{example}
\label{ex:constell}
Consider the graph $\AF = (\{A,B\}, \{(A,B)\}$.  Its subgraphs are 
$\AF_1 =(\{ A, B \}, \{ (A,B)\}$,  	
$\AF_2=(\{ A, B \},\emptyset)$,   		
$\AF_3=(\{ A \}, \emptyset)$,  		
$\AF_4=(\{ B \}, \emptyset)$ and 		
$\AF_5 =(\emptyset, \emptyset)$. 	 
Out of them, $\AF_1$, $\AF_3$, $\AF_4$ and $\AF_5$ are full, and $\AF_1$ and $\AF_2$ are spanning. 
 Consider the following subgraph distribution $\prob^c$: $\prob^c(\AF_1) = 0.09$, 
$\prob^c(\AF_2) = 0.81$, $\prob^c(\AF_3) = 0.01$ and $\prob^c(\AF_4) = 0.09$ and $\prob^c(\AF_5) = 0$. 
The probability of a given set being a grounded extension is as follows:
$\prob_{\sf gr}(\{A,B\})$ =	$\prob^c(\AF_2)$ = 0.81; 
$\prob_{\sf gr}(\{A\})$  =	$\prob^c(\AF_1) + \prob^c(\AF_3)$ =  0.1; 
$\prob_{\sf gr}(\{B\})$  =	$\prob^c(\AF_4)$	=  0.09;
and 
$\prob_{\sf gr}(\{\})$ = $\prob^c(\AF_5)$ =  0.
Therefore $\prob_{\gr}(A) = 0.91$ and $\prob_{\gr}(B) = 0.9$.
\end{example} 

For further reading concerning the constellation approach to probabilistic argumentation we refer the readers to \cite{Hunter:2012,Dung:2010,Li:2011,Dondio:2014a}. 
Computational results can be found in \cite{dondio2014,Fazzinga:2013,Fazzinga:2015}, and in 
\cite{Li:2013,Polberg2014} we can find approaches combining the constellation probabilities with generalizations of Dung's framework.  Finally, in \cite{Doder:2014a} we can find a characterization of one of the versions of the constellation approach in terms of 
probabilistic logic.

\section{Experiment Description}
\label{sec:experiment}

In this section we  explain how the experiment was carried out. The purpose of our study was to gather the opinions on 
dialogues concerning the topic of flu vaccines. 
To this end, we prepared two separate dialogues and asked the participants we recruited online to judge at least one of them. 
In the following sections we will show the dialogues we have used, explain the tasks given to the participants
and describe the recruitment process.  

\subsection{Dialogues \& Tasks}
\label{sec:dialogues}

The dialogues presented to the participants can be seen in Tables \ref{tabd1} and \ref{tabd2}. 
The first dialogue discusses the use of the flu vaccine by hospital staff
members, while the second concerns the safety of the children flu vaccine.
Every dialogue is split into five steps in which two fictional discussants, person 1 (\textbf{P1}) and person 2 (\textbf{P2}), take turns in presenting their opinions. A given statement, once 
uttered, is visible throughout the rest of the dialogue. 
Although the general topic is the same, i.e. concerns flu vaccinations, we can observe that 
the statements 
used in the dialogues do not overlap. The dialogues have been created based on the information found on the Centers for Disease Control and 
Prevention (CDC USA) and National Health Services (NHS UK) websites.
 
\begin{table}[!h]
\renewcommand*{\arraystretch}{1.1}
\renewcommand{\tabularxcolumn}[1]{m{#1}}
\centering
\caption{Dialogue 1 between people P1 and P2. This dialogue starts with P1 claiming that hospital staff
do not need to receive flu shots, to which P2 objects. The two counterarguments of P1 are then defeated by P2. The table presents 
at which steps a given statement was visible, who uttered it and what was its content. }
\label{tabd1}
\begin{tabularx}{\textwidth}{|P{25pt}|P{30pt}|P{40pt}|X|}
\hline
Steps & Person & Statement & Content \\
\hline
\hline
1 to 5 & \textbf{P1} & \textbf{A} & Hospital staff members do not need to receive flu shots. \\
\hline
1 to 5 & \textbf{P2} & \textbf{B} & Hospital staff members are exposed to the flu virus a lot. Therefore, it would be good for them to receive flu shots
in order to stay healthy. \\
\hline
2 to 5 & \textbf{P1} & \textbf{C} & The virus is only airborne and it is sufficient to wear a mask in order to protect yourself. Therefore, a vaccination
is not necessary. \\
\hline
3 to 5 & \textbf{P2} & \textbf{D} & The flu virus is not just airborne, it can be transmitted through touch as well. Hence, a mask is insufficient to
protect yourself against the virus. \\
\hline
4 to 5 & \textbf{P1} & \textbf{E} & The flu vaccine causes flu in order to gain immunity. Making people sick, who otherwise might have stayed
healthy, is unreasonable. \\
\hline
5 & \textbf{P2} &\textbf{F} & The flu vaccine does not cause flu. It only has some side effects, such as headaches, that can be mistaken for flu
symptoms. \\
\hline
\end{tabularx}
\end{table}

\begin{table}[!h]
\renewcommand*{\arraystretch}{1.1}
\renewcommand{\tabularxcolumn}[1]{m{#1}}
\centering
\caption{Dialogue 2 between people  P1 and P2. This dialogue starts with P1 claiming that the vaccine is not safe, to which P2 objects, 
and the discussion proceeds to revolve around 
whether it contains mercury--based compounds or not. The table presents 
at which steps a given statement was visible, who uttered it and what was its content. }
\label{tabd2}
\begin{tabularx}{\textwidth}{|P{25pt}|P{30pt}|P{40pt}|X|}
\hline
Steps & Person & Statement & Content \\
\hline
\hline
1 to 5 & \textbf{P1} & \textbf{A} & The flu vaccine is not safe to use by children.\\
\hline
1 to 5 & \textbf{P2} & \textbf{B} & The flu vaccine does not contain poisonous components and is safe to use. \\
\hline
\multirow{2}{*}{2 to 5} & \multirow{2}{*}{\textbf{P1}} & \textbf{C} & The vaccine contains some mercury compounds. \\
	&	 &  \textbf{D} & The mercury compounds are poisonous and therefore the vaccine is not safe to use. \\
\hline
\multirow{2}{*}{3 to 5} & \multirow{2}{*}{\textbf{P2}} &  \textbf{E} & The child vaccine does not contain any mercury compounds. \\
	 &	 & \textbf{F} & The virus is only accompanied by stabilizers and possibly trace amounts of antibiotics used in its production.\\
\hline
4 to 5 & \textbf{P1} &\textbf{G}&  The vaccine contains a preservative called thimerosal which is a mercury-based compound.\\
\hline
5 & \textbf{P2} & \textbf{H} & Children receive the nasal spray vaccine and thimerosal has been removed from it over 15 years ago. \\
\hline
\end{tabularx}
\end{table} 
 
During the experiment, at every stage of the dialogue the participants were presented with 
three tasks - \textbf{Agreement}, \textbf{Explanation}
and \textbf{Relation} - with an additional \textbf{Awareness} question at the end of the dialogue: 

\begin{itemize}
\item \textbf{Agreement}: the participants were asked to state how much they agree or disagree with a given statement. They were allowed
to choose one of the seven options (\textit{Strongly Agree}, \textit{Agree}, \textit{Somewhat Agree}, \textit{Neither Agree nor Disagree}, 
\textit{Somewhat Disagree}, \textit{Disagree}, \textit{Strongly Disagree}) or select the answer \textit{Don't Know}.

\item \textbf{Explanation}: the participants were  asked to explain the chosen level of agreement for every statement. 
In particular, we requested them to provide us 
with any reason they may have for disagreement that was not mentioned in the dialogue and to explain reasons for changing 
their opinions compared to the previous step in the discussion (if applicable). 

\item \textbf{Relation}: the participants were asked to state how they viewed the relation between the statements. 
For every listed pair, they could say whether one statement was  \textit{A good reason against}, \textit{A somewhat good reason against}, 
\textit{Somewhat related, but can't say how}, \textit{A somewhat good reason for}, \textit{A good reason for} the other statement 
or select the answer \textit{N/A} (i.e. that the statements were unrelated). The questions were kept with the flow of the dialogue, i.e. the 
source of a given relation is preceded by the target in the presented discussion. For example, we would ask how statement \textbf{F} is related
to \textbf{A}, but not the other way around. Thus, the temporal aspects of the dialogues are taken into account 
by the graph, similarly as is done in the analysis of online discussions \cite{BoscCV16,CabrioVillata14}.
This also reduced the number of questions we had to ask and therefore simplified the task for the participants.

\item \textbf{Awareness}: the participants were asked which of the presented statements they were familiar with prior to the experiment. 
By this we understand that, for example, someone they know may have expressed the opinion contained in a given statement, they might have read it in 
some source, or generally heard about it before. To put it simply, we wanted to know what statements the participants were aware of, independently
of whether they agreed or disagreed with them. 
\end{itemize}

\subsection{Recruitment}
\label{sec:recruit}

The recruitment of the participants was done using the Amazon Mechanical Turk (AMT for short) and the survey itself was carried out using the SurveyMonkey, which 
are both common platforms for experiments of this type. In addition to the tasks explained in the previous section, the participants were subjected 
to an additional language exercise, two attention checks and a comprehension test in order
to ensure that they had sufficient skills to complete the test and that they worked honestly. 
The language exercise was on an intermediate level in terms of difficulty. 
The attention checks presented the participants with two sentences requesting them to choose particular answers. 
They were meant to disqualify the participants who were too distracted during the experiment 
or simply resorted to random clicking 
in order to complete the survey. 
Finally, in the comprehension exercise, they were asked to select the answers that described what the statements presented in the dialogue concerned. 
Prior to the survey, the participants were warned of such tests. We also provided them with explanations and examples of the tasks 
they would be asked to complete, including what should be understood as a reason for or against or what we meant by being aware of a given statement. 
We also requested the participants not to use Google or Wikipedia in order to verify the statements in the dialogues. 
At the very end of these instructions 
the participants obtained the code used to unlock the actual questionnaire without which they could not proceed. Moreover, it was also 
necessary for them to accept the terms and conditions of our experiment, which included familiarizing themselves with these explanations. 
Hence, we took all reasonable steps in informing the participants what to expect. 

We ran the survey until we had collected 80 survey responses (40 per dialogue) in which the participants had sufficiently high scores in the language, attention and comprehension tests. 
As we have noted previously, the participants - if they wished to do so - were allowed to judge both of the dialogues. 
We have found 11 people that have completed both tests, which means we have recruited a total of $69$ unique participants.  
The total number of entries (not participants) was equal to 156, this brings us to an acceptability rate of around 51\%.  
The analysis of the demographics data on our participants can be found in the auxiliary appendix at \cite{resources}.  
This research project has been approved by the designated ethics officer in the Computer Science Department at 
University College London. The participants could withdraw from the experiment at any point in time and could refuse to provide 
any piece of information they deemed too private. 

\section{Results}
\label{sec:results}

In this section we will review the results of our study. We will first describe the argument graphs obtained from our participants as well as 
the ones we had in mind when creating the dialogues. Given these structures, we will discuss if and how the declared levels of agreement 
satisfy the epistemic postulates we have
recalled in Section \ref{sec:epistemic}. Finally, we will discuss the changes in participants' beliefs throughout the dialogues.

Before we start, we would like to introduce some necessary notions. As we have explained previously, in addition to declaring 
that one statement is a good 
reason for or against another statement, a possible answer in the relation task
was \textit{Somewhat related, but can't say how}. In order to be able to represent this dependency that a participant could not classify, 
we propose the notion of a tripolar argument graph:

\begin{definition}
The \textbf{tripolar argument graph} is a tuple $\TAF = (\args, \atts, \sups, \deps)$ where $\args$ is a set of arguments, 
$\atts \subseteq \args \times \args$ is the set 
of attacks, $\sups \subseteq \args \times \args$ is the set of supports and $\deps \subseteq \args \times \args$ is the set of dependencies. 
\end{definition}

Please note that we introduce the tripolar graph only for representation purposes and we do not define any semantics for it. 
When required, we can simply extract a Dung's or bipolar graph from it. Its primary purpose is simply to mark relations 
which, potentially due to confusion or the difficulty of the experiment, the participants had problems classifying. 
Consequently, unlike in \cite{Polberg2016,CeruttiPRST16,CMDKLSM07}, it does not necessarily mean that the relation is really neither 
an attack nor a support, or that the relation might not exist. 

In the following sections the ability to compare two graphs will be important for us. To this end, we introduce various types of subgraphs. 
To put it simply, if all arguments and links contained
in one framework are present in another framework, we say that the first framework is a subgraph of another. If the links are perceived in the same 
manner (e.g. if it is an attack in one graph, it is also attack in the other), we are dealing with a correct subgraph, which is the most commonly considered type. We can then relax this notion in 
a number of ways. For example, if we allow links such as attacking or supporting to become classified as dependent, we can use the notion 
of a confusion subgraph. In a dual fashion, by allowing the dependent links to be classified as attacking or supporting, we create a precision subgraph. 
Finally, if we only concern ourselves with verifying whether the arguments connected in one framework are connected in another, independently 
of what the nature of that connection might be, we can use the notion of a lenient subgraph: 

\begin{definition}
\label{def:subgraphs}
Let $\TAF = (\args, \atts, \sups, \deps)$ and $\TAF' = (\args', \atts', \sups', \deps')$ be two tripolar argument graphs. We say that:
\begin{itemize}
\item $\TAF'$ is a \textbf{correct subgraph} of $\TAF$ iff $\args' \subseteq \args$, $\atts' \subseteq \atts$, $\sups' \subseteq \sups$ 
and $\deps' \subseteq \deps$. 

\item $\TAF'$ is a \textbf{confusion subgraph} of $\TAF$ iff $\args' \subseteq \args$, $\atts' \subseteq \atts \cup \deps$ and 
$\sups' \subseteq \sups \cup \deps$ and $\deps' \subseteq \deps$. 

\item $\TAF'$ is a \textbf{precision subgraph} of $\TAF$ iff $\args' \subseteq \args$, $\atts' \subseteq \atts$ and 
$\sups' \subseteq \sups$ and $\deps' \subseteq \deps \cup \atts \cup \sups$. 

\item $\TAF'$ is a \textbf{lenient subgraph} of $\TAF$ iff $\args' \subseteq \args$ and 
$(\atts' \cup \sups' \cup \deps') \subseteq (\atts \cup \sups \cup \deps)$. 
\end{itemize}
We say that a tripolar framework $\TAF$ is \textbf{clarified} if $\deps = \emptyset$. 
\end{definition}

Given the fact that the new types of subgraphs were introduced as a way to relax the classical correct subgraph notion, 
there certain relations between them. We can observe that every correct subgraph is also a confusion and a precision subgraph, 
and every confusion subgraph or precision subgraph is a lenient subgraph. However, not every confusion subgraph 
meets precision requirements and not every precision subgraph meets confusion requirements.  

\begin{example}
Let us consider the graphs 
$\TAF_1 = (\{A,B,C,D\}, \{(A, B)$, $(B, C)\}$, $\{(A, C)\}$, $\emptyset)$, 
$\TAF_2 = (\{A,B,C,D\}$, $\{(A, C)\}$, $\{(A, B)$, $(B, C)\}$, $\emptyset)$, 
$\TAF_3 = (\{A,B,C,D\}$, $\{(A, B)$, $(B, C)\}$, $\{(A, C)$, $(A,D)\}$, $\emptyset)$ and
$\TAF_4 = (\{A,B,C,D\}$, $\{(A, B)\}$, $\{(A, C)$, $(A,D)\}$, $\{(B, C)\})$. 
We can observe that any framework is its own subgraph of any type. 
In addition to that, $\TAF_1$ is a correct subgraph of $\TAF_3$. It is a confusion subgraph of both $\TAF_3$ and $\TAF_4$, a precision 
subgraph of $\TAF_3$  and a lenient subgraph of all of the listed frameworks. 
On the other hand, $\TAF_2$ is not a correct, confusion or precision subgraph of any other framework apart from itself. It is however a lenient subgraph 
of all of the listed structures. 
$\TAF_3$ is not a correct subgraph of any other framework apart from itself and is a confusion and lenient subgraph of $\TAF_4$. 
$\TAF_4$ is only its own correct and confusion subgraph, and a precision and lenient subgraph of $\TAF_3$ and $\TAF_4$. 
\end{example}

In addition to being able to represent the answers provided by the participants, we would also like to be able to measure the disagreement between 
the various frameworks we obtain at a given stage of the dialogue. For this purpose, we create a notion distance between 
frameworks 
conceptually similar to the ones from \cite{CMDKLSM07,BoothCPR12,CLS11}.
The distance is measured in terms of links by which two given frameworks
differ. In particular, if the given arguments are seen as unrelated in one structure, but as attacking/supporting or dependent in the other, we set the difference
to $2$ and $1$ respectively. 
If a link has opposing polarities (i.e. attacking vs supporting) in the frameworks, the distance is set again to $2$. If it is attacking 
or supporting in one and dependent in another, the distance is set to $1$. Finally, if the nature of a given relation is seen in the same way, then the 
difference is naturally $0$. 
The distance between the frameworks is then simply a sum of all such differences: 

\begin{definition}
\label{def:dist}
Let $\TAF = (\args, \atts, \sups, \deps)$ and $\TAF' = (\args, \atts', \sups', \deps')$ be two tripolar frameworks defined over the same set 
of arguments s.t. $\atts \cap \sups \cap \deps = \emptyset$ and $\atts' \cap \sups' \cap \deps' =\emptyset$.
Let $Rel^{\TAF} = \atts \cup \sups \cup \deps$ and $Rel^{\TAF'} = \atts' \cup \sups' \cup \deps'$ be the sets of all relations in 
both structures. 
The difference between $\TAF$ and $\TAF'$ on edge $\alpha$ is defined in the following manner:
\begin{itemize}

\item If $ \alpha \notin Rel^{\TAF} \cap Rel^{\TAF'}$:  
\begin{equation*}
   \fdiff(\TAF, \TAF', \alpha) = \begin{cases}
					  0		  & $ if $\alpha \notin Rel^{\TAF} \cup Rel^{\TAF'} \\
             				  1               & $ if $\alpha \in \deps \cup \deps' \\
             				  2               & $ otherwise $ 
         			  	 \end{cases}
\end{equation*}  
\item If $ \alpha \in Rel^{\TAF} \cap Rel^{\TAF'}$: \\
\begin{equation*}
   \fdiff(\TAF, \TAF', \alpha) = \begin{cases} 
               				  1               &  $ if $\alpha \in \deps \land \alpha \notin \deps' \\
					  1               &  $ if $\alpha \notin \deps \land \alpha \in \deps' \\
					  2               & $ if $\alpha \in \sups \land \alpha \in \atts' \\
					  2               & $ if $\alpha \in \atts \land \alpha \in \sups' \\
					  0 		   & $otherwise$ 
         			  	 \end{cases}
\end{equation*}  
\end{itemize}
The distance between $\TAF$ and $\TAF'$ is then defined as: 
\begin{equation*}
d(\TAF, \TAF') = \sum_{\alpha \in Rel^{\TAF} \cup Rel^{\TAF'}} \fdiff(\TAF, \TAF', \alpha) 
\end{equation*}  
\end{definition}

\begin{figure}[!ht]
\centering
\begin{tikzpicture}
[-,>=stealth,shorten >=1pt,auto,node distance=2cm,
  thick,main node/.style={fill=none,draw=none,minimum size = 0.5cm,font=\small\bfseries},every edge/.append style={font=\small}]

\node[main node] (D) at (1,1) {$\deps$};
\node[main node] (A) at (-0.5,0) {$\atts$}; 
\node[main node] (S) at (2.5,0) {$\sups$};
\node[main node] (N) at (1,2.5) {$N/A$};

 \path
	(S) edge node  {2} (A) 
	(D) edge node  {1} (A) 
	(S) edge node  {1} (D)
	(D) edge node  {1} (N)
	(A) edge node  {2} (N)
	(N) edge node  {2} (S);
\end{tikzpicture}
\label{fig:distances}
        \caption{Difference measure between attacking, supporting, dependent and nonexistent relations}
\end{figure}
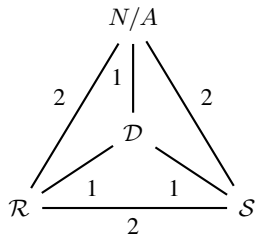

The introduced notion of a distance between the frameworks meets the requirements of a metric, i.e. the following holds:
\begin{restatable}{lemma}{dmetric}
Let $\TAF_1 =  (\args, \atts_1, \sups_1, \deps_1)$, $\TAF_2=(\args, \atts_2, \sups_2, \deps_2)$ and $\TAF_3=(\args, \atts_3, \sups_3, \deps_3)$ 
be tripolar frameworks defined over the same sets of arguments and s.t. for $i \in \{1,2,3\}$, $\atts_i \cap \sups_i \cap \deps_i = \emptyset$. 
The following holds:
\begin{enumerate}
\item $d(\TAF_1, \TAF_2) \geq 0$.
\item $d(\TAF_1, \TAF_2) = 0 \iff \TAF_1 = \TAF_2$\footnote{In this context we say that $\TAF_1 = \TAF_2$ iff $\atts_1 = \atts_2$, $\sups_1 = \sups_2$ 
and $\deps_1 = \deps_2$.}.
\item $d(\TAF_1, \TAF_2) = d(\TAF_2, \TAF_1)$.
\item $d(\TAF_1, \TAF_3) \leq d(\TAF_1, \TAF_2) + d(\TAF_2, \TAF_3)$.
\end{enumerate}
\end{restatable}
 
Finally, in order to be able to analyze  which postulates may or may not be satisfied by the participants, 
we need to map the declared levels of agreement to actual numerical values. By this we understand that if an argument 
$X$ is assigned an answer such as \textit{Strongly Disagree} by a participant, 
then there is a numerical value $y$ associated with \textit{Strongly Disagree} s.t. for the purpose of this analysis we can 
say that $\prob(X) = y$. From such assignments, a full probability distribution over the powerset of arguments can be 
reproduced \cite{HunterThimm17}. 
However, given the way the probabilities are used by the epistemic postulates that we have recalled, 
knowing that $\prob(X) = y$ is sufficient for our purposes. 
We propose the following assignments:  
\begin{itemize}
\label{distributions}
\item \textit{Strongly Agree} -- $6/6$.
\item \textit{Agree} -- $5/6$.
\item \textit{Somewhat Agree} -- $4/6$.
\item \textit{Neither Agree nor Disagree} -- $3/6$.
\item \textit{Somewhat Disagree} -- $2/6$.
\item \textit{Disagree} -- $1/6$.
\item \textit{Strongly Disagree} -- $0/6$. 
\end{itemize}

Additionally, the \textit{Don't know} answer is available in the questionnaire. 
Unless stated otherwise, for calculation purposes we assign it the same value as \textit{Neither Agree nor Disagree} (i.e. $3/6$). 
It is a possible way of shifting from a four--valued setting (in, out, undecided and unknown) to a three--valued one (in, out, undecided), where inability 
to make a decision is treated similarly as indecisiveness. 

Finally, we create the graphs based on the answers in relation task in the following manner. The \textit{Reason for} answer we will mark as support,
 \textit{Reason against} as attack, and the \textit{Somewhat related, but can't say how}
as dependent. The declared strength of the attack or support relation will only become relevant once we analyze its connection to the agreement level
assigned to its source in Section \ref{sec:relcorel}. 

\subsection{Argument Graphs}
\label{sec:graphanalysis}

In this section we will analyze the argument graphs we have received from the participants based on their responses to the relation tasks. 
In what follows we will distinguish the following types of graphs: 
\begin{itemize}
\item the \textbf{intended graph} is meant to depict the minimal set of relations we considered reasonable for a given set of arguments.
\item the \textbf{augmented graph} is obtained by adding to the intended graph the indirect relations from the prudent/careful or bipolar argumentation 
approaches (Definitions \ref{def:prudent} and \ref{def:indiratts}).

\item the \textbf{participant--sourced graphs}: 
\begin{itemize}
\item the \textbf{declared graph} is constructed from the answers given to us by the participants in the Relation tasks, 
where the \textit{Reason for} answer we will mark as support, \textit{Reason against} as attack, and the 
\textit{Somewhat related, but can't say how} as dependent. The arguments correspond to the statements appearing at a given step of the dialogue.

\item the \textbf{expanded graph} is constructed from the declared graph, extended with the statements extracted from the answers that the 
participants have provided in the Explanation task.

\item the \textbf{common graph} is the declared graph created by the highest number of participants at a given step in the dialogue. 
\end{itemize}
\end{itemize}

The intended graph simply depicts certain minimal constraints we had in mind for this experiment. For example, we had intended  
the statement \enquote{\textit{Hospital staff members are exposed to the flu virus a lot. Therefore, it would be good for them to receive flu shots
in order to stay healthy.}} to attack \enquote{\textit{Hospital staff members do not need to receive flu shots.}}. Please note that this graph 
does not necessarily contain all of the possible interactions between the statements. For example, we decided against putting any supporting links 
in. The purpose of this graph was merely to encompass the minimal constraints we had intended 
the participants to recognize. Consequently, this does not necessarily mean that it contains all the possible 
relations that could be extracted from the dialogue.

The augmented graph is built from the intended graph through the addition of additional relations, similarly as indirect edges 
can be added to bipolar graphs for evaluation (see Section \ref{sec:bipolar} and \cite{report:trans} for further details). 
The purpose of this graph is to check whether the relations declared by the participants in addition to the ones stated in 
the intended graph can in fact be reproduced through the use of notions from Definitions \ref{def:prudent} and \ref{def:indiratts}. 
In particular, we will augment the intended graph with attacks using the definitions of indirect conflicts 
from the aforementioned definitions and with supports using the definition of indirect defense. We will then observe that 
these notions do not necessarily give us all the possible relations defined by the participants and analyze how the ones 
that are not accounted for could be explained.

At this point we would like to stress that in this report, we will follow the abstract interpretation of supports, i.e. one in which 
it is understood simply as a positive relation between arguments without any further requirements \cite{incoll:bipolar}. As stated in 
Section \ref{sec:bipolar}, we will not impose any restrictions as to which indirect conflicts should be used. The reason 
for this is two--fold. First of all, there is no consensus as to which indirect conflicts should be associated with a given specialized form 
of support, such as deductive or necessary \cite{report:trans,axiomatic,Prakken14}. 
Second of all, we are not aware 
of any empirical study verifying that people indeed associate with a given type of support the indirect attacks ascribed to it in theory. 
There also appears to be no study testing whether a given supporting edge is assigned the same interpretation both by theory 
and by actual people. All of these are extremely interesting questions, however, they are beyond the scope of this particular study.

We would also like to explain our approach to reproducing supports obtained from the participants through defense 
in the augmented graph. First of all, both defenses and supports are forms of positive relations between arguments. Hence, interactions 
between them are unavoidable, particularly given the results in \cite{report:trans,axiomatic}. Taking into account that our participants are not argumentation experts and their task was to mark positive relations in the dialogue, what they have identified could fall into either support or defense category. 
This will always be the case in any experiment that takes into account people from all walks of life, not just specialists in our area, 
and we cannot discard parts of their answers and extract as supports only those relations that in our opinion are \enquote{real}. 
What we therefore show in Sections \ref{sec:dial1daianalysis} and \ref{sec:dial2usergraph} 
is that even though many of the positive links identified by the participants can be reproduced through 
defense, there are support edges that cannot be explained in this way. 
 

The expanded graph contains an auxiliary statement $P$ that embodies the additional reasons for and against the statements in the dialogues 
that the participants
may have given in their explanations and that have not appeared in the dialogue. For example, $P$ could contain a statement such as 
\enquote{\textit{Masks are insufficient to prevent the flu, gloves are also needed.}} or \enquote{\textit{The vaccine does cause flu, everyone I know 
gets very sick after the shot.}}.    

In this section we will perform two types of analyses. We first consider the intended graphs created for both of the dialogues (Sections 
\ref{sec:intended1} and \ref{sec:intended2}) and see how they are reflected by the graphs obtained from the participants. Based on this information 
we create the notions of the total and core samples, used throughout the rest of this paper. 
We then focus on looking at the declared and common graphs in Sections \ref{sec:dial1daianalysis} and \ref{sec:dial2usergraph}. In particular, we explain  
up to what extent the common graphs can be reproduced from the intended graphs by the use of indirect defenses and attacks from the 
prudent/careful and bipolar setting (see Definitions \ref{def:prudent} and \ref{def:indiratts}). 

\subsubsection{Dialogue 1}

\paragraph{Intended Graph Analysis}\hfill
\label{sec:intended1}

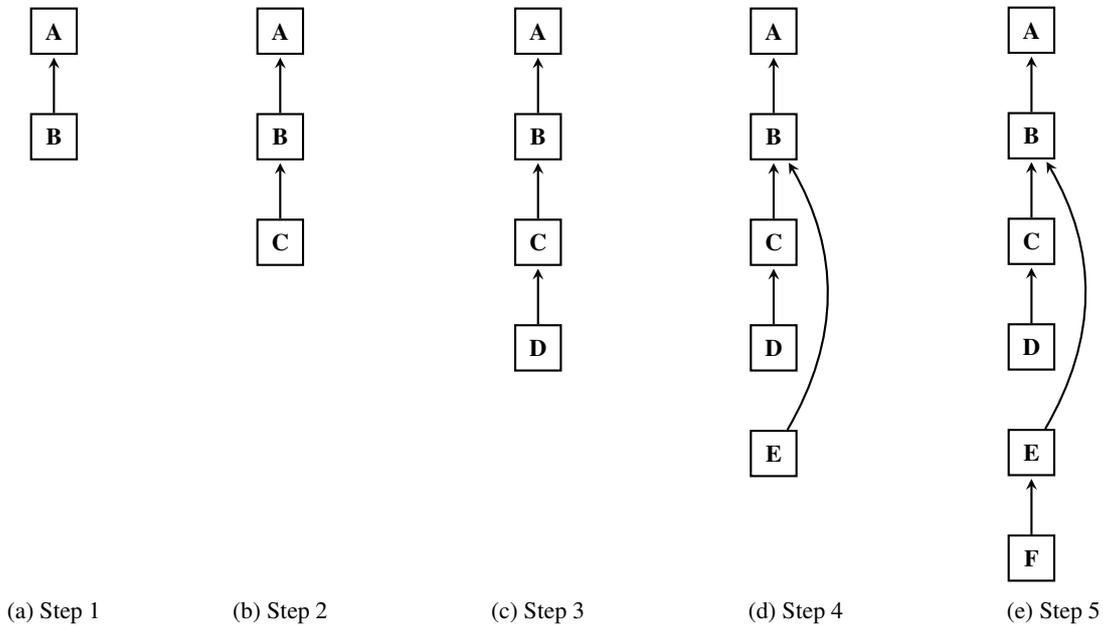
\begin{figure}[!ht] 
 \centering
\begin{subfigure}[b]{0.15\textwidth}
\centering
\begin{tikzpicture}[->,>=stealth,shorten >=1pt,auto,node distance=1.4cm,
thick,main node/.style={rectangle,fill=none,draw,minimum size = 0.6cm,font=\small\bfseries}]
\node[main node] (a) {A};
\node[main node] (b) [below of=a] {B};
\node[main node] (c) [draw=none,below of=b] {};
\node[main node] (d) [draw=none,below of=c] {};
\node[main node] (e) [draw=none,below of=d] {};
\node[main node] (f) [draw=none, below of=e] {};
\path 
(b) edge node {} (a)
;
\end{tikzpicture}
\caption{Step 1}
\end{subfigure}
\begin{subfigure}[b]{0.2\textwidth}
\centering
\begin{tikzpicture}[->,>=stealth,shorten >=1pt,auto,node distance=1.4cm,
thick,main node/.style={rectangle,fill=none,draw,minimum size = 0.6cm,font=\small\bfseries}]
\node[main node] (a) {A};
\node[main node] (b) [below of=a] {B};
\node[main node] (c) [below of=b] {C};
\node[main node] (d) [draw=none,below of=c] {};
\node[main node] (e) [draw=none,below of=d] {};
\node[main node] (f) [draw=none, below of=e] {};
\path 
(b) edge node {} (a)
(c) edge  node {}  (b) 
;
\end{tikzpicture}
\caption{Step 2}
\end{subfigure}
\begin{subfigure}[b]{0.2\textwidth}
\centering
\begin{tikzpicture}[->,>=stealth,shorten >=1pt,auto,node distance=1.4cm,
thick,main node/.style={rectangle,fill=none,draw,minimum size = 0.6cm,font=\small\bfseries}]
\node[main node] (a) {A};
\node[main node] (b) [below of=a] {B};
\node[main node] (c) [below of=b] {C};
\node[main node] (d) [below of=c] {D}; 
\node[main node] (e) [draw=none,below of=d] {};
\node[main node] (f) [draw=none, below of=e] {};
\path 
(b) edge node {} (a)
(c) edge  node {}  (b)
(d) edge node {}  (c) 
;
\end{tikzpicture}
\caption{Step 3}
\end{subfigure}
\begin{subfigure}[b]{0.2\textwidth}
\centering
\begin{tikzpicture}[->,>=stealth,shorten >=1pt,auto,node distance=1.4cm,
thick,main node/.style={rectangle,fill=none,draw,minimum size = 0.6cm,font=\small\bfseries}]
\node[main node] (a) {A};
\node[main node] (b) [below of=a] {B};
\node[main node] (c) [below of=b] {C};
\node[main node] (d) [below of=c] {D}; 
\node[main node] (e) [below of=d] {E};
\node[main node] (f) [draw=none, below of=e] {};
\path 
(b) edge node {} (a)
(c) edge  node {}  (b)
(d) edge node {}  (c)
(e) edge [bend right] node  {}  (b) 
;
\end{tikzpicture}
\caption{Step 4}
\end{subfigure}
\begin{subfigure}[b]{0.2\textwidth}
\centering
\begin{tikzpicture}[->,>=stealth,shorten >=1pt,auto,node distance=1.4cm,
thick,main node/.style={rectangle,fill=none,draw,minimum size = 0.6cm,font=\small\bfseries}]
\node[main node] (a) {A};
\node[main node] (b) [below of=a] {B};
\node[main node] (c) [below of=b] {C};
\node[main node] (d) [below of=c] {D}; 
\node[main node] (e) [below of=d] {E};
\node[main node] (f) [below of=e] {F};
\path 
(b) edge node {}  (a)
(c) edge  node  {} (b)
(d) edge node {}  (c)
(e) edge [bend right] node {}  (b)
(f) edge node {}  (e) 
;
\end{tikzpicture}
\caption{Step 5}
\end{subfigure}
\caption{The intended argument graphs for Dialogue 1. Solid edges represent the attack relation. } 
 \label{fig:intended}	
\end{figure}

The minimal graphs we had wanted the participants to acknowledge at every stage of Dialogue 1 are presented in Figure \ref{fig:intended}.  
The first thing we would like to analyze is how the intended graphs are related to the ones 
declared by the participants. 

\begin{figure}[!ht]
\centering 
\pgfplotsset{compat=1.9}
\begin{tikzpicture} 
\begin{axis}[name=ccount,height=0.3\textwidth, width = 0.44\textwidth, ybar stacked, bar width = 10pt, point meta = explicit, 
nodes near coords, every node near coord/.append style={font=\footnotesize,anchor=east,xshift=-4pt},ytick ={0,10,20,30,40},
xlabel={Step Number},ylabel={Number of Participants},xtick ={1,2,3,4,5},ymin=0, ymax=45, ymajorgrids, yminorgrids,  
enlargelimits=true, 
enlarge x limits = 0.2, enlarge y limits = 0.04, 
legend style={at={(1.1,0.5)},anchor=west, font=\scriptsize,align=center}, legend columns=1, 
legend image code/.code={%
      \draw[#1] (0cm,-0.1cm) rectangle (0.3cm,0.3cm);
    },title style={align=center}, 
title = {Number of participants whose declared \\ graphs contain the intended graphs\\ at a given step}
]

\addplot[pattern = north east lines] coordinates  
{(1,18) [18]
(2,20) [20]
(3,20) [20]
(4,14) [14]
(5,12) [12] }; 
\addplot[fill = black]  coordinates  
{(1,4) [22]
(2,5) [25]
(3,3) [23]
(4,5) [19]
 (5,7) [19]
};
\addplot[pattern = north west lines]  coordinates 
{(1,18) [40]
(2,14) [39]
(3,16) [39]
(4,17) [36]
(5,17) [36]
}; 

\end{axis}  
\begin{axis}[at=(ccount.right of south east), title style={align=center},
title = {Number of participants whose declared \\ graphs contain the intended graphs\\ in a given number of steps},
width=0.6\textwidth, height = 0.3\textwidth,ybar, xtick align=inside,
nodes near coords, every node near coord/.append style={font=\footnotesize},   
xlabel={Number of Steps},
yticklabels={,,,,},
xtick ={0,1,2,3,4,5},ymax=45,
enlarge x limits = 0.1, enlarge y limits = 0.04, ytick ={0,10,20,30,40},
bar width=8pt, ymajorgrids, yminorgrids,  
legend style={at={(-0.5,1.8)},anchor=west, font=\scriptsize,align=center}, legend columns=3, 
legend image code/.code={%
      \draw[#1] (0cm,-0.1cm) rectangle (0.3cm,0.3cm);
    }
]

\addplot[pattern = north east lines] coordinates
{(0,15)  (1,5) (2,0) (3,8) (4,5) (5,7)}; 
\addplot[fill = black] coordinates
{(0,9)  (1,5) (2,3) (3,7) (4,4) (5,12)};
\addplot[pattern = north west lines] coordinates
{(0,0)  (1,1) (2,0) (3,3) (4,0) (5,36)}; 
 
 \legend{Correct subgraph, Confusion  subgraph, Lenient subgraph}
 
\end{axis}
\end{tikzpicture}
\caption{Analysis of intended graph containment in Dialogue 1} 
\label{fig:contained}
\end{figure}
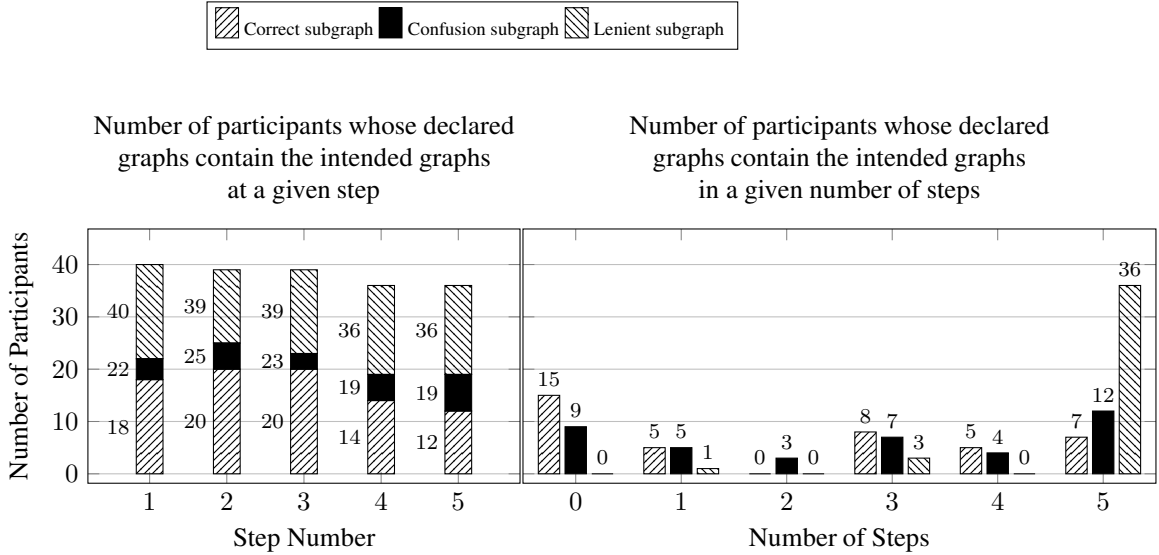

In the first chart in Figure \ref{fig:contained} we can see for how many participants the intended graph 
was a subgraph of a given type of their declared graph. The portion of the participants who have correctly recognized the intended relations at a given 
step of the dialogue does not exceed 50\%, with the worst performance of 30\% at the final step of the dialogue. The results look 
somewhat more interesting once we look at the confusion subgraph, which - as the name suggests - assumes the participants may become confused 
and specify the link as simply dependent rather than attacking or supporting. The biggest difference in the number of participants between
the correct and confusion subgraphs can be seen in the last step of the dialogue, which might point to participants' exhaustion or to the complexity 
of the pieces of information presented in the dialogue.  
Finally, we can consider looking at the lenient subgraphs, in which it only matters that the statements we have seen as related are also 
seen as such by the participants, independently of the nature of this relation. The high number of participants whose declared graphs contained
the intended graph in this lenient approach tells us 
that the participants had no major problems in recognizing the related statements, however, marking the type of this connection has caused some issues. 
All of the declared graphs that have not met the leniency 
requirements missed at least one of the $(C, B)$, $(E,B)$ or $(F, E)$ attacks. Nevertheless, in all 
of the approaches we can see that the lowest number of participants satisfying a given subgraph relation can be seen in the last
two steps of the dialogue in which statements $E$ and $F$ were presented. 
 
Although the first chart in Figure \ref{fig:contained} presents how many participants recognized the intended graphs at a given step 
in the dialogue, it does not show how many participants recognized the intended graphs at a given number of steps. 
For example, even though 18 (resp. 20) participants recognized 
the intended graph correctly at the first step (resp. second step) of the dialogue, 
it does not necessarily mean that 18 participants recognized the intended graph correctly at both of these steps. 
Consequently, in the second chart in Figure \ref{fig:contained} 
we present at how many steps the participants have recognized the intended graph in a particular way (correct, confusion, lenient). 
The lenient requirements are quite easily satisfied, just like in the previous case. 
However, we can observe that a relatively small number of participants (7 w.r.t. the correct and 12 w.r.t. the confusion subgraph)
recognized the intended graph in all dialogue stages. We believe 
this indicates that the relation task might have been more confusing than we had anticipated. 
Consequently, 
we have decided to report the results from all of the participants -- we will refer to them as the \textbf{total sample} 
from now on -- as well as from those that have 
recognized the intended framework in at least four out of five dialogue stages w.r.t. the confusion subgraph approach -- we will refer to 
them as the \textbf{core sample}. The samples contain respectively 40 and 16 participants. 
  
\paragraph{Participant--Sourced Graphs Analysis}\hfill
\label{sec:dial1daianalysis}  

In this section we would like to take a closer look at the graphs declared by the participants throughout the first dialogue. 
Although we have already analyzed whether 
the participants have recognized the intended graph, we have not yet seen what relations - in addition to the 
ones in Figure \ref{fig:intended} -- 
they have also observed. 
We can observe that at every step of the dialogue, the number of unique declared graphs we have 
obtained is smaller than the number of participants (auxiliary data appendix is available at \cite{resources}). 
This is a natural effect of the fact that different participants could 
answer the relation tasks in the same manner at some of the dialogue stages. Nevertheless, the further we are in the dialogue, 
the more common the graphs that have only been declared by one participant become. 
As seen in Figure \ref{fig:clarified1}, also the bigger the chances that the framework declared by a participant 
will not be clarified (i.e. it will contain dependent links, see Definition \ref{def:subgraphs}), 
which was to be expected given the increasing 
number of edges at every stage. 
Although at every step we can find a graph that has been declared by the largest number of participants, the obtained 
structure rarely accounts for more than 50\% of the (total) sample. Thus, not every participant perceives the dialogue 
in the same manner, in part because they do not appear to be always aware of the same facts and arguments (we will discuss 
this issue further in  Section \ref{sec:change}). This disparity in knowledge and perception 
lends support to the use of the constellation approach in opponent modelling. In particular, if we treat the portion of participants 
declaring a given graph as a probability, 
what we obtain at every stage of the dialogue is actually a subgraph distribution.  
  
\begin{figure}[!ht]
\centering
\pgfplotsset{compat=1.9,every axis/.style={width=0.8\textwidth, height = 0.25\textheight, 
grid=both, 
xlabel = {Step Number}, ylabel = {\% of Declared Graphs},
yticklabel style={
        /pgf/number format/fixed,
        /pgf/number format/precision=4
}, bar width = 19pt,
scaled y ticks=false,
ybar=4pt, ymin=0, ymax=80, nodes near coords = {\pgfmathprintnumber\pgfplotspointmeta\%}, 
every node near coord/.append style={font=\scriptsize},
xtick = {1,2,3,4,5}, 
legend style={at={(1.05,0.5)},anchor=west, font=\scriptsize,align=center}, legend columns=1, 
legend image code/.code={%
      \draw[#1] (0cm,-0.1cm) rectangle (0.3cm,0.3cm);}
}}
\begin{tikzpicture} 
  
\begin{axis}[name = clartot, xticklabels = {1,2,3,4,5}]   
\addplot[fill=black] coordinates{(1,10) (2,20) (3,25) (4,42.5) (5,67.5)}; 

\addplot[fill=white] coordinates{(1,18.75) (2,18.75) (3,31.25) (4,31.25) (5,62.5)}; 
\legend{Total \\ sample, Core \\ sample}
\end{axis}
\end{tikzpicture}
\caption{Portion of unclarified graphs (i.e. with nonempty set of dependencies) obtained in Dialogue 1} 
\label{fig:clarified1}
\end{figure}
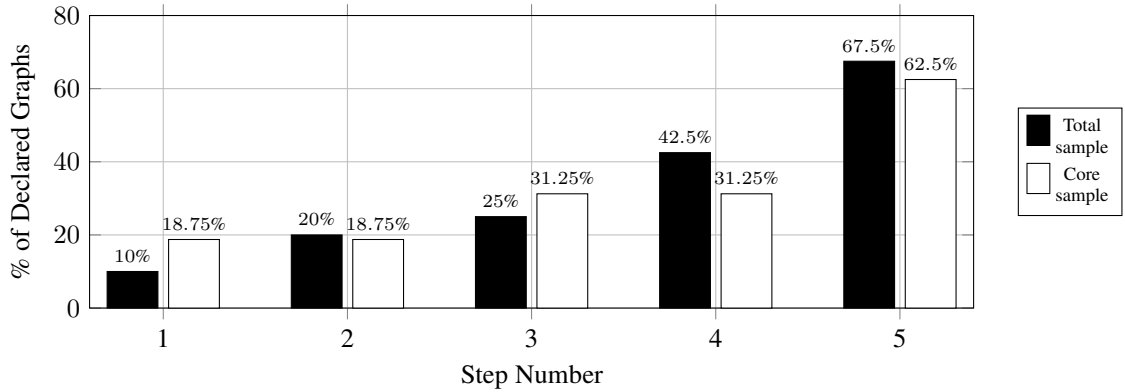 
 
\begin{figure}[!ht] 
 \centering
\begin{subfigure}[b]{0.12\textwidth}
\centering
\begin{tikzpicture}
[->,>=stealth,shorten >=1pt,auto,node distance=1.4cm,
thick,main node/.style={rectangle,fill=none,draw,minimum size = 0.6cm,font=\small\bfseries}]
\node[main node] (a) {A};
\node[main node] (b) [below of=a] {B};
\node[main node] (c) [draw=none,below of=b] {};
\node[main node] (d) [draw=none,below of=c] {};
\node[main node] (e) [below of=d] {A};
\node[main node] (f) [below of=e] {B};
\path 
(b) edge [mega thick] node {} (a)
(f) edge [color=black, dashed] node {} (e)
;
\end{tikzpicture}
\caption{Step 1}
\end{subfigure}
\begin{subfigure}[b]{0.14\textwidth}
\centering
\begin{tikzpicture}[->,>=stealth,shorten >=1pt,auto,node distance=1.4cm,
thick,main node/.style={rectangle,fill=none,draw,minimum size = 0.6cm,font=\small\bfseries}]
\node[main node] (a) {A};
\node[main node] (b) [below of=a] {B};
\node[main node] (c) [below of=b] {C};
\node[main node] (d) [draw=none,below of=c] {};
\node[main node] (e) [draw=none,below of=d] {};
\node[main node] (f) [draw=none, below of=e] {};
\path 
(b) edge [mega thick] node {} (a)
(c) edge  [mega thick] node {}  (b) 
(c) edge [color=black, bend right, dashed] (a)
;
\end{tikzpicture}
\caption{Step 2}
\end{subfigure}
\begin{subfigure}[b]{0.15\textwidth}
\centering
\begin{tikzpicture}[->,>=stealth,shorten >=1pt,auto,node distance=1.4cm,
thick,main node/.style={rectangle,fill=none,draw,minimum size = 0.6cm,font=\small\bfseries}]
\node[main node] (a) {A};
\node[main node] (b) [below of=a] {B};
\node[main node] (c) [below of=b] {C};
\node[main node] (d) [below of=c] {D}; 
\node[main node] (e) [draw=none,below of=d] {};
\node[main node] (f) [draw=none, below of=e] {};
\path 
(b) edge  [mega thick]node {} (a)
(c) edge  [mega thick] node {}  (b)
(d) edge  [mega thick] node {}  (c) 
(d) edge [color=black, bend left=40] (a)
(c) edge [color=black, bend right, dashed] (a)
(d) edge [color=black, bend left, dashed] (b)
;
\end{tikzpicture}
\caption{Step 3}
\end{subfigure}
\begin{subfigure}[b]{0.22\textwidth}
\centering
\begin{tikzpicture}[->,>=stealth,shorten >=1pt,auto,node distance=1.4cm,
thick,main node/.style={rectangle,fill=none,draw,minimum size = 0.6cm,font=\small\bfseries}]
\node[main node] (a) {A};
\node[main node] (b) [below of=a] {B};
\node[main node] (c) [below of=b] {C};
\node[main node] (d) [below of=c] {D}; 
\node[main node] (e) [below of=d] {E};
\node[main node] (f) [draw=none, below of=e] {};
\path 
(b) edge  [mega thick] node {} (a)
(c) edge   [mega thick] node {}  (b)
(d) edge  [mega thick] node {}  (c)
(e) edge [bend right=40, mega thick] node  {}  (b) 

(e) edge [color=black] node  {}  (d) 
(d) edge [color=black, bend left=40] (a)

(c) edge [color=black, bend right, dashed] (a)
(d) edge [color=black, bend left, dashed] (b)
(e) edge [color=black, bend left=45, dashed] (a)
(e) edge [color=black, bend left, dashed] (c)
;
\end{tikzpicture}
\caption{Step 4}
\end{subfigure}
\begin{subfigure}[b]{0.30\textwidth}
\centering
\begin{tikzpicture}[->,>=stealth,shorten >=1pt,auto,node distance=1.4cm,
thick,main node/.style={rectangle,fill=none,draw,minimum size = 0.6cm,font=\small\bfseries}]
\node[main node] (a) {A};
\node[main node] (b) [below of=a] {B};
\node[main node] (c) [below of=b] {C};
\node[main node] (d) [below of=c] {D}; 
\node[main node] (e) [below of=d] {E};
\node[main node] (f) [below of=e] {F};
\path 
(b) edge  [mega thick]  node {}  (a)
(c) edge   [mega thick]  node  {} (b)
(d) edge  [mega thick]  node {}  (c)
(e) edge [bend right=40, mega thick] node {}  (b)
(f) edge   [mega thick] node {}  (e) 

(e) edge [color=black] node  {}  (d) 
(d) edge [color=black, bend left=40] (a)
(f) edge [color=black, bend left=55] (a)
(f) edge [color=black, bend right=40] (c) 

(c) edge [color=black, bend right, dashed] (a)
(d) edge [color=black, bend left, dashed] (b)
(e) edge [color=black, bend left=45, dashed] (a)
(e) edge [color=black, bend right, dashed] (c)
(f) edge [color=black, bend right=50, dashed] (b)
(f) edge [color=black, bend left, dashed] (d)

;
\end{tikzpicture}
\caption{Step 5}
\end{subfigure}
\caption{The common argument graphs for Dialogue 1 based on the total sample. With the exception of the bottom framework 
in Step 1, they are also common in the core sample.  The thicker edges represent the relations appearing in the intended graph. Solid edges
stand for attack and dashed for support links.} 
 \label{fig:declared}	
\end{figure}
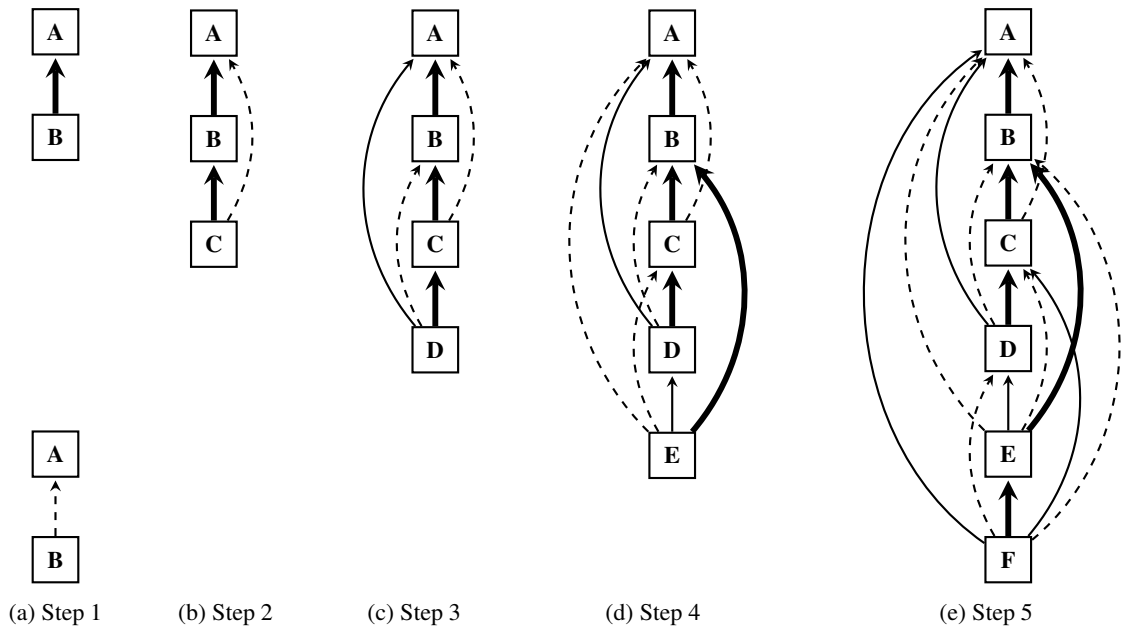

As we have stated before, at every stage of the dialogue we can find a framework that has been declared by the 
largest number of participants. We present these structures in Figure \ref{fig:declared}. 
Only in the first step of the dialogue we obtain 2 equally common graphs, one in which $(B,A)$ edge 
is seen as attacking and one in which it is supporting (both are declared by 18 participants). These proportions change notably once
we look at the core sample, where the $(B,A)$ link is primarily attacking. We believe this behaviour can be put down to the initial confusion 
caused by the experiment, particularly that afterwards the edge is seen mostly as attacking by the participants from both samples. 

We can observe that w.r.t. the intended graphs, the participants have declared significantly more relations in the common graphs. 
However, as we will see, many of these edges can in fact be explained by the existing notions of indirect relations. 
We will first look at the common graphs from the perspective of the prudent/careful semantics (Definition \ref{def:prudent}). 
In other words, we perform a conflict--centered analysis, where the positive edges between the statements are seen 
only as a result of the interplay between the attacks. 
By extending the intended graphs with additional conflicts stemming from the indirect attacks and additional supports 
corresponding to the indirect defenses, we obtain the augmented graphs visible in Figure \ref{fig:intendedcomp}. 
The augmented graphs obtained in the first three steps are identical with the common frameworks declared by the participants. 
The differences between them start appearing once the statement $E$ 
is introduced in the fourth step of the dialogue. In particular, in the augmented framework, 
the $(E, D)$ attack and $(E, C)$ support are unaccounted for. Once $F$ is presented in the fifth step of the dialogue, 
also the $(F, D)$ support and $(F, C)$ 
attack are missing. 

This situation could be interpreted as a sign that the intended framework is too conservative and that the missing edges should have been included 
in it in the first place. However, there are also other possible interpretations that do not resort to modifying the intended graph. 
According to Definition \ref{def:prudent}, 
$E$ is controversial w.r.t. $D$ 
as it is an attacker of $B$, which is defended by $D$. 
Similarly, $F$ is controversial w.r.t. $C$, as $F$ attacks $A$ which is defended by $C$. 
 Consequently, it might be the case that the $(E, D)$ and $(F, C)$ attacks are manifestations of these issues
\footnote{Please note that controversy could 
also lead the $D$ attacking $E$ and $C$ attacking $F$, but such relations could not have been defined by the participants due to the experiment set up.}.
The missing $(E, C)$ and $(F, D)$ supports could then be explained in two ways. If we were to use the aforementioned controversy
as a basis for additional conflicts and add them to the (intended or augmented) graph, the absent supports could potentially be created by 
allowing indirect defense to also take into account 
these new conflicts in their definition. This could perhaps give rise to \enquote{controversial defense}, the existence of which we are not aware of in 
the literature. Another possible, and perhaps more natural approach, 
is to consider the statements that indirectly attack (defend) the same statement as positively related. In this 
respect, we can see $E$ and $C$ as working towards the same goal, which is defending $A$ and attacking $B$. Nevertheless, we are not aware 
of this aspect being explored in the context of attack--based graphs.  

\begin{figure}[!ht] 
 \centering
\begin{subfigure}[b]{0.12\textwidth}
\centering
\begin{tikzpicture}[->,>=stealth,shorten >=1pt,auto,node distance=1.4cm,
thick,main node/.style={rectangle,fill=none,draw,minimum size = 0.6cm,font=\small\bfseries}]
\node[main node] (a) {A};
\node[main node] (b) [below of=a] {B};
\node[main node] (c) [draw=none,below of=b] {};
\node[main node] (d) [draw=none,below of=c] {};
\node[main node] (e) [draw=none,below of=d] {};
\node[main node] (f) [draw=none, below of=e] {};
\path 
(b) edge [mega thick] node {} (a)
;
\end{tikzpicture}
\caption{Step 1}
\end{subfigure}
\begin{subfigure}[b]{0.14\textwidth}
\centering
\begin{tikzpicture}[->,>=stealth,shorten >=1pt,auto,node distance=1.4cm,
thick,main node/.style={rectangle,fill=none,draw,minimum size = 0.6cm,font=\small\bfseries}]
\node[main node] (a) {A};
\node[main node] (b) [below of=a] {B};
\node[main node] (c) [below of=b] {C};
\node[main node] (d) [draw=none,below of=c] {};
\node[main node] (e) [draw=none,below of=d] {};
\node[main node] (f) [draw=none, below of=e] {};
\path 
(b) edge [mega thick] node {} (a)
(c) edge [mega thick] node {}  (b) 
(c) edge [color=black, bend right, dashed] (a)
;
\end{tikzpicture}
\caption{Step 2}
\end{subfigure}
\begin{subfigure}[b]{0.15\textwidth}
\centering
\begin{tikzpicture}[->,>=stealth,shorten >=1pt,auto,node distance=1.4cm,
thick,main node/.style={rectangle,fill=none,draw,minimum size = 0.6cm,font=\small\bfseries}]
\node[main node] (a) {A};
\node[main node] (b) [below of=a] {B};
\node[main node] (c) [below of=b] {C};
\node[main node] (d) [below of=c] {D}; 
\node[main node] (e) [draw=none,below of=d] {};
\node[main node] (f) [draw=none, below of=e] {};
\path 
(b) edge [mega thick]node {} (a)
(c) edge [mega thick] node {}  (b)
(d) edge [mega thick] node {}  (c) 

(d) edge [color=black, bend left=40] (a)
(c) edge [color=black, bend right, dashed] (a)
(d) edge [color=black, bend left, dashed] (b)

;
\end{tikzpicture}
\caption{Step 3}
\end{subfigure}
\begin{subfigure}[b]{0.22\textwidth}
\centering
\begin{tikzpicture}[->,>=stealth,shorten >=1pt,auto,node distance=1.4cm,
thick,main node/.style={rectangle,fill=none,draw,minimum size = 0.6cm,font=\small\bfseries}]
\node[main node] (a) {A};
\node[main node] (b) [below of=a] {B};
\node[main node] (c) [below of=b] {C};
\node[main node] (d) [below of=c] {D}; 
\node[main node] (e) [below of=d] {E};
\node[main node] (f) [draw=none, below of=e] {};
\path 
(b) edge [mega thick] node {} (a)
(c) edge  [mega thick] node {}  (b)
(d) edge [mega thick] node {}  (c)
(e) edge [bend right, mega thick] node  {}  (b)

(e) edge [color=ggray, bend right=25, dash dot,very thick] (c)
(e) edge [color=black] node  {}  (d) 
(d) edge [color=black, bend left=40] (a)
(c) edge [color=black, bend right, dashed] (a)
(d) edge [color=black, bend left, dashed] (b)
(e) edge [color=black, bend left=45, dashed] (a)
;
\end{tikzpicture}
\caption{Step 4}
\end{subfigure}
\begin{subfigure}[b]{0.30\textwidth}
\centering
\begin{tikzpicture}[->,>=stealth,shorten >=1pt,auto,node distance=1.4cm,
thick,main node/.style={rectangle,fill=none,draw,minimum size = 0.6cm,font=\small\bfseries}]
\node[main node] (a) {A};
\node[main node] (b) [below of=a] {B};
\node[main node] (c) [below of=b] {C};
\node[main node] (d) [below of=c] {D}; 
\node[main node] (e) [below of=d] {E};
\node[main node] (f) [below of=e] {F};
\path 
(b) edge [mega thick] node {}  (a)
(c) edge [mega thick]  node  {} (b)
(d) edge [mega thick] node {}  (c)
(e) edge [bend right, mega thick] node {}  (b)
(f) edge [mega thick] node {}  (e)

(e) edge [color=black] node  {}  (d) 
(d) edge [color=black, bend left=40] (a)
(c) edge [color=black, bend right, dashed] (a)
(d) edge [color=black, bend left, dashed] (b)
(e) edge [color=black, bend left=45, dashed] (a)
(f) edge [color=black, bend right=40] (c) 
(f) edge [color=black, bend right=50, dashed] (b)
(f) edge [color=black, bend left=55] (a)

(e) edge [color=ggray, bend right=25, dash dot, very thick] (c)
(f) edge [color=ggray, bend left, dash dot, very thick] (d)
;
\end{tikzpicture}
\caption{Step 5}
\end{subfigure}
\caption{The intended argument graphs for Dialogue 1 extended with support coming from indirect defenses from Definition \ref{def:prudent}, 
secondary/supported attacks and super--mediated attacks following the dialogue flow. 
Dash dotted gray edges represent the additional supports obtainable if we again add defense--based support. 
Solid edges represent the attack relation and dashed edges represent the support relation. Thick edges represent the relations appearing in 
the intended graphs.} 
\label{fig:intendedcompbip}	
\end{figure}
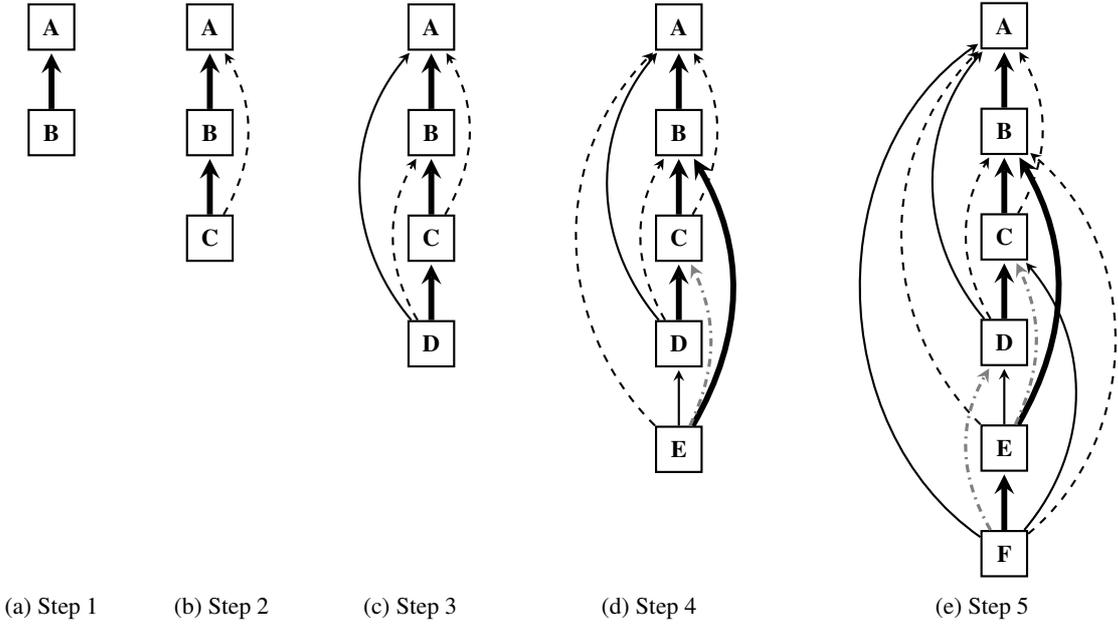

Let us now look again at the common graphs, but from the perspective of bipolar argumentation. Like in the previous case, we extend 
the intended graph with support relations corresponding to the indirect defenses from Definition \ref{def:prudent}. 
However, instead of extending the graph with the indirect conflicts 
in the prudent/careful sense, we will use the ones offered by bipolar argumentation. The obtained augmented 
graphs are visible in Figure \ref{fig:intendedcompbip}. 
By adding the secondary or supported attacks 
(in this particular case, they overlap) to the intended graphs, we recreate the augmented structures visible in Figure \ref{fig:intendedcomp}. 
By considering the (super) mediated attacks that adhere to the flow of the dialogue (i.e. the source of the attack has to appear later than the target), 
we can create the $(E, D)$ and $(F, C)$ attacks, which were missing in the previous approach. Consequently, we are now only missing 
the $(E, C)$ and $(F, D)$ supports. This issue can be addressed in two ways. One method is to repeat the process of 
adding defense--based 
support to the graph. This turns $E$ into a supporter of $C$ (it provides defense against $D$) and $F$ into a supporter of $D$ (it provides defense 
against $E$). Another approach is to consider $(E, C)$ as a supporting link that should have been included in the intended graph 
in the first place. 
In such a situation, the $(F, C)$ conflict can be reproduced by a secondary attack, while the $(F, D)$ support would still require a repetition of the step in 
which we add the defense--based support.  

We can observe that both approaches offer a way of reconstructing the common graphs 
(with the exception of the $(\{A\}, \emptyset, \{(B,A)\}, \emptyset)$ at step 1) from the intended one. The purely conflict--based approach 
would require the introduction of certain auxiliary notions. The bipolar approach uses methods already available in the literature, but in turn requires some of the procedures to be repeated. 

\begin{figure}[!ht] 
 \centering
\begin{subfigure}[b]{0.12\textwidth}
\centering
\begin{tikzpicture}[->,>=stealth,shorten >=1pt,auto,node distance=1.4cm,
thick,main node/.style={rectangle,fill=none,draw,minimum size = 0.6cm,font=\small\bfseries}]
\node[main node] (a) {A};
\node[main node] (b) [below of=a] {B};
\node[main node] (c) [draw=none,below of=b] {};
\node[main node] (d) [draw=none,below of=c] {};
\node[main node] (e) [draw=none,below of=d] {};
\node[main node] (f) [draw=none, below of=e] {};
\path 
(b) edge [mega thick] node {} (a)
;
\end{tikzpicture}
\caption{Step 1}
\end{subfigure}
\begin{subfigure}[b]{0.14\textwidth}
\centering
\begin{tikzpicture}[->,>=stealth,shorten >=1pt,auto,node distance=1.4cm,
thick,main node/.style={rectangle,fill=none,draw,minimum size = 0.6cm,font=\small\bfseries}]
\node[main node] (a) {A};
\node[main node] (b) [below of=a] {B};
\node[main node] (c) [below of=b] {C};
\node[main node] (d) [draw=none,below of=c] {};
\node[main node] (e) [draw=none,below of=d] {};
\node[main node] (f) [draw=none, below of=e] {};
\path 
(b) edge [mega thick] node {} (a)
(c) edge [mega thick] node {}  (b) 
(c) edge [color=black, bend right, dashed] (a)
;
\end{tikzpicture}
\caption{Step 2}
\end{subfigure}
\begin{subfigure}[b]{0.15\textwidth}
\centering
\begin{tikzpicture}[->,>=stealth,shorten >=1pt,auto,node distance=1.4cm,
thick,main node/.style={rectangle,fill=none,draw,minimum size = 0.6cm,font=\small\bfseries}]
\node[main node] (a) {A};
\node[main node] (b) [below of=a] {B};
\node[main node] (c) [below of=b] {C};
\node[main node] (d) [below of=c] {D}; 
\node[main node] (e) [draw=none,below of=d] {};
\node[main node] (f) [draw=none, below of=e] {};
\path 
(b) edge [mega thick]node {} (a)
(c) edge [mega thick] node {}  (b)
(d) edge [mega thick] node {}  (c) 

(d) edge [color=black, bend left=40] (a)
(c) edge [color=black, bend right, dashed] (a)
(d) edge [color=black, bend left, dashed] (b)

;
\end{tikzpicture}
\caption{Step 3}
\end{subfigure}
\begin{subfigure}[b]{0.22\textwidth}
\centering
\begin{tikzpicture}[->,>=stealth,shorten >=1pt,auto,node distance=1.4cm,
thick,main node/.style={rectangle,fill=none,draw,minimum size = 0.6cm,font=\small\bfseries}]
\node[main node] (a) {A};
\node[main node] (b) [below of=a] {B};
\node[main node] (c) [below of=b] {C};
\node[main node] (d) [below of=c] {D}; 
\node[main node] (e) [below of=d] {E};
\node[main node] (f) [draw=none, below of=e] {};
\path 
(b) edge [mega thick] node {} (a)
(c) edge  [mega thick] node {}  (b)
(d) edge [mega thick] node {}  (c)
(e) edge [bend right, mega thick] node  {}  (b) 

(d) edge [color=black, bend left=40] (a)
(c) edge [color=black, bend right, dashed] (a)
(d) edge [color=black, bend left, dashed] (b)
(e) edge [color=black, bend left=45, dashed] (a)
;
\end{tikzpicture}
\caption{Step 4}
\end{subfigure}
\begin{subfigure}[b]{0.30\textwidth}
\centering
\begin{tikzpicture}[->,>=stealth,shorten >=1pt,auto,node distance=1.4cm,
thick,main node/.style={rectangle,fill=none,draw,minimum size = 0.6cm,font=\small\bfseries}]
\node[main node] (a) {A};
\node[main node] (b) [below of=a] {B};
\node[main node] (c) [below of=b] {C};
\node[main node] (d) [below of=c] {D}; 
\node[main node] (e) [below of=d] {E};
\node[main node] (f) [below of=e] {F};
\path 
(b) edge [mega thick] node {}  (a)
(c) edge [mega thick]  node  {} (b)
(d) edge [mega thick] node {}  (c)
(e) edge [bend right, mega thick] node {}  (b)
(f) edge [mega thick] node {}  (e)

(d) edge [color=black, bend left=40] (a)
(c) edge [color=black, bend right, dashed] (a)
(d) edge [color=black, bend left, dashed] (b)
(e) edge [color=black, bend left=45, dashed] (a)
(f) edge [color=black, bend right=50, dashed] (b)
(f) edge [color=black, bend left=55] (a)
;
\end{tikzpicture}
\caption{Step 5}
\end{subfigure}
\caption{The intended argument graphs for Dialogue 1 extended with indirect conflicts and defenses from Definition \ref{def:prudent}. 
Solid edges represent the attack relation and dashed edges represent the defense--generated support relation. 
Thick edges represent the relations appearing in the intended graphs.} 
 \label{fig:intendedcomp}	
\end{figure}
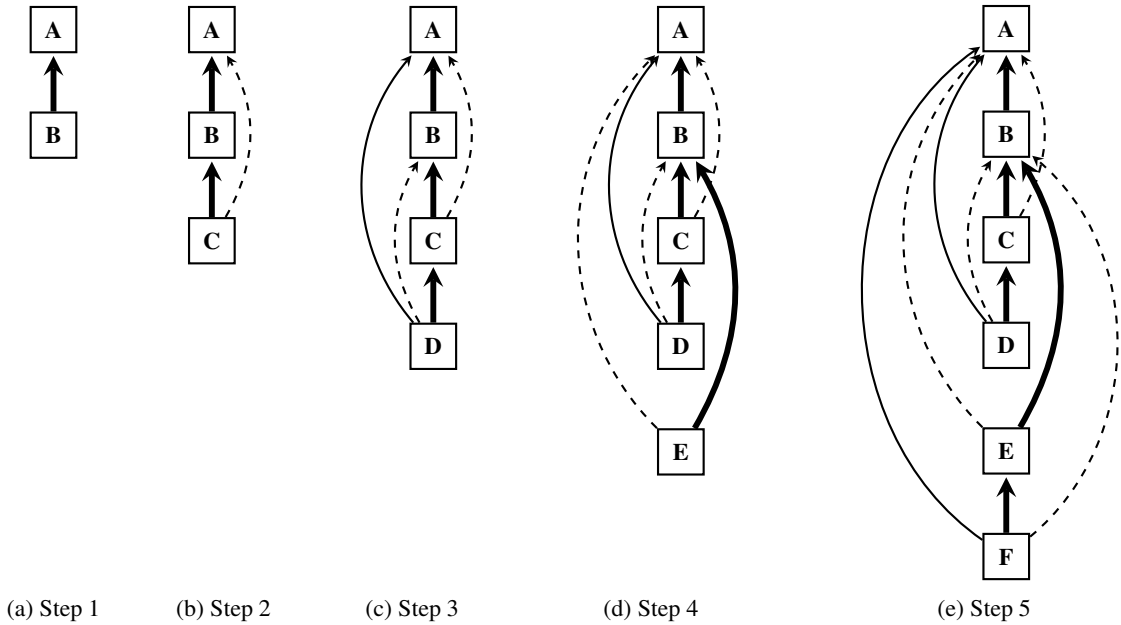

In addition to the presented graphs, in Table \ref{tab:d1rels} we can see how often a given relation was declared 
as attacking, supporting, dependent or nonexistent (in \cite{resources} we can find the same data w.r.t the dialogue steps). 
There are various differences between how a given relation was perceived by the participants belonging to the core 
and total samples. However, we can observe that for a given link, the most commonly associated type by one sample is the same as in the other. 
The only exceptions are the $(F, A)$ and $(F, C)$ links, which were considered primarily attacking in the core sample and dependent in the total sample. 
There appear to be more differences in what is the second and the third most common type. For example, 
a relation primarily seen as attacking and secondly as supporting in the total sample is more likely to be seen as primarily attacking and
secondly dependent in the core one. However, we believe these differences are an effect of the method we have used to select 
the core sample and are most likely not indicative of any particular behaviour patterns. Finally, we can note that if we extracted 
graphs based on the data found in Table \ref{tab:d1rels} for the core sample, we would recreate the common graphs for that sample. 
In the total case, we would not obtain the additional support graph from step 1, and due to the issue with the $(F,A)$ and $(F, C)$ links, 
the common graphs would be precision subgraphs of the structures extracted from the table. 
 
 \begin{table}[!ht]
\centering
\begin{tabular}{|c||c|c|c|c||c|c|c|c|} 
\cline{2-9}
\multicolumn{1}{c|}{}		&\multicolumn{4}{c||}{Core Sample}			&\multicolumn{4}{c|}{Total Sample} \\
\hline
Relation &	 in $\atts$ 	&	 in $\sups$ 	&	 in $\deps$ 	&	 N/A 	&	 in $\atts$ 	&	 in $\sups$ 	&	 in $\deps$ 	&	 N/A 	\\ \hline																			 
(B, A) 	&	85		&	1.25		&	13.75	&	0	&	60.50	&	31.50	&	8		&	0	\\
(C, A) 	&	12.50	&	84.38	&	3.13		&	0	&	29.38	&	63.75	&	5.63		&	1.25	\\
(C, B) 	&	89.06	&	3.13		&	7.81		&	0	&	68.75	&	20		&	7.50		&	3.75	\\
(D, A) 	&	81.25	&	6.25		&	12.50	&	0	&	53.33	&	30.83	&	15		&	0.83	\\
(D, B) 	&	2.08		&	97.92	&	0		&	0	&	10		&	84.17	&	5.83		&	0	\\
(D, C) 	&	91.67	&	0		&	8.33		&	0	&	66.67	&	25		&	8.33		&	0	\\
(E, A) 	&	21.88	&	62.50	&	15.63	&	0	&	30		&	45		&	16.25	&	8.75	\\
(E, B) 	&	81.25	&	3.13		&	15.63	&	0	&	53.75	&	16.25	&	21.25	&	8.75	\\
(E, C) 	&	12.50	&	71.88	&	15.63	&	0	&	28.75	&	43.75	&	20		&	7.5	\\
(E, D) 	&	68.75	&	3.13		&	28.13	&	0	&	53.75	&	10		&	27.50	&	8.75	\\
(F, A) 	&	50		&	6.25		&	43.75	&	0	&	27.50	&	30		&	40		&	2.5	\\
(F, B) 	&	12.50	&	68.75	&	18.75	&	0	&	10		&	62.50	&	25		&	2.5	\\
(F, C) 	&	62.50	&	6.25		&	31.25	&	0	&	30		&	27.50	&	37.50	&	5	\\
(F, D) 	&	0		&	68.75	&	31.25	&	0	&	0		&	55		&	37.50	&	7.5	\\
(F, E) 	&	81.25	&	0		&	18.75	&	0	&	52.50	&	22.50	&	20		&	5	\\
\hline
\end{tabular}
\caption{Occurrences of the declared relations in Dialogue 1 (values are expressed as $\%$)}
\label{tab:d1rels}
\end{table} 
  
In \cite{resources} we have also calculated the average distance from one framework to other declared frameworks obtained from the participants, 
understood as the average of all distances from a given structure to the remaining 39 ones in the total sample and 15 in the core one, i.e.: 

\begin{definition}
\label{def:avg}
Let $n$ be the number of participants and let $\{G_1, \ldots, G_n\}$ be the frameworks they have declared. The average distance from a framework
$G_i$, where $1 \leq i \leq n$, to other frameworks is defined as: 
\begin{equation}
avg\_dist(G_i) = \frac{ \sum_{j=1}^n d(G_j, G_i) }{n-1}
\end{equation}
\end{definition}

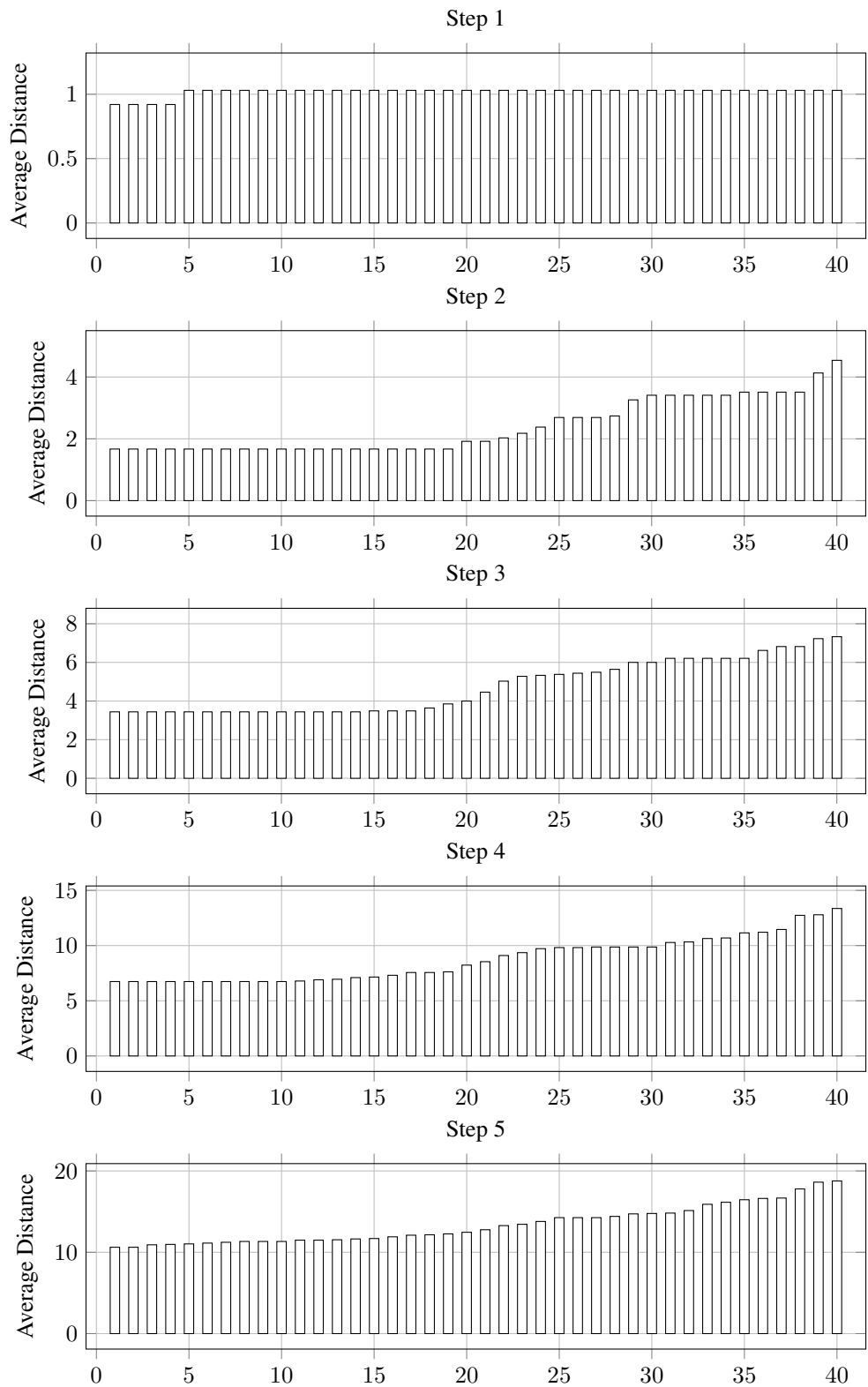
\begin{figure}[!ht]
\centering
\pgfplotsset{compat=1.9,every axis/.style={width=0.8\textwidth, height = 0.2\textheight, xmin=1, xmax=40,
grid=both, 
xlabel = {}, ylabel = {Average Distance},
yticklabel style={
        /pgf/number format/fixed,
        /pgf/number format/precision=4
}, bar width = 4pt,
scaled y ticks=false,
ybar=2pt, 
enlargelimits=true,
enlarge x limits = 0.04,
every node near coord/.append style={font=\scriptsize},
legend style={at={(1.05,0.5)},anchor=west, font=\scriptsize,align=center}, legend columns=1, 
legend image code/.code={%
      \draw[#1] (0cm,-0.1cm) rectangle (0.3cm,0.3cm);}
}}
\begin{tikzpicture} 
\begin{axis}[ymin=0, ymax=1.2, title = {Step 1}, name = dist1]    
\addplot[fill =white] table [y=Step 1] {\dialonedistfiletot};   
\end{axis}
\begin{axis}[ymin=0, ymax=5,  title = {Step 2}, name = dist2, at = (dist1.below south west), anchor = above north west]    
\addplot[fill =white] table [y=Step 2] {\dialonedistfiletot};   
\end{axis}
\begin{axis}[ymin=0, ymax=8,  title = {Step 3}, name = dist3, at = (dist2.below south west), anchor = above north west]    
\addplot[fill =white] table [y=Step 3] {\dialonedistfiletot};  
\end{axis}
\begin{axis}[ymin=0, ymax=14,  title = {Step 4}, name = dist4, at = (dist3.below south west), anchor = above north west]    
\addplot[fill =white] table [y=Step 4] {\dialonedistfiletot};    
\end{axis}
\begin{axis}[ymin=0, ymax=19,  title = {Step 5}, name = dist5, at = (dist4.below south west), anchor = above north west]    
\addplot[fill =white] table [y=Step 5] {\dialonedistfiletot};   
\end{axis}
\end{tikzpicture}
\caption{The list of average distances from the graph declared by a given participant to the graphs of the remaining participants at a given step 
in Dialogue 1.} 
\label{fig:avgdist1}
\end{figure}

The results 
for the total sample can be seen in Figure \ref{fig:avgdist1}, while the overall summary of our findings is presented in Table \ref{tab:dist1}. 
In this table we present the average distance from the common framework 
at a given step to all of the other frameworks as well as the general analysis of all of the obtained average distances. We provide the minimum, maximum 
and median of the obtained values, and the overall average distance (i.e. the average of the averages). 

We can observe that the further we are in the dialogue, the higher these values become, which is natural given the increasing size of the frameworks. 
The average distance calculated for the common framework tends to be smaller than the overall average and less or equal to the obtained median. 
In the core sample, all of the averages for the common frameworks are also the minimum ones from all of the obtained values. We can therefore 
observe that the fact that this framework is defined by the largest number of participants and is close to or identical with the framework induced by Table \ref{tab:d1rels} is reflected in the obtained distances between the structures declared 
by the participants. 

\begin{table}[!ht]
\centering 
  \begin{tabular}{ |c|c|c|c|c|c|c|c|c|c|c| }
\cline{2-11}
\multicolumn{1}{c|}{}	&	\multicolumn{5}{c|}{Total Sample} &	\multicolumn{5}{c|}{Core Sample}								\\
\cline{2-11}
\multicolumn{1}{c|}{}	&	\rot{\parbox{1.9cm}{Common\\Framework}}	&	\rot{Minimum}	&	\rot{Maximum}	&	\rot{Average}	&	\rot{Median}	&	\rot{\parbox{1.9cm}{Common\\Framework}}	&	\rot{Minimum}	&	\rot{Maximum}	&	\rot{Average}	&	\rot{Median}	\\
\hline
\hline
Step 1	&	1.03	&	0.92	&	1.03	&	1.02	&	1.03	&	0.33	&	0.33	&	1.80	&	0.52	&	0.33	\\
Step 2	&	1.67	&	1.67	&	4.54	&	2.40	&	1.92	&	0.40	&	0.40	&	3.87	&	0.72	&	0.40	\\
Step 3	&	3.44	&	3.44	&	7.33	&	4.75	&	4.23	&	1.33	&	1.33	&	7.60	&	2.22	&	1.33	\\
Step 4	&	6.74	&	6.74	&	13.36	&	8.78	&	8.39	&	2.07	&	2.07	&	11.93	&	3.44	&	2.07	\\
Step 5	&	11.33	&	10.62	&	18.77	&	13.40	&	12.62	&	5.13	&	5.13	&	13.13	&	7.17	&	6.87	\\
   \hline
  \end{tabular}
\caption{Analysis of the average distance values for the declared frameworks for Dialogue 1, where average distance is formulated in Definition \ref{def:avg}. 
We include the average distance from the common 
framework to other frameworks, minimum and maximum average distances amongst all the averages, median, and overall average. }
\label{tab:dist1}
 \end{table}

\pgfplotstableread
[col sep=&,row sep=\\]
{
ID&Step 1&Step 2&Step 3&Step 4&Step 5\\
1&0.92&1.67&3.44&6.74&10.62\\
2&0.92&1.67&3.44&6.74&10.62\\
3&0.92&1.67&3.44&6.74&10.92\\
4&0.92&1.67&3.44&6.74&10.97\\
5&1.03&1.67&3.44&6.74&11.03\\
6&1.03&1.67&3.44&6.74&11.13\\
7&1.03&1.67&3.44&6.74&11.23\\
8&1.03&1.67&3.44&6.74&11.33\\
9&1.03&1.67&3.44&6.74&11.33\\
10&1.03&1.67&3.44&6.74&11.33\\
11&1.03&1.67&3.44&6.79&11.49\\
12&1.03&1.67&3.44&6.9&11.49\\
13&1.03&1.67&3.44&6.95&11.54\\
14&1.03&1.67&3.44&7.1&11.64\\
15&1.03&1.67&3.49&7.15&11.69\\
16&1.03&1.67&3.49&7.31&11.9\\
17&1.03&1.67&3.49&7.56&12.1\\
18&1.03&1.67&3.64&7.56&12.15\\
19&1.03&1.67&3.85&7.62&12.26\\
20&1.03&1.92&4&8.23&12.46\\
21&1.03&1.92&4.46&8.54&12.77\\
22&1.03&2.03&5.03&9.1&13.28\\
23&1.03&2.18&5.28&9.36&13.44\\
24&1.03&2.38&5.33&9.72&13.79\\
25&1.03&2.69&5.38&9.82&14.26\\
26&1.03&2.69&5.44&9.82&14.26\\
27&1.03&2.69&5.49&9.87&14.26\\
28&1.03&2.74&5.64&9.87&14.41\\
29&1.03&3.26&6&9.87&14.72\\
30&1.03&3.41&6&9.87&14.77\\
31&1.03&3.41&6.21&10.28&14.82\\
32&1.03&3.41&6.21&10.33&15.13\\
33&1.03&3.41&6.21&10.64&15.9\\
34&1.03&3.41&6.21&10.69&16.15\\
35&1.03&3.51&6.21&11.15&16.46\\
36&1.03&3.51&6.62&11.21&16.62\\
37&1.03&3.51&6.82&11.46&16.67\\
38&1.03&3.51&6.82&12.74&17.79\\
39&1.03&4.13&7.23&12.79&18.62\\
40&1.03&4.54&7.33&13.36&18.77\\
}\dialonedistfiletot

\clearpage
\subsubsection{Dialogue 2}

\paragraph{Intended Graph Analysis}\hfill
\label{sec:intended2}

\begin{figure}[!ht] 
 \centering
\begin{subfigure}[b]{0.15\textwidth}
\centering
\begin{tikzpicture}
[->,>=stealth,shorten >=1pt,auto,node distance=1.4cm,
thick,main node/.style={rectangle,fill=none,draw,minimum size = 0.6cm,font=\small\bfseries}]
\node[main node] (a) {A};
\node[main node] (b) [below of=a] {B};
\node[main node] (c) [draw=none,below of=b] {};
\node[main node] (d) [draw=none,below of=c] {};
\node[main node] (e) [draw=none,below of=d] {};
\node[main node] (f) [draw=none,below of=e] {};
\node[main node] (g) [draw=none,below of=f] {};
\node[main node] (h) [draw=none,below of=g] {};
\path[every node/.style={sloped,anchor=south,auto=false}]
(b) edge node {} (a)
;
\end{tikzpicture}
\caption{Step 1}
\end{subfigure}
\begin{subfigure}[b]{0.2\textwidth}
\centering
\begin{tikzpicture}
[->,>=stealth,shorten >=1pt,auto,node distance=1.4cm,
thick,main node/.style={rectangle,fill=none,draw,minimum size = 0.6cm,font=\small\bfseries}]
\node[main node] (a) {A};
\node[main node] (b) [below of=a] {B};
\node[main node] (c) [below of=b] {C};
\node[main node] (d) [below of=c] {D};
\node[main node] (e) [draw=none,below of=d] {};
\node[main node] (f) [draw=none,below of=e] {};
\node[main node] (g) [draw=none,below of=f] {};
\node[main node] (h) [draw=none,below of=g] {};
\path[every node/.style={sloped,anchor=south,auto=false}]
(b) edge node {} (a) 
(d) edge[bend right=45] node {} (b) 
;
\end{tikzpicture}
\caption{Step 2}
\end{subfigure}
\begin{subfigure}[b]{0.2\textwidth}
\centering
\begin{tikzpicture}
[->,>=stealth,shorten >=1pt,auto,node distance=1.4cm,
thick,main node/.style={rectangle,fill=none,draw,minimum size = 0.6cm,font=\small\bfseries}]
\node[main node] (a) {A};
\node[main node] (b) [below of=a] {B};
\node[main node] (c) [below of=b] {C};
\node[main node] (d) [below of=c] {D};
\node[main node] (e) [below of=d] {E};
\node[main node] (f) [below of=e] {F};
\node[main node] (g) [draw=none,below of=f] {};
\node[main node] (h) [draw=none,below of=g] {};
\path[every node/.style={sloped,anchor=south,auto=false}]
(b) edge node {} (a) 
(d) edge[bend right=45] node {} (b)
(e) edge node {} (d)
(e) edge[bend left=45] node {} (c)
 
;
\end{tikzpicture}
\caption{Step 3}
\end{subfigure}
\begin{subfigure}[b]{0.2\textwidth}
\centering
\begin{tikzpicture}
[->,>=stealth,shorten >=1pt,node distance=1.4cm,
thick,main node/.style={rectangle,fill=none,draw,minimum size = 0.6cm,font=\small\bfseries}]
\node[main node] (a) {A};
\node[main node] (b) [below of=a] {B};
\node[main node] (c) [below of=b] {C};
\node[main node] (d) [below of=c] {D};
\node[main node] (e) [below of=d] {E};
\node[main node] (f) [below of=e] {F};
\node[main node] (g) [below of=f] {G};
\node[main node] (h) [draw=none,below of=g] {};
\path[every node/.style={sloped,anchor=south,auto=false}]
(b) edge node {} (a) 
(d) edge[bend right=45] node {} (b)
(e) edge node {} (d)
(e) edge[bend left=45] node {} (c)
(g) edge[bend right=55, ] node {} (e)
;
 \draw[->,color=ggray, very thick ] (g) -- node[strike out,draw,-]{} (f);
\end{tikzpicture}
\caption{Step 4}
\end{subfigure}
\begin{subfigure}[b]{0.2\textwidth}
\centering
\begin{tikzpicture}
[->,>=stealth,shorten >=1pt, node distance=1.4cm,
thick,main node/.style={rectangle,fill=none,draw,minimum size = 0.6cm,font=\small\bfseries}]
\node[main node] (a) {A};
\node[main node] (b) [below of=a] {B};
\node[main node] (c) [below of=b] {C};
\node[main node] (d) [below of=c] {D};
\node[main node] (e) [below of=d] {E};
\node[main node] (f) [below of=e] {F};
\node[main node] (g) [below of=f] {G};
\node[main node] (h) [below of=g] {H};
\path[every node/.style={sloped,anchor=south,auto=false}]
(b) edge node {} (a) 
(d) edge[bend right=45] node {} (b)
(e) edge node {} (d)
(e) edge[bend left=45] node {} (c)
(g) edge[bend right=55, ] node {} (e)
(h) edge node {} (g) 
;
 \draw[->,color=ggray, very thick ] (g) -- node[strike out,draw,-]{} (f);
\end{tikzpicture}
\caption{Step 5}
\end{subfigure}
\caption{The intended argument graphs for  Dialogue 2. Solid edges represent the attack relation. The crossed gray 
edges represent attacks 
that some participants have interpreted as supports in a way that can be considered justified. } 
 \label{fig:intended2}	
\end{figure}
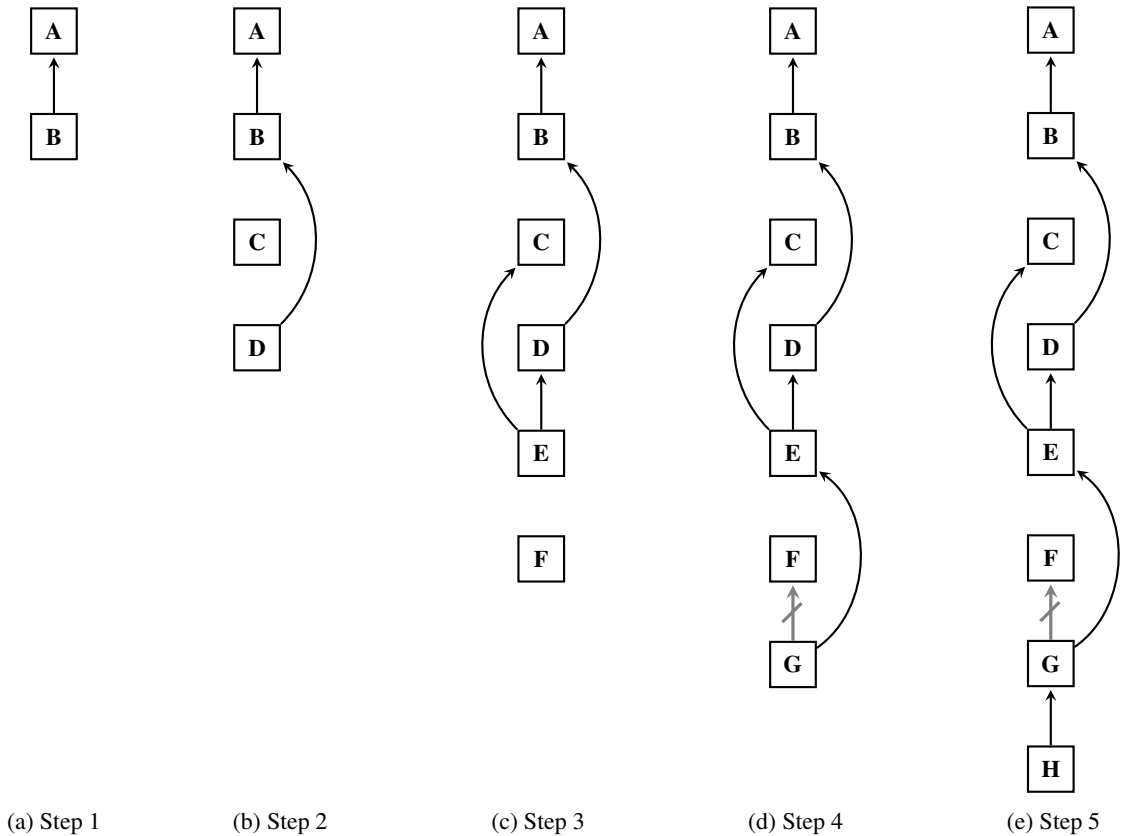

The graphs depicting the minimal set of relations we had intended the participants to recognize in the second dialogue are presented in Figure \ref{fig:intended2}. 
We will now discuss the reason why the $(G,F)$ edge is grayed out.  
Argument $G$ was meant to be a counterargument for $E$ and $F$. In the first case the contradiction is more obvious. In the 
second instance, 
the fact that the virus is accompanied only by stabilizers and antibiotics means it is not accompanied by thimerosal, which is only a preservative. 
Thus, the contradiction depends on the distinction between stabilizers and preservatives. Based on the explanations 
provided by some of the participants we can however observe that these two notions were occasionally confused, which additional research has shown 
to be a common situation in reality.  
Consequently, thimerosal was seen as an example of a stabilizer and as a result, $G$ could have been in fact understood as supporting $F$ rather than 
attacking it. Hence, declaring this relation differently was a conscious and somewhat justified decision, not a result of misunderstanding the 
exercise or an unintentional choice (a \enquote{misclick}). Nevertheless, removing this link from the intended graph does not significantly affect the core sample and would 
allow one more person in w.r.t. the original approach. This indicates that the majority of people marking $(G, F)$ differently than intended have also 
had other issues that prevented them from entering the core sample. Thus, this issue deserves attention on its own, and for now we will evaluate 
the answers without modifying the initially set methodology.

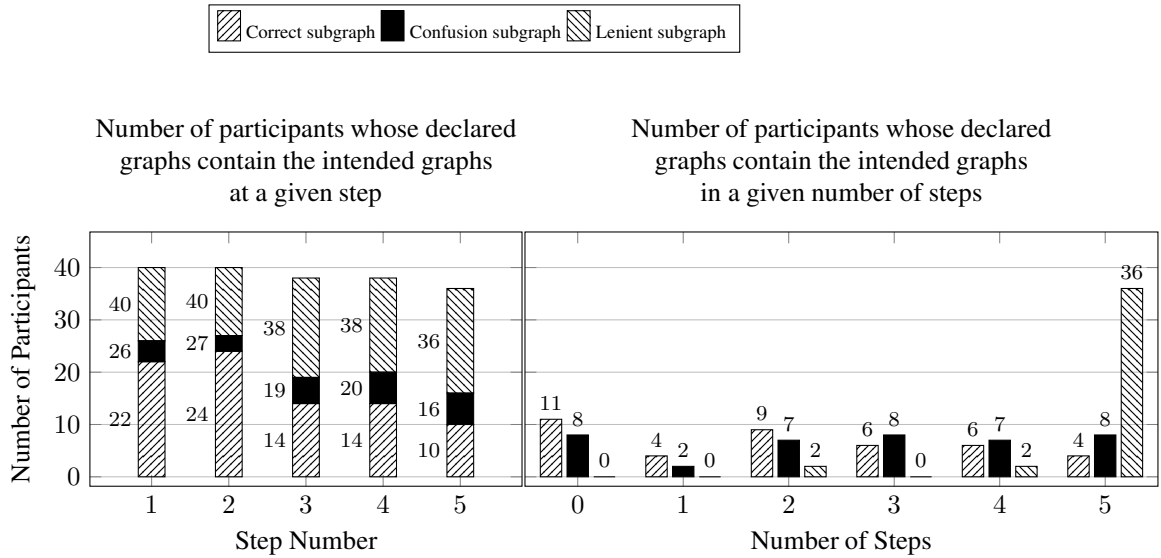
\begin{figure}[!ht]
\centering 
\pgfplotsset{compat=1.9}
\begin{tikzpicture} 
\begin{axis}[name=ccount,height=0.3\textwidth, width = 0.44\textwidth, ybar stacked, bar width = 10pt, point meta = explicit, 
nodes near coords, every node near coord/.append style={font=\footnotesize,anchor=east,xshift=-4pt},ytick ={0,10,20,30,40},
xlabel={Step Number},ylabel={Number of Participants},xtick ={1,2,3,4,5},ymin=0, ymax=45, ymajorgrids, yminorgrids,  
enlargelimits=true, 
enlarge x limits = 0.2, enlarge y limits = 0.04, 
legend style={at={(1.1,0.5)},anchor=west, font=\scriptsize,align=center}, legend columns=1, 
legend image code/.code={%
      \draw[#1] (0cm,-0.1cm) rectangle (0.3cm,0.3cm);
    },title style={align=center}, 
title = {Number of participants whose declared \\ graphs contain the intended graphs\\ at a given step}
]

\addplot[pattern = north east lines]  coordinates
{(1,22) [22]
(2,24) [24]
(3,14) [14]
 (4,14) [14]
(5,10) [10]
}; 
\addplot[fill=black]  coordinates
{(1,4) [26]
(2,3)  [27]
(3,5) [19]
(4,6) [20]
(5,6) [16]
}; 
\addplot[pattern = north west lines]  coordinates
{(1,14) [40]
(2,13) [40]
 (3,19) [38]
(4,18) [38]
(5,20) [36]
};  

\end{axis}  
\begin{axis}[at=(ccount.right of south east), title style={align=center},
title = {Number of participants whose declared \\ graphs contain the intended graphs\\ in a given number of steps},
width=0.6\textwidth, height = 0.3\textwidth,ybar, xtick align=inside,
nodes near coords, every node near coord/.append style={font=\footnotesize},   
xlabel={Number of Steps},
yticklabels={,,,,},
xtick ={0,1,2,3,4,5},ymax=45,
enlarge x limits = 0.1, enlarge y limits = 0.04, ytick ={0,10,20,30,40},
bar width=8pt, ymajorgrids, yminorgrids,  
legend style={at={(-0.5,1.8)},anchor=west, font=\scriptsize,align=center}, legend columns=3, 
legend image code/.code={%
      \draw[#1] (0cm,-0.1cm) rectangle (0.3cm,0.3cm);
    }
]

\addplot[pattern = north east lines]  coordinates
{(0,11)  (1,4) (2,9) (3,6) (4,6) (5,4)}; 
\addplot[fill=black]  coordinates
{(0,8)  (1,2) (2,7) (3,8) (4,7) (5,8)};
\addplot[pattern = north west lines]  coordinates
{(0,0)  (1,0) (2,2) (3,0) (4,2) (5,36)};

 \legend{Correct subgraph, Confusion  subgraph, Lenient subgraph}
 
\end{axis}
\end{tikzpicture}
\caption{Analysis of intended graph containment in Dialogue 2} 
\label{fig:contained2}
\end{figure}

Similarly as in the case of Dialogue 1, in Figure \ref{fig:contained2} we present how many participants recognized
the intended graph at a given step of the dialogue, and how many participants recognized the intended graph in a given number of steps. 
We create the core sample for Dialogue 2 by gathering participants who have acknowledged the intended graph w.r.t. the confusion subgraph relation 
in at least 4 stages. Hence, from now on we will be working with the total sample of size 40 and the core sample of size 15.

\paragraph{Participant--Sourced Graphs Analysis}\hfill 
\label{sec:dial2usergraph} 

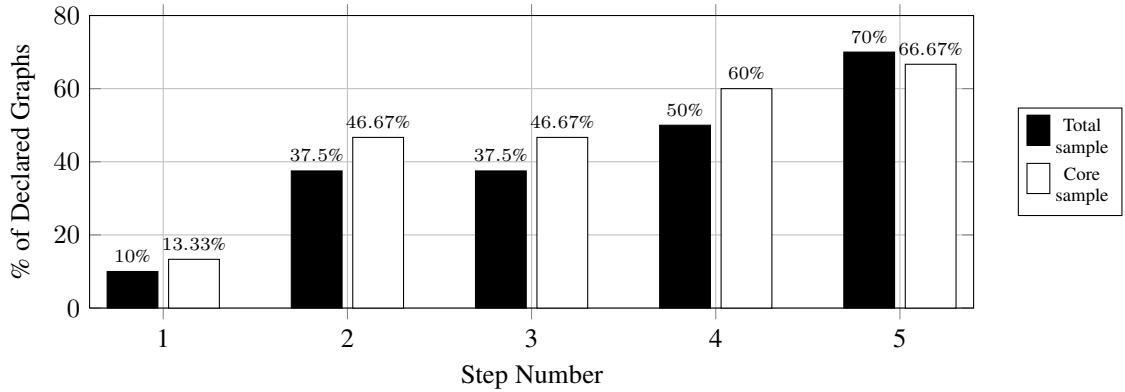
\begin{figure}[!ht]
\centering
\pgfplotsset{compat=1.9,every axis/.style={width=0.8\textwidth, height = 0.25\textheight,  grid=both, 
xlabel = {Step Number}, ylabel = {\% of Declared Graphs},
yticklabel style={
        /pgf/number format/fixed,
        /pgf/number format/precision=4
}, bar width = 19pt,
scaled y ticks=false, ybar=4pt, ymin=0, ymax=80, nodes near coords = {\pgfmathprintnumber\pgfplotspointmeta\%}, 
every node near coord/.append style={font=\scriptsize},
xtick = {1,2,3,4,5}, 
legend style={at={(1.05,0.5)},anchor=west, font=\scriptsize,align=center}, legend columns=1, 
legend image code/.code={%
      \draw[#1] (0cm,-0.1cm) rectangle (0.3cm,0.3cm);}
}}
\begin{tikzpicture} 

\begin{axis}[name = clartot, xticklabels = {1,2,3,4,5}]   
\addplot[fill=black] coordinates{(1,10) (2,37.5) (3,37.5) (4,50) (5,70)}; 
\addplot[fill = white] coordinates{(1,13.33) (2,46.67) (3,46.67) (4,60) (5,66.67)}; 
\legend{Total \\ sample, Core \\ sample}
\end{axis}
\end{tikzpicture}
\caption{Portion of unclarified graphs (i.e. with nonempty set of dependencies) obtained in Dialogue 2} 
\label{fig:clarified2}
\end{figure}

Let us now look at the graphs given to us by the participants of Dialogue 2 (full data can be found in the data appendix in \cite{resources}). 
Just like in the previous dialogue, the further we are in the discussion, the more single--participant graphs we obtain. 
However, in this case, the maximal dispersion occurs during the fourth rather than fifth step, in which the participants tend to agree a bit more again.  
As the dialogue progresses, the number of clarified graphs decreases, as seen in Figure \ref{fig:clarified2}. Although there are some differences 
between the decrease in this dialogue and in Dialogue 1, particularly concerning the second stage, the end states are not that different. 
It therefore appears that participants in both samples had similar issues in deciding on the nature of some of the edges. 

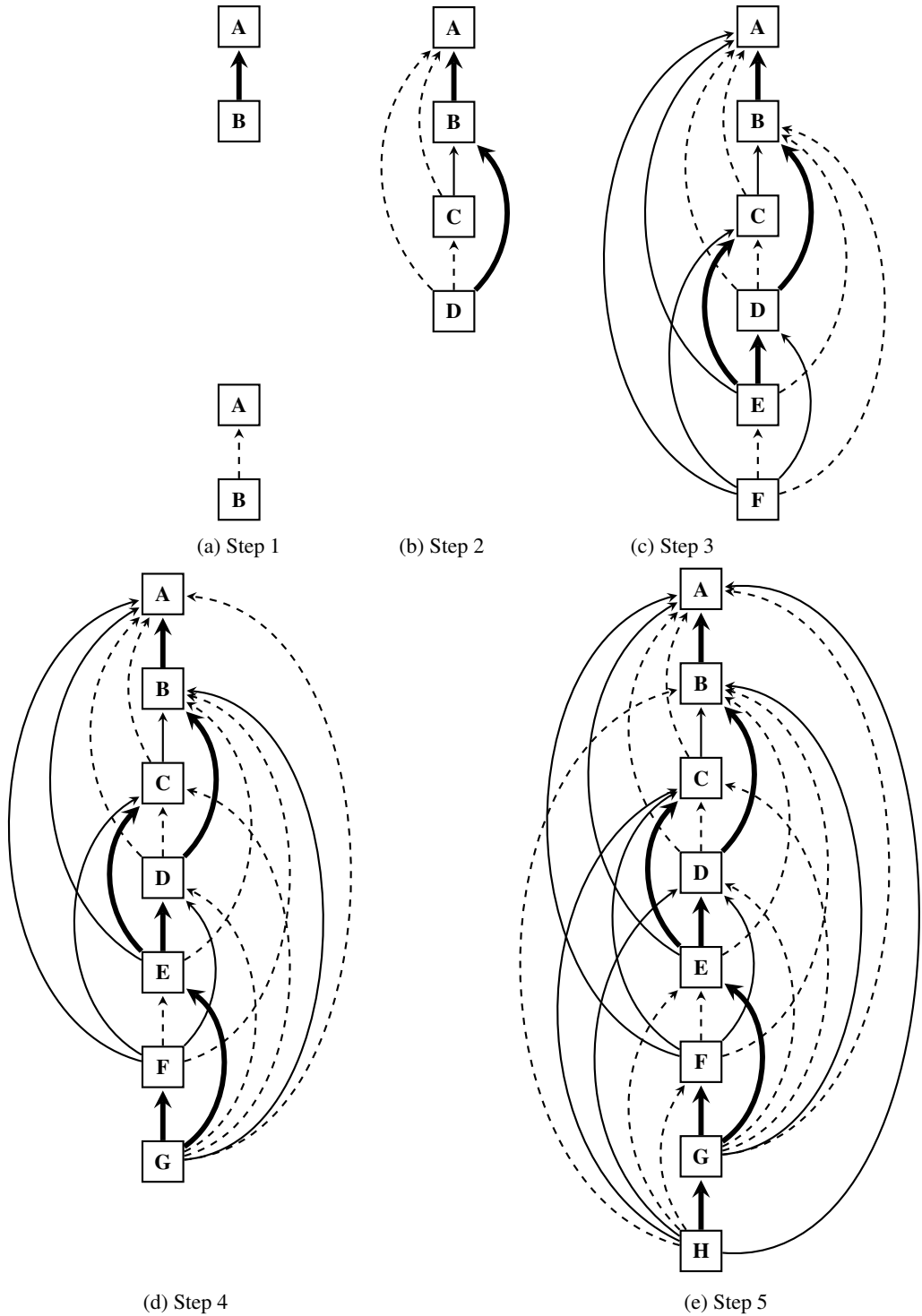
\begin{figure}[p] 
 \centering
\begin{subfigure}[b]{0.15\textwidth}
\centering
\begin{tikzpicture}
[->,>=stealth,shorten >=1pt,auto,node distance=1.4cm,
thick,main node/.style={rectangle,fill=none,draw,minimum size = 0.6cm,font=\small\bfseries}]
\node[main node] (a) {A};
\node[main node] (b) [below of=a] {B};
\node[main node] (c) [draw=none,below of=b] {};
\node[main node] (d) [draw=none,below of=c] {};
\node[main node] (e) [below of=d] {A};
\node[main node] (f) [below of=e] {B};
\path[every node/.style={sloped,anchor=south,auto=false}]
(b) edge [mega thick] node {} (a)
(f) edge [color=black, dashed] node {} (e)
;
\end{tikzpicture}
\caption{Step 1}
\end{subfigure}
\begin{subfigure}[b]{0.2\textwidth}
\centering
\begin{tikzpicture}
[->,>=stealth,shorten >=1pt,auto,node distance=1.4cm,
thick,main node/.style={rectangle,fill=none,draw,minimum size = 0.6cm,font=\small\bfseries}]
\node[main node] (a) {A};
\node[main node] (b) [below of=a] {B};
\node[main node] (c) [below of=b] {C};
\node[main node] (d) [below of=c] {D};
\node[main node] (e) [draw=none,below of=d] {};
\node[main node] (f) [draw=none,below of=e] {};
\path[every node/.style={sloped,anchor=south,auto=false}]
(b) edge  [mega thick] node {} (a) 
(d) edge[bend right=45, mega thick] node {} (b)

(c) edge[color=black ] node {} (b)
(d) edge[color=black,  dashed] node {}(c)
(c) edge[bend left=30, color=black, dashed] node {} (a)
(d) edge[bend left=45, color=black, dashed] node {} (a)
;
\end{tikzpicture}
\caption{Step 2}
\end{subfigure}
\begin{subfigure}[b]{0.2\textwidth}
\centering
\begin{tikzpicture}
[->,>=stealth,shorten >=1pt,auto,node distance=1.4cm,
thick,main node/.style={rectangle,fill=none,draw,minimum size = 0.6cm,font=\small\bfseries}]
\node[main node] (a) {A};
\node[main node] (b) [below of=a] {B};
\node[main node] (c) [below of=b] {C};
\node[main node] (d) [below of=c] {D};
\node[main node] (e) [below of=d] {E};
\node[main node] (f) [below of=e] {F};
\path[every node/.style={sloped,anchor=south,auto=false}]
(b) edge  [mega thick] node {} (a) 
(d) edge[bend right=45, mega thick] node {} (b)
(e) edge  [mega thick]node {} (d)
(e) edge[bend left=45, mega thick] node {} (c)

(c) edge[color=black] node {} (b)
(e) edge[bend left=60, color=black ] node {} (a)
(f) edge[bend left=75, color=black ] node {} (a)
(f) edge[bend left=60, color=black ] node {} (c)
(f) edge[bend right=45, color=black ] node {} (d)

(c) edge[bend left=30, color=black , dashed] node {} (a)
(d) edge[bend left=45, color=black,  dashed] node {} (a)
(d) edge[color=black,  dashed] node {}(c)
(e) edge[bend right=60, color=black,  dashed] node {} (b)
(f) edge[bend right=75, color=black, dashed] node {} (b)
(f) edge[color=black, dashed] node {} (e)
 
;
\end{tikzpicture}
\caption{Step 3}
\end{subfigure}

\begin{subfigure}[b]{0.45\textwidth}
\centering
\begin{tikzpicture}
[->,>=stealth,shorten >=1pt,auto,node distance=1.4cm,
thick,main node/.style={rectangle,fill=none,draw,minimum size = 0.6cm,font=\small\bfseries}]
\node[main node] (a) {A};
\node[main node] (b) [below of=a] {B};
\node[main node] (c) [below of=b] {C};
\node[main node] (d) [below of=c] {D};
\node[main node] (e) [below of=d] {E};
\node[main node] (f) [below of=e] {F};
\node[main node] (g) [below of=f] {G};
\node[main node] (h) [draw=none,below of=g] {};
\path[every node/.style={sloped,anchor=south,auto=false}]
(b) edge  [mega thick] node {} (a) 
(d) edge[bend right=45, mega thick] node {} (b)
(e) edge  [mega thick]node {} (d)
(e) edge[bend left=45, mega thick] node {} (c)
(g) edge[bend right=55, mega thick] node {} (e)
(g) edge  [mega thick] node {} (f)

(c) edge[color=black] node {} (b)
(e) edge[bend left=60, color=black ] node {} (a)
(f) edge[bend left=75, color=black ] node {} (a)
(f) edge[bend left=60, color=black ] node {} (c)
(f) edge[bend right=45, color=black ] node {} (d)
(g) edge[bend right=85, color=black] node {} (b)

(c) edge[bend left=30, color=black , dashed] node {} (a)
(d) edge[bend left=45, color=black,  dashed] node {} (a)
(d) edge[color=black,  dashed] node {}(c)
(e) edge[bend right=60, color=black,  dashed] node {} (b)
(f) edge[bend right=75, color=black, dashed] node {} (b)
(f) edge[color=black, dashed] node {} (e)
(g) edge[bend right=85, color=black,dashed] node {} (a)
(g) edge[bend right=75, color=black, dashed] node {} (c)
(g) edge[bend right=65, color=black, dashed] node {} (d) 
;
\end{tikzpicture}
\caption{Step 4}
\end{subfigure}
\begin{subfigure}[b]{0.5\textwidth}
\centering
\begin{tikzpicture}
[->,>=stealth,shorten >=1pt,auto,node distance=1.4cm,
thick,main node/.style={rectangle,fill=none,draw,minimum size = 0.6cm,font=\small\bfseries}]
\node[main node] (a) {A};
\node[main node] (b) [below of=a] {B};
\node[main node] (c) [below of=b] {C};
\node[main node] (d) [below of=c] {D};
\node[main node] (e) [below of=d] {E};
\node[main node] (f) [below of=e] {F};
\node[main node] (g) [below of=f] {G};
\node[main node] (h) [below of=g] {H};
\path[every node/.style={sloped,anchor=south,auto=false}]
(b) edge  [mega thick] node {} (a) 
(d) edge[bend right=45, mega thick] node {} (b)
(e) edge  [mega thick]node {} (d)
(e) edge[bend left=45, mega thick] node {} (c)
(g) edge[bend right=55, mega thick] node {} (e)
(g) edge  [mega thick] node {} (f)
(h) edge  [mega thick] node {} (g)

(c) edge[color=black] node {} (b)
(e) edge[bend left=60, color=black ] node {} (a)
(f) edge[bend left=75, color=black ] node {} (a)
(f) edge[bend left=60, color=black ] node {} (c)
(f) edge[bend right=45, color=black ] node {} (d)
(g) edge[bend right=85, color=black] node {} (b)
(h) edge[bend right=95, color=black] node {} (a)
(h) edge[bend left=65, color=black] node {} (c)
(h) edge[bend left=55, color=black] node {} (d)

(c) edge[bend left=30, color=black , dashed] node {} (a)
(d) edge[bend left=45, color=black,  dashed] node {} (a)
(d) edge[color=black,  dashed] node {}(c)
(e) edge[bend right=60, color=black,  dashed] node {} (b)
(f) edge[bend right=75, color=black, dashed] node {} (b)
(f) edge[color=black, dashed] node {} (e)
(g) edge[bend right=85, color=black,dashed] node {} (a)
(g) edge[bend right=75, color=black, dashed] node {} (c)
(g) edge[bend right=65, color=black, dashed] node {} (d)
(h) edge[bend left=75, color=black, dashed] node {} (b)
(h) edge[bend left=45, color=black, dashed] node {} (e)
(h) edge[bend left=35, color=black,dashed] node {} (f)
;
\end{tikzpicture}
\caption{Step 5}
\end{subfigure}
\caption{The common argument graphs for Dialogue 2 based on the total sample. With the exception of the bottom framework 
in Step 1, they are also  common in the core sample.  The thicker edges represent the relations appearing in the intended graph. Solid edges
stand for attack and dashed for support links.  } 
 \label{fig:declared2}	
\end{figure}

In Figure \ref{fig:declared2} we have depicted the  common graphs at every stage of the dialogue, where thick edges represent 
the relations contained in the intended graph.  Just like in Dialogue 1, at the very first step the participants are asked only about the $(B,A)$ relation. 
Although more people have seen the $(B, A)$ relation as attacking rather than supporting (22 vs 14 in the total sample), both groups are important enough 
to be reported. Many people who have marked $(B, A)$
as supporting did not make it to the core sample, in which the attacking interpretation is dominant. 

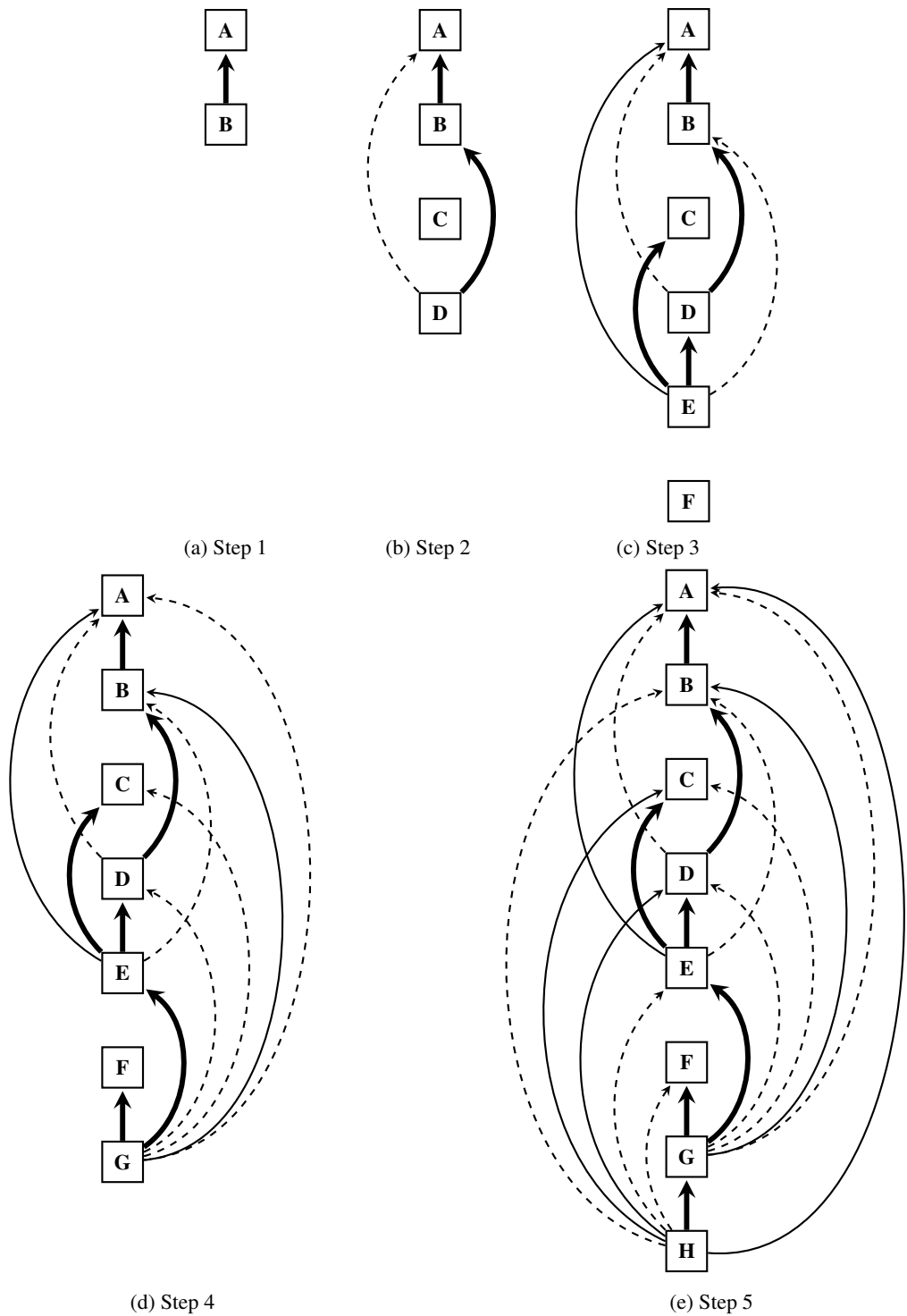
\begin{figure}[p] 
 \centering
\begin{subfigure}[b]{0.15\textwidth}
\centering
\begin{tikzpicture}
[->,>=stealth,shorten >=1pt,auto,node distance=1.4cm,
thick,main node/.style={rectangle,fill=none,draw,minimum size = 0.6cm,font=\small\bfseries}]
\node[main node] (a) {A};
\node[main node] (b) [below of=a] {B};
\node[main node] (c) [draw=none,below of=b] {};
\node[main node] (d) [draw=none,below of=c] {};
\node[main node] (e) [draw=none,below of=d] {};
\node[main node] (f) [draw=none,below of=e] {};
\path[every node/.style={sloped,anchor=south,auto=false}]
(b) edge [mega thick] node {} (a) 
;
\end{tikzpicture}
\caption{Step 1}
\end{subfigure}
\begin{subfigure}[b]{0.2\textwidth}
\centering
\begin{tikzpicture}
[->,>=stealth,shorten >=1pt,auto,node distance=1.4cm,
thick,main node/.style={rectangle,fill=none,draw,minimum size = 0.6cm,font=\small\bfseries}]
\node[main node] (a) {A};
\node[main node] (b) [below of=a] {B};
\node[main node] (c) [below of=b] {C};
\node[main node] (d) [below of=c] {D};
\node[main node] (e) [draw=none,below of=d] {};
\node[main node] (f) [draw=none,below of=e] {};
\path[every node/.style={sloped,anchor=south,auto=false}]
(b) edge  [mega thick] node {} (a) 
(d) edge[bend right=45, mega thick] node {} (b)
 
(d) edge[bend left=45, color=black, dashed] node {} (a)
;
\end{tikzpicture}
\caption{Step 2}
\end{subfigure}
\begin{subfigure}[b]{0.2\textwidth}
\centering
\begin{tikzpicture}
[->,>=stealth,shorten >=1pt,auto,node distance=1.4cm,
thick,main node/.style={rectangle,fill=none,draw,minimum size = 0.6cm,font=\small\bfseries}]
\node[main node] (a) {A};
\node[main node] (b) [below of=a] {B};
\node[main node] (c) [below of=b] {C};
\node[main node] (d) [below of=c] {D};
\node[main node] (e) [below of=d] {E};
\node[main node] (f) [below of=e] {F};
\path[every node/.style={sloped,anchor=south,auto=false}]
(b) edge  [mega thick] node {} (a) 
(d) edge[bend right=45, mega thick] node {} (b)
(e) edge  [mega thick]node {} (d)
(e) edge[bend left=45, mega thick] node {} (c)

(d) edge[bend left=45, color=black,  dashed] node {} (a)
(e) edge[bend left=60, color=black ] node {} (a)
(e) edge[bend right=60, color=black,  dashed] node {} (b)
;
\end{tikzpicture}
\caption{Step 3}
\end{subfigure}

\begin{subfigure}[b]{0.45\textwidth}
\centering
\begin{tikzpicture}
[->,>=stealth,shorten >=1pt,auto,node distance=1.4cm,
thick,main node/.style={rectangle,fill=none,draw,minimum size = 0.6cm,font=\small\bfseries}]
\node[main node] (a) {A};
\node[main node] (b) [below of=a] {B};
\node[main node] (c) [below of=b] {C};
\node[main node] (d) [below of=c] {D};
\node[main node] (e) [below of=d] {E};
\node[main node] (f) [below of=e] {F};
\node[main node] (g) [below of=f] {G};
\node[main node] (h) [draw=none,below of=g] {};
\path[every node/.style={sloped,anchor=south,auto=false}]
(b) edge  [mega thick] node {} (a) 
(d) edge[bend right=45, mega thick] node {} (b)
(e) edge  [mega thick]node {} (d)
(e) edge[bend left=45, mega thick] node {} (c)
(g) edge[bend right=55, mega thick] node {} (e)
(g) edge  [mega thick] node {} (f)

(d) edge[bend left=45, color=black,  dashed] node {} (a)
(e) edge[bend left=60, color=black ] node {} (a)
(e) edge[bend right=60, color=black,  dashed] node {} (b)
(g) edge[bend right=85, color=black,dashed] node {} (a)
(g) edge[bend right=85, color=black] node {} (b)
(g) edge[bend right=75, color=black, dashed] node {} (c)
(g) edge[bend right=65, color=black, dashed] node {} (d)  
;
\end{tikzpicture}
\caption{Step 4}
\end{subfigure}
\begin{subfigure}[b]{0.5\textwidth}
\centering
\begin{tikzpicture}
[->,>=stealth,shorten >=1pt,auto,node distance=1.4cm,
thick,main node/.style={rectangle,fill=none,draw,minimum size = 0.6cm,font=\small\bfseries}]
\node[main node] (a) {A};
\node[main node] (b) [below of=a] {B};
\node[main node] (c) [below of=b] {C};
\node[main node] (d) [below of=c] {D};
\node[main node] (e) [below of=d] {E};
\node[main node] (f) [below of=e] {F};
\node[main node] (g) [below of=f] {G};
\node[main node] (h) [below of=g] {H};
\path[every node/.style={sloped,anchor=south,auto=false}]
(b) edge  [mega thick] node {} (a) 
(d) edge[bend right=45, mega thick] node {} (b)
(e) edge  [mega thick]node {} (d)
(e) edge[bend left=45, mega thick] node {} (c)
(g) edge[bend right=55, mega thick] node {} (e)
(g) edge  [mega thick] node {} (f)
(h) edge  [mega thick] node {} (g)

(d) edge[bend left=45, color=black,  dashed] node {} (a)
(e) edge[bend left=60, color=black ] node {} (a)
(e) edge[bend right=60, color=black,  dashed] node {} (b)
(g) edge[bend right=85, color=black,dashed] node {} (a)
(g) edge[bend right=85, color=black] node {} (b)
(g) edge[bend right=75, color=black, dashed] node {} (c)
(g) edge[bend right=65, color=black, dashed] node {} (d)  
(h) edge[bend left=75, color=black, dashed] node {} (b)
(h) edge[bend left=45, color=black, dashed] node {} (e)
(h) edge[bend left=35, color=black,dashed] node {} (f)
(h) edge[bend right=95, color=black] node {} (a)
(h) edge[bend left=65, color=black] node {} (c)
(h) edge[bend left=55, color=black] node {} (d)
;
\end{tikzpicture}
\caption{Step 5}
\end{subfigure}
\caption{The intended argument graphs for  Dialogue 2 extended with indirect conflicts and defenses 
from Definition \ref{def:prudent}. 
The thicker edges represent the relations appearing in the intended graph. Solid edges
stand for attack and dashed ones for defense--generated support.} 
 \label{fig:intendedcomp2}	
\end{figure}

Similarly as we had done in the case of the first dialogue, we will now analyze whether the relations contained in the common graphs, but not in the intended ones, 
could possibly be explained with additional notions such as indirect attacks and defenses. Let us first consider creating the augmented graphs with the 
use of the indirect attacks and defenses from the prudent/careful setting (see Definition \ref{def:prudent}). 
The results are visible in Figure \ref{fig:intendedcomp2}. 
Starting from step 2, there are differences between the common and the augmented frameworks. 
They are primarily caused by $C$ and $F$ being initially disconnected from other statements. 
In total, we are missing eight edges -- four attacks $(C, B)$, $(F, A)$, $(F, C)$ and $(F, D)$, and four supports -- $(C, A)$, $(D, C)$, $(F, B)$ and 
$(F, E)$. We will now try to explain them following the conflict--centered perspective. 

Let us start with the $(C, B)$ conflict and the $(D, C)$ and $(C, A)$ supports. Statement $C$ simply presents a \enquote{fact} that vaccine 
contains a particular substance, which in this case is a mercury compound. As such, it does not contradict any previous information. However, 
it becomes conflicting once it is paired with the information that mercury compounds 
are poisonous, contained in statement $D$. This supporting $(D, C)$ link possibly reflects this pairing and makes the $(C, B)$ attack perfectly 
reasonable. 
One can therefore argue that either $(C, B)$ or both $(C, B)$ and $(D, C)$ should have been included in 
the intended graph. If only $(C, B)$ was added to the graph, we could potentially 
reproduce the $(D, C)$ support by treating statements working towards the same goal 
as positively related, similarly as proposed in the case of $(E, C)$ and $(F, D)$ supports in Dialogue 1. Adding 
only $(C, B)$ or both $(C, B)$ and $(D, C)$ would also 
lead to the introduction of the $(C, A)$ support through indirect defense. 

A similar analysis could be carried out for the relations associated with $F$. Statement $F$ serves as a further backing for $E$, which was recognized as support by the participants. 
Consequently, including the $(F, E)$ support and $(F, C)$ and $(F, D)$ attacks in the intended graph (or, based on the aforementioned explanation, 
just the $(F, C)$ and $(F, D)$ attacks) would lead to the reproduction of the remaining missing relations through the use of indirect attack or defense. 

Let us now look at the intended graphs augmented with supports corresponding to the indirect defenses from the prudent/careful setting and with indirect attacks associated with the bipolar setting. 
By using secondary/supported attacks and defense--induced support, we obtain the augmented frameworks identical to the 
ones present in Figure \ref{fig:intendedcomp2}. Thus, again we are missing the same set of relations as in the conflict--centered analysis: four attacks -- $(C, B)$, $(F, A)$, $(F, C)$ and $(F, D)$ -- and four supports -- $(C, A)$, $(D, C)$, $(F, B)$ and 
$(F, E)$.
In this case, the (super) mediated attacks do not offer any additional insight. 
 By considering (super) extended attacks, we can retrieve the $(C, B)$ attack, though only starting from the fourth step 
of the dialogue. In the same fashion we can also obtain the $(F, A)$, $(F, C)$ and $(F, D)$ conflicts at the last stage of 
the discussion. Nevertheless, these changes occur later than the arguments are introduced and our main issue 
is not resolved. 
 In the conflict--centered analysis, 
we had proposed adding 
certain attacks (or both attacks and supports) to the intended graph in order to bridge the gap between the common graph and the augmented graph. 
In this, more support--centered, analysis, we can consider adding the $(D, C)$ and $(F, E)$ supports to the intended graph. 
Adding the $(D, C)$ support would allow us to recreate the $(C, B)$ conflict as an extended attack. The inclusion of the $(F, E)$ support 
would lead to the recreation of the missing $(F, C)$ and $(F, D)$ attacks as supported attacks in the augmented graph. 

We are thus left with the recreation of the $(F, A)$ attack and the $(C, A)$ and $(F, B)$ supports. In order to do so, we can repeat the procedure 
of adding indirect defenses and conflicts, i.e. we extend the augmented graph in the way we extended the intended one. 
Then $F$ becomes a supported and secondary attacker of $A$ 
through the secondary/supported attack $(E, A)$ or the supported attack $(F, D)$. Additionally, $F$ is now an indirect 
defender of $B$ through the supported attack $(F, D)$. Finally,  
$(C, A)$ can be recreated through indirect defense generated by the the extended conflict $(C, B)$. 

We can observe that although certain supports can be recreated through indirect defense, 
this is not the only possible reason why the participants have declared these relations. For example, 
we can also observe the similarity between statement $A$ and the conclusion of $D$ and the fact that $G$ is a more detailed reiteration of $C$. 
If we view $(F, E)$ support as a relation that should have been in the intended graph, then given the fact that 
$E$ is a defense--mimicking supporter of $B$, the $(F, B)$ support could be seen as a particular mixture stemming from the support transitivity.  
Consequently, it is possible to be recreate some of the missing supports using a different methodology. 

To summarize, we can observe that recreation of the common graph from the intended one in Dialogue 2 requires the addition of 
at least 2 relations carried out by $C$ and $F$. However, just like in Dialogue 1, we would still need to introduce auxiliary notions when recreating 
the common graphs via the prudent/careful approach only and perform a two--step recreation via the bipolar approach.

In addition to the graph reproduction analysis, in Table \ref{tab:d2rels} we list how often a given relation was perceived as 
attacking, supporting, dependent or nonexistent during the dialogue (more detailed tables can be found in \cite{resources}). 
Similarly as in the case of Dialogue 1, we can observe that the dominating assignment 
for a given edge is the same in both core and total samples, though the second and third most common choices often differ. By focusing on the 
most common edges, we can reconstruct the graphs from Figure \ref{fig:declared2}.

 \begin{table}[!ht]
\centering
\begin{tabular}{|c||c|c|c|c||c|c|c|c|}
\cline{2-9}
\multicolumn{1}{c|}{}		&\multicolumn{4}{c||}{Core Sample}			&\multicolumn{4}{c|}{Total Sample} \\
\hline
Relation 	&	 in $\atts$ 	&	 in $\sups$ 	&	 in $\deps$ 	&	 N/A 	&	 in $\atts$ 	&	 in $\sups$ 	&	 in $\deps$ 	&	 N/A 	\\ \hline	 
(B, A)	&	85.33	&	2.67	&	12	&	0	&	66.50	&	26.50	&	7	&	0	\\
(C, A)	&	13.33	&	73.33	&	13.33	&	0	&	27.50	&	58.75	&	13.13	&	0.63	\\
(C, B)	&	75	&	3.33	&	21.67	&	0	&	68.13	&	12.50	&	16.25	&	3.13	\\
(D, A)	&	11.67	&	81.67	&	6.67	&	0	&	27.50	&	65.63	&	6.88	&	0	\\
(D, B)	&	85	&	3.33	&	11.67	&	0	&	76.25	&	12.50	&	9.38	&	1.88	\\
(D, C)	&	11.67	&	60	&	25	&	3.33	&	27.50	&	51.88	&	16.25	&	4.38	\\
(E, A)	&	73.33	&	8.89	&	17.78	&	0	&	58.33	&	26.67	&	14.17	&	0.83	\\
(E, B)	&	8.89	&	86.67	&	4.44	&	0	&	18.33	&	73.33	&	8.33	&	0	\\
(E, C)	&	91.11	&	0	&	8.89	&	0	&	65	&	17.50	&	12.50	&	5	\\
(E, D)	&	84.44	&	0	&	15.56	&	0	&	61.67	&	21.67	&	13.33	&	3.33	\\
(F, A)	&	73.33	&	11.11	&	15.56	&	0	&	58.33	&	29.17	&	11.67	&	0.83	\\
(F, B)		&	8.89	&	84.44	&	6.67	&	0	&	13.33	&	75.83	&	10	&	0.83	\\
(F, C)		&	82.22	&	0	&	17.78	&	0	&	53.33	&	30.83	&	12.50	&	3.33	\\
(F, D)	&	82.22	&	0	&	17.78	&	0	&	55.83	&	29.17	&	12.50	&	2.50	\\
(F, E)		&	2.22	&	88.89	&	8.89	&	0	&	11.67	&	75.83	&	9.17	&	3.33	\\
(G, A)	&	16.67	&	63.33	&	20	&	0	&	23.75	&	58.75	&	16.25	&	1.25	\\
(G, B)	&	60	&	10	&	30	&	0	&	61.25	&	16.25	&	21.25	&	1.25	\\
(G, C)	&	10	&	83.33	&	6.67	&	0	&	25	&	63.75	&	8.75	&	2.50	\\
(G, D)	&	16.67	&	70	&	13.33	&	0	&	21.25	&	65	&	11.25	&	2.50	\\
(G, E)	&	86.67	&	3.33	&	10	&	0	&	70	&	16.25	&	10	&	3.75	\\
(G, F)	&	80	&	6.67	&	13.33	&	0	&	63.75	&	18.75	&	15	&	2.50	\\
(H, A)	&	73.33	&	0	&	26.67	&	0	&	45	&	25	&	27.50	&	2.50	\\
(H, B)	&	13.33	&	73.33	&	13.33	&	0	&	12.50	&	67.50	&	17.50	&	2.50	\\
(H, C)	&	86.67	&	0	&	13.33	&	0	&	60	&	20	&	12.50	&	7.50	\\
(H, D)	&	73.33	&	6.67	&	20	&	0	&	52.50	&	22.50	&	22.50	&	2.50	\\
(H, E)	&	0	&	93.33	&	6.67	&	0	&	5	&	77.50	&	12.50	&	5	\\
(H, F)	&	0	&	73.33	&	26.67	&	0	&	7.50	&	72.50	&	15	&	5	\\
(H, G)	&	80	&	6.67	&	13.33	&	0	&	50	&	32.50	&	10	&	7.50	\\							 
\hline  
\end{tabular}
\caption{Occurrences of the declared relations in Dialogue 2 (expressed as $\%$)}
\label{tab:d2rels}
\end{table}

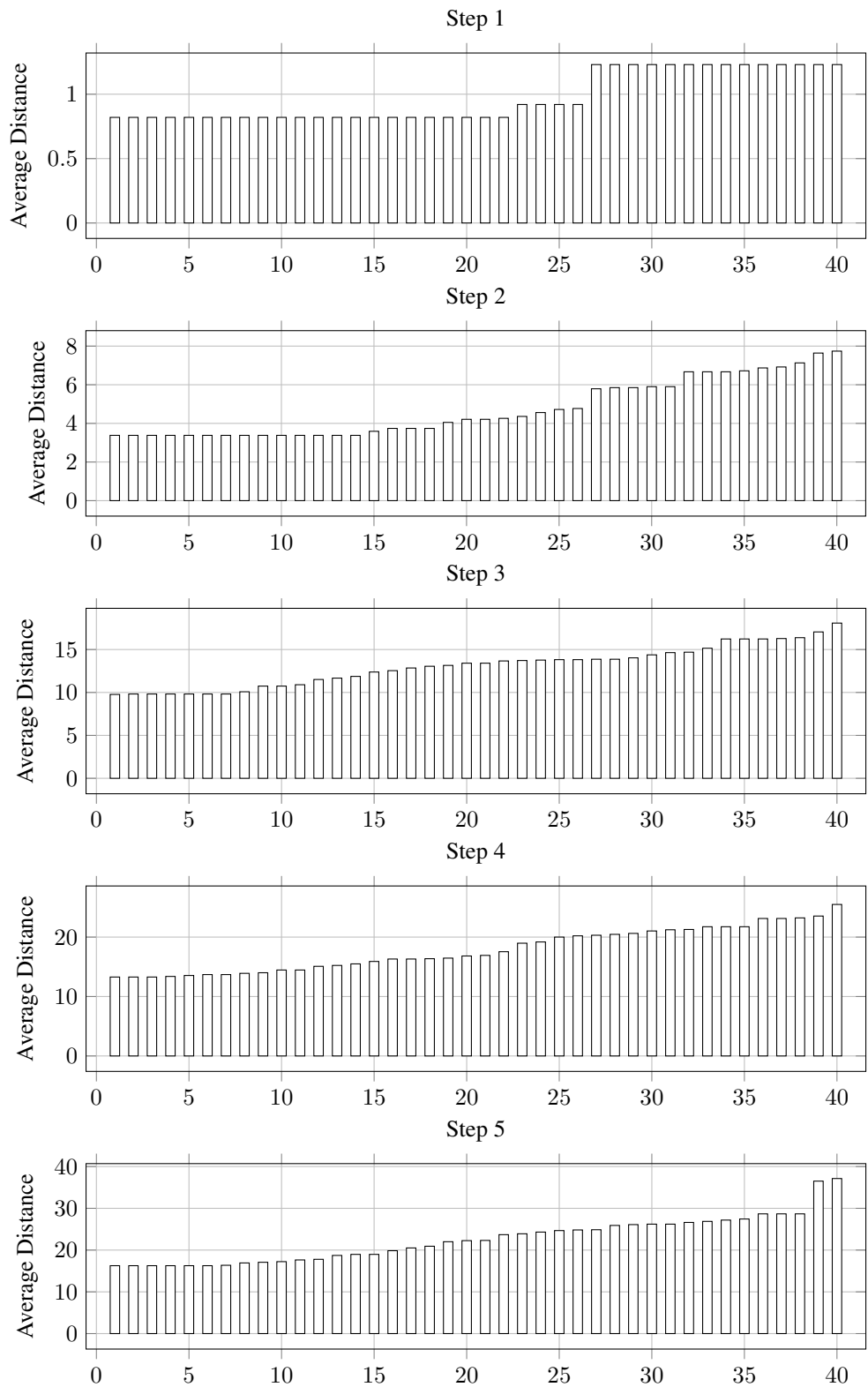
\begin{figure}[!ht]
\centering
\pgfplotsset{compat=1.9,every axis/.style={width=0.8\textwidth, height = 0.2\textheight, xmin=1, xmax=40,
grid=both, 
xlabel = {}, ylabel = {Average Distance},
yticklabel style={
        /pgf/number format/fixed,
        /pgf/number format/precision=4
}, bar width = 4pt,
scaled y ticks=false,
ybar=2pt, 
enlargelimits=true,
enlarge x limits = 0.04,
every node near coord/.append style={font=\scriptsize},
legend style={at={(1.05,0.5)},anchor=west, font=\scriptsize,align=center}, legend columns=1, 
legend image code/.code={%
      \draw[#1] (0cm,-0.1cm) rectangle (0.3cm,0.3cm);}
}}
\begin{tikzpicture} 
\begin{axis}[ymin=0, ymax=1.2, title = {Step 1}, name = dist1]    
\addplot[fill =white] table [y=Step 1] {\dialtwodistfiletot};   
\end{axis}
\begin{axis}[ymin=0, ymax=8,  title = {Step 2}, name = dist2, at = (dist1.below south west), anchor = above north west]    
\addplot[fill =white] table [y=Step 2] {\dialtwodistfiletot};   
\end{axis}
\begin{axis}[ymin=0, ymax=18,  title = {Step 3}, name = dist3, at = (dist2.below south west), anchor = above north west]    
\addplot[fill =white] table [y=Step 3] {\dialtwodistfiletot};   
\end{axis}
\begin{axis}[ymin=0, ymax=26,  title = {Step 4}, name = dist4, at = (dist3.below south west), anchor = above north west]    
\addplot[fill =white] table [y=Step 4] {\dialtwodistfiletot};   
\end{axis}
\begin{axis}[ymin=0, ymax=37,  title = {Step 5}, name = dist5, at = (dist4.below south west), anchor = above north west]    
\addplot[fill =white] table [y=Step 5] {\dialtwodistfiletot};   
\end{axis}
\end{tikzpicture}
\caption{The list of average distances from the graph declared by a given participant to the graphs of the remaining participants at a given step 
in Dialogue 2.} 
\label{fig:avgdist2}
\end{figure}

In \cite{resources} we have also calculated the average distance from one framework to other declared frameworks obtained from the participants, 
understood as the average of all distances from a given structure to the remaining 39 ones in the total sample and 14 in the core one 
(see also Definition \ref{def:avg}).  
The results in the case of the total sample are visible in Figure \ref{fig:avgdist2}, with the overall summary of our findings visible in Table \ref{tab:dist2}. 
In this table we present the average distance from the common framework 
at a given step to all of the other frameworks as well as the general analysis of all of the obtained average distances. We provide the minimum, maximum 
and median of the obtained values, and the overall average distance (i.e. the average of the averages). 

The same observations as in the case of Dialogue 1 can be made for Dialogue 2.  
The average distance calculated for the common framework tends to be smaller than the overall average and less or equal to the obtained median. 
In the case of the core sample, all of the averages for the common frameworks are also the minimum ones from all of the obtained values. This is also true 
in the case of the total sample, with the exception of the third step of the dialogue. The fact that the common 
framework is defined by the largest number of participants and is identical with the framework induced by Table \ref{tab:d2rels} 
is reflected in the obtained distances between the structures declared 
by the participants.
 
\begin{table}[!ht]
\centering 
  \begin{tabular}{ |c|c|c|c|c|c|c|c|c|c|c| }
\cline{2-11}
\multicolumn{1}{c|}{}	&	\multicolumn{5}{c|}{Total Sample} &	\multicolumn{5}{c|}{Core Sample}								\\
\cline{2-11}
\multicolumn{1}{c|}{}	&	\rot{\parbox{1.9cm}{Common\\Framework}}	&	\rot{Minimum}	&	\rot{Maximum}	&	\rot{Average}	&	\rot{Median}	&	\rot{\parbox{1.9cm}{Common\\Framework}}	&	\rot{Minimum}	&	\rot{Maximum}	&	\rot{Average}	&	\rot{Median}	\\
\hline
\hline
Step 1	&	0.82	&	0.82	&	1.23	&	0.97	&	0.82	&	0.29	&	0.29	&	1.86	&	0.48	&	0.29	\\
Step 2	&	3.38	&	3.38	&	7.74	&	4.74	&	4.21	&	1.93	&	1.93	&	8.14	&	2.95	&	2.00	\\
Step 3	&	9.82	&	9.77	&	18.08	&	13.07	&	13.41	&	5.00	&	5.00	&	15.57	&	7.70	&	5.79	\\
Step 4	&	13.28	&	13.28	&	25.49	&	17.91	&	16.87	&	6.64	&	6.64	&	21.64	&	10.15	&	8.29	\\
Step 5	&	16.26	&	16.26	&	37.13	&	22.59	&	22.29	&	7.21	&	7.21	&	25.64	&	11.43	&	8.36	\\

   \hline
  \end{tabular}
\caption{Analysis of the average distance values for the declared frameworks for Dialogue 2, where average distance is formulated in Definition \ref{def:avg}. 
We include the average distance from the common 
framework to other frameworks, minimum and maximum average distances amongst all the averages, median, and overall average. }
\label{tab:dist2}
 \end{table}

\pgfplotstableread
[col sep=&,row sep=\\]
{
ID&Step 1&Step 2&Step 3&Step 4&Step 5\\
1&0.82&3.38&9.77&13.28&16.26\\
2&0.82&3.38&9.82&13.28&16.26\\
3&0.82&3.38&9.82&13.28&16.26\\
4&0.82&3.38&9.82&13.38&16.26\\
5&0.82&3.38&9.82&13.54&16.26\\
6&0.82&3.38&9.82&13.69&16.26\\
7&0.82&3.38&9.82&13.69&16.36\\
8&0.82&3.38&10.08&13.9&16.92\\
9&0.82&3.38&10.74&14&17.08\\
10&0.82&3.38&10.74&14.46&17.23\\
11&0.82&3.38&10.9&14.46&17.64\\
12&0.82&3.38&11.51&15.08&17.79\\
13&0.82&3.38&11.67&15.23&18.72\\
14&0.82&3.38&11.87&15.49&18.97\\
15&0.82&3.59&12.38&15.9&18.97\\
16&0.82&3.74&12.54&16.31&19.85\\
17&0.82&3.74&12.85&16.31&20.51\\
18&0.82&3.74&13.05&16.36&20.92\\
19&0.82&4.05&13.15&16.46&22\\
20&0.82&4.21&13.41&16.82&22.26\\
21&0.82&4.21&13.41&16.92&22.31\\
22&0.82&4.26&13.67&17.54&23.69\\
23&0.92&4.36&13.72&18.97&23.9\\
24&0.92&4.56&13.77&19.18&24.31\\
25&0.92&4.72&13.82&20&24.67\\
26&0.92&4.77&13.82&20.21&24.82\\
27&1.23&5.79&13.87&20.31&24.87\\
28&1.23&5.85&13.87&20.46&25.9\\
29&1.23&5.85&14.03&20.62&26.1\\
30&1.23&5.9&14.38&21.03&26.21\\
31&1.23&5.9&14.64&21.23&26.21\\
32&1.23&6.67&14.69&21.28&26.62\\
33&1.23&6.67&15.15&21.74&26.87\\
34&1.23&6.67&16.23&21.74&27.18\\
35&1.23&6.72&16.23&21.74&27.44\\
36&1.23&6.87&16.23&23.13&28.67\\
37&1.23&6.92&16.28&23.13&28.67\\
38&1.23&7.13&16.38&23.23&28.67\\
39&1.23&7.64&17.05&23.54&36.51\\
40&1.23&7.74&18.08&25.49&37.13\\
}\dialtwodistfiletot

\clearpage
\subsection{Postulate Satisfaction} 
\label{sec:postsat}

In this section we analyze if and how the postulates we recalled in Section \ref{sec:epistemic} are reflected in the behaviours
of the participants. In order to do so, we will evaluate the probability distributions extracted from the agreement tasks 
against the intended, expanded, declared and common graphs (see the end of the introduction of Section \ref{sec:results} 
on page \pageref{distributions} concerning how these distributions 
were obtained).
Given the fact that the postulates are defined only for Dung's 
frameworks, we need to consider the attack subgraphs of the aforementioned graphs in order to perform the analysis. 
Although this means that the frameworks contain information not yet harnessed by the postulates, 
in Sections \ref{sec:dial1daianalysis} and \ref{sec:dial2usergraph} 
we could have observed that many of the supporting links could be explained with indirect defence or lead to additional indirect conflicts. 
Thus, we believe that the effect of the positive relations in the obtained graphs to be reasonably, even if not completely, 
reflected by the negative relations. Therefore, the 
analysis performed only on the attack graph is still valuable. 

The results will be presented in the following manner. 
For every participant, we calculate the proportion of steps on which a given postulate was satisfied according to a given graph. For example, if we have five dialogue steps and the obtained 
statement probabilities conformed to the rational postulate on four of them, we obtain the adherence rate of 80\% (or $0.8$). The values we report are the averages of these scores 
on the total and core samples. We then analyze the differences between the results obtained between the graphs and samples.
The additional statistical analysis can be found in Section \ref{sec:stats}. 

\subsubsection{Dialogue 1} 

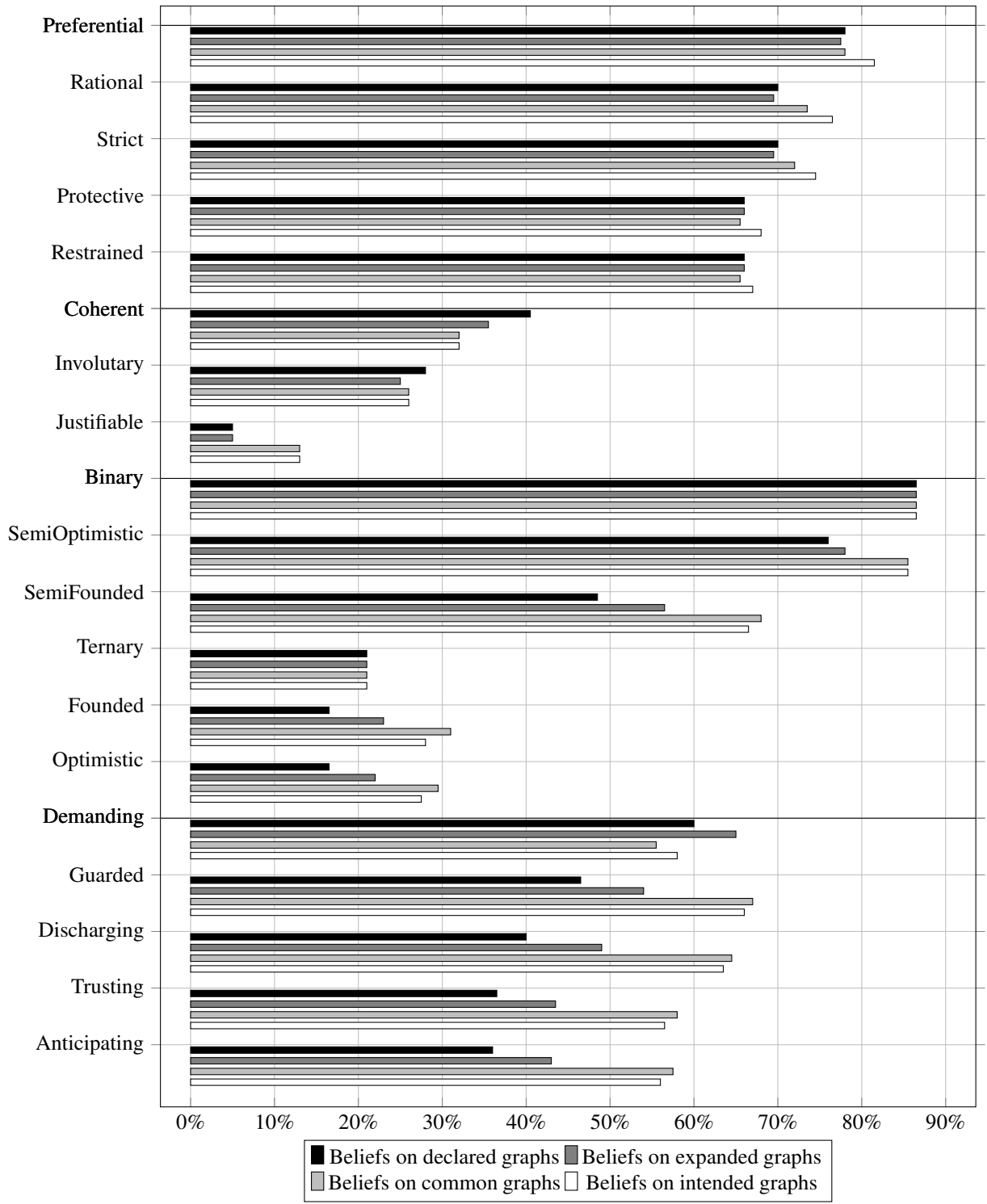
\begin{figure}[p]
\centering
\begin{tikzpicture} 
\begin{axis}[xbar, bar width=3pt,grid=both, symbolic y coords={Preferential, Rational, Strict,  Protective,  Restrained, Coherent, Involutary, Justifiable, 
Binary, SemiOptimistic, SemiFounded, Ternary, Founded, Optimistic,  Demanding, Guarded,  Discharging, Trusting,  Anticipating},
ytick = {Preferential, Rational, Strict,  Protective,  Restrained, Coherent, Involutary, Justifiable, 
Binary, SemiOptimistic, SemiFounded, Ternary, Founded, Optimistic,  Demanding, Guarded,  Discharging, Trusting,  Anticipating},  
yticklabel style = {yshift=10pt, font=\normalsize},ytick style = {yshift=10pt}, 
enlargelimits = true,
enlarge y limits = 0.04,
enlarge x limits =0.04,
y grid style = {yshift = 10pt},  extra y ticks={Preferential, Coherent, Binary, Demanding}, extra y tick style={grid=major, grid style={black}},   
 width=0.9\textwidth, height=0.9\textheight, y dir = reverse, reverse legend, 
legend style={at={(0.5,-0.03)},anchor=north}, legend columns=2,
 xmin = 0.0, xmax = 0.9, xtick = {0,0.1,0.2,0.3,0.4,0.5,0.6,0.7,0.8,0.9},
xticklabels = {0\%,10\%,20\%,30\%,40\%,50\%,60\%,70\%,80\%,90\%},
legend image code/.code={%
      \draw[#1] (0cm,-0.1cm) rectangle (0.2cm,0.2cm);}]

\addplot[fill = white] table [y=Postulate, x=Total] {\postulatesoneauthor};  

\addplot[fill=lightgray] table [y=Postulate, x=Total] {\postulatesonecommon}; 

\addplot[fill=gray] table [y=Postulate, x=Total] {\postulatesoneexpanded};  

\addplot[fill=black] table [y=Postulate, x=Total]  {\postulatesonedeclared};

\legend{Beliefs on intended graphs, Beliefs on common graphs, Beliefs on expanded graphs, Beliefs on declared graphs}  
\end{axis} 
\end{tikzpicture}
\caption{Postulate satisfaction on the total sample in Dialogue 1 }
\label{fig:postulatetotal}
\end{figure}

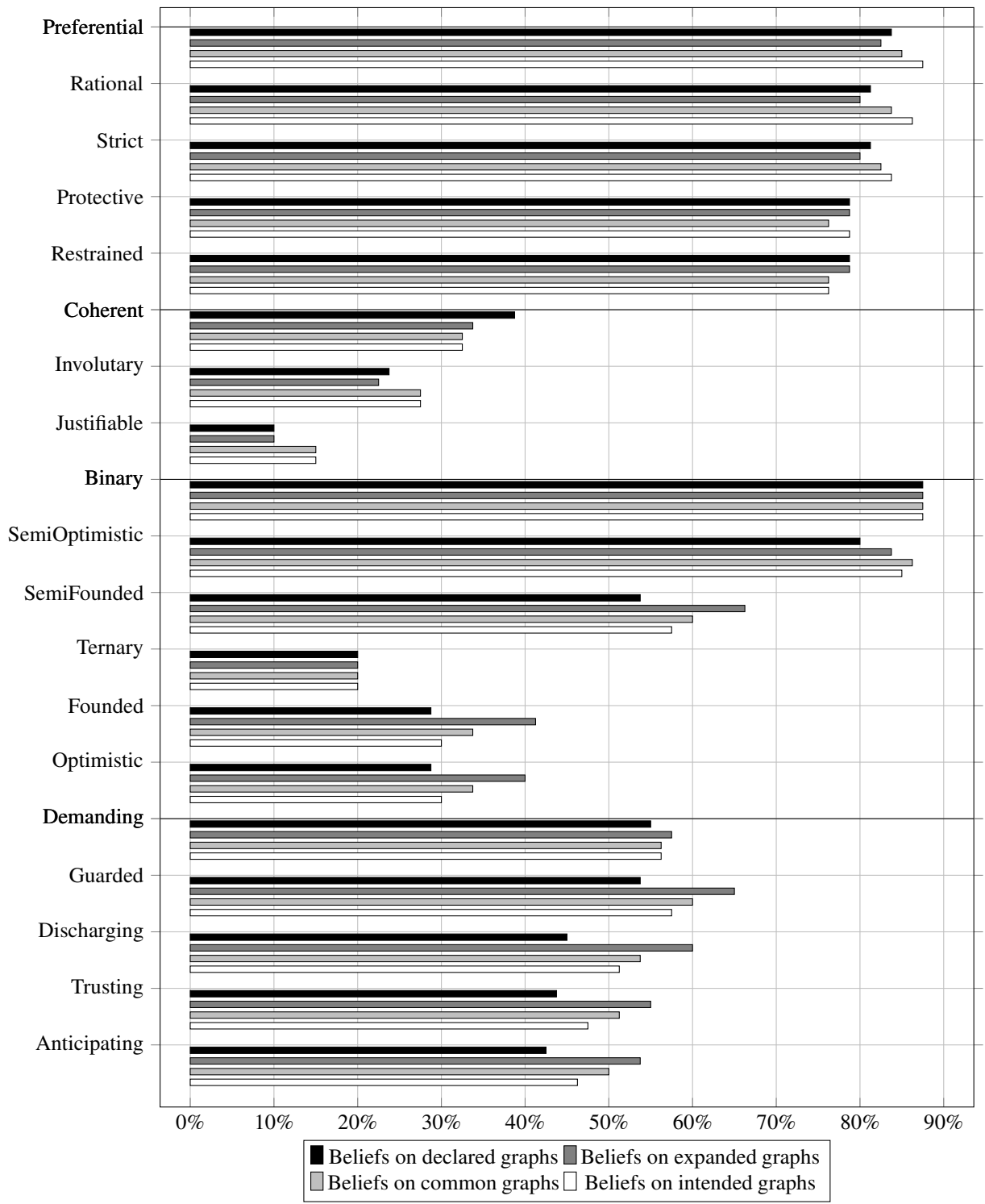
\begin{figure}[p]
\centering
\begin{tikzpicture} 
\begin{axis}[xbar, bar width=3pt,grid=both, symbolic y coords={Preferential, Rational, Strict,  Protective,  Restrained, Coherent, Involutary, Justifiable, 
Binary, SemiOptimistic, SemiFounded, Ternary, Founded, Optimistic,  Demanding, Guarded,  Discharging, Trusting,  Anticipating},
ytick = {Preferential, Rational, Strict,  Protective,  Restrained, Coherent, Involutary, Justifiable, 
Binary, SemiOptimistic, SemiFounded, Ternary, Founded, Optimistic,  Demanding, Guarded,  Discharging, Trusting,  Anticipating},  
yticklabel style = {yshift=10pt, font=\normalsize},ytick style = {yshift=10pt}, 
enlargelimits = true,
enlarge y limits = 0.04,
enlarge x limits =0.04,
y grid style = {yshift = 10pt},  extra y ticks={Preferential, Coherent, Binary, Demanding}, extra y tick style={grid=major, grid style={black}},   
 width=0.9\textwidth, height=0.9\textheight, y dir = reverse, reverse legend, 
legend style={at={(0.5,-0.03)},anchor=north}, legend columns=2,
 xmin = 0.0, xmax = 0.9, xtick = {0,0.1,0.2,0.3,0.4,0.5,0.6,0.7,0.8,0.9},
xticklabels = {0\%,10\%,20\%,30\%,40\%,50\%,60\%,70\%,80\%,90\%},
legend image code/.code={%
      \draw[#1] (0cm,-0.1cm) rectangle (0.2cm,0.2cm);}]
  
\addplot[fill=white] table [y=Postulate, x=Core] {\postulatesoneauthor};  

\addplot[fill=lightgray] table [y=Postulate, x=Core] {\postulatesonecommon}; 

\addplot[fill=gray] table [y=Postulate, x=Core]{\postulatesoneexpanded};  

\addplot[fill=black]  table [y=Postulate, x=Core] {\postulatesonedeclared};

\legend{Beliefs on intended graphs, Beliefs on common graphs, Beliefs on expanded graphs, Beliefs on declared graphs} 
\end{axis}
\end{tikzpicture}
\caption{Postulate satisfaction on core sample Dialogue 1 }
\label{fig:postulatecore}
\end{figure}

The results concerning the average adherence rates of the postulates we have recalled in Section \ref{sec:epistemic} 
on the intended, declared, expanded and  common graphs are presented in 
Figures \ref{fig:postulatetotal} (total sample) and \ref{fig:postulatecore} (core sample). 
Let us first look at the total sample. 
In the majority of cases, the results obtained on the intended and common graphs do not differ significantly, which 
can be seen as supporting our graph augmentation approach from Section \ref{sec:dial1daianalysis} 
(see also Table \ref{tab:postulatewilcox1} in Section \ref{stats:postulates1}). 
All of the results appear to be closely related 
on postulates that, in general, have a high satisfaction rate or belong to the preferential family. 
Thus, the adherence rates between the graphs are closely connected for postulates such as preferential, strict, protective and restrained, coherent, involutary and semi--optimistic. Since the ternary and binary postulates do not depend on the structure of the graph in question, they do not contribute to this particular 
observation. 
Nevertheless, once the satisfaction rate goes down and we look at the valued or explanatory postulates, 
the results begin to diverge and it is mostly the intended and common graphs that still behave similarly (see Table \ref{tab:postulatewilcox1} in Section \ref{stats:postulates1}).  

We can thus observe good adherence to the postulates and a similarity between the results on all of the graphs in the case
the preferential branch of our epistemic postulates. The obtained satisfaction rates slowly decrease as the more specialized versions are considered, 
reaching the minimum on 
the justifiable postulate. Low adherence to this property is most likely caused
by the optimistic property, which is also rarely satisfied. In principle, postulates considered shared between preferential and value based families do not 
perform as well as the ones classified as only preferential. We would also like to observe that the preferential postulate is 
highly satisfied not just
due to its \enquote{leniency}, but also because it reflects the reasoning of various participants. For example, some of the participants satisfying the preferential, 
but not the rational postulate in the first step of the dialogue, justified their agreement with both of the statements by saying that e.g. 
\enquote{\textit{A: I agree that Hospital staff members don't need to receive flu shots unless they want to. They should not be required to. }}
and \enquote{\textit{B: It is also true that hospital workers are exposed to the flu a lot so if they choose to 
it would be beneficial for them to receive the shot.}}, 
or \enquote{\textit{It's been shown, I think, that mysteriously, nurses and doctors are less prone to catching the diseases they treat in general.}}
and \enquote{\textit{Extra precautions couldn't hurt during a particularly nasty flu season.}}. 
To summarize, these results mean that participants are in general capable of reasoning well when it comes to considering the 
conflicts between two statements.   

Lower satisfaction rate of the postulates and/or bigger dissimilarities between the results on different graphs (see Table \ref{tab:postulatewilcox1} in Section \ref{stats:postulates1}) 
can be observed particularly in the case of postulates that are highly affected by the personal knowledge 
of the participants and by accuracy of modelling the perceived attacks. 
For example, these are the cases in which the results for the expanded graph -- which is precisely the declared graph 
augmented with statements extracted 
from explanations - differ from the results for the declared graph. What was considered an initial argument 
may not have been seen in the same way by the participants, thus reducing the adherence to the optimistic postulate. 
This, along with the reluctance 
of the participants to choose the extreme values (see the ternary postulate), may explain the lower performance of the 
founded property. In particular, by relaxing 
the founded restriction of complete belief to the semi--founded one, which simply requests non--disbelief, we obtain much better results. 
All of the guarded, discharging, trusting and anticipating postulates are intended to grasp the idea of believing an argument when there is no reason against 
it and disbelieving an argument only if there is a reason for it. Consequently, they are highly dependent on the accuracy of the graphs we create for 
the participants in terms of their knowledge, and are as such more difficult to measure and estimate than the postulates of the preferential family. 
The fact that the demanding postulate appears to be the least affected by it might be connected to the ternary postulate -- again, if the extreme 
values are not chosen, this property is satisfied more easily.  

We can observe that the semi--optimistic postulate is commonly satisfied, even twice as often as the coherent one. This has certain  important implications. 
The purpose of these properties is to set respectively lower and upper bounds on the belief we have in a statement in the face of its attackers. 
For example, given the performance of the protective and strict postulates, it appears that while the participants are capable of choosing what to believe 
and what to disbelieve, they can be more \enquote{generous} with their belief than anticipated by some of the postulates.  

We can now consider the results obtained for the core sample. We can observe that the average satisfaction rate of the postulates 
that the participants appeared to adhere to in the total sample increased in the core sample. 
In other words, the preferential, rational, 
strict, protective and restrained postulates are, on average, satisfied more often. 
The results obtained between the graphs appear to be much more closely connected than in the total sample (see Table \ref{tab:postulatewilcox1} in Section \ref{stats:postulates1}). 
Significant differences between the adherence rates can be observed primarily in the case of declared and expanded graph. 
They are also bigger than in the case of the total sample, i.e. the average of the differences between adherence rates of participants 
between the declared and expanded graphs is bigger in the total sample only in the case of the involutary and demanding postulates. 
This may be caused by the participants
being more forthcoming with their personal knowledge and thus the expanded graphs being more precise. It would also explain 
the somewhat better results on the explanatory and optimistic, founded and justifiable postulates, which are more dependent on such factors.  
Finally, we can consider the binary and ternary postulates. In both samples, the binary postulate has a high adherence rate, 
which means that the participants were capable of deciding 
whether to believe or disbelieve the presented statements using their own knowledge and the information found in the dialogue. Thus,  the text itself could not have been overly complicated for the participants. 
The low adherence to the ternary postulate in both of the samples, 
combined with the low 
indecisiveness of the participants, shows that people often exhibit various degrees of belief and disbelief and are reluctant to completely 
accept or completely reject a given statement. 
 
\begin{figure}[!hb]
\centering
\pgfplotsset{compat=1.9,
every axis/.style={width=0.75\textwidth, height=0.28\textwidth, grid=both, 
yticklabel style={
        /pgf/number format/fixed,
        /pgf/number format/precision=4
}, 
bar width = 16pt,
scaled y ticks=false, ybar=6pt, ymin=0, ymax=50, nodes near coords = {\pgfmathprintnumber\pgfplotspointmeta\%}, 
every node near coord/.append style={font=\scriptsize}, 
xtick = {2,3,4,5,6,7}, ytick = {10,20,30,40},
xlabel ={Number of Values},
legend style={at={(1.03,0.5)},anchor=west, font=\scriptsize}, legend columns=1, 
legend image code/.code={%
      \draw[#1] (0cm,-0.1cm) rectangle (0.3cm,0.3cm);}
}  
}
\begin{tikzpicture}

\begin{axis}[name = clartot, xticklabels = {2,3,4,5,6,7},ylabel={\% of Participants},yticklabels = {10,20,30,40}]   
\addplot[fill=black] coordinates{ (2,5) (3,17.5) (4,35) (5,17.5) (6,22.5) (7,2.5)};  
\addplot[fill=white] coordinates{ (2,12.5) (3,12.5) (4,43.75) (5,18.75) (6,6.25) (7,6.5)};  
\legend{Total sample, Core sample}
\end{axis}
\end{tikzpicture}
\caption{Usage of different levels of agreement in Dialogue 1} 
\label{fig:vald1}
\end{figure}
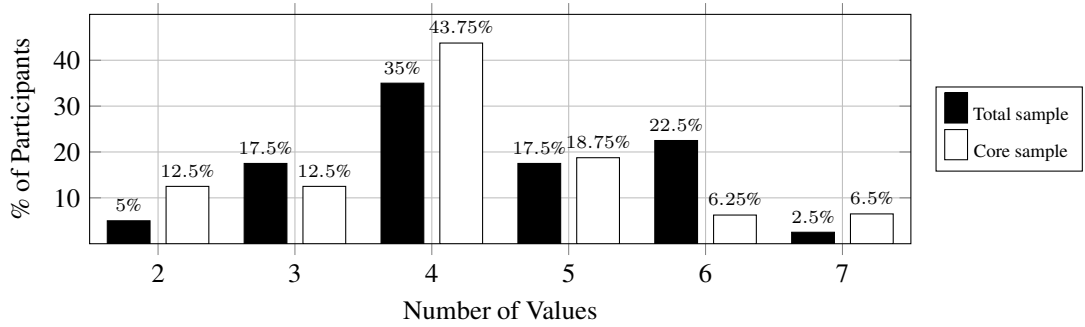

In addition to the postulates available in Figures \ref{fig:postulatetotal} and \ref{fig:postulatecore}, we have also considered a 
slight modification of the VAL$^n$ property. This postulate tells us how many distinct probabilities have been used in a given 
probability distribution. We can recall that from every user we obtain five distributions, one per every step of the dialogue. 
However, in some dialogue steps the number of used arguments is not sufficient to make any important observations 
w.r.t. the VAL$^n$ postulate. 
Consequently, rather than counting the distinct values in a single distribution, we count the distinct values used in all 
five distributions defined by a participant \footnote{In this analysis, we treat \textit{Don't know} the same as \textit{Neither Agree nor Disagree} due to the fact that they both evaluate to $3/6$.}. 
The results are presented in Figure \ref{fig:vald1} (see also the auxiliary data appendix at \cite{resources} for a more detailed table). We can observe that 
the vast majority of the participants did not use 3 or less values during the dialogue, independently of whether we consider the total or the core sample. 
The preferred options appear to be 4 and 6 (total sample) or 4 and 5 (core sample) values. These results are 
significantly different from what we would expect if the participants were providing us with random answers and show that 
what we obtained were in fact conscious choices\footnote{Under the assumption that every value is equally likely to be selected, 
if the participants were answering randomly throughout Dialogue 1 then the 
probability of them using exactly $7$ values throughout the dialogue is 
~$0.704$, probability of them using exactly $6$ values is ~$0.272$, 
of using $5$ values is ~$0.024$, of using $4$ is ~$0.00048$, and of $3$ or less is 
~$0.0000015$. Exact multinomial test shows there are significant differences between the obtained and the expected 
probabilities (obtained p--value was small enough that it was rounded up to $0$ by the used software).}. 
 
The above observations indicate that the two (or three) valued perception of acceptance exhibited by the classical Dung's semantics 
oversimplifies human modelling and might be insufficient to model the beliefs of the majority of participants. 
This also confirms the findings of \cite{Rahwan2011}.
We may also recall that rational, strict, protective, discharging and trusting postulates 
were related to classical conflict--free, admissible and complete semantics. While the performance of the preferential branch indicates 
that in general the statements accepted by the participants are not conflicting, the discharging and trusting results show 
that admissibility and completeness may be more problematic to satisfy. 
Thus, breaking down the classical semantics into 
smaller, separate properties, might give us a more detailed insight into the reasoning of the participants and provide 
more feedback in situations in which classical semantics are not satisfied.  
Consequently, more fine--grained approaches than Dung's semantics should be considered in order to provide a more 
accurate human modelling. 

\pgfplotstableread
[col sep=&,row sep=\\]
{
Postulate&Total&Core\\
Anticipating&0.56&0.4625\\
Binary&0.865&0.875\\
Coherent&0.32&0.325\\
Demanding&0.58&0.5625\\
Discharging&0.635&0.5125\\
Founded&0.28&0.3\\
Guarded&0.66&0.575\\
Involutary&0.26&0.275\\
Justifiable&0.13&0.15\\
Optimistic&0.275&0.3\\
Preferential&0.815&0.875\\
Protective&0.68&0.7875\\
Rational&0.765&0.8625\\
Restrained&0.67&0.7625\\
SemiFounded&0.665&0.575\\
SemiOptimistic&0.855&0.85\\
Strict&0.745&0.8375\\
Ternary&0.21&0.2\\
Trusting&0.565&0.475\\
}\postulatesoneauthor

\pgfplotstableread
[col sep=&,row sep=\\]
{
Postulate&Total&Core\\
Anticipating&0.36&0.425\\
Binary&0.865&0.875\\
Coherent&0.405&0.3875\\
Demanding&0.6&0.55\\
Discharging&0.4&0.45\\
Founded&0.165&0.2875\\
Guarded&0.465&0.5375\\
Involutary&0.28&0.2375\\
Justifiable&0.05&0.1\\
Optimistic&0.165&0.2875\\
Preferential&0.78&0.8375\\
Protective&0.66&0.7875\\
Rational&0.7&0.8125\\
Restrained&0.66&0.7875\\
SemiFounded&0.485&0.5375\\
SemiOptimistic&0.76&0.8\\
Strict&0.7&0.8125\\
Ternary&0.21&0.2\\
Trusting&0.365&0.4375\\
}\postulatesonedeclared

\pgfplotstableread
[col sep=&,row sep=\\]
{
Postulate&Total&Core\\
Anticipating&0.575&0.5\\
Binary&0.865&0.875\\
Coherent&0.32&0.325\\
Demanding&0.555&0.5625\\
Discharging&0.645&0.5375\\
Founded&0.31&0.3375\\
Guarded&0.67&0.6\\
Involutary&0.26&0.275\\
Justifiable&0.13&0.15\\
Optimistic&0.295&0.3375\\
Preferential&0.78&0.85\\
Protective&0.655&0.7625\\
Rational&0.735&0.8375\\
Restrained&0.655&0.7625\\
SemiFounded&0.68&0.6\\
SemiOptimistic&0.855&0.8625\\
Strict&0.72&0.825\\
Ternary&0.21&0.2\\
Trusting&0.58&0.5125\\
}\postulatesonecommon

\pgfplotstableread
[col sep=&,row sep=\\]
{
Postulate&Total&Core\\
Anticipating&0.43&0.5375\\
Binary&0.865&0.875\\
Coherent&0.355&0.3375\\
Demanding&0.65&0.575\\
Discharging&0.49&0.6\\
Founded&0.23&0.4125\\
Guarded&0.54&0.65\\
Involutary&0.25&0.225\\
Justifiable&0.05&0.1\\
Optimistic&0.22&0.4\\
Preferential&0.775&0.825\\
Protective&0.66&0.7875\\
Rational&0.695&0.8\\
Restrained&0.66&0.7875\\
SemiFounded&0.565&0.6625\\
SemiOptimistic&0.78&0.8375\\
Strict&0.695&0.8\\
Ternary&0.21&0.2\\
Trusting&0.435&0.55\\
}\postulatesoneexpanded
 
%

\clearpage
\subsubsection{Dialogue 2}   

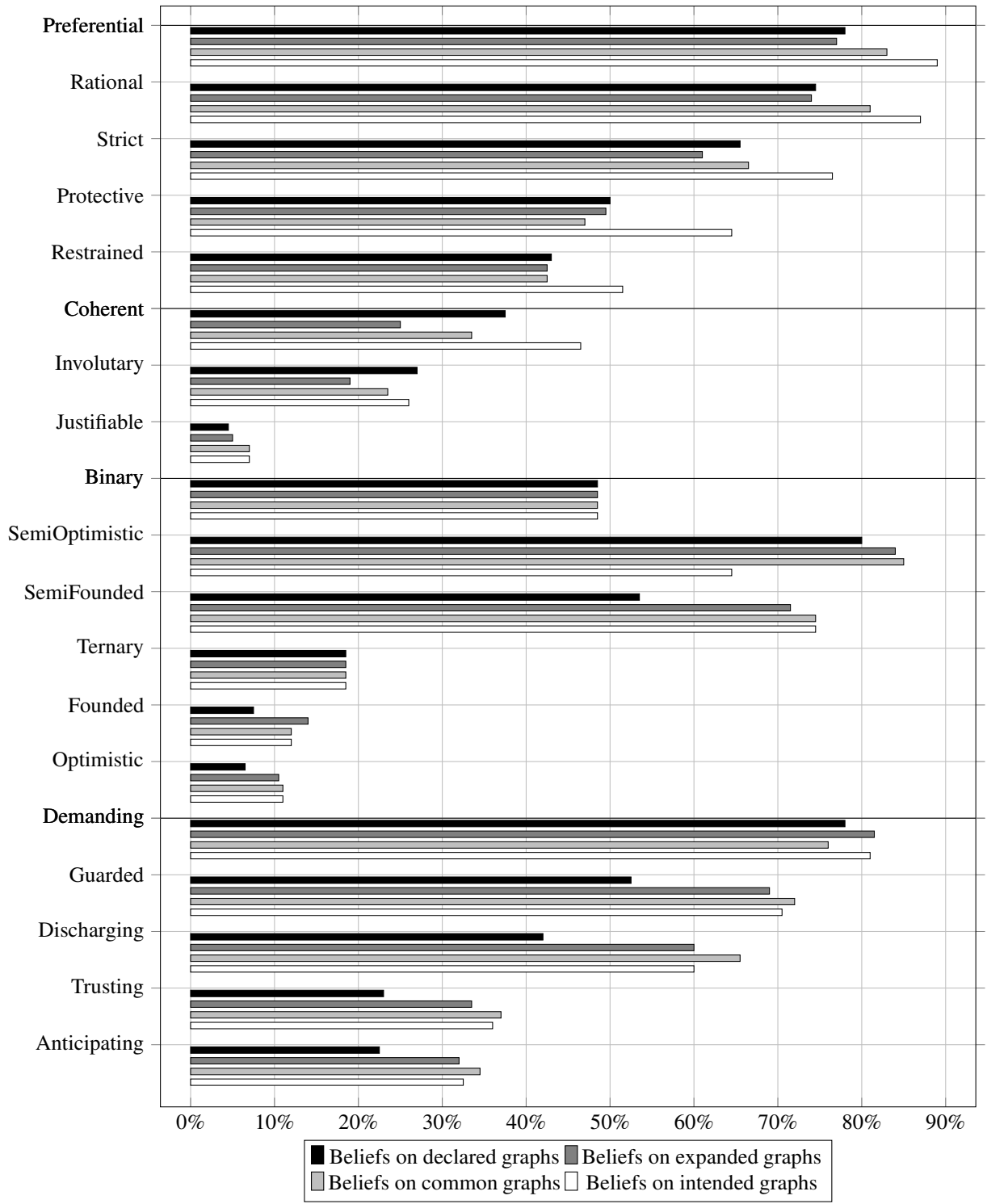
\begin{figure}[p]
\centering
\begin{tikzpicture} 
\begin{axis}[xbar, bar width=3pt,grid=both, symbolic y coords={Preferential, Rational, Strict,  Protective,  Restrained, Coherent, Involutary, Justifiable, 
Binary, SemiOptimistic, SemiFounded, Ternary, Founded, Optimistic,  Demanding, Guarded,  Discharging, Trusting,  Anticipating},
ytick = {Preferential, Rational, Strict,  Protective,  Restrained, Coherent, Involutary, Justifiable, 
Binary, SemiOptimistic, SemiFounded, Ternary, Founded, Optimistic,  Demanding, Guarded,  Discharging, Trusting,  Anticipating},  
yticklabel style = {yshift=10pt, font=\normalsize},ytick style = {yshift=10pt}, 
enlargelimits = true,
enlarge y limits = 0.04,
enlarge x limits =0.04,
y grid style = {yshift = 10pt},  extra y ticks={Preferential, Coherent, Binary, Demanding}, extra y tick style={grid=major, grid style={black}},   
 width=0.9\textwidth, height=0.9\textheight, y dir = reverse, reverse legend, 
legend style={at={(0.5,-0.03)},anchor=north}, legend columns=2, 
 xmin = 0.0, xmax = 0.9, xtick = {0,0.1,0.2,0.3,0.4,0.5,0.6,0.7,0.8,0.9},
xticklabels = {0\%,10\%,20\%,30\%,40\%,50\%,60\%,70\%,80\%,90\%},
legend image code/.code={%
      \draw[#1] (0cm,-0.1cm) rectangle (0.2cm,0.2cm);}]
 
\addplot[fill=white] table [y=Postulate, x=Total] {\postulatestwoauthor};  

\addplot[fill=lightgray] table [y=Postulate, x=Total] {\postulatestwocommon}; 

\addplot[fill=gray] table [y=Postulate, x=Total] {\postulatestwoexpanded};  

\addplot[fill=black] table [y=Postulate, x=Total]  {\postulatestwodeclared};  

\legend{Beliefs on intended graphs, Beliefs on common graphs, Beliefs on expanded graphs, Beliefs on declared graphs} 
\end{axis} 
\end{tikzpicture}
\caption{Postulate satisfaction on the total sample in Dialogue 2}
\label{fig:postulatetotal2}
\end{figure}

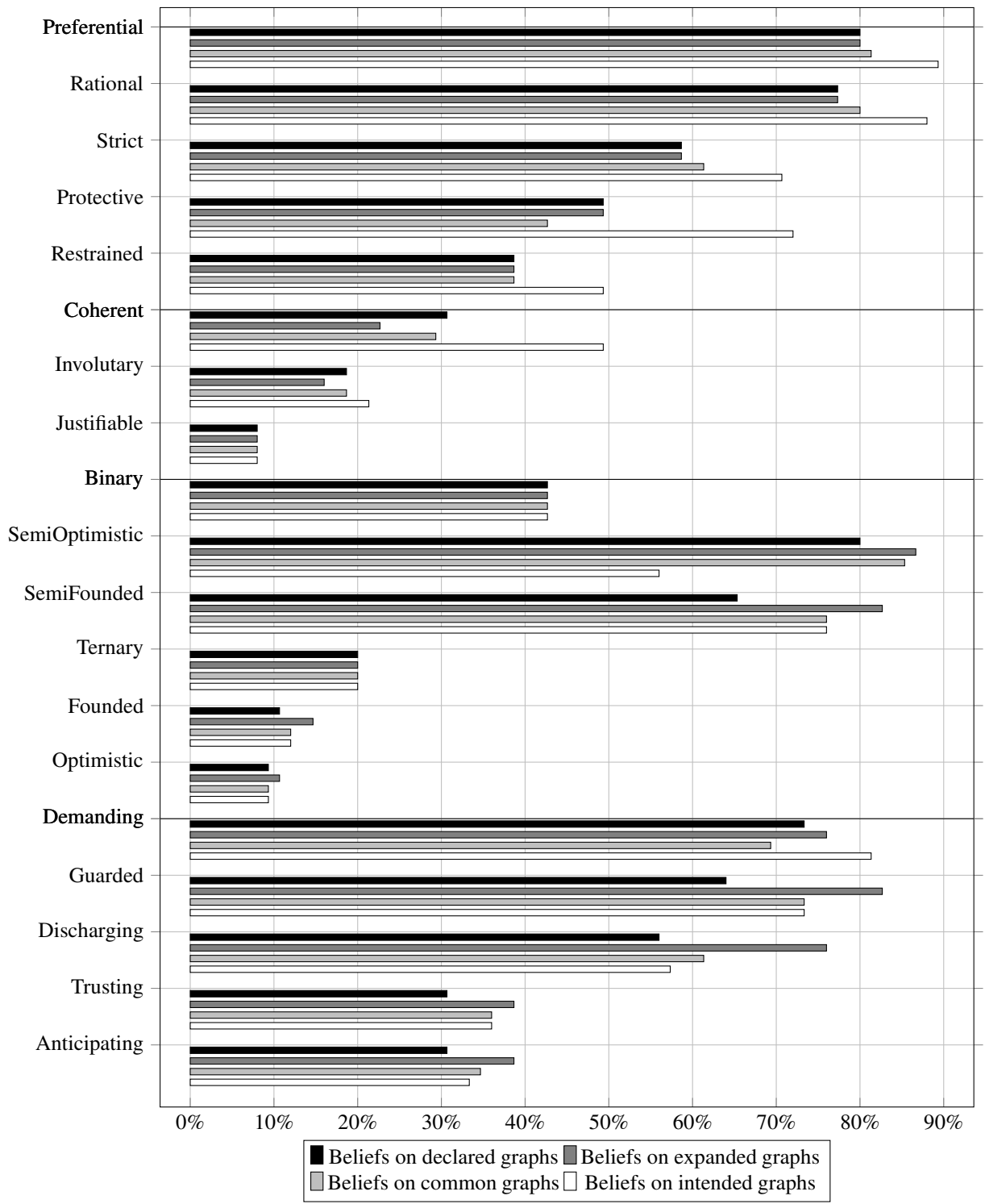
\begin{figure}[p]
\centering
\begin{tikzpicture} 
\begin{axis}[xbar, bar width=3pt,grid=both, symbolic y coords={Preferential, Rational, Strict,  Protective,  Restrained, Coherent, Involutary, Justifiable, 
Binary, SemiOptimistic, SemiFounded, Ternary, Founded, Optimistic,  Demanding, Guarded,  Discharging, Trusting,  Anticipating},
ytick = {Preferential, Rational, Strict,  Protective,  Restrained, Coherent, Involutary, Justifiable, 
Binary, SemiOptimistic, SemiFounded, Ternary, Founded, Optimistic,  Demanding, Guarded,  Discharging, Trusting,  Anticipating},  
yticklabel style = {yshift=10pt, font=\normalsize},ytick style = {yshift=10pt}, 
enlargelimits = true,
enlarge y limits = 0.04,
enlarge x limits =0.04,
y grid style = {yshift = 10pt},  extra y ticks={Preferential, Coherent, Binary, Demanding}, extra y tick style={grid=major, grid style={black}},   
 width=0.9\textwidth, height=0.9\textheight, y dir = reverse, reverse legend, 
legend style={at={(0.5,-0.03)},anchor=north}, legend columns=2, 
 xmin = 0.0, xmax = 0.9, xtick = {0,0.1,0.2,0.3,0.4,0.5,0.6,0.7,0.8,0.9},
xticklabels = {0\%,10\%,20\%,30\%,40\%,50\%,60\%,70\%,80\%,90\%},
legend image code/.code={%
      \draw[#1] (0cm,-0.1cm) rectangle (0.2cm,0.2cm);}]

\addplot[fill=white] table [y=Postulate, x=Core] {\postulatestwoauthor};  

\addplot[fill=lightgray] table [y=Postulate, x=Core] {\postulatestwocommon}; 

\addplot[fill=gray] table [y=Postulate, x=Core] {\postulatestwoexpanded};  

\addplot[fill=black] table [y=Postulate, x=Core]  {\postulatestwodeclared};  

\legend{Beliefs on intended graphs, Beliefs on common graphs, Beliefs on expanded graphs, Beliefs on declared graphs} 
\end{axis}
\end{tikzpicture}
\caption{Postulate satisfaction on core sample in Dialogue 2}
\label{fig:postulatecore2}
\end{figure}

In Figures \ref{fig:postulatetotal2} and \ref{fig:postulatecore2} we can observe the average adherence rates 
to the epistemic postulates 
in the total and core samples in Dialogue 2. Let us focus on the total sample first. 
Some of the observations we have made in the case of Dialogue 1 carry over to Dialogue 2. 
Similarly as in the first dialogue, the results obtained for the declared, expanded and common graphs appear to be similar in the 
case of preferential family (see Table \ref{tab:postulatewilcox2}). 
However, unlike in the previous case, the behaviour of the intended graph on this family appears to 
be more distinct from the rest. Fortunately, with the exception of the preferential postulates, 
the similarities between the results on the common and intended graphs, 
present in first dialogue, 
hold here as well. This behaviour may indicate that the intended graphs were too conservative in terms of the considered attacks w.r.t. 
the common graphs (see also Section \ref{sec:dial2usergraph}).
We can also note that similarly as in the first dialogue, 
the adherence to the rational family of postulates decreases the more specialized the property is. Again, the justifiable 
postulate is barely adhered to.

The differences between the results on the declared graph 
and other graphs in the case of explanatory and value--based families are more visible 
than in the first dialogue (see also Table \ref{tab:postulatewilcox1} in Section \ref{stats:postulates1} and Table \ref{tab:postulatewilcox2} in Section \ref{stats:postulates2}).
In principle, the largest number of differences can be observed between the intended and declared graphs, 
and exceeds our findings in the first dialogue. 
Additionally, many of the postulates we have considered do not perform as well as in the first dialogue in terms of the declared graphs. 
Nevertheless, the rational, semi--optimistic, semi--founded, guarded, discharging and particularly demanding properties appear to achieve better
results than in the previous dialogue. 
We believe some of this behaviour (including also the lower performance of the strict and restrained postulates) can be explained with the behaviour 
of the binary property. The low satisfaction rate of this particular postulate means that the participants had problems in deciding 
whether to believe or disbelieve the presented statements using their own knowledge and the information found in the dialogue. 
This dialogue presented more specialized information than the first dialogue and without access to, for example, Wikipedia, the participants had problems verifying 
the statements. The presence of the undecided assignments appears to have benefited some of the postulates 
that are easily satisfied with neutral distributions. We can observe that all of the properties that have experienced an 
increase in adherence belong to this group. 
At the same time the undecided assignments harmed those postulates
that try to force more decisiveness, such as strict or protective properties, or force belief or disbelief in the face of undecided attackers (e.g. restrained 
or anticipating postulates). Consequently, in addition to the postulates vulnerable to participants who do not share all of their knowledge 
(see the analysis of Dialogue 1), we can also
identify those that are vulnerable or benefiting from the indecisive participants. 

We can now consider the results obtained for the core sample. The average adherence rates of the postulates in the core sample are quite close to the ones 
from the total sample. The results obtained for different graphs also appear to be more closely connected  (see Table  \ref{tab:postulatewilcox2} in Section \ref{stats:postulates2}). 
The most visible changes can be observed in the case of the explanatory family of postulates 
(with the exception of the demanding property), which have all increased their values. The results w.r.t. the expanded graphs 
on the guarded and discharging postulates become even more distinct from the declared graphs in terms of the average difference
of participant responses between these graphs. It therefore appears that the 
participants were more forthcoming with their personal knowledge, in particular with the additional reasons against believing certain statements. 
We therefore again identify the properties more vulnerable to such information.

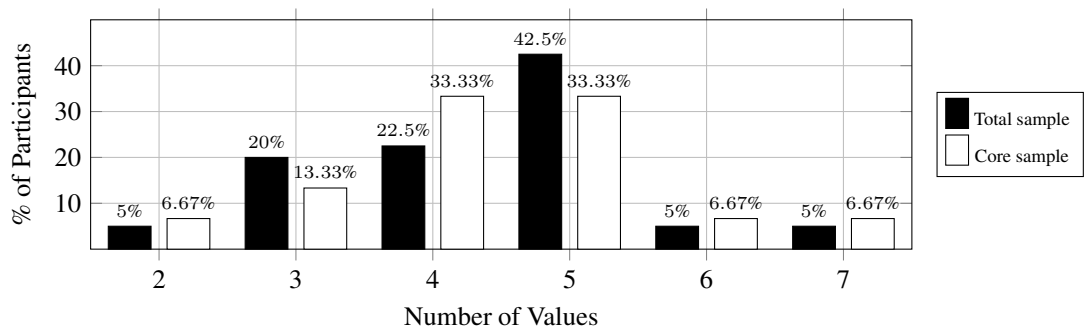
\begin{figure}[!ht]
\centering
\pgfplotsset{compat=1.9,
every axis/.style={width=0.75\textwidth, height=0.28\textwidth, grid=both, 
yticklabel style={
        /pgf/number format/fixed,
        /pgf/number format/precision=4
}, 
bar width = 16pt,
scaled y ticks=false, ybar=6pt, ymin=0, ymax=50, nodes near coords = {\pgfmathprintnumber\pgfplotspointmeta\%}, 
every node near coord/.append style={font=\scriptsize}, 
xtick = {2,3,4,5,6,7}, ytick = {10,20,30,40},
xlabel ={Number of Values},
legend style={at={(1.03,0.5)},anchor=west, font=\scriptsize}, legend columns=1, 
legend image code/.code={%
      \draw[#1] (0cm,-0.1cm) rectangle (0.3cm,0.3cm);}
}  
}
\clearpage
\begin{tikzpicture} 

\begin{axis}[name = clartot,  xticklabels = {2,3,4,5,6,7},ylabel={\% of Participants},yticklabels = {10,20,30,40},]   
\addplot[fill=black] coordinates{ (2,5) (3,20) (4,22.5) (5,42.5) (6,5) (7,5)};  
\addplot[fill=white] coordinates{ (2,6.67) (3,13.33) (4,33.33) (5,33.33) (6,6.67) (7,6.67)}; 
\legend{Total sample, Core sample} 
\end{axis}
\end{tikzpicture}
\caption{Usage of different levels of agreement in Dialogue 2} 
\label{fig:vald2}
\end{figure}

In Figure \ref{fig:vald2} we can observe the number of participants who have chosen a given number of distinct 
probabilities throughout the dialogue
\footnote{The presented results treated \textit{Don't know} as the same as \textit{Neither Agree nor Disagree}.}
(see also the auxiliary data appendix at \cite{resources} for a more detailed table).
Similarly to Dialogue 1, the majority of participants choose 4 or more values, though in this case 
5 values appear more often in the total sample than 4 values, and they are picked equally often in the core sample. 
This increase may be related to a wider use of the \textit{Neither Agree nor Disagree} and \textit{Don't know} values, which 
is visible in the difference of the adherence to the binary postulate between the dialogues. 
Also in this case it is easy to see that the selection of the values by the participants is clearly not random\footnote{Under the assumption that every value is equally likely to be selected, 
if the participants were answering randomly throughout Dialogue 2 then the 
probability of them using exactly $7$ values throughout the dialogue is 
~$0.876$, probability of them using exactly $6$ values is ~$0.121$, 
of using $5$ values is ~$0.003$, of using $4$ is ~$0.00002$, and of $3$ or less is 
~$0.0000000095$. 
Exact multinomial test shows there are significant differences between the obtained and the expected 
probabilities (obtained p--value was small enough that it was rounded up to $0$ by the used software).}
 and thus the tendency 
of the participants to choose more than 3 values is their conscious choice. 
%
%
%

\pgfplotstableread
[col sep=&,row sep=\\]
{
Postulate&Total&Core\\
Anticipating&0.325&0.333333333\\
Binary&0.485&0.426666667\\
Coherent&0.465&0.493333333\\
Demanding&0.81&0.813333333\\
Discharging&0.6&0.573333333\\
Founded&0.12&0.12\\
Guarded&0.705&0.733333333\\
Involutary&0.26&0.213333333\\
Justifiable&0.07&0.08\\
Optimistic&0.11&0.093333333\\
Preferential&0.89&0.893333333\\
Protective&0.645&0.72\\
Rational&0.87&0.88\\
Restrained&0.515&0.493333333\\
SemiFounded&0.745&0.76\\
SemiOptimistic&0.645&0.56\\
Strict&0.765&0.706666667\\
Ternary&0.185&0.2\\
Trusting&0.36&0.36\\
}\postulatestwoauthor

\pgfplotstableread
[col sep=&,row sep=\\]
{
Postulate&Total&Core\\
Anticipating&0.225&0.306666667\\
Binary&0.485&0.426666667\\
Coherent&0.375&0.306666667\\
Demanding&0.78&0.733333333\\
Discharging&0.42&0.56\\
Founded&0.075&0.106666667\\
Guarded&0.525&0.64\\
Involutary&0.27&0.186666667\\
Justifiable&0.045&0.08\\
Optimistic&0.065&0.093333333\\
Preferential&0.78&0.8\\
Protective&0.5&0.493333333\\
Rational&0.745&0.773333333\\
Restrained&0.43&0.386666667\\
SemiFounded&0.535&0.653333333\\
SemiOptimistic&0.8&0.8\\
Strict&0.655&0.586666667\\
Ternary&0.185&0.2\\
Trusting&0.23&0.306666667\\
}\postulatestwodeclared

\pgfplotstableread
[col sep=&,row sep=\\]
{
Postulate&Total&Core\\
Anticipating&0.345&0.346666667\\
Binary&0.485&0.426666667\\
Coherent&0.335&0.293333333\\
Demanding&0.76&0.693333333\\
Discharging&0.655&0.613333333\\
Founded&0.12&0.12\\
Guarded&0.72&0.733333333\\
Involutary&0.235&0.186666667\\
Justifiable&0.07&0.08\\
Optimistic&0.11&0.093333333\\
Preferential&0.83&0.813333333\\
Protective&0.47&0.426666667\\
Rational&0.81&0.8\\
Restrained&0.425&0.386666667\\
SemiFounded&0.745&0.76\\
SemiOptimistic&0.85&0.853333333\\
Strict&0.665&0.613333333\\
Ternary&0.185&0.2\\
Trusting&0.37&0.36\\
}\postulatestwocommon

\pgfplotstableread
[col sep=&,row sep=\\]
{
Postulate&Total&Core\\
Anticipating&0.32&0.386666667\\
Binary&0.485&0.426666667\\
Coherent&0.25&0.226666667\\
Demanding&0.815&0.76\\
Discharging&0.6&0.76\\
Founded&0.14&0.146666667\\
Guarded&0.69&0.826666667\\
Involutary&0.19&0.16\\
Justifiable&0.05&0.08\\
Optimistic&0.105&0.106666667\\
Preferential&0.77&0.8\\
Protective&0.495&0.493333333\\
Rational&0.74&0.773333333\\
Restrained&0.425&0.386666667\\
SemiFounded&0.715&0.826666667\\
SemiOptimistic&0.84&0.866666667\\
Strict&0.61&0.586666667\\
Ternary&0.185&0.2\\
Trusting&0.335&0.386666667\\
}\postulatestwoexpanded

\clearpage
\subsection{Statements vs Relations}
\label{sec:relcorel}

Relations in argumentation frameworks are often defined in terms of the argument structure we are dealing with. 
Consequently, they are 
seen as secondary to arguments, and as such their acceptability is rarely considered. 
One of the exceptions is the attack--based 
semantics for argumentation frameworks, in which the status assigned to a relation corresponds to the status 
assigned to its source \cite{attacksem} and 
vice versa. 
The same behaviour is replicated by other frameworks allowing the attacks to appear in extensions 
\cite{BaroniCGG2011,CohenGGS16} if we 
limit ourselves to non--recursive graphs. 
In this section we would like to verify this view by analyzing the answers to the agreement tasks and relation tasks. 

We will perform our analysis in the following manner. 
For every relation of a given type, by which we understand \textit{A good reason against}, 
\textit{A somewhat good reason against}, \textit{A good reason for}, \textit{A somewhat good reason for} and \textit{Somehow related, but can't say how},  
we create a distribution based on the levels of agreement assigned to the statements carrying them out. Moreover, for every possible level of agreement, 
we gather the types of relations that the statements assigned a given level carry out. This allows us to observe whether by knowing the value 
ascribed to a relation we can guess the value assigned to its source and vice versa. In particular, this allows us to verify whether the connection 
between the values assigned to arguments and relations is as strong as we might think based on some of the research in abstract argumentation. 
Therefore, what we would expect to obtain is either 
\begin{inparaenum}[\itshape 1\upshape)]
\item relations considered \enquote{somewhat good} to be carried out primarily by the statements that the 
participants disagree with and \enquote{good} relations to be sourced primarily by the statements that the participants agree with, or
\item  relations considered \enquote{somewhat good} to be carried out primarily by the statements that have low levels of agreement or disagreement 
and relations considered \enquote{good} to be sourced primarily by the statements that have high levels of agreement or disagreement. 
\end{inparaenum}
In both cases we expect the \textit{Somewhat related, but can't say how} edges to be primarily associated with statements of an  
undecided or unknown status. 
Our analysis will be carried out on the original as well as pooled data, i.e. one in which certain categories can be merged. Both relation types and agreement levels 
can be grouped by strength or polarity:
\begin{itemize}
\item Agreement grouping:
\begin{itemize}
\item By strength: we group the agreement levels according to how strongly believed or disbelieved they are. We 
distinguish between \textit{Strong Belief} (created from \textit{Strongly Agree} and \textit{Strongly Disagree}), \textit{Moderate Belief} (combining \textit{Agree} and \textit{Disagree}), \textit{Weak Belief} (grouping \textit{Somewhat Agree} and 
\textit{Somewhat Disagree}), and \textit{Neither} (created from \textit{Neither Agree nor Disagree} and \textit{Don't Know}) 
categories.

\item By polarity:  we group the agreement levels according to whether they represent belief or disbelief. We distinguish 
between \textit{Believed} (grouping the \textit{Somewhat Agree},  \textit{Agree} and  \textit{Strongly Agree} levels), 
 \textit{Disbelieved} (created from \textit{Somewhat Disagree},  \textit{Disagree} and  \textit{Strongly Disagree}), 
and  \textit{Neither} (formed from  \textit{Neither Agree nor Disagree} and  \textit{Don't Know}) categories.

\end{itemize}
\item Relation grouping:

\begin{itemize}
\item By strength: we group relations according to whether they are good or somewhat good reasons. We distinguish between 
 \textit{Strong Relations} (formed from \textit{A good reason for} and  \textit{A good reason against}), 
 \textit{Normal Relations} (grouping  \textit{A somewhat good reason for} and  \textit{A somewhat good reason against}), 
and   \textit{Dependencies} (representing  \textit{Somehow related, but can't say how}).

\item By polarity: we group relations according to whether they are positive or negative. We distinguish between 
 \textit{Attacks} (formed from \textit{A good reason against} and and  \textit{A somewhat good reason against}), 
 \textit{Support} (grouping  \textit{A good for} and  \textit{A somewhat good reason for}), 
and  \textit{Dependencies} (representing  \textit{Somehow related, but can't say how}).

\end{itemize}
\end{itemize}

We consider pooling data in order to be able to more clearly investigate certain trends that may become more apparent with 
fewer categories.  
  
\subsubsection{Dialogue 1} 

\paragraph{Effect of a Relation on its Source} \hfill
 
\begin{figure}[!ht] 
\centering
\pgfplotsset{every axis/.style={ybar, bar width=4pt,grid=both, , height=0.32\textwidth, width=0.40\textwidth,align =center,
x label style ={at={(axis description cs:0.5,0.04)}},
y label style ={at={(axis description cs:0.06,0.5)}}, 
symbolic x coords={Strongly agree, Agree, Somewhat Agree, Neither Agree nor Disagree, Don't know, Somewhat Disagree, Disagree, Strongly disagree},
xtick={Strongly agree, Agree, Somewhat Agree, Neither Agree nor Disagree, Don't know, Somewhat Disagree, Disagree, Strongly disagree}, 
xticklabels={SA, A, SoA, NAD, DK, SoD, D, SD}, 
enlargelimits=true,  ymin = 0, ymax = 62,  ytick = {0,10,20,30,40,50,60},
xticklabel style={font = \small, align=center, rotate=60,anchor=east},
yticklabels = {0\%,10\%,20\%,30\%,40\%,50\%,60\%},
legend style={at={(0.0,1.3)},anchor=south, font=\scriptsize}, legend columns=2, 
legend image code/.code={%
      \draw[#1] (0cm,-0.1cm) rectangle (0.2cm,0.2cm);
    },}}
\begin{tikzpicture}

\begin{axis}[name=sap,title = {A good reason against}]
\addplot[fill=black] table [y=Strong attack, x=TYPE] {\donetotbyrel};   
\addplot[fill = white] table [y=Strong attack, x=TYPE] {\donecorbyrel};  
\end{axis} 

\begin{axis}[name=ssp, title = {A good reason for} , yticklabels={,,,,},  at=(sap.right of south east)]
\addplot[fill=black] table [y=Strong support, x=TYPE] {\donetotbyrel};   
\addplot[fill = white] table [y=Strong support, x=TYPE] {\donecorbyrel};   
\legend{Total sample, Core sample} 
\end{axis} 

\begin{axis}[name=attp, title = {A somewhat good \\reason against},   at=(sap.below south west), anchor = above north]
\addplot[fill=black] table [y=Attack, x=TYPE] {\donetotbyrel};   
\addplot[fill = white] table [y=Attack, x=TYPE] {\donecorbyrel};   

\end{axis} 

\begin{axis}[name=supp, title = {A somewhat good\\ reason for},yticklabels={,,,,},   at=(attp.right of south east)]
\addplot[fill=black] table [y=Support, x=TYPE] {\donetotbyrel};   
\addplot[fill = white] table [y=Support, x=TYPE] {\donecorbyrel};   
\end{axis} 

\begin{axis}[name=depp, title = {Somewhat related,\\ but can't say how},at=(supp.right of south east), yticklabels={,,,}]
\addplot[fill=black] table [y=Dependency, x=TYPE] {\donetotbyrel};   
\addplot[fill = white] table [y=Dependency, x=TYPE] {\donecorbyrel};   
\end{axis}  
\end{tikzpicture}
\caption{The levels of agreements assigned to the sources of relations of a given type in Dialogue 1. We use the following abbreviations: 
Strongly Agree (SA), Agree (A), Somewhat Agree (SoA), Neither Agree nor Disagree (NAD), Somewhat Disagree (SoD), 
Disagree (D), Strongly Disagree (SD), Don't Know (DK)}
\label{fig:d1totbyrel}
\end{figure}
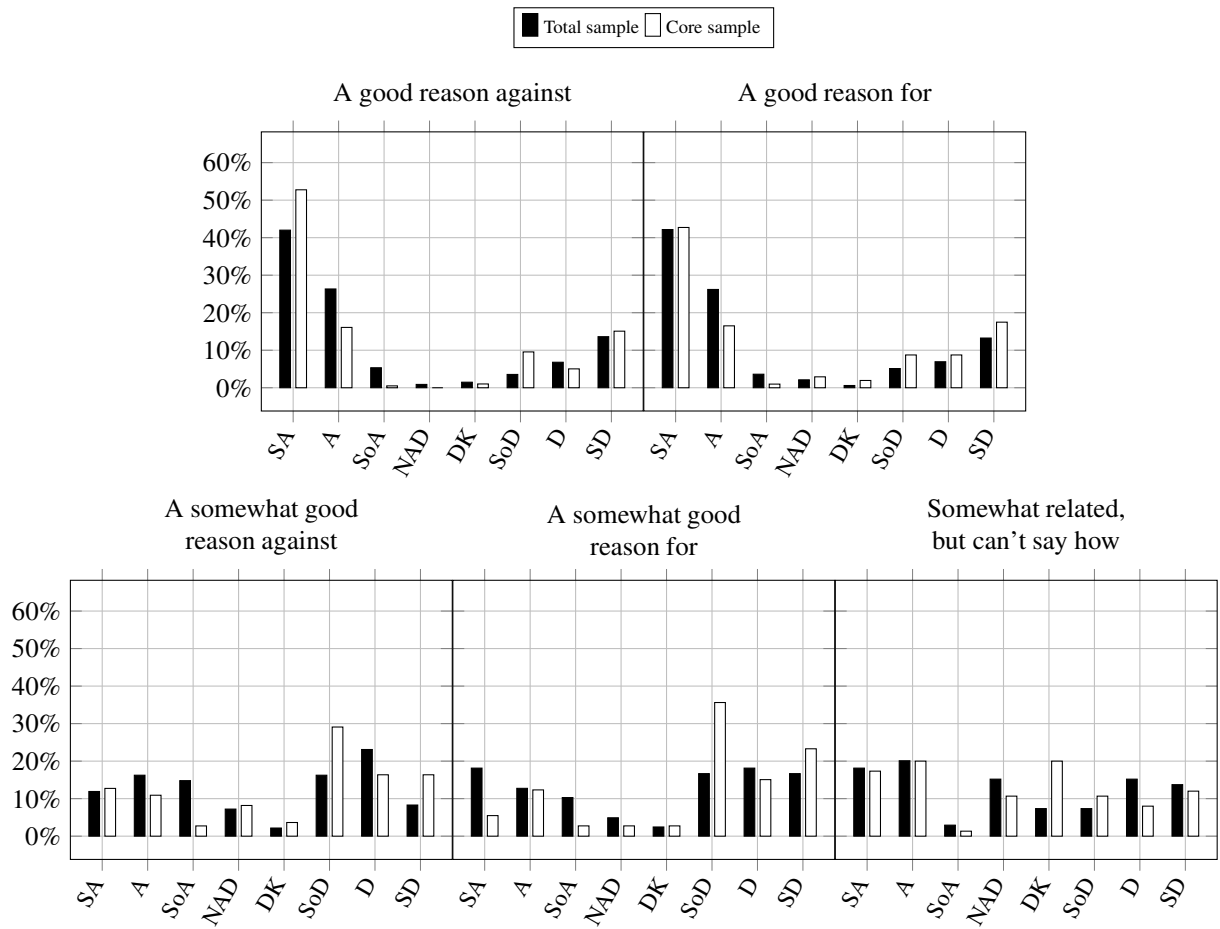

In Figure \ref{fig:d1totbyrel} we can observe how the levels of agreements of the sources of relations of a given kind are distributed. 
The results show that this agreement distribution is in principle dependent on the nature of the relation in question \footnote{G--test for independence 
yields G--value $317.77$ 
with $28$ degrees of freedom and p--value less than $2.2 \times 10^{-16}$. 
This result was obtained using the library Deducer (likelihood.test function) in R.}.  
Nevertheless, the distributions visible in Figure \ref{fig:d1totbyrel} 
do show certain similarities and we carry out additional pairwise--analysis. 

We can observe that the distributions of relations marked as being good reasons are relatively similar, as also supported 
by the results in Table \ref{tab:d1indeprel} in Section \ref{stats:relsource1}. 
In both cases, independently of whether 
we choose the core or total sample, the dominating belief assigned to the source is \textit{Strongly Agree}, 
with \textit{Agree} and \textit{Strongly Disagree} next in line. 

Although some similarities can also be observed between the relations considered as somewhat good reasons 
for or against a given statement, they are more visible in case of the core sample (Table \ref{tab:d1indeprel} in Section \ref{stats:relsource1}). 
In the total sample, the proportions of source agreement levels are in principle distinct. 
It appears that more participants 
have chosen disagreement rather than agreement types of sources, though it does not appear that one particular belief is 
substantially more 
common than the other in the total sample. In the core sample, the \textit{Somewhat Disagree} option appears to be chosen 
slightly more often. No particular pattern appears to hold for the \textit{Somewhat related, but can't say how} relation.  

Although none of the agreement distributions is random (see Table \ref{tab:d1goodnessrel} in Section \ref{stats:relsource1}), we can observe that 
the \enquote{weaker} the relation, the more blurred the picture becomes. Consequently, if we were given 
a relation, informed of its type and then asked to guess what should the belief in its source be, our best chances 
of answering correctly 
are in the case of edges marked as good reasons. 

\pgfplotstableread
[col sep=&,row sep=\\]
{
TYPE&Strong attack&Attack&Strong support&Support&Dependency\\
Strongly agree&42.01&11.91&42.17&18.14&18.14\\
Agree&26.33&16.25&26.2&12.75&20.1\\
Somewhat Agree&5.33&14.8&3.61&10.29&2.94\\
Neither Agree nor Disagree&0.89&7.22&2.11&4.9&15.2\\
Disagree&6.8&23.1&6.93&18.14&15.2\\
Somewhat Disagree&3.55&16.25&5.12&16.67&7.35\\
Strongly disagree&13.61&8.3&13.25&16.67&13.73\\
Don't know&1.48&2.17&0.6&2.45&7.35\\
}\donetotbyrel

\pgfplotstableread
[col sep=&,row sep=\\]
{
TYPE&Strong attack&Attack&Strong support&Support&Dependency\\
Strongly agree&52.76&12.73&42.72&5.48&17.33\\
Agree&16.08&10.91&16.5&12.33&20\\
Somewhat Agree&0.5&2.73&0.97&2.74&1.33\\
Neither Agree nor Disagree&0&8.18&2.91&2.74&10.67\\
Disagree&5.03&16.36&8.74&15.07&8\\
Somewhat Disagree&9.55&29.09&8.74&35.62&10.67\\
Strongly disagree&15.08&16.36&17.48&23.29&12\\
Don't know&1.01&3.64&1.94&2.74&20\\
}\donecorbyrel

\pgfplotstableread
[col sep=&,row sep=\\]
{
TYPE&Strongly agree&Agree&Somewhat Agree&Neither Agree nor Disagree&Disagree&Somewhat Disagree&Strongly disagree&Don't know\\
Strong attack&36.5&30.9&18.37&4.23&12.92&9.76&26.29&15.15\\
Attack&8.48&15.63&41.84&28.17&35.96&36.59&13.14&18.18\\
Strong support&35.99&30.21&12.24&9.86&12.92&13.82&25.14&6.06\\
Support&9.51&9.03&21.43&14.08&20.79&27.64&19.43&15.15\\
Dependency&9.51&14.24&6.12&43.66&17.42&12.2&16&45.45\\
}\donetotbyarg

\pgfplotstableread
[col sep=&,row sep=\\]
{
TYPE&Strongly agree&Agree&Somewhat Agree&Neither Agree nor Disagree&Disagree&Somewhat Disagree&Strongly disagree&Don't know\\
Strong attack&58.33&37.65&12.5&0&18.52&20.21&32.61&8\\
Attack&7.78&14.12&37.5&40.91&33.33&34.04&19.57&16\\
Strong support&24.44&20&12.5&13.64&16.67&9.57&19.57&8\\
Support&2.22&10.59&25&9.09&20.37&27.66&18.48&8\\
Dependency&7.22&17.65&12.5&36.36&11.11&8.51&9.78&60\\
}\donecorbyarg

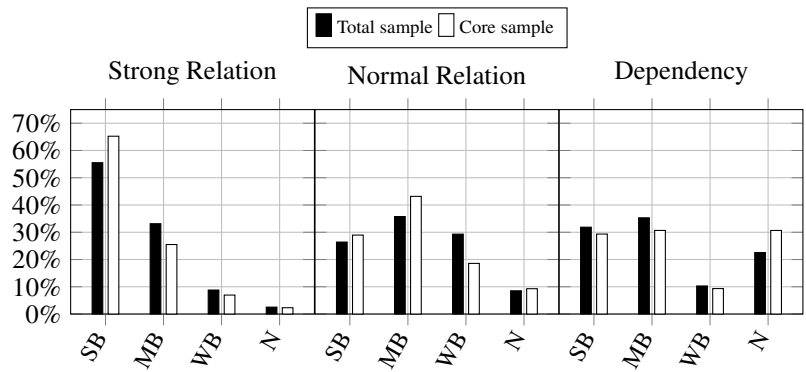
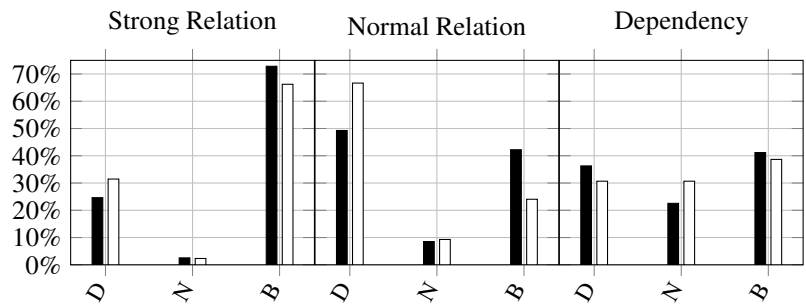
\begin{figure}[!ht] 
\centering
  
\begin{subfigure}[b]{\textwidth}
\centering
\pgfplotsset{every axis/.style={ybar, bar width=4pt,grid=both,height=0.26\textwidth, width=0.29\textwidth, align =center,
x label style ={at={(axis description cs:0.5,0.04)}},
y label style ={at={(axis description cs:0.06,0.5)}}, 
symbolic x coords={Strong Belief, Moderate Belief, Weak Belief, Neither},
xtick={Strong Belief, Moderate Belief, Weak Belief, Neither}, 
xticklabels={SB, MB, WB, N}, 
enlarge x limits=0.2,  ymin = 0, ymax = 75,  ytick = {0,10,20,30,40,50,60,70},
xticklabel style={font = \small, align=center, rotate=60,anchor=east},
yticklabels = {0\%,10\%,20\%,30\%,40\%,50\%,60\%,70\%},
legend style={at={(-0.5,1.3)},anchor=south, font=\scriptsize}, legend columns=2, 
legend image code/.code={%
      \draw[#1] (0cm,-0.1cm) rectangle (0.2cm,0.2cm);
    },}} 
\begin{tikzpicture}
\centering
\begin{axis}[name=strrel,title = {Strong Relation}]
\addplot[fill=black] table [y=Strong Relation, x=Belief] {\strpooldialone};   
\addplot[fill = white] table [y=Strong Relation, x=Belief] {\strpoolcoredialone};  
\end{axis} 

\begin{axis}[name=normrel, title = {Normal Relation} , yticklabels={,,,,},  at=(strrel.right of south east)]

\addplot[fill=black] table [y=Normal Relation, x=Belief] {\strpooldialone};   
\addplot[fill = white] table [y=Normal Relation, x=Belief] {\strpoolcoredialone};  

\end{axis} 

\begin{axis}[name=dep, title = {Dependency}, yticklabels={,,,,},  at=(normrel.right of south east)]
\addplot[fill=black] table [y=Dependency, x=Belief] {\strpooldialone};   
\addplot[fill = white] table [y=Dependency, x=Belief] {\strpoolcoredialone};  \legend{Total sample, Core sample} 
\end{axis} 
\end{tikzpicture}
\caption{Arguments and relations pooled by strength}
\label{fig:strpoold1totbyrel}
\end{subfigure}
\begin{subfigure}[b]{\textwidth}
\centering
\pgfplotsset{every axis/.style={ybar, bar width=4pt,grid=both,height=0.26\textwidth, width=0.29\textwidth, align =center,
x label style ={at={(axis description cs:0.5,0.04)}},
y label style ={at={(axis description cs:0.06,0.5)}}, 
symbolic x coords={Disbelieved, Neither, Believed},
xtick={Disbelieved, Neither, Believed}, 
xticklabels={D, N,B}, 
enlarge x limits=0.2,  ymin = 0, ymax = 75,  ytick = {0,10,20,30,40,50,60,70},
xticklabel style={font = \small, align=center, rotate=60,anchor=east},
yticklabels = {0\%,10\%,20\%,30\%,40\%,50\%,60\%,70\%},
legend style={at={(0.5,1.3)},anchor=south, font=\scriptsize}, legend columns=2, 
legend image code/.code={%
      \draw[#1] (0cm,-0.1cm) rectangle (0.2cm,0.2cm);
    },}}
\begin{tikzpicture}

\begin{axis}[name=strrel,title = {Strong Relation}]
\addplot[fill=black] table [y=Strong Relation, x=Belief] {\strpolpooldialone};   
\addplot[fill = white] table [y=Strong Relation, x=Belief] {\strpolpoolcoredialone};  
\end{axis} 

\begin{axis}[name=normrel, title = {Normal Relation} , yticklabels={,,,,},  at=(strrel.right of south east)]

\addplot[fill=black] table [y=Normal Relation, x=Belief] {\strpolpooldialone};   
\addplot[fill = white] table [y=Normal Relation, x=Belief] {\strpolpoolcoredialone};  
\end{axis} 

\begin{axis}[name=dep, title = {Dependency}, yticklabels={,,,,},  at=(normrel.right of south east)]
\addplot[fill=black] table [y=Dependency, x=Belief] {\strpolpooldialone};   
\addplot[fill = white] table [y=Dependency, x=Belief] {\strpolpoolcoredialone};  
\end{axis} 
\end{tikzpicture}
\caption{Arguments pooled by polarity and relations pooled by strength}
\label{fig:strpolpoold1totbyrel}
\end{subfigure}
\caption{The levels of agreements assigned to the sources of relations of a given type in Dialogue 1 according 
to a given type of pooling. We use the following abbreviations: 
Strong Belief (SB), Normal Belief (NB), Weak Belief (WB), Neither (N), Disbelief (D), Belief (B).}
\end{figure}

Let us now look at the results obtained for the pooled data. Given the similarity in the behaviour of the relations
marked as good reasons for and against in both samples and the somewhat good reasons for and against in the core sample, 
we will now pool our data according to its strength. 
We first consider grouping both arguments and relations according to this criterion. 
We obtain the distributions 
visible in Figure \ref{fig:strpoold1totbyrel}. The results show that they depend on the chosen relation type, both in the overall
\footnote{G--test for independence 
yields G--value $208.01$ 
with $6$ degrees of freedom and p--value less than $2.2 \times 10^{-16}$. 
This result was obtained using the library Deducer (likelihood.test function) in R.} 
and pairwise analysis (see Table \ref{tab:d1indeprelpool} in Section \ref{stats:relsource1}), 
and that they are not random (see Table \ref{tab:d1goodnessrelpool} in Section \ref{stats:relsource1}). We can observe
that strong relations are primarily carried out by strong arguments, i.e. those that the participants strongly agree or disagree with. 
The slightly weaker relations appear to be carried out more by moderately believed arguments, however, the obtained 
result accounts for less than 50\% of the cases. The sources of relations marked as dependencies do not seem to follow any particular pattern. We can only observe that weakly believed arguments are the least common sources. 

Let us now consider pooling arguments by polarity and relations by strength. The obtained 
distributions are depicted in Figure \ref{fig:strpolpoold1totbyrel}. We can observe that in this case, strong relations 
are primarily carried out by believed arguments. Normal relations tend to be carried out by disbelieved arguments, though 
only in the core sample such sources account for more than 50\% of the answers. Finally, relations marked as dependencies 
appear to be carried out by all possible types of arguments, to the point that the results obtained in the case of the core sample 
are similar to random (see Table \ref{tab:d1goodnessrelpool} in Section \ref{stats:relsource1}). Nevertheless, results in Table \ref{tab:d1indeprelpool}  in Section \ref{stats:relsource1} show
that despite pooling, the nature of the source is dependent on the type of relation we consider\footnote{G--test for independence 
yields G--value $175.33$ 
with $4$ degrees of freedom and p--value less than $2.2 \times 10^{-16}$. 
This result was obtained using the library Deducer (likelihood.test function) in R.} .  

Let us now briefly consider pooling relations and arguments according to polarity. 
Given the similarities in the behaviour of relations marked as good reasons 
in the original data, it is not surprising that the results obtained for attacks and supports are similar in both core and total samples 
(see also Table \ref{tab:d1indeprelpool} in Section \ref{stats:relsource1}), despite the fact that the overall analysis shows that source distributions depend 
on the chosen relation\footnote{G--test for independence 
yields G--value $62.846$ 
with $4$ degrees of freedom and p--value of $7.312 \times 10^{-13}$. 
This result was obtained using the library Deducer (likelihood.test function) in R.}. 
In both cases, the pooled belief marked as \textit{Neither} is the least common option, 
with \textit{Believed} being the most common choice with the exception of the \textit{Support} relations on the core sample. 
Again, relations marked as \textit{Dependencies} do not follow any particular pattern, even to the point that the answers 
can be considered random (see Table \ref{tab:d1goodnessrelpool} in Section \ref{stats:relsource1}). This similarity between attacks and supports 
persists even if we consider pooling agreement levels according to strength (see Tables \ref{tab:d1indeprelpool} and 
\ref{tab:d1goodnessrelpool} in Section \ref{stats:relsource1}), again despite overall result indicating that all of the relations are distinct
\footnote{G--test for independence 
yields G--value $69.849$ 
with $6$ degrees of freedom and p--value of $ 4.391 \times 10^{-13}$. 
This result was obtained using the library Deducer (likelihood.test function) in R.}. 

To conclude, we can observe that the dependent relations do not appear to favour sources marked as \textit{Neither} 
in any particular way, which does not agree with the models we have recalled at the start of the section. 
Nevertheless, 
certain interesting patterns can be observed for strong and normal relations. In particular, 
strong relations tend to be carried out by arguments that are believed (polarity pooling) and strongly believed (strength pooling). 
Grouping relations by polarity leads to similar behaviour of attack and support relations, which may mean that future experimental 
data on one of them might be generalized to the other. 
 
\pgfplotstableread
[col sep=&,row sep=\\]
{
Belief&Strong Relation&Normal Relation&Dependency\\
Strong Belief&55.522&26.403&31.863 \\
Moderate Belief&33.134&35.759&35.294\\
Weak Belief&8.806&29.314&10.294\\
Neither&2.537&8.524&22.549\\
}\strpooldialone
\pgfplotstableread
[col sep=&,row sep=\\]
{
Belief&Strong Relation&Normal Relation&Dependency\\
Strong Belief&65.232&28.962&29.333\\
Moderate Belief&25.497&43.169&30.667\\
Weak Belief&6.954&18.579&9.333\\
Neither&2.318&9.290&30.667\\
}\strpoolcoredialone
\pgfplotstableread
[col sep=&,row sep=\\]
{
Belief&Strong Relation&Normal Relation&Dependency\\
Disbelieved&24.627&49.272&36.275\\
Believed&72.836&42.204&41.176\\
Neither&2.537&8.524&22.549\\
}\strpolpooldialone

\pgfplotstableread
[col sep=&,row sep=\\]
{
Belief&Strong Relation&Normal Relation&Dependency\\
Disbelieved&31.457&66.667&30.667\\
Believed&66.225&24.044&38.667\\
Neither&2.318&9.290&30.667\\
}\strpolpoolcoredialone

\clearpage
\paragraph{Effect of an Argument on the Relations it Carries Out} \hfill 

\begin{figure}[!ht]
\centering
\pgfplotsset{every axis/.style={ybar, bar width=5pt,grid=both, width=0.32\textwidth,align =center,
x label style ={at={(axis description cs:0.5,0.04)}},
y label style ={at={(axis description cs:0.06,0.5)}}, 
symbolic x coords={Strong attack,Attack,Dependency,Support,Strong support},
xtick={Strong attack,Attack,Dependency,Support,Strong support},
xticklabels={- -, -,?,+,++}, 
enlargelimits=true,  ymin = 0, ymax = 62,  ytick = {0,10,20,30,40,50,60},
xticklabel style={font = \small, align=center, rotate=60,anchor=east},
yticklabels = {0\%,10\%,20\%,30\%,40\%,50\%,60\%},
legend style={at={(0.0,1.3)},anchor=south, font=\scriptsize}, legend columns=2, 
legend image code/.code={%
      \draw[#1] (0cm,-0.1cm) rectangle (0.2cm,0.2cm);
    }}}
\begin{tikzpicture}

\begin{axis}[name=stragr,title = {Strongly Agree},  ]
\addplot[fill = black] table [y=Strongly agree, x=TYPE] {\donetotbyarg};   
\addplot[fill = white] table [y=Strongly agree, x=TYPE] {\donecorbyarg};   
\end{axis} 

\begin{axis}[name=strdagr,title = {Strongly Disagree},yticklabels={,,,}, at=(stragr.right of south east)]
\addplot[fill = black] table [y=Strongly disagree, x=TYPE] {\donetotbyarg};   
\addplot[fill = white] table [y=Strongly disagree, x=TYPE]  {\donecorbyarg}; 

\end{axis} 

\begin{axis}[name=agr, title = {Agree} ,at=(strdagr.right of south east),yticklabels={,,,}]
\addplot[fill = black] table [y=Agree, x=TYPE] {\donetotbyarg};   
\addplot[fill = white] table [y=Agree, x=TYPE] {\donecorbyarg};    \legend{Total sample, Core sample} 
\end{axis} 

\begin{axis}[name=dagr, title = {Disagree}, yticklabels={,,,}, at=(agr.right of south east)]
\addplot[fill = black] table [y=Disagree, x=TYPE] {\donetotbyarg};  
\addplot[fill = white] table [y=Disagree, x=TYPE] {\donecorbyarg}; 
\end{axis}  

\begin{axis}[name=sagr, title = {Somewhat Agree},  at=(stragr.below south west), anchor = above north west, yshift=-10pt]
\addplot[fill = black] table [y=Somewhat Agree, x=TYPE] {\donetotbyarg};   
\addplot[fill = white] table [y=Somewhat Agree, x=TYPE]  {\donecorbyarg}; 
\end{axis} 
  
\begin{axis}[name=sdagr, title = {Somewhat Disagree}, yticklabels={,,,}, at=(sagr.right of south east)]
\addplot[fill = black] table [y=Somewhat Disagree, x=TYPE] {\donetotbyarg};   
\addplot[fill = white] table [y=Somewhat Disagree, x=TYPE] {\donecorbyarg}; 
\end{axis} 

\begin{axis}[name=nagr,yticklabels={,,,},, title = {Neither Agree \\ nor Disagree},  at=(sdagr.right of south east)]
\addplot[fill = black] table [y=Neither Agree nor Disagree, x=TYPE] {\donetotbyarg};   
\addplot[fill = white] table [y=Neither Agree nor Disagree, x=TYPE] {\donecorbyarg};
\end{axis} 

\begin{axis}[name=dunno, title = {Don't know}, yticklabels={,,,},  at=(nagr.right of south east)]  
\addplot[fill = black] table [y=Don't know, x=TYPE] {\donetotbyarg};  
\addplot[fill = white] table [y=Don't know, x=TYPE] {\donecorbyarg}; 
\end{axis}  
\end{tikzpicture}
\caption{The types of relations carried out by statements of a given level of agreement in Dialogue 1. We use the following abbreviations for the relation types:  
A good reason against (- -), A somewhat good reason against (-), A good reason for (++), A somewhat good reason for (+), Somewhat related, but 
can't say how (?)}
\label{fig:d1totbyargtot}
\end{figure}
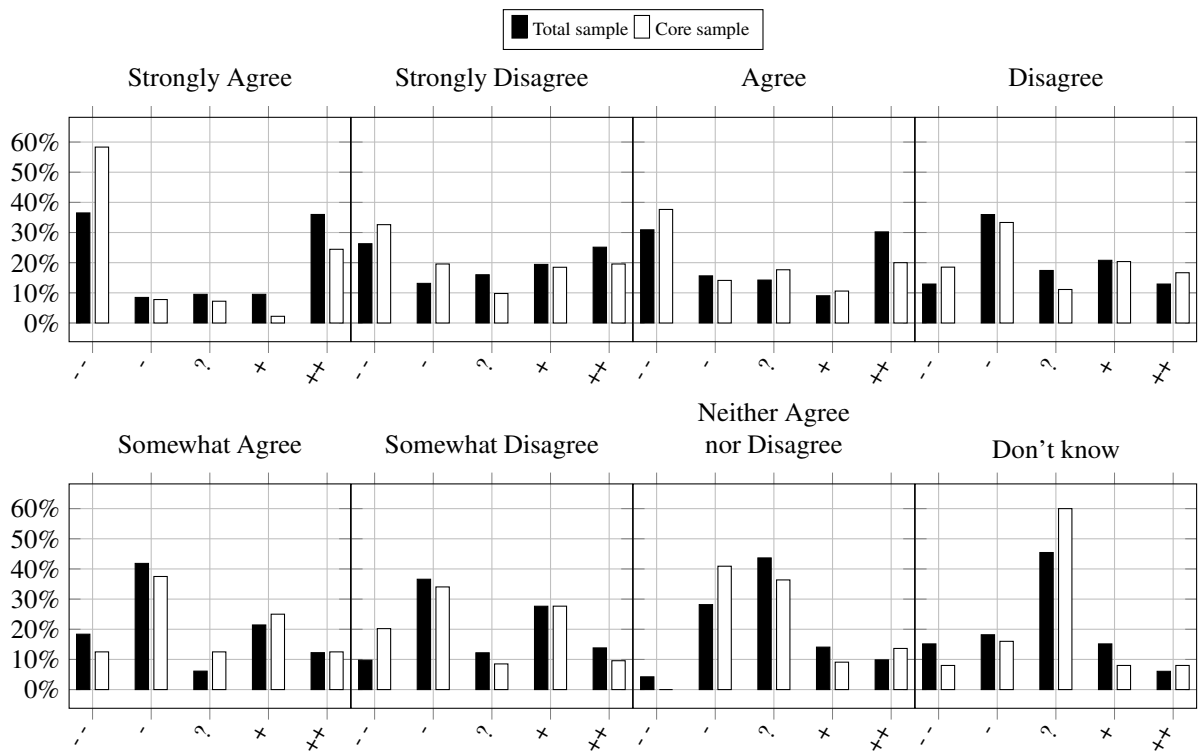

Let us now consider the dual problem i.e. what relations are carried out by an argument assigned a given agreement level? 
In Figure \ref{fig:d1totbyargtot} we can find the data extracted from the answers of our participants. Although in principle 
the relation distribution does depend on the strength of the argument carrying it out \footnote{G--test for independence 
yields G--value $317.77$ 
with $28$ degrees of freedom and p--value less than $2.2 \times 10^{-16}$. 
This result was obtained using the library Deducer (likelihood.test function) in R.}, 
we perform 
additional pairwise analysis. 
We can observe that 
while in the previous case the charts of relations with the same status but opposite polarity shared certain similarities, the grouping in this case 
is somewhat different. 
The graphs of \textit{Strongly Agree}, \textit{Agree} and potentially \textit{Strongly Disagree}
do tend to put more focus on the good reason for and against edges, however, they in general lead to different results 
(see Table \ref{tab:d1indeparg} in Section \ref{stats:sourcerel1}).
The \textit{Somewhat Agree},  \textit{Somewhat Disagree} and 
\textit{Disagree} statements on the other hand appear to lead more to relations marked as somewhat good reasons for or against.
The similarity in the types of relations these three agreement levels carry out is also reflected by the results in 
Table \ref{tab:d1indeparg}  in Section \ref{stats:sourcerel1}. 
The \textit{Somewhat good reason against} and \textit{Somehow related, but can't say how} are the two most common choices
when it comes to the statements that the participants neither agreed nor disagreed with, the latter being the dominant relation resulting from the elements marked as \textit{Don't know}. These two levels of agreement also lead to similar distributions 
(see Table \ref{tab:d1indeparg} in Section \ref{stats:sourcerel1}). 

Based on the results in Table \ref{tab:d1goodnessarg} in Section \ref{stats:sourcerel1}, we can also observe that our results 
are in principle not random. The results for agreement levels marked as \textit{Somewhat} in the core sample 
are not significant, however, 
given the distributions in Figure \ref{fig:d1totbyargtot}, this might be related to the sample size. 
 
\begin{figure}[!ht]
\centering

\begin{subfigure}[b]{\textwidth}
\centering
\pgfplotsset{every axis/.style={ybar, bar width=5pt,grid=both, height=0.26\textwidth, width=0.29\textwidth,align =center,
x label style ={at={(axis description cs:0.5,0.04)}},
y label style ={at={(axis description cs:0.06,0.5)}}, 
symbolic x coords={Strong Relation,Normal Relation,Dependency},
xtick={Strong Relation,Normal Relation,Dependency},
xticklabels={SR,NR,D}, 
enlarge x limits=0.3,  ymin = 0, ymax =75,  ytick = {0,10,20,30,40,50,60,70},
xticklabel style={font = \small, align=center, rotate=60,anchor=east},
yticklabels = {0\%,10\%,20\%,30\%,40\%,50\%,60\%,70\%},
legend style={at={(0.0,1.3)},anchor=south, font=\scriptsize}, legend columns=2, 
legend image code/.code={%
      \draw[#1] (0cm,-0.1cm) rectangle (0.2cm,0.2cm);
    }}}
\begin{tikzpicture}

\begin{axis}[name=stragr,title = {Strong Belief},  ]
\addplot[fill = black] table [y=Strong Belief, x=Relation] {\strpooldialonearg};   
\addplot[fill = white] table [y=Strong Belief, x=Relation] {\strpoolcoredialonearg};   
\end{axis} 

\begin{axis}[name=strdagr,title = {Moderate Belief},yticklabels={,,,}, at=(stragr.right of south east)]
\addplot[fill = black] table [y=Moderate Belief, x=Relation] {\strpooldialonearg};   
\addplot[fill = white] table [y=Moderate Belief, x=Relation] {\strpoolcoredialonearg};   

\end{axis} 

\begin{axis}[name=agr, title = {Weak Belief} ,at=(strdagr.right of south east),yticklabels={,,,}]
\addplot[fill = black] table [y=Weak Belief, x=Relation] {\strpooldialonearg};   
\addplot[fill = white] table [y=Weak Belief, x=Relation] {\strpoolcoredialonearg};   
\legend{Total sample, Core sample} 
\end{axis} 

\begin{axis}[name=dagr, title = {Neither}, yticklabels={,,,}, at=(agr.right of south east)]
\addplot[fill = black] table [y=Neither, x=Relation] {\strpooldialonearg};   
\addplot[fill = white] table [y=Neither, x=Relation] {\strpoolcoredialonearg};   
\end{axis}  
\end{tikzpicture}
\caption{Arguments and relations pooled by strength}
\label{fig:d1totbyargtotpooled}
\end{subfigure}

\begin{subfigure}[b]{\textwidth}
\centering
\pgfplotsset{every axis/.style={ybar, bar width=5pt,grid=both, height=0.26\textwidth, width=0.29\textwidth,align =center,
x label style ={at={(axis description cs:0.5,0.04)}},
y label style ={at={(axis description cs:0.06,0.5)}}, 
symbolic x coords={Strong Relation,Normal Relation,Dependency},
xtick={Strong Relation,Normal Relation,Dependency},
xticklabels={SR,NR,D}, 
enlarge x limits=0.3,  ymin = 0, ymax =75,  ytick = {0,10,20,30,40,50,60,70},
xticklabel style={font = \small, align=center, rotate=60,anchor=east},
yticklabels = {0\%,10\%,20\%,30\%,40\%,50\%,60\%,70\%},
legend style={at={(0.0,1.3)},anchor=south, font=\scriptsize}, legend columns=2, 
legend image code/.code={%
      \draw[#1] (0cm,-0.1cm) rectangle (0.2cm,0.2cm);
    }}}
\begin{tikzpicture}

\begin{axis}[name=stragr,title = {Belief}]
\addplot[fill = black] table [y=Believed, x=Relation] {\polstrpooldialonearg};   
\addplot[fill = white] table [y=Believed, x=Relation] {\polstrpoolcoredialonearg};   
\end{axis} 

\begin{axis}[name=dagr, title = {Neither}, yticklabels={,,,}, at=(stragr.right of south east)]
\addplot[fill = black] table [y=Neither, x=Relation] {\polstrpooldialonearg};   
\addplot[fill = white] table [y=Neither, x=Relation] {\polstrpoolcoredialonearg};    
\end{axis}  

\begin{axis}[name=strdagr,title = {Disbelief}, at=(dagr.right of south east),yticklabels={,,,} ]
\addplot[fill = black] table [y=Disbelieved, x=Relation] {\polstrpooldialonearg};   
\addplot[fill = white] table [y=Disbelieved, x=Relation] {\polstrpoolcoredialonearg};   
\end{axis}  
\end{tikzpicture}
\caption{Arguments pooled by polarity and relations pooled by strength}
\label{fig:d1totbyargtotpooled2}
\end{subfigure}
\caption{The types of relations carried out by statements of a given level of agreement in Dialogue 1 according to a given pooling. 
We use the following abbreviations for relation types:
Strong Relation (SR),Normal Relation (NR), Dependency (D).}
\end{figure}
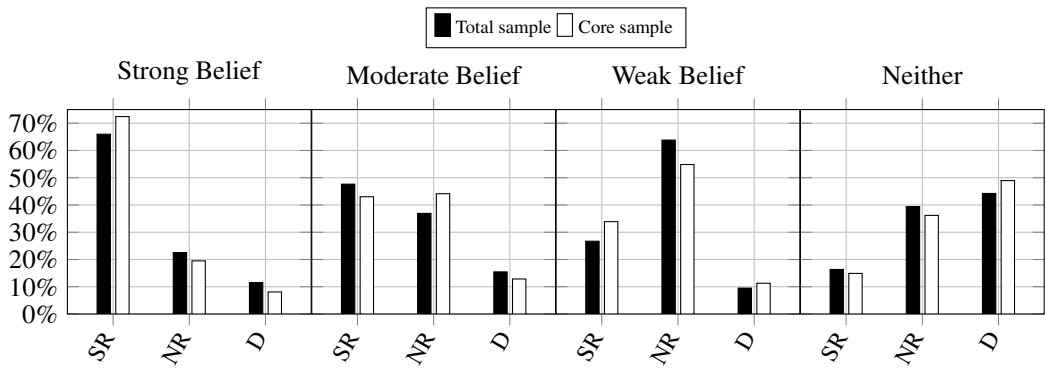
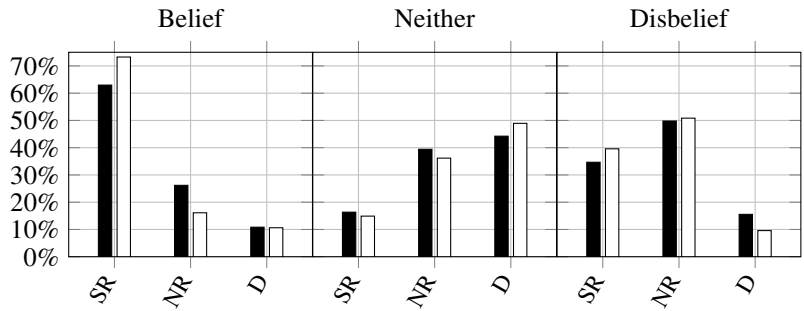

In many cases, this initial analysis points to the statements with a particular level of agreement favouring 
relations of a corresponding \enquote{power}. Thus, we will now consider pooling our results 
by strength, as visible in Figure \ref{fig:d1totbyargtotpooled}. All of the belief strengths affect the associated relation 
distributions in the total sample\footnote{G--test for independence 
yields G--value $208.01$ 
with $6$ degrees of freedom and p--value less than $2.2 \times 10^{-16}$. 
This result was obtained using the library Deducer (likelihood.test function) in R.}, 
although the results for the moderate and weak belief appear to be more closely related 
on the core sample (see Table \ref{tab:d1indepargpool} in Section \ref{stats:sourcerel1}). Nevertheless, the obtained distributions are not random (see Table 
\ref{tab:d1goodnessargpool} in Section \ref{stats:sourcerel1}) and we can observe certain interesting patterns emerging. Strong belief 
leads primarily to strong relations and weak belief leads primarily to normal relations (obtained results account for 55\% to 72\%
of the answers, depending on the sample). Relations caused by the arguments with moderate belief are 
almost evenly split between strong and normal relations. Arguments that are classified as \textit{Neither} 
often carry out dependencies, however, they do not account for 50\% of all of the possible relations. 

Let us now consider pooling agreement levels according to belief and relations according to strength. By doing so, 
we obtain the distributions visible in Figure \ref{fig:d1totbyargtotpooled2}. We can observe that 
believed arguments lead primarily to strong relations, disbelieved to normal relations, and neither to dependencies. 
However, clear dominance is visible only in the case of believed arguments. In all cases, the chosen agreement level 
leads to a different relation distribution\footnote{G--test for independence 
yields G--value $175.33$ 
with $6$ degrees of freedom and p--value less than $2.2 \times 10^{-16}$. 
This result was obtained using the library Deducer (likelihood.test function) in R.} 
that is not random (see Tables \ref{tab:d1indepargpool} and \ref{tab:d1goodnessargpool} in Section \ref{stats:sourcerel1}).  

 Finally, let us briefly consider agreement levels paired with relations pooled
by polarity. Although tests show that the obtained results are in principle distinct, many similarities become apparent in the 
pairwise analysis\footnote{When pooling arguments by strength, G--test for independence 
yields G--value $69.849$ 
with $6$ degrees of freedom and p--value of $ 4.391 \times 10^{-13}$. 
When pooling arguments by polarity, G--test for independence 
yields G--value $62.846$ 
with $4$ degrees of freedom and p--value of $7.312\times 10^{-13}$. 
This result was obtained using the library Deducer (likelihood.test function) in R.}. 
If we group levels by polarity, we obtain the result that the proportions of the relation types carried out by believed 
and disbelieved 
arguments are not significantly different (see Tables \ref{tab:d1indepargpool} and \ref{tab:d1goodnessargpool} in Section \ref{stats:sourcerel1}). 
This similarity may in part be due to the connections between the 
\textit{Somewhat Agree}, \textit{Somewhat Disagree} and \textit{Disagree} agreement levels. 
It might also explain the fact that when we consider merging arguments by strength, the similarities between the weak 
and moderate or strong levels of agreement become visible (see Tables \ref{tab:d1indepargpool} and 
\ref{tab:d1goodnessargpool} in Section \ref{stats:sourcerel1}).

To conclude, we can observe that the tendencies observable in Figures \ref{fig:d1totbyargtotpooled}
and \ref{fig:d1totbyargtotpooled2} seem to agree with what the models we have previously recalled would assume. 
However, the domination of the desired relations is not as significant in all situations as we would expect. Similarly as in the 
previous relation analysis, the tendencies for strongly believed (strength pooling) and believed (polarity pooling) arguments 
are the most visible.  

\pgfplotstableread
[col sep=&,row sep=\\]
{
Relation&Strong Belief&Moderate Belief&Weak Belief&Neither\\
Strong Relation&65.957&47.639&26.697&16.346\\
Normal Relation&22.518&36.910&63.801&39.423\\
Dependency&11.525&15.451&9.502&44.231\\
}\strpooldialonearg
\pgfplotstableread
[col sep=&,row sep=\\]
{
Relation&Strong Belief&Moderate Belief&Weak Belief&Neither\\
Strong Relation&72.426&43.017&33.871&14.894\\
Normal Relation&19.485&44.134&54.839&36.170\\
Dependency&8.088&12.849&11.290&48.936\\
}\strpoolcoredialonearg

\pgfplotstableread
[col sep=&,row sep=\\]
{
Relation&Disbelieved&Believed&Neither\\
Strong Relation&34.664&62.968&16.346\\
Normal Relation&49.790&26.194&39.423\\
Dependency&15.546&10.839&44.231\\
}\polstrpooldialonearg
\pgfplotstableread
[col sep=&,row sep=\\]
{
Relation&Disbelieved&Believed&Neither\\
Strong Relation&39.583&73.260&14.894\\
Normal Relation&50.833&16.117&36.170\\
Dependency&9.583&10.623&48.936\\
}\polstrpoolcoredialonearg

\subsubsection{Dialogue 2}

\paragraph{Effect of a Relation on its Source} \hfill

\begin{figure}[!ht] 
\centering
\pgfplotsset{every axis/.style={ybar, bar width=4pt,grid=both, height=0.32\textwidth, width=0.40\textwidth,align =center,
x label style ={at={(axis description cs:0.5,0.04)}},
y label style ={at={(axis description cs:0.06,0.5)}}, 
symbolic x coords={Strongly agree, Agree, Somewhat Agree, Neither Agree nor Disagree, Don't know, Somewhat Disagree, Disagree, Strongly disagree},
xtick={Strongly agree, Agree, Somewhat Agree, Neither Agree nor Disagree, Don't know, Somewhat Disagree, Disagree, Strongly disagree}, 
xticklabels={SA, A, SoA, NAD, DK, SoD, D, SD}, 
enlargelimits=true,  ymin = 0, ymax = 62,  ytick = {0,10,20,30,40,50,60},
xticklabel style={font = \small, align=center, rotate=60,anchor=east},
yticklabels = {0\%,10\%,20\%,30\%,40\%,50\%,60\%},
legend style={at={(0.0,1.3)},anchor=south, font=\scriptsize}, legend columns=2, 
legend image code/.code={%
      \draw[#1] (0cm,-0.1cm) rectangle (0.2cm,0.2cm);
    },}}
\begin{tikzpicture}

\begin{axis}[name=sap,title = {A good reason against}]
\addplot[fill = black] table [y=Strong attack, x=TYPE] {\dtwototbyrel};   
\addplot[fill = white] table [y=Strong attack, x=TYPE] {\dtwocorbyrel};  
\end{axis} 

\begin{axis}[name=ssp, title = {A good reason for} , yticklabels={,,,,},  at=(sap.right of south east)]
\addplot[fill = black] table [y=Strong support, x=TYPE] {\dtwototbyrel};   
\addplot[fill = white] table [y=Strong support, x=TYPE] {\dtwocorbyrel};   
\legend{Total sample, Core sample} 
\end{axis} 

\begin{axis}[name=attp, title = {A somewhat good \\reason against},   at=(sap.below south west), anchor = above north]
\addplot[fill = black] table [y=Attack, x=TYPE] {\dtwototbyrel};   
\addplot[fill = white] table [y=Attack, x=TYPE] {\dtwocorbyrel};   

\end{axis} 

\begin{axis}[name=supp, title = {A somewhat good\\ reason for},yticklabels={,,,,},   at=(attp.right of south east)]
\addplot[fill = black] table [y=Support, x=TYPE] {\dtwototbyrel};   
\addplot[fill = white] table [y=Support, x=TYPE] {\dtwocorbyrel};   
\end{axis} 

\begin{axis}[name=depp, title = {Somewhat related,\\ but can't say how},at=(supp.right of south east),yticklabels={,,,,}]
\addplot[fill = black] table [y=Dependency, x=TYPE] {\dtwototbyrel};   
\addplot[fill = white] table [y=Dependency, x=TYPE] {\dtwocorbyrel};   
\end{axis}  
\end{tikzpicture}
\caption{The levels of agreements assigned to sources of relations of a given type in Dialogue 2. We use the following abbreviations: 
Strongly Agree (SA), Agree (A), Somewhat Agree (SoA), Neither Agree nor Disagree (NAD), Somewhat Disagree (SoD), 
Disagree (D), Strongly Disagree (SD), Don't Know (DK)}
\label{fig:d2totbyrel}
\end{figure}
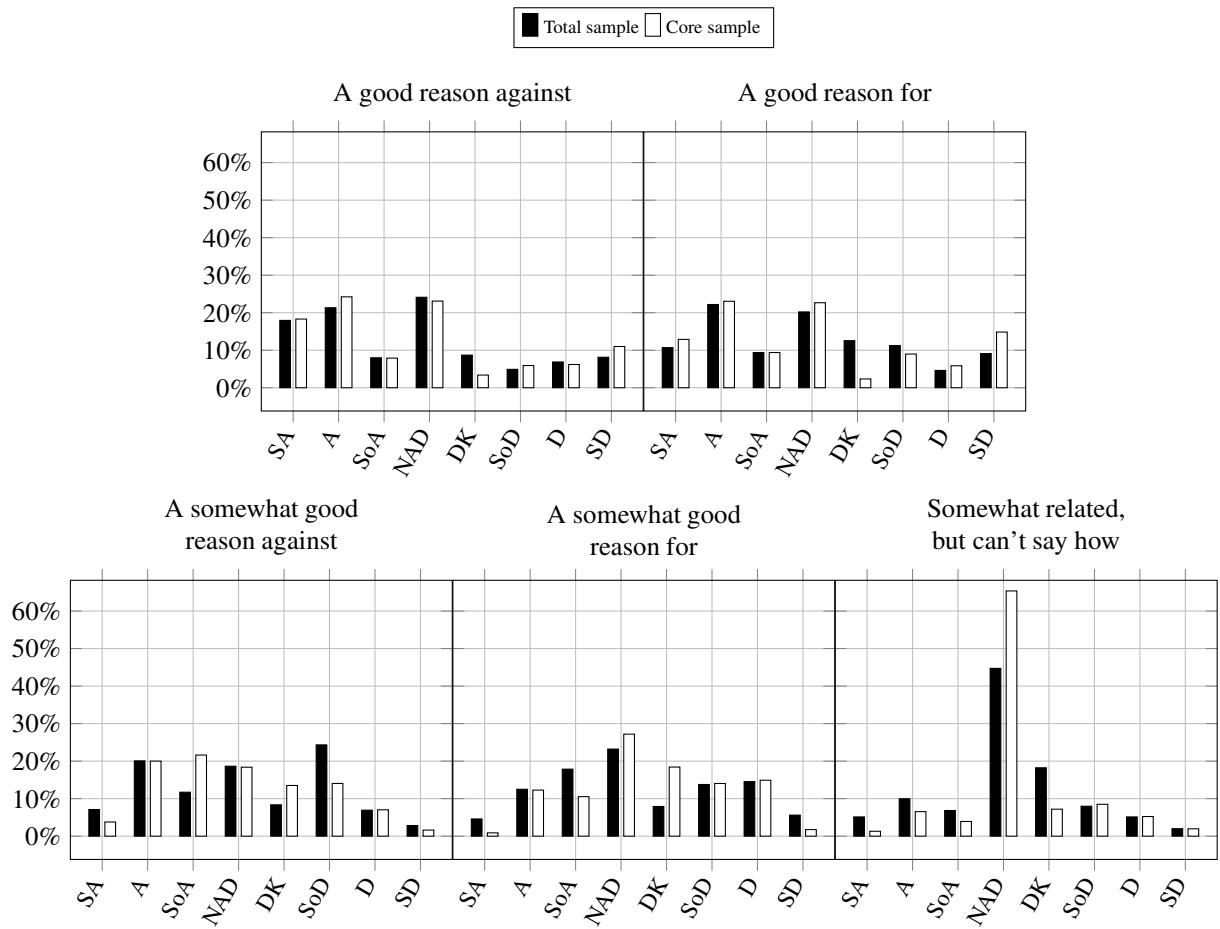

In Figure \ref{fig:d2totbyrel} we can observe how the levels of agreements on the sources of relations are distributed. 
The tests show that the agreement distributions depend on the type of the relation we consider, both in the overall 
\footnote{G--test for independence 
yields G--value $390.02$ 
with $28$ degrees of freedom and p--value less than $2.2 \times 10^{-16}$. 
This result was obtained using the library Deducer (likelihood.test function) in R.} and
pairwise analysis w.r.t. the total sample and in almost all cases w.r.t. the core sample (see Table \ref{tab:d2indeprel} in Section \ref{stats:relsource2}). In all cases, the obtained distributions are not random (Table \ref{tab:d2goodnessrel} in Section \ref{stats:relsource2}). 

As in the previous dialogue, we can observe certain similarities in the distributions of relations marked as good reasons 
for and against, particularly in the case of the core sample (Table \ref{tab:d2indeprel} in Section \ref{stats:relsource2}).
However, it is easy to see that these results are substantially different from the ones observed in 
Dialogue 1 (see Figure \ref{fig:d1totbyrel}). 
In this case, the \textit{Agree} and \textit{Neither Agree nor Disagree} options appear to 
be the two most common choices. Nevertheless, none of them exceeds 25\% of the given answers. 
The options representing agreement appear to be favoured over those representing disagreement, which we will take into consideration 
when analyzing pooled results. 

The \textit{Agree} and \textit{Neither Agree nor Disagree} answers are also common when it comes to relations marked 
as \textit{A somewhat good reason against}. However, depending on whether we are dealing with the total or core sample, the \textit{Somewhat Agree} or \textit{Somewhat Disagree} value 
is more frequent. 
It is worth noticing is that the strongest possible values -- \textit{Strongly Agree} and \textit{Strongly Disagree} -- 
are the least common. These values are also rarely chosen in the case of the \textit{A somewhat good reason for} relation. 
For this edge, the \textit{Neither Agree nor Disagree} 
level of agreement is the most common value, both in the total and in the core sample.  

Finally, we reach the relations marked as \textit{Somewhat related, but can't say how}. This is the only case in which the domination of a single value -- 
\textit{Neither Agree nor Disagree} -- is clearly visible, particularly in the case of the core sample. Additionally, in the total sample we can observe that the \textit{Don't know} option may also be of some importance. 
Thus,  we can observe  that independently of the chosen relation type, the \textit{Neither Agree nor Disagree} agreement level of the source
appears to be quite common, often with \textit{Agree}, \textit{Somewhat Agree} or \textit{Somewhat Disagree} not far behind.  
The observed behaviour of the relations is also notably different from what we have seen in Dialogue 1.

\pgfplotstableread
[col sep=&,row sep=\\]
{
TYPE&Strong attack&Attack&Strong support&Support&Dependency\\
Strongly agree&17.95&7.1&10.7&4.59&5.13\\
Agree&21.32&20.07&22.19&12.5&9.97\\
Somewhat Agree&7.99&11.72&9.38&17.86&6.84\\
Neither Agree nor Disagree&24.12&18.65&20.21&23.21&44.73\\
Disagree&6.87&6.93&4.62&14.54&5.13\\
Somewhat Disagree&4.91&24.33&11.23&13.78&7.98\\
Strongly disagree&8.13&2.84&9.11&5.61&1.99\\
Don't know&8.7&8.35&12.55&7.91&18.23\\
}\dtwototbyrel

\pgfplotstableread
[col sep=&,row sep=\\]
{
TYPE&Strong attack&Attack&Strong support&Support&Dependency\\
Strongly agree&18.31&3.78&12.89&0.88&1.31\\
Agree&24.23&20&23.05&12.28&6.54\\
Somewhat Agree&7.89&21.62&9.38&10.53&3.92\\
Neither Agree nor Disagree&23.1&18.38&22.66&27.19&65.36\\
Disagree&6.2&7.03&5.86&14.91&5.23\\
Somewhat Disagree&5.92&14.05&8.98&14.04&8.5\\
Strongly disagree&10.99&1.62&14.84&1.75&1.96\\
Don't know&3.38&13.51&2.34&18.42&7.19\\
}\dtwocorbyrel

\pgfplotstableread
[col sep=&,row sep=\\]
{
TYPE&Strongly agree&Agree&Somewhat Agree&Neither Agree nor Disagree&Disagree&Somewhat Disagree&Strongly disagree&Don't know\\
Strong attack&44.91&29.4&19.79&25.37&24.75&10.32&33.72&20.74\\
Attack&14.04&21.86&22.92&15.49&19.7&40.41&9.3&15.72\\
Strong support&28.42&32.5&24.65&22.57&17.68&25.07&40.12&31.77\\
Support&6.32&9.48&24.31&13.42&28.79&15.93&12.79&10.37\\
Dependency&6.32&6.77&8.33&23.16&9.09&8.26&4.07&21.4\\
}\dtwototbyarg

\pgfplotstableread
[col sep=&,row sep=\\]
{
TYPE&Strongly agree&Agree&Somewhat Agree&Neither Agree nor Disagree&Disagree&Somewhat Disagree&Strongly disagree&Don't know\\
Strong attack&60.19&41.75&25.45&26.89&29.33&21.21&45.88&16\\
Attack&6.48&17.96&36.36&11.15&17.33&26.26&3.53&33.33\\
Strong support&30.56&28.64&21.82&19.02&20&23.23&44.71&8\\
Support&0.93&6.8&10.91&10.16&22.67&16.16&2.35&28\\
Dependency&1.85&4.85&5.45&32.79&10.67&13.13&3.53&14.67\\
}\dtwocorbyarg

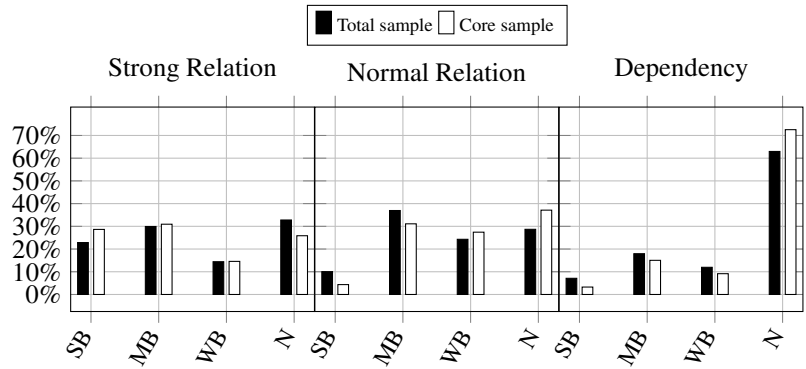
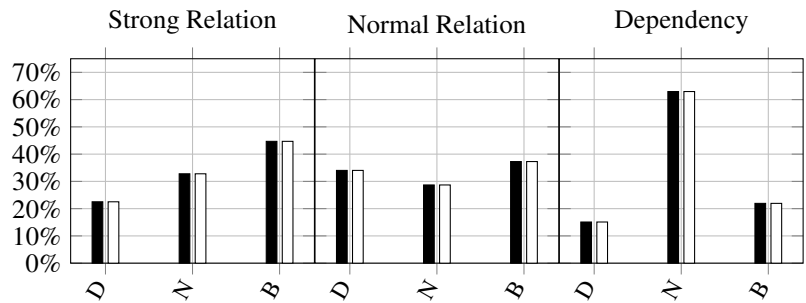
\begin{figure}[!ht] 
\centering
\begin{subfigure}[b]{\textwidth}
\centering
\pgfplotsset{every axis/.style={ybar, bar width=4pt,grid=both,height=0.26\textwidth, width=0.29\textwidth,align =center,
x label style ={at={(axis description cs:0.5,0.04)}},
y label style ={at={(axis description cs:0.06,0.5)}}, 
symbolic x coords={Strong Belief, Moderate Belief, Weak Belief, Neither},
xtick={Strong Belief, Moderate Belief, Weak Belief, Neither}, 
xticklabels={SB, MB, WB, N}, 
enlargelimits=true,  ymin = 0, ymax = 75,  ytick = {0,10,20,30,40,50,60,70},
xticklabel style={font = \small, align=center, rotate=60,anchor=east},
yticklabels = {0\%,10\%,20\%,30\%,40\%,50\%,60\%,70\%},
legend style={at={(-0.5,1.3)},anchor=south, font=\scriptsize}, legend columns=2, 
legend image code/.code={%
      \draw[#1] (0cm,-0.1cm) rectangle (0.2cm,0.2cm);
    },}}
\begin{tikzpicture}

\begin{axis}[name=strrel,title = {Strong Relation}]
\addplot[fill=black] table [y=Strong Relation, x=Belief] {\strpooldialtwo};   
\addplot[fill = white] table [y=Strong Relation, x=Belief] {\strpoolcoredialtwo};  
\end{axis} 

\begin{axis}[name=normrel, title = {Normal Relation} , yticklabels={,,,,},  at=(strrel.right of south east)]

\addplot[fill=black] table [y=Normal Relation, x=Belief] {\strpooldialtwo};   
\addplot[fill = white] table [y=Normal Relation, x=Belief] {\strpoolcoredialtwo};  
\end{axis} 

\begin{axis}[name=dep, title = {Dependency}, yticklabels={,,,,},  at=(normrel.right of south east)]
\addplot[fill=black] table [y=Dependency, x=Belief] {\strpooldialtwo};   
\addplot[fill = white] table [y=Dependency, x=Belief] {\strpoolcoredialtwo};  
\legend{Total sample, Core sample} 
\end{axis} 
\end{tikzpicture}
\caption{Arguments and relations pooled by strength}
\label{fig:strpoold2totbyrel}
\end{subfigure}

\begin{subfigure}[b]{\textwidth}
\centering
\pgfplotsset{every axis/.style={ybar, bar width=4pt,grid=both,height=0.26\textwidth, width=0.29\textwidth, align =center,
x label style ={at={(axis description cs:0.5,0.04)}},
y label style ={at={(axis description cs:0.06,0.5)}}, 
symbolic x coords={Disbelieved, Neither, Believed},
xtick={Disbelieved, Neither, Believed}, 
xticklabels={D, N,B}, 
enlarge x limits=0.2,  ymin = 0, ymax = 75,  ytick = {0,10,20,30,40,50,60,70},
xticklabel style={font = \small, align=center, rotate=60,anchor=east},
yticklabels = {0\%,10\%,20\%,30\%,40\%,50\%,60\%,70\%},
legend style={at={(0.5,1.3)},anchor=south, font=\scriptsize}, legend columns=2, 
legend image code/.code={%
      \draw[#1] (0cm,-0.1cm) rectangle (0.2cm,0.2cm);
    },}}
\begin{tikzpicture}

\begin{axis}[name=strrel,title = {Strong Relation}]
\addplot[fill=black] table [y=Strong Relation, x=Belief] {\strpolpooldialtwo};   
\addplot[fill = white] table [y=Strong Relation, x=Belief] {\strpolpoolcoredialtwo};  
\end{axis} 

\begin{axis}[name=normrel, title = {Normal Relation} , yticklabels={,,,,},  at=(strrel.right of south east)]

\addplot[fill=black] table [y=Normal Relation, x=Belief] {\strpolpooldialtwo};   
\addplot[fill = white] table [y=Normal Relation, x=Belief] {\strpolpoolcoredialtwo};  
\end{axis} 

\begin{axis}[name=dep, title = {Dependency}, yticklabels={,,,,},  at=(normrel.right of south east)]
\addplot[fill=black] table [y=Dependency, x=Belief] {\strpolpooldialtwo};   
\addplot[fill = white] table [y=Dependency, x=Belief] {\strpolpoolcoredialtwo};  
\end{axis} 
\end{tikzpicture}
\caption{Arguments pooled by polarity and relations pooled by strength}
\label{fig:strpolpoold2totbyrel}
\end{subfigure}
\caption{The levels of agreements assigned to the sources of relations of a given type in Dialogue 2 according 
to a given type of pooling. We use the following abbreviations: 
Strong Belief (SB), Normal Belief (NB), Weak Belief (WB), Neither (N), Disbelief (D), Belief (B).}
\end{figure}

Let us now analyze pooled data. We start by grouping relations by strength, and pool agreement levels first by strength 
and afterwards by polarity. We obtain the distributions 
visible in Figures \ref{fig:strpoold2totbyrel} and \ref{fig:strpolpoold2totbyrel}. 
The results show that the way the sources are distributed depends on the chosen relation type, 
both in the overall \footnote{In case both arguments and relations have been pooled by strength, the G--test for independence 
yields G--value $232.25$ 
with $6$ degrees of freedom and p--value less than $2.2 \times 10^{-16}$. 
In pooling arguments by polarity and relations by strength, we obtain G--value $167.86$, 
$4$ degrees of freedom 
and again p--value less than $2.2 \times 10^{-16}$.
This result was obtained using the library Deducer (likelihood.test function) in R.}
and pairwise analysis (see Table \ref{tab:d2indeprelpool} in Section \ref{stats:relsource2}). 
Furthermore, the results are also not random
 (Table \ref{tab:d2goodnessrelpool} in Section \ref{stats:relsource2}). 
Let us now look more closely at the results with strength pooling on both agreement levels and relations. 
We can observe that unlike in the first dialogue, the distribution associated with the strong relations does not appear 
to follow any particular pattern. We can only note that weak belief is the least common type of a source. 
In a similar fashion, normal relations are carried out by strongly believed arguments the least often, with the remaining 
types of sources being spread relatively evenly. 
Finally, while in Dialogue 1 relations marked as \textit{Dependencies} did not appear to follow any patterns, 
in Dialogue 2 the dominance of sources grouped as \textit{Neither} is clear. 

The dominance of \textit{Neither} sources in \textit{Dependencies} is also quite clear if we look at the results 
of arguments pooled by polarity and relations by strength. We can observe that in this analysis, strong relations 
tend to be carried out by believed arguments. Although this repeats the results of Dialogue 1, in this case the dominance 
is not entirely clear and believed sources do not account for 50\% of the relations. Additionally, unlike in Dialogue 1, the disbelieved 
arguments appear to be the least common sources. No particular pattern emerges when we look at normal relations -- we can 
only observe that the sources marked as \textit{Neither} appear to be somewhat less common than other types. 

Let us now briefly consider pooling relations and arguments according to polarity. 
Despite the fact that the overall analysis shows that source distributions depend 
on the chosen relation\footnote{G--test for independence 
yields G--value $132.84$ 
with $4$ degrees of freedom and p--value less than $2.2 \times 10^{-16}$. 
This result was obtained using the library Deducer (likelihood.test function) in R.}, pairwise analysis gives us a different result.  
Although the similarities in the behaviour of relations marked as good reasons 
in the original data were only apparent in the core sample, 
the pooled results obtained for attacks and supports are similar in both core and total samples 
(see also Table \ref{tab:d2indeprelpool} in Section \ref{stats:relsource2}).  
Similarly as in the first dialogue, the \textit{Believed} level is the most common choice. However, in this dialogue, 
the relations marked as \textit{Dependencies} do follow a pattern and are carried out primarily by arguments marked 
as \textit{Neither}. In all cases, our results are not 
random (see Table \ref{tab:d2goodnessrelpool} in Section \ref{stats:relsource2}). 
Let us now pool agreement levels according to strength. The overall test shows that the source distribution depends on 
the chosen relation \footnote{G--test for independence 
yields G--value $ 142.15$ 
with $6$ degrees of freedom and p--value less than $2.2 \times 10^{-16}$.  
This result was obtained using the library Deducer (likelihood.test function) in R.}.  
However, a detailed analysis shows that agreement distributions associated with attacks and supports 
still share certain similarities (see Tables \ref{tab:d2indeprelpool} and 
\ref{tab:d2goodnessrelpool} in Section \ref{stats:relsource2}), even though it becomes apparent mostly on the core sample.

To conclude, although there are certain characteristics present in both dialogues, not all of the findings from Dialogue 2 coincide with those from Dialogue 1. This is particularly the case when we consider 
pooling relations by strength. Results associated with agreements pooled by strength are quite distinct from the results 
in Dialogue 1. It is worth observing that relations marked as \textit{Dependency}, which previously did not follow any particular pattern, now favour agreement levels marked as \textit{Neither}. 
While strong relations in both dialogues do tend to favour believed arguments more than disbelieved or neither in polarity pooling of agreement levels,
the domination of this value in Dialogue 1 is not replicated in Dialogue 2.  

\pgfplotstableread
[col sep=&,row sep=\\]
{
Belief&Strong Relation&Normal Relation&Dependency\\
Strong Belief&22.85714286&10.05235602&7.122507123\\	
Moderate Belief&29.93197279&36.96335079&17.94871795\\
Weak Belief&14.42176871&24.29319372&11.96581197\\
Neither&32.78911565&28.69109948&62.96296296\\
}\strpooldialtwo
\pgfplotstableread
[col sep=&,row sep=\\]
{
Belief&Strong Relation&Normal Relation&Dependency\\
Strong Belief&28.64157119&4.347826087&3.267973856\\	
Moderate Belief&30.93289689&31.10367893&15.03267974\\
Weak Belief&14.56628478&27.42474916&9.150326797\\
Neither&25.85924714&37.12374582&72.54901961\\
}\strpoolcoredialtwo

\pgfplotstableread
[col sep=&,row sep=\\]
{
Belief&Strong Relation&Normal Relation&Dependency\\
Disbelieved&22.517&34.031&15.100\\
Believed&44.694&37.277&21.937\\
Neither&32.789&28.691&62.963\\
}\strpolpooldialtwo
\pgfplotstableread
[col sep=&,row sep=\\]
{
Belief&Strong Relation&Normal Relation&Dependency\\
Disbelieved&22.517&34.031&15.100\\
Believed&44.694&37.277&21.937\\
Neither&32.789&28.691&62.963\\
}\strpolpoolcoredialtwo

\clearpage
\paragraph{Effect of an Argument on the Relations it Carries Out} \hfill

\begin{figure}[!ht]
\centering
\pgfplotsset{every axis/.style={ybar, bar width=5pt,grid=both, width=0.32\textwidth,align =center,
x label style ={at={(axis description cs:0.5,0.04)}},
y label style ={at={(axis description cs:0.06,0.5)}}, 
symbolic x coords={Strong attack,Attack,Dependency,Support,Strong support},
xtick={Strong attack,Attack,Dependency,Support,Strong support},
xticklabels={- -, -,?,+,++}, 
enlargelimits=true,  ymin = 0, ymax = 62,  ytick = {0,10,20,30,40,50,60},
xticklabel style={font = \small, align=center, rotate=60,anchor=east},
yticklabels = {0\%,10\%,20\%,30\%,40\%,50\%,60\%},
legend style={at={(0.0,1.3)},anchor=south, font=\scriptsize}, legend columns=2, 
legend image code/.code={%
      \draw[#1] (0cm,-0.1cm) rectangle (0.2cm,0.2cm);
    }}}
\begin{tikzpicture}

\begin{axis}[name=stragr,title = {Strongly Agree},  ]
\addplot[fill = black] table [y=Strongly agree, x=TYPE] {\dtwototbyarg};   
\addplot[fill = white] table [y=Strongly agree, x=TYPE] {\dtwocorbyarg};   
\end{axis}  

\begin{axis}[name=strdagr,title = {Strongly Disagree},yticklabels={,,,}, at=(stragr.right of south east)]
\addplot[fill = black] table [y=Strongly disagree, x=TYPE] {\dtwototbyarg};   
\addplot[fill = white] table [y=Strongly disagree, x=TYPE]  {\dtwocorbyarg}; 
\end{axis} 

\begin{axis}[name=agr, title = {Agree},at=(strdagr.right of south east), yticklabels={,,,}]
\addplot[fill = black] table [y=Agree, x=TYPE] {\dtwototbyarg};   
\addplot[fill = white] table [y=Agree, x=TYPE] {\dtwocorbyarg};  
\legend{Total sample, Core sample}  
\end{axis} 

\begin{axis}[name=dagr, title = {Disagree},at=(agr.right of south east), yticklabels={,,,}]
\addplot[fill = black] table [y=Disagree, x=TYPE] {\dtwototbyarg};  
\addplot[fill = white] table [y=Disagree, x=TYPE] {\dtwocorbyarg}; 
\end{axis}  

\begin{axis}[name=sagr, title = {Somewhat Agree},   at=(stragr.below south west), anchor = above north west, yshift=-10pt]
\addplot[fill = black] table [y=Somewhat Agree, x=TYPE] {\dtwototbyarg};   
\addplot[fill = white] table [y=Somewhat Agree, x=TYPE]  {\dtwocorbyarg}; 
\end{axis} 

\begin{axis}[name=sdagr, title = {Somewhat Disagree}, yticklabels={,,,}, at = (sagr.right of south east)]
\addplot[fill = black] table [y=Somewhat Disagree, x=TYPE] {\dtwototbyarg};   
\addplot[fill = white] table [y=Somewhat Disagree, x=TYPE] {\dtwocorbyarg}; 
\end{axis} 

\begin{axis}[name=nagr,yticklabels={,,,},, title = {Neither Agree \\ nor Disagree},  at=(sdagr.right of south east)]
\addplot[fill = black] table [y=Neither Agree nor Disagree, x=TYPE] {\dtwototbyarg};   
\addplot[fill = white] table [y=Neither Agree nor Disagree, x=TYPE] {\dtwocorbyarg}; 
\end{axis} 

\begin{axis}[name=dunno, title = {Don't know}, yticklabels={,,,},  at=(nagr.right of south east)]  
\addplot[fill = black] table [y=Don't know, x=TYPE] {\dtwototbyarg};  
\addplot[fill = white] table [y=Don't know, x=TYPE] {\dtwocorbyarg}; 
\end{axis}  
\end{tikzpicture}
\caption{The types of relations carried out by statements of a given level of agreement in Dialogue 2. We use the following abbreviations for the relation types: 
A good reason against (- -), A somewhat good reason against (-), A good reason for (++), A somewhat good reason for (+), Somewhat related, but 
can't say how (?)}
\label{fig:d2totbyargtot}
\end{figure}
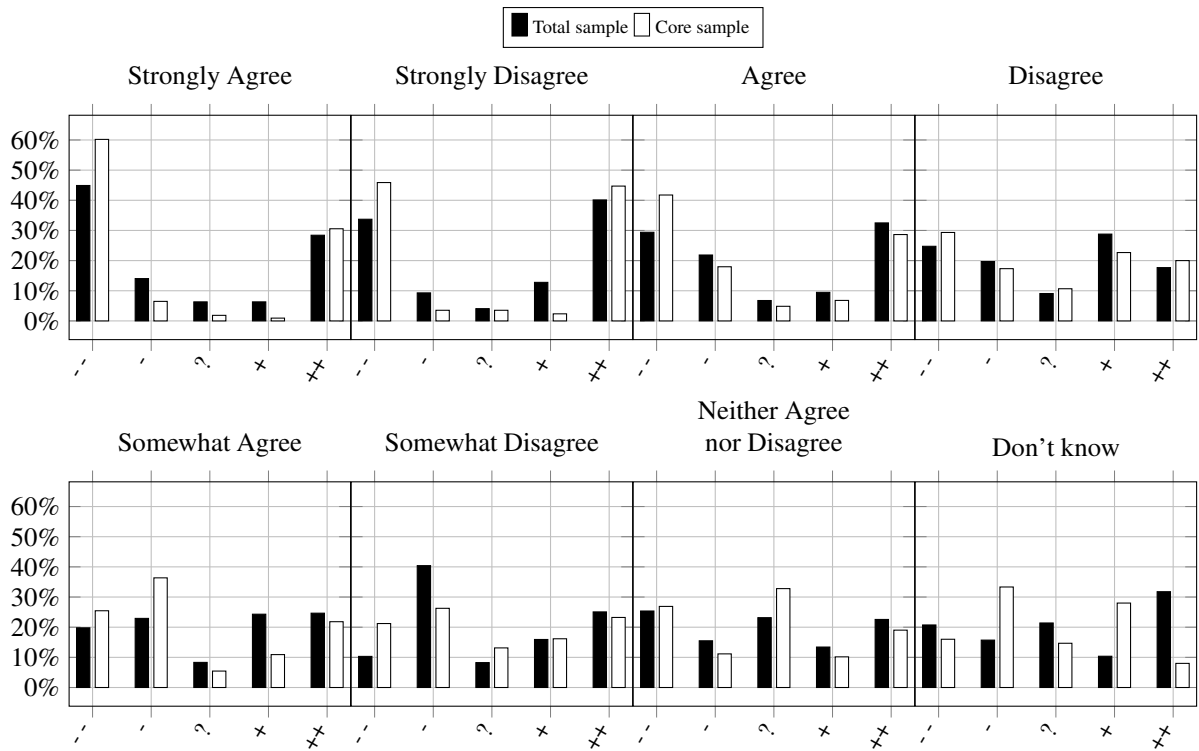

In Figure \ref{fig:d2totbyargtot} we can find the data showing what sort of relations are carried out by the statements with a particular agreement level. In general, the choice of agreement level affects the resulting relation types\footnote{G--test for independence 
yields G--value $390.02$ 
with $28$ degrees of freedom and p--value less than $2.2 \times 10^{-16}$. 
This result was obtained using the library Deducer (likelihood.test function) in R.}. 
However, in the pairwise analysis, certain similarities become more visible (see Table \ref{tab:d2indeparg} in Section \ref{stats:sourcerel2}).

We may notice that some of the observations made in the case of Dialogue 1 carry over to Dialogue 2. In particular, the 
charts created for \textit{Strongly Agree}, \textit{Agree} and \textit{Strongly Disagree} tend to put more focus on the 
good reason for and against edges. Nevertheless, the distributions are technically speaking distinct, 
and the similarity between the strong agreement
and disagreement is detected only in the core sample. 
The \textit{Disagree} level, depending on the sample we are dealing with, favours \textit{A good reason against} 
and \textit{Somewhat good reason for} edges, with \textit{A good reason for} and \textit{A somewhat good reason against} next in line. 

The \textit{Somewhat Agree} and \textit{Somewhat Disagree} charts are quite similar, as also supported by results in 
Table \ref{tab:d2indeparg} in Section \ref{stats:sourcerel2}. 
In this case, the \textit{A somewhat good reason against} agreement level is perhaps more common than other options. 
The \textit{Neither Agree nor Disagree} agreement level appears to favour the \textit{Somewhat related, but can't say how} option the most 
and the \textit{Somewhat good reason for/against} answers the least. Finally, we can observe that the distributions obtained from the total and core samples 
in the case of the \textit{Don't know} chart are quite different; one favours the \textit{A good reason for} answer, while the other 
the \textit{Somewhat good reason for/against} options. 
We can observe that the distributions associated with the agreement levels are in principle not random (see Table \ref{tab:d2goodnessarg} in Section \ref{stats:sourcerel2}), though there appear to be certain difficulties with the \textit{Disagree}
and \textit{Somewhat Disagree} values in the core sample.

\begin{figure}[!ht]
\centering

\begin{subfigure}[b]{\textwidth}
\centering
\pgfplotsset{every axis/.style={ybar, bar width=5pt,grid=both, height=0.26\textwidth, width=0.29\textwidth,align =center,
x label style ={at={(axis description cs:0.5,0.04)}},
y label style ={at={(axis description cs:0.06,0.5)}}, 
symbolic x coords={Strong Relation,Normal Relation,Dependency},
xtick={Strong Relation,Normal Relation,Dependency},
xticklabels={SR,NR,D}, 
enlarge x limits=0.3,  ymin = 0, ymax =95,  ytick = {0,10,20,30,40,50,60,70,80,90},
xticklabel style={font = \small, align=center, rotate=60,anchor=east},
yticklabels = {0\%,10\%,20\%,30\%,40\%,50\%,60\%,70\%,80\%,90\%},
legend style={at={(0.0,1.3)},anchor=south, font=\scriptsize}, legend columns=2, 
legend image code/.code={%
      \draw[#1] (0cm,-0.1cm) rectangle (0.2cm,0.2cm);
    }}}
\begin{tikzpicture}

\begin{axis}[name=stragr,title = {Strong Belief},  ]
\addplot[fill = black] table [y=Strong Belief, x=Relation] {\strpooldialtwoarg};   
\addplot[fill = white] table [y=Strong Belief, x=Relation] {\strpoolcoredialtwoarg};   
\end{axis} 

\begin{axis}[name=strdagr,title = {Moderate Belief},yticklabels={,,,}, at=(stragr.right of south east)]
\addplot[fill = black] table [y=Moderate Belief, x=Relation] {\strpooldialtwoarg};   
\addplot[fill = white] table [y=Moderate Belief, x=Relation] {\strpoolcoredialtwoarg};   

\end{axis} 

\begin{axis}[name=agr, title = {Weak Belief} ,at=(strdagr.right of south east),yticklabels={,,,}]
\addplot[fill = black] table [y=Weak Belief, x=Relation] {\strpooldialtwoarg};   
\addplot[fill = white] table [y=Weak Belief, x=Relation] {\strpoolcoredialtwoarg};   
\legend{Total sample, Core sample} 
\end{axis} 

\begin{axis}[name=dagr, title = {Neither}, yticklabels={,,,}, at=(agr.right of south east)]
\addplot[fill = black] table [y=Neither, x=Relation] {\strpooldialtwoarg};   
\addplot[fill = white] table [y=Neither, x=Relation] {\strpoolcoredialtwoarg};   
\end{axis}  
\end{tikzpicture}
\caption{Arguments and relations pooled by strength}
\label{fig:d2totbyargtotpooled}
\end{subfigure}

\begin{subfigure}[b]{\textwidth}
\centering
\pgfplotsset{every axis/.style={ybar, bar width=5pt,grid=both, height=0.26\textwidth, width=0.29\textwidth,align =center,
x label style ={at={(axis description cs:0.5,0.04)}},
y label style ={at={(axis description cs:0.06,0.5)}}, 
symbolic x coords={Strong Relation,Normal Relation,Dependency},
xtick={Strong Relation,Normal Relation,Dependency},
xticklabels={SR,NR,D}, 
enlarge x limits=0.3,  ymin = 0, ymax =75,  ytick = {0,10,20,30,40,50,60,70},
xticklabel style={font = \small, align=center, rotate=60,anchor=east},
yticklabels = {0\%,10\%,20\%,30\%,40\%,50\%,60\%,70\%},
legend style={at={(0.0,1.3)},anchor=south, font=\scriptsize}, legend columns=2, 
legend image code/.code={%
      \draw[#1] (0cm,-0.1cm) rectangle (0.2cm,0.2cm);
    }}}
\begin{tikzpicture}

\begin{axis}[name=stragr,title = {Belief}]
\addplot[fill = black] table [y=Believed, x=Relation] {\polstrpooldialtwoarg};   
\addplot[fill = white] table [y=Believed, x=Relation] {\polstrpoolcoredialtwoarg};   
\end{axis} 

\begin{axis}[name=dagr, title = {Neither}, yticklabels={,,,}, at=(stragr.right of south east)]
\addplot[fill = black] table [y=Neither, x=Relation] {\polstrpooldialtwoarg};   
\addplot[fill = white] table [y=Neither, x=Relation] {\polstrpoolcoredialtwoarg};    
\end{axis}  

\begin{axis}[name=strdagr,title = {Disbelief}, at=(dagr.right of south east),yticklabels={,,,} ]
\addplot[fill = black] table [y=Disbelieved, x=Relation] {\polstrpooldialtwoarg};   
\addplot[fill = white] table [y=Disbelieved, x=Relation] {\polstrpoolcoredialtwoarg};   
\end{axis}  
\end{tikzpicture}
\caption{Arguments pooled by polarity and relations pooled by strength}
\label{fig:d2totbyargtotpooled2}
\end{subfigure}
\caption{The types of relations carried out by statements of a given level of agreement in Dialogue 2 according to a given pooling. 
We use the following abbreviations for the relation types:
Strong Relation (SR),Normal Relation (NR), Dependency (D).}
\end{figure}
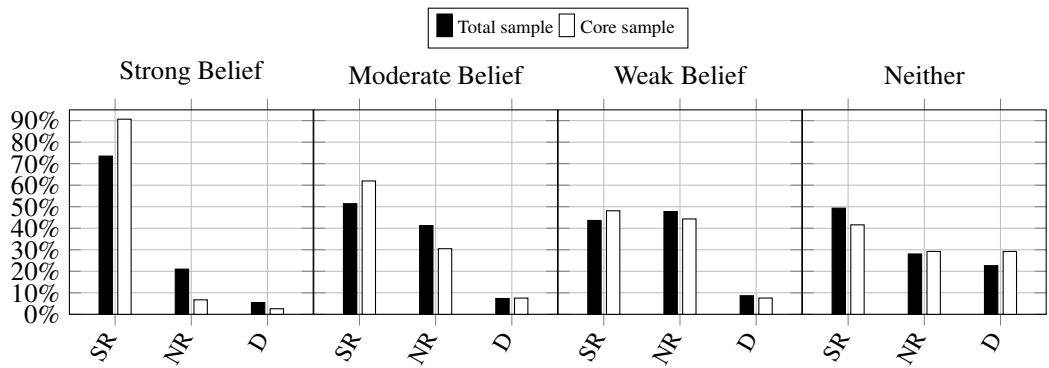
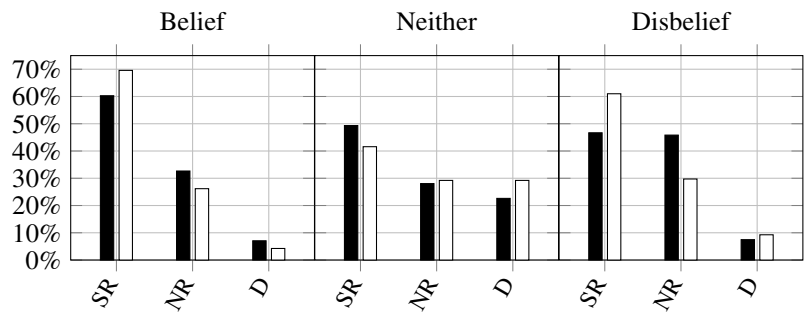

Let us now consider pooling our results. We start by grouping arguments and relations  
by strength, as visible in Figure \ref{fig:d2totbyargtotpooled}. All of the belief strengths affect the associated relation 
distributions, both in the overall\footnote{G--test for independence 
yields G--value $232.25$ 
with $6$ degrees of freedom and p--value less than $2.2 \times 10^{-16}$. 
This result was obtained using the library Deducer (likelihood.test function) in R.} 
and pairwise analysis (see Table \ref{tab:d2indepargpool} in Section \ref{stats:sourcerel2}). Additionally, the answers extracted from the users
are not random (see Table \ref{tab:d2goodnessargpool} in Section \ref{stats:sourcerel2}).
We can observe that the strong relations account for at least 50\% of connections  
carried out by arguments that are strongly or moderately believed. Weak belief appears to be split evenly between 
the strong and normal relations. Arguments that are neither believed nor disbelieved can lead to all three types of relations, 
though strong relations are more favoured than others. Consequently, not all of the observations made in the case
of Dialogue 1 are repeated in Dialogue 2. 

Let us now consider grouping arguments by polarity and relations by strength (see Figure \ref{fig:d2totbyargtotpooled2}). 
Again, all of the belief strengths affect the associated relation 
distributions, both in the overall\footnote{G--test for independence 
yields G--value $167.86$ 
with $4$ degrees of freedom and p--value less than $2.2 \times 10^{-16}$. 
This result was obtained using the library Deducer (likelihood.test function) in R.} 
and pairwise analysis (see Table \ref{tab:d2indepargpool} in Section \ref{stats:sourcerel2}). The answers extracted from the users
are also not random (see Table \ref{tab:d2goodnessargpool} in Section \ref{stats:sourcerel2}). 
Nevertheless, there are certain similarities between the obtained distributions. In particular, we can observe that strong 
relations appear to be the most common relations carried out by arguments marked as \textit{Believed} or \textit{Neither}, 
independently of the sample, and by \textit{Disbelieved} in the core sample. In particular, in the first and the last case 
these relations account for over 60\% of all edges. What is worth observing is that dependent links are the least common when 
it comes to arguments that are believed or disbelieved. Nevertheless, it is easy to see that the results obtained for 
agreement levels marked as \textit{Neither} and \textit{Disbelieved} in the second dialogue are different from the first dialogue.

 We will now now briefly consider agreement levels paired with relations pooled
by polarity. Although tests show that the obtained results are in principle distinct, certain similarities become apparent in the 
pairwise analysis\footnote{When pooling arguments by strength, G--test for independence 
yields G--value $142.15$ 
with $6$ degrees of freedom and p--value less than $2.2 \times 10^{-16}$.  
When pooling arguments by polarity, G--test for independence 
yields G--value $ 132.84$ 
with $4$ degrees of freedom and p--value less than $2.2 \times 10^{-16}$. 
This result was obtained using the library Deducer (likelihood.test function) in R.}. 
If we group agreement levels by polarity, we obtain the result that the proportions of the relation types carried out by the believed 
and disbelieved 
arguments are not significantly different (see Table \ref{tab:d2indepargpool} in Section \ref{stats:sourcerel2}), 
particularly in the case of the total sample. When we consider merging arguments by strength, certain similarities between the weak, moderate and strong levels of agreement become visible (see Tables \ref{tab:d2indepargpool} and 
\ref{tab:d2goodnessargpool} in Section \ref{stats:sourcerel2}).

To conclude, despite various differences between Dialogue 1 and Dialogue 2, it again appears 
that analysis in which relations are pooled by polarity leads to situations where relation distributions might not in all cases 
depend on the agreement level of the source. When pooling relations by strength, we can also observe that similarly to Dialogue 1,  the strongly believed (strength pooling) and believed (polarity pooling) arguments lead primarily to strong relations. However, 
the remaining pooled agreement levels do not appear to follow exactly the same patterns. Even though certain shifts towards 
weaker relations can be observed in Figure \ref{fig:d2totbyargtotpooled}, their progress can be seen as \enquote{slower} than 
in Dialogue 1.  
 
\pgfplotstableread
[col sep=&,row sep=\\]
{
Relation&Strong Belief&Moderate Belief&Weak Belief&Neither\\
Strong Relation&73.523&51.402&43.621&49.335\\
Normal Relation&21.007&41.238&47.737&28.045\\
Dependency&5.470&7.360&8.642&22.620 \\
}\strpooldialtwoarg
\pgfplotstableread
[col sep=&,row sep=\\]
{
Relation&Strong Belief&Moderate Belief&Weak Belief&Neither\\
Strong Relation&90.674&61.967&48.108&41.579\\
Normal Relation&6.736&30.492&44.324&29.211\\
Dependency&2.591&7.541&7.568&29.211\\
}\strpoolcoredialtwoarg

\pgfplotstableread
[col sep=&,row sep=\\]
{
Relation&Disbelieved&Believed&Neither\\
Strong Relation&46.685&60.275&49.335\\
Normal Relation&45.839&32.661&28.045\\
Dependency&7.475&7.064&22.620\\
}\polstrpooldialtwoarg
\pgfplotstableread
[col sep=&,row sep=\\]
{
Relation&Disbelieved&Believed&Neither\\
Strong Relation&61.004&69.575&41.579\\
Normal Relation&29.730&26.179&29.211\\
Dependency&9.266&4.245&29.211\\
}\polstrpoolcoredialtwoarg

\clearpage

\subsection{Changes in Beliefs}
\label{sec:change} 

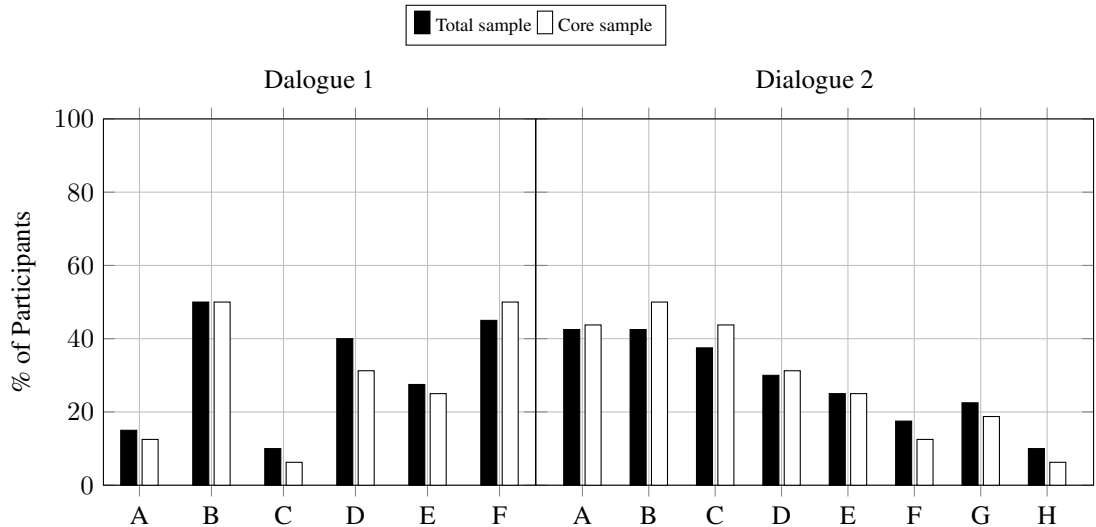
\begin{figure}[!ht]
\centering
\pgfplotsset{compat=1.9, every axis/.style={width = 0.54\textwidth, height=0.39\textwidth, grid=both, 
yticklabel style={
        /pgf/number format/fixed,
        /pgf/number format/precision=4
}, bar width = 6pt,
scaled y ticks=false, ybar, ymin=0, ymax=100,  
legend style={at={(0,1.2)},anchor=south, font=\scriptsize}, legend columns=2, 
legend image code/.code={%
      \draw[#1] (0.0cm,-0.1cm) rectangle (0.2cm,0.2cm);
    }
}}
\begin{tikzpicture}

\begin{axis}[name = awaretot,  title={Dalogue 1}, width = 0.44\textwidth, ylabel = {\% of Participants},symbolic x coords = {A,B,C,D,E,F},
xtick = {A,B,C,D,E,F},xticklabels = {A,B,C,D,E,F}]   
\addplot[fill=black] coordinates{(A,15) (B,50) (C,10) (D,40) (E,27.5) (F,45)};  
\addplot[fill=white] coordinates{(A, 12.5) (B,50) (C,6.25) (D,31.25) (E,25) (F,50)};  
\end{axis}
\begin{axis}[name = awaretot2, title = {Dialogue 2}, at=(awaretot.right of south east),symbolic x coords = {A,B,C,D,E,F,G,H},
xtick = {A,B,C,D,E,F,G,H}, xticklabels = {A,B,C,D,E,F,G,H}, yticklabels={,,,}]   
\addplot[fill=black] coordinates{(A,42.5) (B,42.5) (C,37.5) (D,30) (E,25) (F,17.5) (G,22.5) (H,10)};    
\addplot[fill=white] coordinates{(A,43.75) (B,50) (C,43.75) (D,31.25) (E,25) (F,12.5) (G,18.75) (H,6.25)}; 
\legend{Total sample, Core sample}
\end{axis}
\end{tikzpicture}
\caption{The percentage of participants in the core and total samples in both dialogues who have stated that 
they were aware of a given argument prior to the experiment} 
\label{fig:awareness}
\end{figure}

In this section we  discuss the statement awareness declared by the participants in both of the dialogues and the changes -- or rather, the lack 
of them -- in participants' opinions. In Figure \ref{fig:awareness} we can observe that the statements we have used in the dialogues 
were, in general, not common knowledge. 
 
Nevertheless, this lack of awareness of the presented pieces of information did not result in participants changing their opinions easily. 
For example, in the first dialogue, the majority of participants  -- resp. 90\% 
and 87.5\% in the total and core samples -- agreed with $B$ on all stages of the dialogue, with the remainder disagreeing with $B$ on all stages 
of the dialogue. The majority of participants -- 75\% and 68.75\% 
in the total and core samples -- 
disagreed with $A$ on all stages of the dialogue, with resp. 17.5\% and 25\% agreeing with $A$ on all stages of the dialogue. 
This leaves us only with a couple of participants \enquote{swapping sides} in the dialogue. More such people can be found 
in the second dialogue, where we have 17.5\% and 26.67\% participants changing between agreeing, disagreeing and being undecided at least 
once in the case of statement $B$ in the total and core samples, with 10\% and 20\% changing their opinions in the case of $A$. 
With all the opportunities that the participants were given to change their opinion, the number of people who did not at least once alternate between 
belief and disbelief appears extremely high. Even in a more detailed analysis, where we differentiate between all the 
declarable levels of agreement 
rather than just belief--disbelief, the significant changes appear to primarily concern argument $B$ in the first dialogue (see Table \ref{tab:belchange}). Hence, even though the dialogues appear to have certain effect on the participants, it is more visible 
on the fine--grained level. 

Despite these results, there are few notable changes that have occurred and which we would like to discuss. 
In the first dialogue, we have two participants 
who have changed their position concerning the statement $A$: one who initially disbelieved $A$, but believed it at the end, 
and one who initially believed $A$ and disbelieved it at the end (this particular person was also present in the core sample for in this dialogue). 
In both cases, the change appeared in the second step of the dialogue. The first person initially disagreed with both $A$ and $B$ and 
apparently interpreted the dialogue as discussing whether the hospital staff should be demanded to 
take the vaccine, not whether it would be best for them to take it.  
Once statement $C$ was presented, which in the opinion of this participant 
reinforced the fact that the doctors don't need the shots, he started agreeing with $A$ and no further counterarguments fixed this situation.
The second participant initially agreed with both $A$ and $B$ and provided no real explanation for his opinions. However, the appearance 
of $C$, with which the participant disagreed, led to further disagreement with $A$ throughout the rest of the dialogue. 

In the second dialogue, we have five participants that we would like to discuss: 
\begin{inparaenum}[\itshape 1\upshape)]
\item two for whom initial agreement with $B$ turned to disagreement 
without affecting $A$, 
\item  one for whom agreement (disagreement) with $B$ (with $A$) turned to disagreement (agreement), 
\item one for whom disagreement with $A$ turned into indecisiveness, and
\item one for whom indecisiveness with $B$ turned into disagreement.
\end{inparaenum} The last three participants belong to the core sample. 
In the majority of these cases, the participants, either due to some initial skepticism or lack of knowledge, became swayed by the 
 incorrect 
statements, namely $C$, $D$ and $G$. This was often paired with the inability to verify the correctness of $H$, or with interpreting it 
as an admission that the contained substances were so harmful that they had to be removed for the sake of the children\footnote{The thimerosal 
has been removed from the vaccine primarily due to social pressure and overwhelming distrust, not any scientific evidence of its harmfulness.}.  

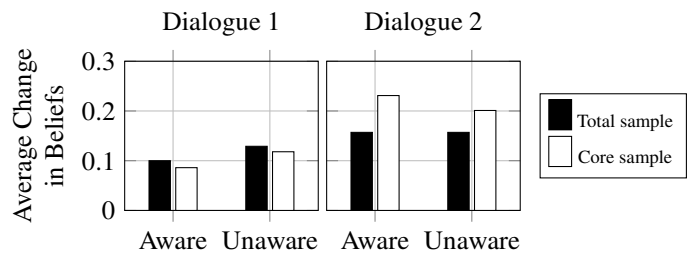
\begin{figure}[!ht]
\centering
\pgfplotsset{every axis/.style = {ybar, bar width =8pt,
symbolic x coords = {Aware, Unaware}, enlarge x limits=0.5, 
xtick ={Aware, Unaware},   
ymin = 0, ymax = 0.3, ytick = {0,0.1, 0.2, 0.3}, yticklabels={0,0.1, 0.2, 0.3},
grid=both, width=0.25\textwidth, ylabel style = {align = {center}},
legend style={at={(1.1,0.5)},anchor=west, font=\scriptsize,align=center}, legend columns=1, 
legend image code/.code={%
      \draw[#1] (0cm,-0.1cm) rectangle (0.3cm,0.3cm);
    }
}}
\begin{tikzpicture}   
\begin{axis}[name = d1, title = {Dialogue 1},  ylabel = {Average Change\\ in Beliefs} ]
\addplot[fill=black] coordinates
{ 
 (Aware,0.1)
(Unaware,0.129) 
};  

\addplot[fill=white] coordinates
{
(Aware,0.086)
(Unaware,0.118)  
};
\end{axis}   
\begin{axis}[at=(d1.right of south east),name = agetot, title = {Dialogue 2},yticklabels={,,,}]
\addplot[fill=black] coordinates
{ 
(Aware,0.157)
(Unaware,0.157) 
};

\addplot[fill=white] coordinates
{
(Aware,0.231)
(Unaware,0.201)  
};
\legend{Total sample, Core sample}
\end{axis}  

\end{tikzpicture}
\caption{Average change in beliefs in dialogues per sample, calculated using the value assignment from page~\pageref{distributions}.}
\label{fig:change}
\end{figure}

In Figure \ref{fig:change} we present the average change in beliefs of the participants, which is calculated in the following manner. 
For every participant, we have summed the differences in their beliefs throughout all the stages in the dialogue on statements they were aware of 
and those there were not w.r.t. the value assignment explained at the beginning of Section \ref{sec:results}. 
By dividing this sum by the number of statements declared as aware or unaware by a given participant 
(assuming this number is non--zero), we have obtained an average change in statements of a given type per person. Based on this we calculate 
the average change in statements of a given type per sample\footnote{Participants who did not have any unaware (resp. aware) statements 
were excluded from the appropriate calculation.}. As we may observe, the changes are quite modest, 
and relatively close in both aware and unaware statements. This is particularly visible in Dialogue 1, where the average change is smaller 
than the single smallest change possible (e.g. the difference between \textit{Strongly Agree} to \textit{Agree}, which are mapped to $6/6$ and $5/6$ 
respectively, is approximately $0.167$). The variability in Dialogue 2 is somewhat larger, though one also has to bear in mind that more statements
were presented to the participants and therefore they were given more opportunities to change their mind per statement. Nevertheless, it is possible that 
some of the observed changes were noise rather than actual modifications in beliefs caused by the dialogue. This is particularly supported by the fact
that various possible explanations for these differences, such as the appearance of an attacker, could have been paired with an increase as well as decrease 
in the beliefs declared by the participants.  

It appears that in many cases, the participants were able to evaluate the statements they were not familiar with 
by using their own knowledge and other facts that they were aware of. Unfortunately, this also happened in the cases where the knowledge of the 
participants was incorrect, such as the vaccines causing autism or flu shots causing flu. However, the complete inability to verify the pieces 
of information contained in the statements with their own knowledge had made some participants of Dialogue 2 choose to neither agree nor 
disagree with the initial statements presented in it. This might mean that depending on the difficulty of the 
dialogue, too much as well as too little personal knowledge might affect how open to change people are. 

We would also like to note that we have found participants who, on at least one occasion,  
gave answers such as \enquote{\textit{I agree/disagree with this statement}}, 
\enquote{\textit{This information is (in)accurate}} or \enquote{\textit{I think/believe this is true/false}} in the explanation tasks and refrained from 
providing  
a reason for their opinions. By considering only those cases in which the participants claimed 
they were not aware of the presented pieces of information and the stage of the dialogue they were at should not have given them sufficient 
reasons for evaluating these statements, we have found respectively 8 and 5 participants for the first and second dialogues resorting to such 
explanations. These participants also exhibited insignificant, if any, changes in their opinions. 
Therefore, it might be the case that for various reasons, they could not or did not want to provide us with further details.

\section{Related Work}

In this section we would like to review the results of other works dealing with empirical evaluation of argumentation approaches 
and  discuss other works dealing with more fine--grained models to argument acceptability. 
We begin by analyzing the work by Rahwan et al \cite{Rahwan2011}, which focuses on the issue of reinstatement. 
The authors have prepared a total of 10 short dialogues in order to study this phenomenon. A sample dialogue for the case of simple 
reinstatement is as follows:

\begin{itemize}
\item[\textbf{A}] Louis applied the brake and the brake was not faulty.
Therefore, the car slowed down.
\item[\textbf{B}] The brake fluid was empty. Therefore, the brake was
faulty.
\item[\textbf{C}] The car had just undergone maintenance service. Therefore, the brake fluid was not empty.
\end{itemize}

The dialogue was split into three stages, where first \textbf{A}, then \textbf{A} and \textbf{B}, and finally all \textbf{A}, \textbf{B} and \textbf{C} 
were presented. For each problem, the participants had to choose a value from a 7--point scale ranging from \textit{Certainly false} 
to \textit{Certainly true} that best describes their confidence in the conclusion of \textbf{A}. The obtained results support the notions of defeat 
and reinstatement, that is, that the belief in \textbf{A} decreases once it is defeated and increases when it is defended, though still remaining 
significantly lower than prior to the defeat. 

This study, similarly to ours, lends support to the use of more fine--grained approaches towards describing the beliefs of the participants, such as
the epistemic approaches. However, as we can observe, the dialogues used in this study were much simpler and shorter than ours, and unlikely 
to be affected by any subjective views of the participants. The results show that reinstatement as such does occur 
and thus support the idea behind semantics such as grounded or preferred. However, at this point it cannot be claimed that they are applicable 
in a more complex setting and that they are a general proof for the use of argumentation semantics in computational persuasion. 
 
The next study we would like to consider is by Cerutti et al \cite{Cerutti2014}, which again studies various forms of reinstatement. 
The authors have prepared two texts -- a base one and an extended one -- on 
four topics (weather forecast, political debate, used car sale and romantic relationship). For example, the texts prepared for the political debate scenario 
are as follows: 

\begin{itemize}
\item[\textbf{Base Case:}]

In a TV debate, the politician AAA argues that if Region X becomes independent then X’s citizens will be poorer than now. Subsequently, financial
expert Dr. BBB presents a document, which scientifically shows that Region X
will not be worse off financially if it becomes independent. 

\item[\textbf{Extended Case:}] 

In a TV debate, the politician AAA argues that if Region X becomes independent then X’s citizens will be poorer than now. Subsequently, financial
expert Dr. BBB presents a document, which scientifically shows that Region X
will not be worse off financially if it becomes independent. After that, the moderator of the debate reminds BBB of more recent research by several important
economists that disputes the claims in that document.
\end{itemize}

The participants are given one of the eight scenarios and then asked to determine which of the following positions they think is accurate:

\begin{itemize}
\item $P_{A}$: I think that AAA’s position is correct (e.g. \enquote{X’s citizens will
be poorer than now}). 
\item   $P_{B}$: I think that BBB’s position is correct (e.g. \enquote{X’s citizens will
not be worse off financially}). 
\item  $P_{U}$: I cannot determine if either AAA’s or BBB’s position is
correct (e.g. \enquote{I cannot conclude anything about Region X’s finances}). 
\end{itemize}

Next, the participants were asked certain questions concerning the text. In particular, concerning the agreement, the participants are 
asked \enquote{How much do you agree with the following statements?} and respond on a 7--point scale from \textit{Disagree} 
to \textit{Agree} for each statement. 

The results show that the majority of participants in the base case scenarios agree with position 
$P_{B}$ and with position $P_{U}$ in the extended scenarios. They suggest a correspondence between the formal theory used by the authors 
and its representation in natural language. However, additional analyses show that there are some significant deviations, 
apparently caused by the personal knowledge of the participants. For example, while $P_{B}$ was the most common choice in the base weather forecast 
scenario, there were participants who chose $P_{U}$ and explained their decision with \enquote{\textit{All weather forecasts are
notoriously inaccurate}}. 

Our study, similarly to this one, points to the personal knowledge of the participants affecting their decisions. However, due to a different methodology, 
it is difficult to compare the other findings. We have presented the participants with an ongoing dialogue and tried to monitor the changes 
in their beliefs at every step of the way, similarly as in \cite{Rahwan2011}. In this study, a given participant receives only one text to evaluate, i.e. 
the base and extended scenarios are judged by different people. We therefore cannot claim that if a person was first shown the base case, and then 
the extended one, the $P_{U}$ choice would still be more common than $P_{B}$, which is what our results would point to. 

Another interesting study worth considering is \cite{Rosenfeld16}, which focuses on exploring the abilities of argumentation theory, 
relevance heuristics, machine and transfer learning for predicting the argument choices of  
participants, with a particular focus on the machine learning. 
Argumentation theory is verified using three experiments, in which the dialogues are used to construct bipolar argumentation frameworks 
and the sets of arguments selected by the participants are contrasted with grounded, preferred and stable extensions. 
In the first experiment consisting of 6 scenarios, the authors create bipolar argumentation frameworks which are not known to the participants, 
present two standpoints from two parties 
and ask the participants to choose which of the additional four arguments they would use next if they were one of the deliberants 
in the discussion. In the second experiment, real life data is annotated and structured into a bipolar argumentation framework that is later analyzed. 
In the third experiment, a chat bot is created, aimed at discussing flu vaccination with the participants. In this case, both the chat and the participant 
can only use arguments from a predefined list. Finally, in an additional experiment, an argument theory agent was implemented in order to provide 
suggestions to the participants during a two--person chat.  

The authors report that a substantial part of the results (or in some cases, even the majority) do not 
conform to the results predicted by the semantics. It is worth mentioning that the  stated adherence to the 
conflict--free extension--based 
semantics is 78\% and is similar to our results concerning the rational postulate, which corresponds to this semantics. 
The created agent was also considered helpful only by a small percentage of the participants. 
Nevertheless, the causes for such 
behaviour of the semantics are not investigated, and the participants were not allowed to explain their 
decisions (the first and the third experiment) or there was no possibility
to ask them for further input (the second experiment). Moreover, unlike in our study, the participants were evaluated against the graphs constructed by
the authors or annotators. As shown by our results, they do not necessarily reflect how the participants view the relations between the arguments. 
Finally, there is no discussion concerning whether these particular bipolar argument framework semantics used in this experiment \cite{amgoud08} 
are applicable. 
The stable and conflict--free semantics in this work are based on direct and supported attacks, and only the direct ones need to be defended from
and can be used for defence. Taking into account the fact that this approach has been superseded by a number of different methods 
since it has been introduced, the presented results indicate that these particular semantics are not useful in modelling of the user behaviour, 
rather than there exists a deeper issue within the argumentation itself. 

The aforementioned studies focused on the empirical evaluation of certain phenomenona in argumentation. However, 
there are also other studies, which focus more on the behavioural methods or computational linguistics perspective
rather than the argumentation perspective and can be seen as complementary to ours. 
These works often use arguments sourced from e.g. social media,  
and analyze the relation between the persuasiveness of an argument and its traits, the persuasiveness of an argument or relations between 
the arguments and the personality or 
the emotions of the participant, and more. Some of them, such as \cite{lukin}, also point to the importance of the 
prior knowledge in changing one's beliefs. These studies are an  
important line of work which can be harnessed in creating computational persuasion methods that are tailored to the participants and which can provide  
guidance in transforming logical arguments to natural language arguments. Therefore, they should be considered in the next steps of our work. 
We refer the readers to \cite{villata17,lukin} for further details. 
  
Although in this work we have mostly focused on the analysis of epistemic probabilities, we would like to note that 
this is not the only approach allowing a more fine--grained perspective on argument acceptability. 
There also are other interesting methods that allow score assignments, 
such as certain forms of preference or weight--based argumentation (see \cite{general,Bonzon16} for an overview).  
However, many of these works share one common problem, namely that 
the values associated with the arguments are quite abstract and do not have a meaning of their own. 
Even though using them we can, for example, 
state that a participant agrees with one argument more than with another, we cannot state whether he agrees with it
in the first place. Let us consider a framework containing a single argument
and no attacks. Giving this argument a preference $0.3$ has no effect, as it will always be accepted in any extension. Giving it the  epistemic probability of $0.3$, which is interpreted as disbelief, will lead it to it being always rejected. Consequently, despite certain 
similarities on the structural level, the semantics of the preference and weight--based frameworks are quite different from 
epistemic probabilities.

\section{Conclusions}  

Our work, through empirically verifying certain prevailing assumptions in argumentation, is about informing the design of formalisms that can model 
how participants think about arguments arising in a dialogue. We observe that handling the uncertainty concerning both the participants' opinions 
about arguments and the structure of the graph is of critical importance, and that a Dung framework equipped only with the standard semantics 
is insufficient to represent the views of the participants. Our results can be summarized in the following manner:  

\begin{itemize}
\item[\textbf{Observation 1}] The data supports the use of the \textbf{constellation approach} to probabilistic argumentation 
for modelling the argument graphs representing the views of dialogue participants.
In particular, in Section \ref{sec:graphanalysis} 
we have seen that people may interpret statements and relations between them differently and without adhering to the 
intended relations. Furthermore, their personal knowledge can affect their perception and evaluation of the dialogue.
Thus, the constellation approach can represent our uncertainty about the argument graphs describing our opponents views.

\item[\textbf{Observation 2}] People may explicitly declare
that two given statements are connected, however, they might not be sure of the exact nature of the relation between them. In Section \ref{sec:graphanalysis} 
we have observed that the portion of the graphs declared by the participants that were not clarified (i.e. contained at least one relation marked as dependent) reached even 70\%. 
We therefore also need to express the uncertainty that a person has about his or her own views. Although for the purpose of the analysis we have introduced 
the notion of a tripolar framework in order to be represent such situations, this issue can potentially be addressed with the constellation approach to 
argumentation. By this we understand that associated with a given unclarified framework we can have a number of graphs which interpret the dependent 
links as attacking or supporting, each with a given probability that could, for example, be obtained from the graph distribution of all participants.

\item[\textbf{Observation 3}] The data supports the use of  the \textbf{epistemic approach} to probabilistic argumentation. 
In Section \ref{sec:postsat} we calculated how many values the participants were using throughout the dialogues, and we could have observed 
that in most of the cases three values were insufficient to represent the participants' opinions. 
Another of our important observations concerns the adherence to the epistemic postulates, analyzed in the aforementioned section.  
While classical semantics tend to represent a number of 
properties at the same time, a single postulate tends to focus on a single aspect, as seen in Section \ref{sec:epistemic}. They therefore allow 
a more detailed view on the participant behaviour and can allow us to analyze the cases in which classic semantics may fail to explain it. 
The epistemic postulates can also provide more feedback to argumentation systems, such as for computational persuasion, than normal semantics do. 
For example, the low performance of the argumentation semantics such as complete does not really inform the system what aspect of the participant 
reasoning does not meet the semantics requirements. In contrast, the high performance of the preferential postulates and low performance 
of the explanatory ones, can inform the system that it has insufficient information 
about the participant's personal knowledge and that it should proceed with querying him or her for further data. 
This is particularly important as our data also shows that people use their own personal knowledge in order to make 
judgments and might not necessarily disclose it. 

The extended epistemic approach \cite{probattack} would also allow us to model situations where the way a given 
relation is perceived is not necessarily tightly related to the way how much we believe or disbelieve its source, 
which as seen in Section \ref{sec:relcorel}, does tend to occur. 
It is likely that there are more properties of an argument or statement, such as how detailed and informative it is, which affect how 
the relations carried out by it are seen by people. 
While our analysis concerned the impact of the source of a relation on the relation and vice versa, 
it is possible that both source and target can affect the relation and thus approaches from \cite{CLS13,KL11} could be
verified.
Thus, these issues need further investigation.  
 
\item[\textbf{Observation 4}] The data supports the use of \textbf{bipolar argumentation frameworks}. 
In Section \ref{sec:graphanalysis} we have observed that the participants explicitly view certain relations as supporting and that the notion of 
defence does not account for all of the positive relations  
that the participants have identified between the presented statements. In particular, we could have observed that there are new support relations 
arising in the context of the dialogue, such as support coming from statements working towards the same goal.  

\item[\textbf{Observation 5}] The data supports the use of \textbf{bipolar argumentation in combination with the prudent/careful approaches}. 
In Section \ref{sec:graphanalysis} we have observed that many additional attacks perceived by the participants can be explained by 
the existing notions of indirect conflicts in these settings. They can therefore be used to model auxiliary conflicts arising in the context of a dialogue, but 
not necessarily created on the logical level.  

\item[\textbf{Observation 6}] The data shows that \textbf{people use their own personal knowledge} in order to make judgments and
might not necessarily disclose it. In Section \ref{sec:postsat} we have shown how the differences between the declared and expanded graphs 
become visible on the postulates highly sensitive to personal knowledge, such as the ones belonging to the explanatory family. Additionally, 
in Section \ref{sec:change} we have noted that a number of participants were able 
to evaluate the presented statements despite the lack of awareness of certain information. This, combined with them not providing any explanations  
for their opinions, may mean that argumentation systems need to handle participants not willing to share their knowledge. 
 
\item[\textbf{Observation 7}] The data shows that presenting a new and correct piece of information that a given person was not aware of \textbf{does not 
necessarily lead to changing that person's beliefs}. In Section \ref{sec:change} we have observed that throughout the dialogues, not many participants 
have changed their opinions in a significant manner. Moreover, quite often those who have, have done so under the influence of incorrect information 
and presenting them with the correct data has not managed to rectify their opinions. 
\end{itemize}
 
Our exploratory study shows that the most common approaches to argumentation might be too simplistic in order to adequately grasp human reasoning. 
However, we do not believe that the argumentation theory as a field is insufficient altogether. In particular, we have highlighted 
the correspondence of the obtained results to various, less common formalisms, such as probabilistic and bipolar frameworks, and prudent and careful approaches. 
Consequently, these methods could be merged in the future in order to develop abstract argumentation tools that can be used 
in dialogical argumentation with more success. Nevertheless, further and more specialized studies concerning our observations should be carried out. In particular, our experiment could be seen as dynamic, as it concerned two dialogues between different parties. It would be interesting to observe whether our findings could be replicated in a more static setting, where arguments are presented randomly and not in the context of a dialogue. For example, this shift could affect the perception of the indirect attacks, defenses and supports between different statements. We believe further studies will provide more insight into this matter.

\newpage

\bibliographystyle{plain}
\bibliography{dialogue} 
\pagenumbering{gobble}
\newpage
\section{Statistics Appendix}
\label{sec:stats}

\subsection{Dialogue 1}

\subsubsection{Postulate Satisfaction}
\label{stats:postulates1}

\begin{table}[h]
\centering
\resizebox{\textwidth}{!}{
\begin{tabular}{|c||P{1cm}|P{1.3cm}|P{1cm}|P{1.3cm}|P{1cm}|P{1.3cm}|P{1cm}|P{1.3cm}|P{1cm}|P{1.3cm}|P{1cm}|P{1.3cm}|} 
\hline
	 \multirow{2}{*}{Total Sample}		&\tframec{Intended vs Declared}	& \tframec{Intended vs Expanded} & \tframec{Intended vs Common}& \tframec{Declared vs Expanded}& \tframec{Declared vs Common}		& \tframec{Expanded vs Common}	\\
\cline{2-13}
 &Z		&		p--value	&Z		&		p--value	&Z		&		p--value	&Z		&		p--value		&Z		&		p-value	&Z		&		p--value	\\
\hline 
Preferential	&1.428	&\cellg{0.149}	&1.472	&\cellg{0.143}	&1.999	&\cellg{0.125}	&1.000	&\cellg{1.000}		&-0.221	&\cellg{0.819}	&-0.544	&\cellg{0.601}	\\
 \hline 
Rational		&2.417	&		0.019	&2.426	&		0.018	&1.732	&\cellg{0.250}	&1.000	&\cellg{1.000}		&-2.017	&\cellg{0.061}	&-2.199	&		0.036	\\
 \hline 
Strict		&1.609	&\cellg{0.115}	&1.795	&\cellg{0.079}	&1.732	&\cellg{0.250}	&1.000	&\cellg{1.000}		&-1.079	&\cellg{0.306}	&-1.364	&\cellg{0.203}	\\
 \hline 
Protective	&1.102	&\cellg{0.289}	&1.102	&\cellg{0.289}	&1.732	&\cellg{0.250}	&NA		&\cellg{NA}			&-0.608	&\cellg{0.577}	&-0.608	&\cellg{0.577}	\\
 \hline 
Restrained	&0.880	&\cellg{0.411}	&0.880	&\cellg{0.411}	&1.414	&\cellg{0.500}	&NA		&\cellg{NA}			&-0.608	&\cellg{0.577}	&-0.608	&\cellg{0.577}	\\
 \hline 
Coherent		&-1.783	&\cellg{0.075}	&-0.252	&\cellg{0.804}	&NA		&\cellg{NA}		&2.447	&		0.031		&1.783	&\cellg{0.075}	&0.252	&\cellg{0.804}	\\
 \hline 
Involutary	&-1.344	&\cellg{0.183}	&-0.461	&\cellg{0.650}	&NA		&\cellg{NA}		&2.235	&\cellg{0.063}		&1.344	&\cellg{0.183}	&0.461	&\cellg{0.650}	\\
 \hline 
Justifiable		&2.917	&		0.004	&2.917	&		0.004	&NA		&\cellg{NA}		&NA		&\cellg{NA}			&-2.917	&		0.004	&-2.917	&		0.004	\\
 \hline 
Semi Optimistic	&1.799	&\cellg{0.064}	&1.420	&\cellg{0.144}	&-0.296	&\cellg{1.000}	&-1.414	&\cellg{0.500}		&-1.462	&\cellg{0.146}	&-0.990	&\cellg{0.331}	\\
 \hline 
Semi Founded	&4.107	&		0.000	&1.847	&\cellg{0.066}	&-1.732	&\cellg{0.250}	&-3.304	&		0.001		&-4.682	&		0.000	&-2.225	&		0.025	\\
 \hline 
Founded		&3.353	&		0.001	&0.956	&\cellg{0.351}	&-2.449	&		0.031	&-3.153	&		0.002		&-4.412	&		0.000	&-1.870	&\cellg{0.062}	\\
 \hline 
Optimistic		&3.213	&		0.001	&1.150	&\cellg{0.260}	&-2.000	&\cellg{0.125}	&-2.823	&		0.008		&-4.084	&		0.000	&-1.735	&\cellg{0.083}	\\
 \hline 
Demanding	&-1.083	&\cellg{0.354}	&-2.383	&		0.016	&1.999	&\cellg{0.125}	&-2.447	&		0.031		&2.299	&		0.026	&3.025	&		0.002	\\
 \hline 
Guarded		&4.220	&		0.000	&2.109	&		0.034	&-1.414	&\cellg{0.500}	&-3.153	&		0.002		&-4.682	&		0.000	&-2.356	&		0.017	\\
 \hline 
Discharging	&4.220	&		0.000	&2.133	&		0.032	&-1.414	&\cellg{0.500}	&-3.153	&		0.002		&-4.682	&		0.000	&-2.363	&		0.017	\\
 \hline 
Trusting		&4.224	&		0.000	&2.203	&		0.026	&-1.732	&\cellg{0.250}	&-3.153	&		0.002		&-4.772	&		0.000	&-2.559	&		0.009	\\
 \hline 
Anticipating	&4.224	&		0.000	&2.203	&		0.026	&-1.732	&\cellg{0.250}	&-3.153	&		0.002		&-4.772	&		0.000	&-2.559	&		0.009	\\
\hline 
\hline
	 \multirow{2}{*}{Core Sample}		&\tframec{Intended vs Declared}	& \tframec{Intended vs Expanded} & \tframec{Intended vs Common}& \tframec{Declared vs Expanded}& \tframec{Declared vs Common}		& \tframec{Expanded vs Common}	\\
\cline{2-13}
  	&Z		&		p--value	&Z		&		p--value	&Z		&		p--value	&Z		&		p--value		&Z		&		p-value	&Z		&		p--value	\\
\hline 
Preferential	&1.732	&\cellg{0.250}	&1.730	&\cellg{0.250}		&1.000		&\cellg{1.000}	&1.000		&\cellg{1.000}		&-0.500		&\cellg{1.000}		&-1.414		&\cellg{0.500}	\\
 \hline                                                                                                                                                                               
Rational		&1.730	&\cellg{0.250}	&1.730	&\cellg{0.250}		&1.000		&\cellg{1.000}	&1.000		&\cellg{1.000}		&-1.414		&\cellg{0.500}		&-1.732		&\cellg{0.250}	\\
 \hline                                                                                                                                                                               
Strict		&1.414	&\cellg{0.500}	&1.732	&\cellg{0.250}		&1.000		&\cellg{1.000}	&1.000		&\cellg{1.000}		&-0.577		&\cellg{1.000}		&-1.414		&\cellg{0.500}	\\
 \hline                                                                                                                                                                               
Protective	&0.318	&\cellg{1.000}	&0.318	&\cellg{1.000}		&1.000		&\cellg{1.000}	&NA			&\cellg{NA}		&0.138		&\cellg{0.750}		&0.138		&\cellg{0.750}	\\
 \hline                                                                                                                                                                               
Restrained	&-0.138	&\cellg{0.750}	&-0.138	&\cellg{0.750}		&NA			&\cellg{NA}	&NA			&\cellg{NA}		&0.138		&\cellg{0.750}		&0.138		&\cellg{0.750}	\\
 \hline                                                                                                                                                                               
Coherent		&-0.138	&\cellg{0.750}	&1.208	&\cellg{0.375}		&NA			&\cellg{NA}	&1.413		&\cellg{0.500}		&0.138		&\cellg{0.750}		&-1.208		&\cellg{0.375}	\\
 \hline                                                                                                                                                                               
Involutary	&0.615	&\cellg{0.750}	&1.033	&\cellg{0.500}		&NA			&\cellg{NA}	&1.000		&\cellg{1.000}		&-0.615		&\cellg{0.750}		&-1.033		&\cellg{0.500}	\\
 \hline                                                                                                                                                                               
Justifiable		&1.730	&\cellg{0.250}	&1.730	&\cellg{0.250}		&NA			&\cellg{NA}	&NA			&\cellg{NA}		&-1.730		&\cellg{0.250}		&-1.730		&\cellg{0.250}	\\
 \hline                                                                                                                                                                               
SemiOptimistic	&0.615	&\cellg{0.750}	&0.034	&\cellg{1.000}		&-1.000		&\cellg{1.000}	&-1.000		&\cellg{1.000}		&-1.033		&\cellg{0.500}		&-0.034		&\cellg{1.000}	\\
 \hline                                                                                                                                                                               
SemiFounded	&0.841	&\cellg{0.500}	&-1.587	&\cellg{0.124}		&-1.413		&\cellg{0.500}	&-2.617		&0.016			&-1.995		&\cellg{0.125}		&1.313		&\cellg{0.227}	\\
 \hline                                                                                                                                                                               
Founded		&0.382	&\cellg{1.000}	&-2.169	& 	0.031		&-1.732		&\cellg{0.250}	&-2.617		&0.016			&-1.996		&\cellg{0.125}		&1.671		&\cellg{0.094}	\\
 \hline                                                                                                                                                                               
Optimistic		&0.382	&\cellg{1.000}	&-1.932	&\cellg{0.055}		&-1.732		&\cellg{0.250}	&-2.431		&0.031			&-1.996		&\cellg{0.125}		&1.361		&\cellg{0.191}	\\
 \hline                                                                                                                                                                               
Demanding	&1.000	&\cellg{1.000}	&-0.046	&\cellg{1.000}		&NA			&\cellg{NA}	&-1.000		&\cellg{1.000}		&-1.000		&\cellg{1.000}		&0.046		&\cellg{1.000}	\\
 \hline                                                                                                                                                                               
Guarded		&0.841	&\cellg{0.500}	&-1.390	&\cellg{0.176}		&-1.413		&\cellg{0.500}	&-2.431		&0.031			&-1.995		&\cellg{0.125}		&1.089		&\cellg{0.313}	\\
 \hline                                                                                                                                                                               
Discharging	&0.841	&\cellg{0.500}	&-1.389	&\cellg{0.176}		&-1.413		&\cellg{0.500}	&-2.431		&0.031			&-1.995		&\cellg{0.125}		&1.115		&\cellg{0.266}	\\
 \hline                                                                                                                                                                               
Trusting		&0.841	&\cellg{0.500}	&-1.442	&\cellg{0.152}		&-1.730		&\cellg{0.250}	&-2.431		&0.031			&-2.226		&\cellg{0.063}		&0.855		&\cellg{0.426}	\\
 \hline                                                                                                                                                                               
Anticipating	&0.841	&\cellg{0.500}	&-1.442	&\cellg{0.152}		&-1.730		&\cellg{0.250}	&-2.431		&0.031			&-2.226		&\cellg{0.063}		&0.855		&\cellg{0.426} 	\\
 \hline 
\end{tabular}
}
\caption{The results of the Wilcoxon signed rank test with Pratt adjustment for a given postulate evaluated on two separate graphs on the total and core samples in Dialogue 1. 
The results have been obtained using R library coin. 
We have highlighted the fields with p--value greater than $0.05$, i.e. those for which we cannot reject the null hypothesis. In case the adherence rate for a given postulate was identical 
for two graphs for all participants, the algorithm has returned NA.}
\label{tab:postulatewilcox1}
\end{table}

\clearpage
\subsubsection{Statements vs. Relations: Effect of a Relation on its Source}
\label{stats:relsource1}

\begin{table}[h]
\centering
\resizebox{\textwidth}{!}{
\begin{tabular}{|c||c|c|c|c|c|c|c|c|c|c|c|c|} 
\hline
 \multirow{2}{*}{Total Sample}	 &	\multicolumn{3}{c|}{-} & \multicolumn{3}{c|}{+ +} & \multicolumn{3}{c|}{+} & \multicolumn{3}{c|}{?}\\
\cline{2-13}
 &	G--value	&	DF	&	p--value	&	G--value	&	DF	&	p--value	&	G--value	&	DF	&	p--value	&	G--value	&	DF	&	p--value	\\ \hline
- -	&	153.455	&	7	&	0.000	&	5.076	&	7	&	\cellg{0.651}	&	89.014	&	7	&	0.000	&	94.698	&	7	&	0.000	\\\hline
-	&	 	&	 	&	 	&	145.156	&	7	&	0.000	&	15.290	&	7	&	0.032	&	52.819	&	7	&	0.000	\\\hline
+ +	&	 	&	 	&	 	&	 	&	 	&	 	&	82.477	&	7	&	0.000	&	84.490	&	7	&	0.000	\\\hline
+	&	 	&	 	&	 	&	 	&	 	&	 	&	 	&	 	&	 	&	37.405	&	7	&	0.000	\\
\hline
\hline
	 \multirow{2}{*}{Core Sample}	&	\multicolumn{3}{c|}{-} & \multicolumn{3}{c|}{+ +} & \multicolumn{3}{c|}{+} & \multicolumn{3}{c|}{?}\\
\cline{2-13}
&	G--value	&	DF	&	p--value	&	G--value	&	DF	&	p--value	&	G--value	&	DF	&	p--value	&	G--value	&	DF	&	p--value	\\ \hline
- -	&	85.098	&	7	&	0.000	&	10.231	&	7	&	\cellg{0.176}	&	77.735	&	7	&	0.000	&	69.700	&	7	&	0.000	\\\hline
-	&	 	&	 	&	 	&	38.538	&	7	&	0.000	&	6.837	&	7	&	\cellg{0.446}	&	26.354	&	7	&	0.000	\\\hline
+ +	&	 	&	 	&	 	&	 	&	 	&	 	&	45.756	&	7	&	0.000	&	30.842	&	7	&	0.000	\\\hline
+	&	 	&	 	&	 	&	 	&	 	&	 	&	 	&	 	&	 	&	35.980	&	7	&	0.000	\\
\hline
\end{tabular}
}
\caption{Results of G--test for independence between relations of given type on total and core samples. 
 We use the following abbreviations for the relation types:  
A good reason against (- -), A somewhat good reason against (-), A good reason for (++), A somewhat good reason for (+), Somewhat related, but 
can't say how (?). DF stands for degrees of freedom. 
We have highlighted the fields with p--value greater than $0.05$, i.e. those for which we cannot reject the null hypothesis. 
These results were obtained using library Deducer (likelihood.test function) in R.}
\label{tab:d1indeprel}
\end{table}

\begin{table}[h]
\centering
\resizebox{\textwidth}{!}{
\begin{tabular}{|c||c|c|c|c|c|c|c|c|c|c|} 
\cline{2-11} 
\multicolumn{1}{c|}{} 	&	\multicolumn{5}{c|}{Total Sample}    	&\multicolumn{5}{c|}{Core Sample} \\			
\cline{2-11} 				
\multicolumn{1}{c|}{} & - -	&	-	&	+ +	&	+	&	?	&	- -	&	-	&	+ +	&	+	&	?	\\ \hline
$\chi^2$	&	401.219	&	66.134	&	393.783	&	42.745	&	40.784	&	340.297	&	44.036	&	108.651	&	57.959	&	17.267	\\ \hline
DF	&	7	&	7	&	7	&	7	&	7	&	7	&	7	&	7	&	7	&	7	\\ \hline
p--value	&	0.000	&	0.000	&	0.000	&	0.000	&	0.000	&	0.000	&	0.000	&	0.000	&	0.000	&	0.016	\\ \hline 
\end{tabular}
}
\caption{Results of chi--squared goodness of fit test on levels of agreement of sources of a relation of a given type 
in Dialogue 1. 
 We use the following abbreviations for the relation types:  
A good reason against (- -), A somewhat good reason against (-), A good reason for (++), A somewhat good reason for (+), Somewhat related, but 
can't say how (?). DF stands for degrees of freedom. 
We have highlighted the fields with p--value greater than $0.05$, i.e. those for which we cannot reject the null hypothesis. 
These results were obtained using R.}
\label{tab:d1goodnessrel}
\end{table}

\begin{table}[h]
\centering
\resizebox{\textwidth}{!}{
\begin{tabular}{|c||c|c|c|c|c|c|c|c|c|c|c|c|c|}  
\hline
\multicolumn{13}{|c|}{Relations pooled according to strength}\\
\hline 
\multirow{3}{*}{\shortstack[c]{Agreement pooled \\according to strength}} &	\multicolumn{6}{c|}{Total Sample} & \multicolumn{6}{c|}{Core Sample} \\
 \cline{2-13}
 &	\multicolumn{3}{c|}{Normal Relation} & \multicolumn{3}{c|}{Dependency} &	\multicolumn{3}{c|}{Normal Relation} & \multicolumn{3}{c|}{Dependency} \\
\cline{2-13}
  &	G--value	&	DF	&	p--value	&	G--value	&	DF	&	p--value	&	G--value	&	DF	&	p--value	&	G--value	&	DF	&	p--value	\\ \hline 
Strong Relation	&	145.715	&	3	&	0.000	&	89.373	&	3	&	0.000	&	66.202	&	3	&	0.000	&	61.424	&	3	&	0.000	\\ \hline
Normal Relation&	 	&	 	&	 	&	47.221	&	3	&	0.000	&	 	&	 	&	 	&	19.351	&	3	&	0.000	
\\ \hline
\hline 
\multirow{3}{*}{\shortstack[c]{Agreement pooled \\according to polarity}}  &	\multicolumn{6}{c|}{Total Sample} & \multicolumn{6}{c|}{Core Sample} \\
 \cline{2-13}
 &	\multicolumn{3}{c|}{Normal Relation} & \multicolumn{3}{c|}{Dependency} &	\multicolumn{3}{c|}{Normal Relation} & \multicolumn{3}{c|}{Dependency} \\
\cline{2-13}
&	G--value	&	DF	&	p--value	&	G--value	&	DF	&	p--value	&	G--value	&	DF	&	p--value	&	G--value	&	DF	&	p--value	\\ \hline 
Strong Relation	&	113.155	&	2	&	0.000	&	103.253	&	2	&	0.000	&	86.140	&	2	&	0.000	&	53.169	&	2	&	0.000	\\ \hline 
Normal Relation	&		&		&		&	25.713	&	2	&	0.000	&		&		&		&	31.549	&	2	&	0.000	\\ \hline 
\hline
\multicolumn{13}{|c|}{Relations pooled according to polarity} \\
\hline 
\multirow{3}{*}{\shortstack[c]{Agreement pooled \\according to polarity}} &	\multicolumn{6}{c|}{Total Sample} & \multicolumn{6}{c|}{Core Sample} \\
 \cline{2-13}
&	\multicolumn{3}{c|}{Support} & \multicolumn{3}{c|}{Dependency} &	\multicolumn{3}{c|}{Support} & \multicolumn{3}{c|}{Dependency} \\
\cline{2-13}
&	G--value	&	DF	&	p--value	&	G--value	&	DF	&	p--value	&	G--value	&	DF	&	p--value	&	G--value	&	DF	&	p--value	\\ \hline 
Attack 	&	0.673	&	2	&\cellg{0.714}&	48.684	&	2	&	0.000	&	4.906	&	2	&\cellg{0.086}&	35.441	&	2	&	0.000	\\ \hline
Support 	&	 	&		&		&	54.445	&	2	&	0.000	&		&		&		&	29.528	&	2	&	0.000	\\ \hline
 \hline
\multirow{3}{*}{\shortstack[c]{Agreement pooled \\according to strength}}&	\multicolumn{6}{c|}{Total Sample} & \multicolumn{6}{c|}{Core Sample} \\
 \cline{2-13}
  &	\multicolumn{3}{c|}{Support} & \multicolumn{3}{c|}{Dependency} &	\multicolumn{3}{c|}{Support} & \multicolumn{3}{c|}{Dependency} \\
\cline{2-13}
 &	G--value	&	DF	&	p--value	&	G--value	&	DF	&	p--value	&	G--value	&	DF	&	p--value	&	G--value	&	DF	&	p--value	\\ \hline 
Attack 	&	7.553	&	3	&	\cellg{0.056}	&	48.315	&	3	&	0.000	&	2.302	&	3	&\cellg{0.512}	&	39.198	&	3	&	0.000	\\ \hline
Support 	&	 	&		&		&	56.617	&	3	&	0.000	&		&		&		&	29.107	&	3	&	0.000	\\ \hline
\end{tabular}
}
\caption{Results of G--test for independence between relations of given type in Dialogue 1 on total and core samples according 
to a given pooling. DF stands for degrees of freedom. 
We have highlighted the fields with p--value greater than $0.05$, i.e. those for which we cannot reject the null hypothesis. 
These results were obtained using library Deducer (likelihood.test function) in R.}
\label{tab:d1indeprelpool}
\end{table}
 
\begin{table}[h]
\centering
\resizebox{\textwidth}{!}{
\begin{tabular}{|c||c|c|c|c|c|c|} 
\hline
\multicolumn{7}{|c|}{Relations pooled according to strength}\\
\hline 
\multirow{2}{*}{\shortstack[c]{Agreement pooled \\according to strength}}	&	\multicolumn{3}{c|}{Total Sample}    	&\multicolumn{3}{c|}{Core Sample} \\			
\cline{2-7} 				
 &	Strong Relation	&	Normal Relation	&	Dependency	&	Strong Relation	&	Normal Relation	&	Dependency	\\ \hline	
$\chi^2$	&	472.913	&	78.459	&	30.627	&	297.046	&	46.399	&	9.853	\\ \hline	
DF	&	3	&	3	&	3	&	3	&	3	&	3	\\ \hline	
p--value	&	0.000	&	0.000	&	0.000	&	0.000	&	0.000	&	0.020	\\ \hline	
\hline 
\multirow{2}{*}{\shortstack[c]{Agreement pooled \\according to polarity}}	&	\multicolumn{3}{c|}{Total Sample}    	&\multicolumn{3}{c|}{Core Sample} \\			
\cline{2-7} 		
 &	Strong Relation	&	Normal Relation	&	Dependency	&	Strong Relation	&	Normal Relation	&	Dependency	\\ \hline	
$\chi^2$	&	519.513	&	136.832	&	11.412	&	185.490	&	97.475	&	0.960	\\	\hline
DF	&	2	&	2	&	2	&	2	&	2	&	2	\\	\hline
p--value	&	0.000	&	0.000	&	0.003	&	0.000	&	0.000	&\cellg{0.619}	\\	\hline
\hline
\multicolumn{7}{|c|}{Relations pooled according to polarity}\\
\hline 
\multirow{2}{*}{\shortstack[c]{Agreement pooled \\according to polarity}} 	&	\multicolumn{3}{c|}{Total Sample}    	&\multicolumn{3}{c|}{Core Sample} \\			
\cline{2-7} 				
 &	Attack	&	Support	&	Dependency	&	Attack	&	Support	&	Dependency	\\ \hline	
$\chi^2$	&	272.556	&	251.086	&	11.412	&	120.544	&	64.511	&	0.960	\\ \hline
DF	&	2	&	2	&	2	&	2	&	2	&	2	\\ \hline
p--value	&	0.000	&	0.000	&	0.003	&	0.000	&	0.000	& \cellg{0.619}	\\ \hline
\hline 
\multirow{2}{*}{\shortstack[c]{Agreement pooled \\according to strength}}	&	\multicolumn{3}{c|}{Total Sample}    	&\multicolumn{3}{c|}{Core Sample} \\			
\cline{2-7} 				
 &	Attack	&	Support	&	Dependency	&	Attack	&	Support	&	Dependency	\\ \hline	
$\chi^2$	&	184.929	&	229.567	&	30.627	&	185.019	&	79.000	&	9.853	\\ \hline
DF	&	3	&	3	&	3	&	3	&	3	&	3	\\ \hline
p--value	&	0.000	&	0.000	&	0.000	&	0.000	&	0.000	&	0.020	\\ \hline
\end{tabular}
}
\caption{Results of chi--squared goodness of fit test on levels of agreement of sources of a relation of a given type 
in Dialogue 1 on total and core samples according to a given pooling. DF stands for degrees of freedom. 
We have highlighted the fields with p--value greater than $0.05$, i.e. those for which we cannot reject the null hypothesis. 
These results were obtained using R.}
\label{tab:d1goodnessrelpool}
\end{table}

\clearpage
\subsubsection{Statements vs. Relations: Effect of an Argument on the Relations it Carries Out}
\label{stats:sourcerel1}
\begin{table}[h]
\centering
\resizebox{\textwidth}{!}{
\begin{tabular}{|c||c|c|c|c|c|c|c|c|c|c|c|c|c|c|} 
\hline
\multirow{2}{*}{\shortstack[c]{Total\\sample}} &\multicolumn{2}{c|}{D}&\multicolumn{2}{c|}{SoD}&\multicolumn{2}{c|}{NAD}&\multicolumn{2}{c|}{DK}&	\multicolumn{2}{c|}{SoA}&\multicolumn{2}{c|}{A}&\multicolumn{2}{c|}{SA}\\ 
\cline{2-15} 
 &	G--value	&	p--value	&	G--value	&	p--value	&	G--value	&	p--value	&	G--value	&	p--value	&	G--value	&	p--value	&	G--value	&	p--value	&	G--value	&	p--value	\\ 
\hline
SD	&	34.873	&0.000	&35.761	&	0.000		&44.077	&0.000	&17.742	&	0.001		&33.778	&	0.000		&11.239	&	0.024	&	23.965	&	0.000	\\ \hline
D	&	 		&	 	&3.452	&\cellg{0.485}	&19.668	&0.001	&13.085	&	0.011		&8.647	&\cellg{0.071}	&61.659	&	0.000	&	118.932	&	0.000	\\ \hline
SoD	&	 		&	 	&	 	&	 			&25.352	&0.000	&19.199	&	0.001		&6.513	&\cellg{0.164}	&63.386	&	0.000	&	105.564	&	0.000	\\ \hline
NAD	&	 		&	 	&	 	&	 			&	 	&	 	&4.571	&\cellg{0.334}	&38.749	&	0.000		&59.589	&	0.000	&	97.628	&	0.000	\\ \hline
DK	&			&		&		&				&		&		&		&				&25.826	&	0.000		&25.004	&	0.000	&	40.261	&	0.000	\\ \hline
SoA	&			&		&		&				&		&		&		&				&		&				&47.747	&	0.000	&	80.135	&	0.000	\\ \hline
A	&			&		&		&				&		&		&		&				&		&				&		&			&	13.623	&	0.009	\\ \hline
\hline
\multirow{2}{*}{\shortstack[c]{Core\\sample}} &\multicolumn{2}{c|}{D}&\multicolumn{2}{c|}{SoD}&\multicolumn{2}{c|}{NAD}&\multicolumn{2}{c|}{DK}&	\multicolumn{2}{c|}{SoA}&\multicolumn{2}{c|}{A}&\multicolumn{2}{c|}{SA}\\ 
\cline{2-15} 
 &	G--value	&	p--value	&	G--value	&	p--value	&	G--value	&	p--value	&	G--value	&	p--value	&	G--value	&	p--value	&	G--value	&	p--value	&	G--value	&	p--value	\\ 
\hline
SD	&11.457	&0.022	&5.418	&\cellg{0.247}	&23.947	&0.000	&28.026	&	0.000		&2.570	&\cellg{0.632}	&	5.042	&\cellg{0.283}	&	36.282	&	0.000	\\ \hline
D	&	 	&	 	&2.413	&\cellg{0.660}	&19.445	&0.001	&29.463	&	0.000		&0.477	&\cellg{0.976}	&	25.615	&	0.000		&	89.884	&	0.000	\\ \hline
SoD	&	 	&	 	&	 	&	 			&13.304	&0.010	&20.222	&	0.000		&0.351	&\cellg{0.986}	&	13.042	&	0.011		&	51.616	&	0.000	\\ \hline
NAD	&	 	&	 	&	 	&	 			&	 	&	 	&6.920	&\cellg{0.140}	&4.976	&\cellg{0.290}	&	22.987	&	0.000		&	50.419	&	0.000	\\ \hline
DK	&		&		&		&				&		&		&		&				&6.329	&\cellg{0.176}	&	19.897	&	0.001		&	50.310	&	0.000	\\ \hline
SoA	&		&		&		&				&		&		&		&				&		&				&	4.931	&\cellg{0.294}	&	14.571	&	0.006	\\ \hline
A	&		&		&		&				&		&		&		&				&		&				&			&				&	20.806	&	0.000	\\ \hline
\end{tabular}
}
\caption{Results of G--test for independence between different argument acceptance levels in Dialogue 1 on total and core samples. 
In all cases we have obtained 4 degrees of freedom. 
We use the following abbreviations: 
Strongly Agree (SA), Agree (A), Somewhat Agree (SoA), Neither Agree nor Disagree (NAD), Somewhat Disagree (SoD), 
Disagree (D), Strongly Disagree (SD), Don't Know (DK). DF stands for degrees of freedom. 
We have highlighted the fields with p--value greater than $0.05$, i.e. those for which we cannot reject the null hypothesis. 
These results were obtained using library Deducer (likelihood.test function) in R.}
\label{tab:d1indeparg}
\end{table}

\begin{table}[h]
\centering
\resizebox{\textwidth}{!}{
\begin{tabular}{|c||c|c|c|c|c|c|c|c|c|c|c|c|c|c|c|c|} 
\cline{2-17} 
\multicolumn{1}{c|}{} 	&	\multicolumn{8}{c|}{Total Sample}    	&\multicolumn{8}{c|}{Core Sample} \\			
\cline{2-17} 				
\multicolumn{1}{c|}{} &	SD	&	D	&	SoD	&	NAD	&	DK	&	SoA	&	A	&	SA	&	SD	&	D	&	SoD	&	NAD	&	DK	&	SoA	&	A	&	SA	\\ \hline
$\chi^2$	&	11.314	&	32.225	&	33.057	&	35.972	&	14.727	&	35.980	&	57.000	&	171.296	&	12.239	&	23.340	&	7.296	&	13.909	&	25.600	&	2.000	&	18.706	&	190.611	\\ \hline
DF	&	4	&	4	&	4	&	4	&	4	&	4	&	4	&	4	&	4	&	4	&	4	&	4	&	4	&	4	&	4	&	4	\\ \hline
p--value	&	0.023	&	0.000	&	0.000	&	0.000	&	0.005	&	0.000	&	0.000	&	0.000	&	0.016	&	0.000	&	\cellg{0.121}&	0.008	&	0.000	&\cellg{0.736}&	0.001	&	0.000	\\ \hline
\end{tabular}
}
\caption{Results of chi--squared goodness of fit test on relations carried out by arguments of a given acceptance level in Dialogue 1. 
We use the following abbreviations: 
Strongly Agree (SA), Agree (A), Somewhat Agree (SoA), Neither Agree nor Disagree (NAD), Somewhat Disagree (SoD), 
Disagree (D), Strongly Disagree (SD), Don't Know (DK). DF stands for degrees of freedom. 
We have highlighted the fields with p--value greater than $0.05$, i.e. those for which we cannot reject the null hypothesis. 
These results were obtained using R.}
\label{tab:d1goodnessarg}
\end{table}

\begin{table}[h]
\centering
\resizebox{\textwidth}{!}{
\begin{tabular}{|c||c|c|c|c|c|c|c|c|c|c|c|c|c|c|c|c|c|c|}  
\hline
\multicolumn{13}{|c|}{Agreement pooled according to strength} \\
\hline 
\multirow{3}{*}{\shortstack[c]{Relations pooled \\according to strength}} &	\multicolumn{6}{c|}{Total Sample} & \multicolumn{6}{c|}{Core Sample} \\
 \cline{2-13}  
 &\multicolumn{2}{c|}{Moderate Belief}&\multicolumn{2}{c|}{Weak Belief}&\multicolumn{2}{c|}{Neither} &\multicolumn{2}{c|}{Moderate Belief}&\multicolumn{2}{c|}{Weak Belief}&\multicolumn{2}{c|}{Neither}	\\  \cline{2-13} 

				&G--value	&p--value	&G--value	&p--value	&G--value	&p--value	&G--value	&p--value	&G--value	&	p--value	&G--value	&	p--value	\\ \hline

Strong Belief	&36.110		&0.000		&	122.613	&0.000		&	100.750	&	0.000	&40.240		&0.000		&33.854		&	0.000		&	65.784	&	0.000	\\ \hline

Moderate Belief	&			&			&	44.053	&0.000		&	52.513	&	0.000	&			&			&2.165		&\cellg{0.339}	&	29.473	&	0.000	\\ \hline

Weak Belief		&			&			&			&			&	49.153	&	0.000	&			&			&			&				&	20.024	&	0.000	\\ \hline
 
\hline 
\multirow{3}{*}{\shortstack[c]{Relations pooled \\according to polarity}} &	\multicolumn{6}{c|}{Total Sample} & \multicolumn{6}{c|}{Core Sample} \\
 \cline{2-13}  
&\multicolumn{2}{c|}{Moderate Belief}&\multicolumn{2}{c|}{Weak Belief}&\multicolumn{2}{c|}{Neither	}				&\multicolumn{2}{c|}{Moderate Belief}&\multicolumn{2}{c|}{Weak Belief}&\multicolumn{2}{c|}{Neither}	\\  \cline{2-13} 

				&G--value	&p--value	&G--value	&	p--value	&G--value	&p--value	&G--value	&	p--value	&G--value	&	p--value	&G--value	&	p--value	\\ \hline
Strong Belief	&	7.965	&	0.019	&	5.437	&\cellg{0.066}	&	56.972	&	0.000	&	4.124	& \cellg{0.127}	&	2.060	&\cellg{0.357}	&	41.822	&	0.000	\\ \hline
Moderate Belief	&			&			&	4.981	&\cellg{0.083}	&	37.544	&	0.000	&			&				&	0.229	&\cellg{0.892}	&	25.974	&	0.000	\\ \hline
Weak Belief		&			&			&			&				&	49.160	&	0.000	&			&				&			&				&	19.549	&	0.000	\\ \hline
\end{tabular} 
}
\resizebox{0.7\textwidth}{!}{
\begin{tabular}{|c||c|c|c|c|c|c|c|c|}  
\hline
\multicolumn{9}{|c|}{Agreement pooled according to polarity} \\
\hline 
\multirow{3}{*}{\shortstack[c]{Relations pooled \\according to strength}}   &	\multicolumn{4}{c|}{Total Sample} & \multicolumn{4}{c|}{Core Sample} \\
 \cline{2-9}  
&\multicolumn{2}{c|}{Believed}&\multicolumn{2}{c|}{Neither}&\multicolumn{2}{c|}{Believed}&\multicolumn{2}{c|}{Neither} \\  \cline{2-9}  

			&G--value	&p--value	&G--value	&p--value	&G--value	&p--value	&G--value	&	p--value	\\ \hline
Disbelieved	&98.123		&	0.000	&	40.289	&	0.000	&	74.909	&	0.000	&	37.864	&	0.000	\\ \hline
Believed	&			&			&	100.516	&	0.000	&			&			&	62.282	&	0.000	\\ \hline
\hline 
\multirow{3}{*}{\shortstack[c]{Relations pooled \\according to polarity}}  &	\multicolumn{4}{c|}{Total Sample} & \multicolumn{4}{c|}{Core Sample} \\ \cline{2-9}  
 &\multicolumn{2}{c|}{Believed}&\multicolumn{2}{c|}{Neither}&\multicolumn{2}{c|}{Believed}&\multicolumn{2}{c|}{Neither} \\  \cline{2-9}  
			&G--value	&	p--value	&G--value	&p--value	&G--value	&	p--value	&G--value	&	p--value	\\ \hline
Disbelieved	&	5.811	&\cellg{0.055}	&	37.922	&	0.000	&	5.042	&\cellg{0.080}	&	36.045	&	0.000	\\ \hline
Believed	&	 		&				&	62.659	&	0.000	&			&				&	34.388	&	0.000	\\ \hline
\end{tabular}
}
\caption{Results of G--test for independence between different argument acceptance levels in Dialogue 1 on total and core samples 
according to a given pooling. In all cases we have obtained 2 degrees of freedom. 
We have highlighted the fields with p--value greater than $0.05$, i.e. those for which we cannot reject the null hypothesis. 
These results were obtained using library Deducer (likelihood.test function) in R.}
\label{tab:d1indepargpool}
\end{table}
 
\begin{table}[h]
\centering
\resizebox{\textwidth}{!}{
\begin{tabular}{|c||c|c|c|c|c|c|c|c|} 
\hline
\multicolumn{9}{|c|}{Agreement pooled according to strength} \\
\hline 
\multirow{2}{*}{\shortstack[c]{Relations pooled \\according to strength}} &	\multicolumn{4}{c|}{Total Sample}    	&\multicolumn{4}{c|}{Core Sample} \\			
\cline{2-9} &	Strong Belief	&	Moderate Belief	&	Weak Belief	&	Neither	&	Strong Belief	&	Moderate Belief	&	Weak Belief	&	Neither	\\  \hline 
$\chi^2$	&	280.351	&	75.107	&	102.118	&	13.865	&	192.360	&	33.832	&	17.645	&	8.340	\\ \hline
DF	&	2	&	2	&	2	&	2	&	2	&	2	&	2	&	2	\\ \hline
p--value	&	0.000	&	0.000	&	0.000	&	0.001	&	0.000	&	0.000	&	0.000	&	0.015	\\ \hline
\hline 
\multirow{2}{*}{\shortstack[c]{Relations pooled \\according to polarity}} &\multicolumn{4}{c|}{Total Sample}    	&\multicolumn{4}{c|}{Core Sample} \\			
\cline{2-9} 				
 &	Strong Belief	&	Moderate Belief	&	Weak Belief	&	Neither	&	Strong Belief	&	Moderate Belief	&	Weak Belief	&	Neither	\\   \hline  
$\chi^2$	&	121.032	&	74.476	&	63.430	&	7.000	&	116.919	&	43.486	&	15.516	&	6.298	\\ \hline
DF	&	2	&	2	&	2	&	2	&	2	&	2	&	2	&	2	\\ \hline
p--value	&	0.000	&	0.000	&	0.000	&	0.030	&	0.000	&	0.000	&	0.000	&	0.043	\\ \hline
\end{tabular}
 }
\resizebox{0.7\textwidth}{!}{
\begin{tabular}{|c||c|c|c|c|c|c|} 
\hline
\multicolumn{7}{|c|}{Agreement pooled according to polarity} \\
\hline 
\multirow{2}{*}{\shortstack[c]{Relations pooled \\according to strength}}  &	\multicolumn{3}{c|}{Total Sample}    	&\multicolumn{3}{c|}{Core Sample} \\			
\cline{2-7} 		
 &	Disbelieved	&	Believed	&	Neither	&	Disbelieved	&	Believed	&	Neither	\\ \hline
$\chi^2$	&	84.105	&	333.680	&	13.865	&	65.475	&	197.077	&	8.340	\\ \hline
DF	&	2	&	2	&	2	&	2	&	2	&	2	\\ \hline
p--value	&	0.000	&	0.000	&	0.001	&	0.000	&	0.000	&	0.015	\\ \hline\hline 
\multirow{2}{*}{\shortstack[c]{Relations pooled \\according to polarity}} & \multicolumn{3}{c|}{Total Sample}    	&\multicolumn{3}{c|}{Core Sample} \\			
\cline{2-7} 				
&	Disbelieved	&	Believed	&	Neither	&	Disbelieved	&	Believed	&	Neither	\\ \hline
$\chi^2$	&	69.584	&	180.390	&	7.000	&	69.475	&	107.868	&	6.298	\\ \hline
DF	&	2	&	2	&	2	&	2	&	2	&	2	\\ \hline
p--value	&	0.000	&	0.000	&	0.030	&	0.000	&	0.000	&	0.043	\\ \hline
\end{tabular} 
}
\caption{Results of chi--squared goodness of fit test on relations carried out by arguments of a given acceptance level in Dialogue 1 
on total and core samples according to a given pooling. DF stands for degrees of freedom. 
We have highlighted the fields with p--value greater than $0.05$, i.e. those for which we cannot reject the null hypothesis. 
These results were obtained using R.}
\label{tab:d1goodnessargpool}
\end{table}

\clearpage

\subsection{Dialogue 2}

\subsubsection{Postulate Satisfaction}
\label{stats:postulates2}

\begin{table}[h]
\centering
\resizebox{\textwidth}{!}{
\begin{tabular}{|c||P{1cm}|P{1.3cm}|P{1cm}|P{1.3cm}|P{1cm}|P{1.3cm}|P{1cm}|P{1.3cm}|P{1cm}|P{1.3cm}|P{1cm}|P{1.3cm}|} 
\hline
	 \multirow{2}{*}{Total Sample}		&\tframec{Intended vs Declared}	& \tframec{Intended vs Expanded} & \tframec{Intended vs Common}& \tframec{Declared vs Expanded}& \tframec{Declared vs Common}		& \tframec{Expanded vs Common}	\\
\cline{2-13}
 &Z		&		p--value	&Z		&		p--value	&Z		&		p--value	&Z		&		p--value		&Z		&		p-value	&Z		&		p--value	\\
\hline 
Preferential	&3.231	&0.001			&3.256	&0.001			&2.995	&0.004			&1.414	&\cellg{0.500}		&-1.632	&\cellg{0.132}		&-1.849	&\cellg{0.080}	\\
 \hline 
Rational		&3.021	&0.002			&3.035	&0.002			&2.642	&0.016			&1.000	&\cellg{1.000}		&-2.029	&\cellg{0.050}		&-2.044	&0.048		\\
 \hline 
Strict		&1.957	&\cellg{0.050}		&2.550	&0.010			&3.585	&0.000			&1.999	&\cellg{0.125}		&-0.143	&\cellg{0.892}		&-0.941	&\cellg{0.351}	\\
 \hline 
Protective	&2.410	&0.015			&2.485	&0.012			&4.295	&0.000			&1.000	&\cellg{1.000}		&0.378	&\cellg{0.712}		&0.307	&\cellg{0.766}	\\
 \hline 
Restrained	&2.019	&0.043			&2.026	&0.042			&3.718	&0.000			&1.000	&\cellg{1.000}		&-0.131	&\cellg{0.905}		&-0.182	&\cellg{0.863}	\\
 \hline 
Coherent		&2.019	&0.043			&3.794	&0.000			&4.190	&0.000			&3.845	&0.000			&0.472	&\cellg{0.640}		&-2.421	&0.015		\\
 \hline 
Involutary	&0.465	&\cellg{0.641}		&2.425	&0.015			&2.235	&\cellg{0.063}		&3.154	&0.002			&0.497	&\cellg{0.613}		&-1.914	&\cellg{0.063}	\\
 \hline 
Justifiable		&2.236	&\cellg{0.063}		&2.000	&\cellg{0.125}		&NA		&\cellg{NA}		&-1.000	&\cellg{1.000}		&-2.236	&\cellg{0.063}		&-2.000	&\cellg{0.125}	\\
 \hline 
SemiOptimistic	&-3.196	&0.001			&-3.834	&0.000			&-4.498	&0.000			&-2.643	&0.016			&-1.062	&\cellg{0.308}		&0.416	&\cellg{0.757}	\\
 \hline 
SemiFounded	&4.291	&0.000			&0.389	&\cellg{0.704}		&NA		&\cellg{NA}		&-3.963	&0.000			&-4.291	&0.000			&-0.389	&\cellg{0.704}	\\
 \hline 
Founded		&2.825	&0.008			&-0.971	&\cellg{0.315}		&NA		&\cellg{NA}		&-3.156	&0.002			&-2.825	&0.008			&0.971	&\cellg{0.315}	\\
 \hline 
Optimistic		&2.827	&0.008			&0.284	&\cellg{0.918}		&NA		&\cellg{NA}		&-2.644	&0.016			&-2.827	&0.008			&-0.284	&\cellg{0.918}	\\
 \hline 
Demanding	&1.363	&\cellg{0.250}		&-0.508	&\cellg{0.684}		&1.999	&\cellg{0.125}		&-1.867	&\cellg{0.094}		&-0.255	&\cellg{1.000}		&1.275	&\cellg{0.173}	\\
 \hline 
Guarded		&4.072	&0.000			&0.194	&\cellg{0.850}		&-1.732	&\cellg{0.250}		&-3.842	&0.000			&-4.298	&0.000			&-0.631	&\cellg{0.535}	\\
 \hline 
Discharging	&3.365	&0.000			&-0.157	&\cellg{0.880}		&-2.824	&0.008			&-3.841	&0.000			&-4.500	&0.000			&-0.888	&\cellg{0.381}	\\
 \hline 
Trusting		&4.208	&0.000			&0.916	&\cellg{0.363}		&-1.414	&\cellg{0.500}		&-3.588	&0.000			&-4.330	&0.000			&-1.259	&\cellg{0.213}	\\
 \hline 
Anticipating	&3.636	&0.000			&0.346	&\cellg{0.735}		&-1.999	&\cellg{0.125}		&-3.305	&0.001			&-4.104	&0.000			&-0.986	&\cellg{0.331}	\\
 \hline 
\hline
	 \multirow{2}{*}{Core Sample}		&\tframec{Intended vs Declared}	& \tframec{Intended vs Expanded} & \tframec{Intended vs Common}& \tframec{Declared vs Expanded}& \tframec{Declared vs Common}		& \tframec{Expanded vs Common}	\\
\cline{2-13}
 	&Z		&		p--value	&Z		&		p--value	&Z		&		p--value	&Z		&		p--value		&Z		&		p-value	&Z		&		p--value	\\
\hline 
 Preferential	&	2.226	&\cellg{0.063}	&	2.226	&\cellg{0.063}	&	1.994	&\cellg{0.125}		&	NA		&\cellg{NA}	&	-1.000	&\cellg{1.000}	&	-1.000	&\cellg{1.000}	\\ \hline
Rational		&	2.224	&\cellg{0.063}	&	2.224	&\cellg{0.063}	&	1.729	&\cellg{0.250}		&	NA		&\cellg{NA}	&	-1.413	&\cellg{0.500}	&	-1.413	&\cellg{0.500}	\\ \hline
Strict		&	2.025	&\cellg{0.063}	&	2.025	&\cellg{0.063}	&	1.993	&\cellg{0.125}		&	NA		&\cellg{NA}	&	-0.961	&\cellg{0.625}	&	-0.961	&\cellg{0.625}	\\ \hline
Protective	&	2.124	&	0.035	&	2.124	&	0.035	&	2.924	&	0.004		&	NA		&\cellg{NA}	&	0.435	&\cellg{0.703}	&	0.435	&\cellg{0.703}	\\ \hline
Restrained	&	1.983	&\cellg{0.051}	&	1.983	&\cellg{0.051}	&	2.229	&\cellg{0.063}		&	NA		&\cellg{NA}	&	-0.411	&\cellg{0.813}	&	-0.411	&\cellg{0.813}	\\ \hline
Coherent		&	2.215	&	0.027	&	2.647	&	0.006	&	2.777	&	0.008		&	2.229	&\cellg{0.063}	&	-0.494	&\cellg{1.000}	&	-1.679	&\cellg{0.125}	\\ \hline
Involutary	&	0.961	&\cellg{0.625}	&	1.594	&\cellg{0.219}	&	1.414	&\cellg{0.500}		&	1.414	&\cellg{0.500}	&	0.049	&\cellg{1.000}	&	-0.961	&\cellg{0.625}	\\ \hline
Justifiable		&	NA		&\cellg{NA}	&	NA		&\cellg{NA}	&	NA		&\cellg{NA}		&	NA		&\cellg{NA}	&	NA		&\cellg{NA}	&	NA		&\cellg{NA}		\\ \hline
SemiOptimistic	&	-2.647	&	0.008	&	-3.056	&	0.002	&	-3.172	&	0.001		&	-1.994	&\cellg{0.125}	&	-0.960	&\cellg{0.500}	&	0.535	&\cellg{1.000}	\\ \hline
SemiFounded	&	2.224	&\cellg{0.063}	&	-1.061	&\cellg{0.359}	&	NA		&\cellg{NA}		&	-2.433	&	0.031	&	-2.224	&\cellg{0.063}	&	1.061	&\cellg{0.359}	\\ \hline
Founded		&	1.000	&\cellg{1.000}	&	-0.659	&\cellg{0.500}	&	NA		&\cellg{NA}		&	-1.413	&\cellg{0.500}	&	-1.000	&\cellg{1.000}	&	0.659	&\cellg{0.500}	\\ \hline
Optimistic		&	NA		&\cellg{NA}	&	-1.000	&\cellg{1.000}	&	NA		&\cellg{NA}		&	-1.000	&\cellg{1.000}	&	NA		&\cellg{NA}	&	1.000	&\cellg{1.000}	\\ \hline
Demanding	&	1.729	&\cellg{0.250}	&	0.618	&\cellg{0.750}	&	1.730	&\cellg{0.250}		&	-1.000	&\cellg{1.000}	&	0.049	&\cellg{1.000}	&	0.659	&\cellg{0.500}	\\ \hline
Guarded		&	1.993	&\cellg{0.125}	&	-1.393	&\cellg{0.188}	&	NA		&\cellg{NA}		&	-2.429	&	0.031	&	-1.993	&\cellg{0.125}	&	1.393	&\cellg{0.188}	\\ \hline
Discharging	&	0.745	&\cellg{0.594}	&	-1.687	&\cellg{0.102}	&	-1.413	&\cellg{0.500}		&	-2.427	&	0.031	&	-1.994	&\cellg{0.125}	&	1.359	&\cellg{0.191}	\\ \hline
Trusting		&	1.730	&\cellg{0.250}	&	-0.130	&\cellg{0.906}	&	NA		&\cellg{NA}		&	-1.729	&\cellg{0.250}	&	-1.730	&\cellg{0.250}	&	0.130	&\cellg{0.906}	\\ \hline
Anticipating	&	0.659	&\cellg{0.500}	&	-0.809	&\cellg{0.531}	&	-1.000	&\cellg{1.000}		&	-1.729	&\cellg{0.250}	&	-1.413	&\cellg{0.500}	&	0.513	&\cellg{0.688}	\\ \hline  
\end{tabular}
}
\caption{The results of the Wilcoxon signed rank test with Pratt adjustment for a given postulate evaluated on two separate graphs on the total and core samples in Dialogue 2. 
The results have been obtained using R library coin. 
We have highlighted the fields with p--value greater than $0.05$, i.e. those for which we cannot reject the null hypothesis. In case the adherence rate for a given postulate was identical 
for two graphs for all participants, the algorithm has returned NA.}
\label{tab:postulatewilcox2}
\end{table}

\clearpage
\subsubsection{Statements vs. Relations: Effect of a Relation on its Source}
\label{stats:relsource2}

\begin{table}[h]
\centering
\resizebox{\textwidth}{!}{
\begin{tabular}{|c||c|c|c|c|c|c|c|c|c|c|c|c|} 
\hline
 \multirow{2}{*}{Total Sample}	 &	\multicolumn{3}{c|}{-} & \multicolumn{3}{c|}{+ +} & \multicolumn{3}{c|}{+} & \multicolumn{3}{c|}{?}\\
\cline{2-13}
 &	G--value	&	DF	&	p--value	&	G--value	&	DF	&	p--value	&	G--value	&	DF	&	p--value	&	G--value	&	DF	&	p--value	\\ \hline
- -	&	146.771	&	7	&	0.000	&	44.558	&	7	&	0.000	&	112.838	&	7	&	0.000	&	122.577	&	7	&	0.000	\\ \hline
-	&	 	&	 	&	 	&	70.229	&	7	&	0.000	&	49.829	&	7	&	0.000	&	125.574	&	7	&	0.000	\\ \hline
+ +	&	 	&	 	&	 	&	 	&	 	&		&	82.165	&	7	&	0.000	&	110.601	&	7	&	0.000	\\ \hline
+	&	 	&	 	&	 	&	 	&	 	&	 	&	 	&	 	&	 	&	90.918	&	7	&	0.000	\\ \hline
\hline
\multirow{2}{*}{Core Sample}		 &	\multicolumn{3}{c|}{-} & \multicolumn{3}{c|}{+ +} & \multicolumn{3}{c|}{+} & \multicolumn{3}{c|}{?}\\
\cline{2-13}
 &	G--value	&	DF	&	p--value	&	G--value	&	DF	&	p--value	&	G--value	&	DF	&	p--value	&	G--value	&	DF	&	p--value	\\ \hline
- -	&	86.269	&	7	&	0.000	&	7.506	&	7	&	\cellg{0.378}	&	83.920	&	7	&	0.000	&	123.81499	&	7	&	0.000	\\ \hline
-	&	 	&	 	&	 	&	70.549	&	7	&	0.000	&	18.330	&	7	&	0.011	&	89.734	&	7	&	0.000	\\ \hline
+ +	&	 	&	 	&	 	&	 	&	 	&		&	74.209	&	7	&	0.000	&	110.2263	&	7	&	0.000	\\ \hline
+	&	 	&	 	&	 	&	 	&	 	&	 	&	 	&	 	&	 	&	42.580	&	7	&	0.000	\\ \hline
\end{tabular}
}
\caption{Results of G--test for independence between relations of given type in Dialogue 2 on total and core samples. 
 We use the following abbreviations for the relation types:  
A good reason against (- -), A somewhat good reason against (-), A good reason for (++), A somewhat good reason for (+), Somewhat related, but 
can't say how (?). DF stands for degrees of freedom. 
We have highlighted the fields with p--value greater than $0.05$, i.e. those for which we cannot reject the null hypothesis. 
These results were obtained using library Deducer (likelihood.test function) in R.}
\label{tab:d2indeprel}
\end{table} 

\begin{table}[h]
\centering
\resizebox{\textwidth}{!}{
\begin{tabular}{|c||c|c|c|c|c|c|c|c|c|c|} 
\cline{2-11} 
\multicolumn{1}{c|}{} 	&	\multicolumn{5}{c|}{Total Sample}    	&\multicolumn{5}{c|}{Core Sample} \\			
\cline{2-11} 				
\multicolumn{1}{c|}{} & - -	&	-	&	+ +	&	+	&	?	&	- -	&	-	&	+ +	&	+	&	?	\\ \hline
$\chi^2$	&	220.015	&	183.075	&	146.260	&	87.918	&	378.960	&	134.442	&	59.454	&	79.750	&	46.842	&	396.177	\\ \hline
DF	&	7	&	7	&	7	&	7	&	7	&	7	&	7	&	7	&	7	&	7	\\ \hline
p--value	&	0.000	&	0.000	&	0.000	&	0.000	&	0.000	&	0.000	&	0.000	&	0.000	&	0.000	&	0.000	\\ \hline
\end{tabular}
}
\caption{Results of chi--squared goodness of fit test on levels of agreement of sources of a relation of a given type 
in Dialogue 2. 
 We use the following abbreviations for the relation types:  
A good reason against (- -), A somewhat good reason against (-), A good reason for (++), A somewhat good reason for (+), Somewhat related, but 
can't say how (?). DF stands for degrees of freedom. 
We have highlighted the fields with p--value greater than $0.05$, i.e. those for which we cannot reject the null hypothesis. 
These results were obtained using R.}
\label{tab:d2goodnessrel}
\end{table}

\begin{table}[h]
\centering
\resizebox{\textwidth}{!}{
\begin{tabular}{|c||c|c|c|c|c|c|c|c|c|c|c|c|c|}  
\hline
\multicolumn{13}{|c|}{Relations pooled according to strength}\\
\hline 
\multirow{3}{*}{\shortstack[c]{Agreement pooled \\according to strength}} &	\multicolumn{6}{c|}{Total Sample} & \multicolumn{6}{c|}{Core Sample} \\
 \cline{2-13}
 &	\multicolumn{3}{c|}{Normal Relation} & \multicolumn{3}{c|}{Dependency} &	\multicolumn{3}{c|}{Normal Relation} & \multicolumn{3}{c|}{Dependency} \\
\cline{2-13}
 &	G--value	&	DF	&	p--value	&	G--value	&	DF	&	p--value	&	G--value	&	DF	&	p--value	&	G--value	&	DF	&	p--value	\\ \hline 
Strong Relation	&	99.442	&	3	&	0.000	&	120.905	&	3	&	0.000	&	98.797	&	3	&	0.000	&	127.352	&	3	&	0.000	\\ \hline
Normal Relation	&	 	&	 	&	 	&	127.914	&	3	&	0.000	&	 	&	 	&	 	&	54.266	&	3	&	0.000	\\ \hline
\hline 
\multirow{3}{*}{\shortstack[c]{Agreement pooled \\according to polarity}}  &	\multicolumn{6}{c|}{Total Sample} & \multicolumn{6}{c|}{Core Sample} \\
 \cline{2-13}
 &	\multicolumn{3}{c|}{Normal Relation} & \multicolumn{3}{c|}{Dependency} &	\multicolumn{3}{c|}{Normal Relation} & \multicolumn{3}{c|}{Dependency} \\
\cline{2-13}
  &	G--value	&	DF	&	p--value	&	G--value	&	DF	&	p--value	&	G--value	&	DF	&	p--value	&	G--value	&	DF	&	p--value	\\ \hline 
Strong Relation	&	38.622	&	2	&	0.000	&	108.870	&	2	&	0.000	&	14.090	&	2	&	0.001	&	120.824	&	2	&	0.000	\\ \hline
Normal Relation	&		&		&		&	127.904	&	2	&	0.000	&		&		&		&	55.808	&	2	&	0.000	\\ \hline
\hline
\multicolumn{13}{|c|}{Relations pooled according to polarity} \\
\hline 
\multirow{3}{*}{\shortstack[c]{Agreement pooled \\according to polarity}} &	\multicolumn{6}{c|}{Total Sample} & \multicolumn{6}{c|}{Core Sample} \\
 \cline{2-13}
 &	\multicolumn{3}{c|}{Support} & \multicolumn{3}{c|}{Dependency} &	\multicolumn{3}{c|}{Support} & \multicolumn{3}{c|}{Dependency} \\
\cline{2-13}
  &	G--value	&	DF	&	p--value	&	G--value	&	DF	&	p--value	&	G--value	&	DF	&	p--value	&	G--value	&	DF	&	p--value	\\ \hline 
Attack 	&	3.595	&	2	&\cellg{0.166}&	122.975	&	2	&	0.000	&	9.883	&	2	&	0.007	&	107.440	&	2	&	0.000	\\ \hline
Support 	&	 	&		&		&	104.779	&	2	&	0.000	&		&		&		&	78.490	&	2	&	0.000	\\ \hline
 \hline
\multirow{3}{*}{\shortstack[c]{Agreement pooled \\according to strength}}&	\multicolumn{6}{c|}{Total Sample} & \multicolumn{6}{c|}{Core Sample} \\
 \cline{2-13}
  &	\multicolumn{3}{c|}{Support} & \multicolumn{3}{c|}{Dependency} &	\multicolumn{3}{c|}{Support} & \multicolumn{3}{c|}{Dependency} \\
\cline{2-13}
  &	G--value	&	DF	&	p--value	&	G--value	&	DF	&	p--value	&	G--value	&	DF	&	p--value	&	G--value	&	DF	&	p--value	\\ \hline 
Attack 	&	9.338	&	3	&	0.025	&	128.599	&	3	&	0.000	&	0.965	&	3	&\cellg{0.810}&	104.209	&	3	&	0.000	\\ \hline
Support 	&	 	&		&		&	106.480	&	3	&	0.000	&		&		&		&	82.158	&	3	&	0.000	\\ \hline
\end{tabular}
}
\caption{Results of G--test for independence between relations of given type in Dialogue 2 on total and core samples according to 
a given pooling. DF stands for degrees of freedom. 
We have highlighted the fields with p--value greater than $0.05$, i.e. those for which we cannot reject the null hypothesis. 
These results were obtained using library Deducer (likelihood.test function) in R.}
\label{tab:d2indeprelpool}
\end{table}

\begin{table}[h]
\centering
\resizebox{\textwidth}{!}{
\begin{tabular}{|c||c|c|c|c|c|c|} 
\hline
\multicolumn{7}{|c|}{Relations pooled according to strength}\\
\hline 
\multirow{2}{*}{\shortstack[c]{Agreement pooled \\according to strength}}	&	\multicolumn{3}{c|}{Total Sample}    	&\multicolumn{3}{c|}{Core Sample} \\			
\cline{2-7} 				
&	Strong Relation	&	Normal Relation	&	Dependency	&	Strong Relation	&	Normal Relation	&	Dependency	\\ \hline	
$\chi^2$	&	118.474	&	145.419	&	278.048	&	38.630	&	73.749	&	188.726	\\ \hline
DF	&	3	&	3	&	3	&	3	&	3	&	3	\\ \hline
p--value	&	0.000	&	0.000	&	0.000	&	0.000	&	0.000	&	0.000	\\ \hline
\hline 
\multirow{2}{*}{\shortstack[c]{Agreement pooled \\according to polarity}}	&	\multicolumn{3}{c|}{Total Sample}    	&\multicolumn{3}{c|}{Core Sample} \\			
\cline{2-7} 				
 &	Strong Relation	&	Normal Relation	&	Dependency	&	Strong Relation	&	Normal Relation	&	Dependency	\\ \hline	
$\chi^2$	&	108.641	&	10.771	&	141.128	&	61.437	&	7.732	&	106.235	\\ \hline
DF	&	2	&	2	&	2	&	2	&	2	&	2	\\ \hline
p--value	&	0.000	&	0.005	&	0.000	&	0.000	&	0.021	&	0.000	\\ \hline
 \hline
\multicolumn{7}{|c|}{Relations pooled according to polarity}\\
\hline 
\multirow{2}{*}{\shortstack[c]{Agreement pooled \\according to polarity}} 	&	\multicolumn{3}{c|}{Total Sample}    	&\multicolumn{3}{c|}{Core Sample} \\			
\cline{2-7} 				
&	Attack	&	Support	&	Dependency	&	Attack	&	Support	&	Dependency	\\ \hline	
$\chi^2$	&	63.392	&	24.454	&	141.128	&	59.744	&	4.805	&	106.235	\\ \hline
DF	&	2	&	2	&	2	&	2	&	2	&	2	\\ \hline
p--value	&	0.000	&	0.000	&	0.000	&	0.000	&	\cellg{0.090}	&	0.000	\\ \hline
\hline 
\multirow{2}{*}{\shortstack[c]{Agreement pooled \\according to strength}}	&	\multicolumn{3}{c|}{Total Sample}    	&\multicolumn{3}{c|}{Core Sample} \\			
\cline{2-7} 				
 &	Attack	&	Support	&	Dependency	&	Attack	&	Support	&	Dependency	\\ \hline	
$\chi^2$	&	112.872	&	83.463	&	278.048	&	22.326	&	20.270	&	188.726	\\ \hline
DF	&	3	&	3	&	3	&	3	&	3	&	3	\\ \hline
p--value	&	0.000	&	0.000	&	0.000	&	0.000	&	0.000	&	0.000	\\ \hline
\end{tabular}
}
\caption{Results of chi--squared goodness of fit test on levels of agreement of sources of a relation of a given type 
in Dialogue 2 on total and core samples according to a given type of pooling. DF stands for degrees of freedom. 
We have highlighted the fields with p--value greater than $0.05$, i.e. those for which we cannot reject the null hypothesis. 
These results were obtained using R.}
\label{tab:d2goodnessrelpool}
\end{table} 
 
\clearpage
\subsubsection{Statements vs. Relations: Effect of an Argument on the Relations it Carries Out}
\label{stats:sourcerel2}

\begin{table}[h]
\centering
\resizebox{\textwidth}{!}{
\begin{tabular}{|c||c|c|c|c|c|c|c|c|c|c|c|c|c|c|} 
\hline
\multirow{2}{*}{\shortstack[c]{Total\\sample}} &\multicolumn{2}{c|}{D}&\multicolumn{2}{c|}{SoD}&\multicolumn{2}{c|}{NAD}&\multicolumn{2}{c|}{DK}&\multicolumn{2}{c|}{SoA}&\multicolumn{2}{c|}{A}&\multicolumn{2}{c|}{SA}\\ 
\cline{2-15} 
&	G--value	&	p--value	&	G--value	&	p--value	&	G--value	&	p--value	&	G--value	&	p--value	&	G--value	&	p--value	&	G--value	&	p--value	&	G--value	&	p--value	\\ 
\hline
SD	&88.802	&0.000	&41.240	&0.000	&57.479	&0.000	&	39.781	&	0.000	&	39.381	&	0.000		&	18.469	&	0.001	&	15.769	&	0.003	\\ \hline
D	&	 	&	 	&46.449	&0.000	&116.328&0.000	&	72.405	&	0.000	&	30.161	&	0.000		&	72.721	&	0.000	&	129.046	&	0.000	\\ \hline
SoD	&	 	&	 	&	 	&	 	&40.291	&0.000	&	45.535	&	0.000	&	5.529	&\cellg{0.237}	&	46.450	&	0.000	&	60.854	&	0.000	\\ \hline
NAD	&	 	&	 	&	 	&	 	&	 	&	 	&	10.459	&	0.033	&	50.145	&	0.000		&	77.095	&	0.000	&	74.098	&	0.000	\\ \hline
DK	&		&		&		&		&		&		&			&			&	41.020	&	0.000		&	41.392	&	0.000	&	55.609	&	0.000	\\ \hline
SoA	&		&		&		&		&		&		&			&			&			&				&	37.557	&	0.000	&	68.731	&	0.000	\\ \hline
A	&		&		&		&		&		&		&			&			&			&				&			&			&	21.771	&	0.000	\\ \hline
\hline
\multirow{2}{*}{\shortstack[c]{Core\\sample}}&\multicolumn{2}{c|}{D}&\multicolumn{2}{c|}{SoD}&\multicolumn{2}{c|}{NAD}&\multicolumn{2}{c|}{DK}&\multicolumn{2}{c|}{SoA}&\multicolumn{2}{c|}{A}&\multicolumn{2}{c|}{SA}\\  
\cline{2-15} 
&	G--value	&	p--value	&	G--value	&	p--value	&	G--value	&	p--value	&	G--value	&	p--value	&	G--value	&	p--value	&	G--value	&	p--value	&	G--value	&	p--value	\\ 
\hline 
SD	&48.190	&0.000	&37.146	&	0.000		&64.903	&0.000	&83.274	&0.000	&48.583	&	0.000		&	19.065	&	0.001	&	6.361	&	\cellg{0.174}\\ \hline
D	&	 	&	 	&4.049	&\cellg{0.399}	&26.022	&0.000	&10.662	&0.031	&6.652	&\cellg{0.155}	&	22.830	&	0.000	&	61.635	&	0.000	\\ \hline
SoD	&	 	&	 	&	 	&	 			&21.170	&0.000	&11.723	&0.020	&11.731	&	0.019		&	17.529	&	0.002	&	46.062	&	0.000	\\ \hline
NAD	&	 	&	 	&	 	&	 			&	 	&	 	&43.599	&0.000	&56.900	&	0.000		&	72.986	&	0.000	&	87.561	&	0.000	\\ \hline
DK	&		&		&		&				&		&		&		&		&18.957	&	0.001		&	53.371	&	0.000	&	94.664	&	0.000	\\ \hline
SoA	&		&		&		&				&		&		&		&		&		&				&	17.817	&	0.001	&	55.203	&	0.000	\\ \hline
A	&		&		&		&				&		&		&		&		&		&				&			&			&	20.983	&	0.000	\\ \hline

\end{tabular}
}
\caption{Results of G--test for independence between different argument acceptance levels in Dialogue 2 on total and core samples. In 
all cases we have obtained 4 degrees of freedom. 
We use the following abbreviations: 
Strongly Agree (SA), Agree (A), Somewhat Agree (SoA), Neither Agree nor Disagree (NAD), Somewhat Disagree (SoD), 
Disagree (D), Strongly Disagree (SD), Don't Know (DK). DF stands for degrees of freedom. 
We have highlighted the fields with p--value greater than $0.05$, i.e. those for which we cannot reject the null hypothesis. 
These results were obtained using library Deducer (likelihood.test function) in R.}
\label{tab:d2indeparg}
\end{table}

\begin{table}[h]
\centering
\resizebox{\textwidth}{!}{
\begin{tabular}{|c||c|c|c|c|c|c|c|c|c|c|c|c|c|c|c|c|} 
\cline{2-17} 
\multicolumn{1}{c|}{} 	&	\multicolumn{8}{c|}{Total Sample}    	&\multicolumn{8}{c|}{Core Sample} \\			
\cline{2-17} 				
\multicolumn{1}{c|}{} &	SD	&	D	&	SoD	&	NAD	&	DK	&	SoA	&	A	&	SA	&	SD	&	D	&	SoD	&	NAD	&	DK	&	SoA	&	A	&	SA	\\ \hline
$\chi^2$	&	87.128	&	117.032	&	22.202	&	36.956	&	37.706	&	26.618	&	137.961	&	156.983	&	90.706	&	5.596	&	7.067	&	59.016	&	16.133	&	32.727	&	98.417	&	140.519	\\ \hline
DF	&	4	&	4	&	4	&	4	&	4	&	4	&	4	&	4	&	4	&	4	&	4	&	4	&	4	&	4	&	4	&	4	\\ \hline
p--value	&	0.000	&	0.000	&	0.000	&	0.000	&	0.000	&	0.000	&	0.000	&	0.000	&	0.000	&\cellg{0.231}&\cellg{0.132}	&	0.000	&	0.003	&	0.000	&	0.000	&	0.000	\\ \hline

\end{tabular}
}
\caption{Results of chi--squared goodness of fit test on relations carried out by arguments of a given acceptance level in Dialogue 2. We use the following abbreviations: 
Strongly Agree (SA), Agree (A), Somewhat Agree (SoA), Neither Agree nor Disagree (NAD), Somewhat Disagree (SoD), 
Disagree (D), Strongly Disagree (SD), Don't Know (DK). DF stands for degrees of freedom. 
We have highlighted the fields with p--value greater than $0.05$, i.e. those for which we cannot reject the null hypothesis. 
These results were obtained using R.}
\label{tab:d2goodnessarg}
\end{table}

\begin{table}[h]
\centering
\resizebox{\textwidth}{!}{
\begin{tabular}{|c||c|c|c|c|c|c|c|c|c|c|c|c|}  
\hline
\multicolumn{13}{|c|}{Agreement pooled according to strength} \\
\hline 
\multirow{3}{*}{\shortstack[c]{Relations pooled \\according to strength}} &	\multicolumn{6}{c|}{Total Sample} & \multicolumn{6}{c|}{Core Sample} \\
 \cline{2-13}  
 &\multicolumn{2}{c|}{Moderate Belief}&\multicolumn{2}{c|}{Weak Belief}&\multicolumn{2}{c|}{Neither} &\multicolumn{2}{c|}{Moderate Belief}&\multicolumn{2}{c|}{Weak Belief}&\multicolumn{2}{c|}{Neither}	\\  \cline{2-13} 
				&G--value	&p--value	&G--value	&p--value	&G--value	&p--value	&G--value	&p--value	&G--value	&	p--value	&G--value	&	p--value	\\ \hline

Strong Belief	&64.174	&0.000	&	89.900	&0.000	&	101.892	&	0.000	&55.723	&0.000	&	88.655	&	0.000	&	146.970	&	0.000	\\ \hline
Moderate Belief	&		&		&	7.532	&0.023	&	97.014	&	0.000	&		&		&	10.005	&	0.007	&	59.026	&	0.000	\\ \hline
Weak Belief		&		&		&			&		&	76.925	&	0.000	&		&		&			&			&	40.821	&	0.000	\\ \hline
\hline 
\multirow{3}{*}{\shortstack[c]{Relations pooled \\according to polarity}} &	\multicolumn{6}{c|}{Total Sample} & \multicolumn{6}{c|}{Core Sample} \\
 \cline{2-13}  
&\multicolumn{2}{c|}{Moderate Belief}&\multicolumn{2}{c|}{Weak Belief}&\multicolumn{2}{c|}{Neither	}				&\multicolumn{2}{c|}{Moderate Belief}&\multicolumn{2}{c|}{Weak Belief}&\multicolumn{2}{c|}{Neither}	\\  \cline{2-13} 
				&G--value	&p--value	&G--value	&	p--value	&G--value	&p--value	&G--value	&	p--value	&G--value	&	p--value	&G--value	&	p--value	\\ \hline
				
Strong Belief	&	1.847	&\cellg{0.397}	&	9.971	&	0.007	&	78.619	&	0.000	&6.068	&0.048	&	5.081	&\cellg{0.079}	&	72.506	&	0.000	\\ \hline
Moderate Belief	&			&				&	7.261	&	0.026	&	88.552	&	0.000	&		&		&	0.000	&\cellg{1.000}	&	55.628	&	0.000	\\ \hline
Weak Belief		&			&				&			&			&	49.026	&	0.000	&		&		&			&				&	39.409	&	0.000	\\ \hline
\end{tabular}
}
\resizebox{0.7\textwidth}{!}{
\begin{tabular}{|c||c|c|c|c|c|c|c|c|}  
\hline
\multicolumn{9}{|c|}{Agreement pooled according to polarity} \\
\hline 
\multirow{3}{*}{\shortstack[c]{Relations pooled \\according to strength}}   &	\multicolumn{4}{c|}{Total Sample} & \multicolumn{4}{c|}{Core Sample} \\
 \cline{2-9}  
&\multicolumn{2}{c|}{Believed}&\multicolumn{2}{c|}{Neither}&\multicolumn{2}{c|}{Believed}&\multicolumn{2}{c|}{Neither} \\  \cline{2-9}  
			&G--value	&p--value	&G--value	&p--value	&G--value	&p--value	&G--value	&	p--value	\\ \hline
			
Disbelieved	&	34.173	&	0.000	&	100.464&	0.000	&	8.874	&	0.012	&	43.919	&	0.000	\\ \hline
Believed	&			&			&	104.099&	0.000	&			&			&	114.249	&	0.000	\\ \hline
\hline 
\multirow{3}{*}{\shortstack[c]{Relations pooled \\according to polarity}}  &	\multicolumn{4}{c|}{Total Sample} & \multicolumn{4}{c|}{Core Sample} \\ \cline{2-9}  
 &\multicolumn{2}{c|}{Believed}&\multicolumn{2}{c|}{Neither}&\multicolumn{2}{c|}{Believed}&\multicolumn{2}{c|}{Neither} \\  \cline{2-9}  
			&G--value	&	p--value	&G--value	&p--value	&G--value	&	p--value	&G--value	&	p--value	\\ \hline
Disbelieved	&	2.630	&\cellg{0.268}	&	74.999	&	0.000	&	15.702	&	0.000	&	40.888	&	0.000	\\ \hline
Believed	&	 		&				&	106.419	&	0.000	&			&			&	104.415	&	0.000	\\ \hline		
\end{tabular}
}
\caption{Results of G--test for independence between different argument acceptance levels in Dialogue 2 on total and core samples according to a given type of pooling. In all cases we have obtained 2 degrees of freedom. 
We have highlighted the fields with p--value greater than $0.05$, i.e. those for which we cannot reject the null hypothesis. 
These results were obtained using library Deducer (likelihood.test function) in R.}
\label{tab:d2indepargpool}
\end{table} 

\begin{table}[h]
\centering
\resizebox{\textwidth}{!}{
\begin{tabular}{|c||c|c|c|c|c|c|c|c|} 
\hline
\multicolumn{9}{|c|}{Agreement pooled according to strength} \\
\hline 
\multirow{2}{*}{\shortstack[c]{Relations pooled \\according to strength}} &	\multicolumn{4}{c|}{Total Sample}    	&\multicolumn{4}{c|}{Core Sample} \\			
\cline{2-9} &	Strong Belief	&	Moderate Belief	&	Weak Belief	&	Neither	&	Strong Belief	&	Moderate Belief	&	Weak Belief	&	Neither	\\  \hline 

$\chi^2$	&	348.713	&	273.129	&	134.568	&	116.882	&	286.052	&	136.630	&	55.665	&	11.626	\\ \hline
DF	&	2	&	2	&	2	&	2	&	2	&	2	&	2	&	2	\\ \hline
p--value	&	0.000	&	0.000	&	0.000	&	0.000	&	0.000	&	0.000	&	0.000	&	0.003	\\ \hline 
\hline 
\multirow{2}{*}{\shortstack[c]{Relations pooled \\according to polarity}} &\multicolumn{4}{c|}{Total Sample}    	&\multicolumn{4}{c|}{Core Sample} \\			
\cline{2-9} 	
 &	Strong Belief	&	Moderate Belief	&	Weak Belief	&	Neither	&	Strong Belief	&	Moderate Belief	&	Weak Belief	&	Neither	\\ \hline
$\chi^2$	&	168.530	&	271.362	&	134.827	&	50.852	&	94.518	&	107.849	&	65.200	&	8.311	\\ \hline
DF	&	2	&	2	&	2	&	2	&	2	&	2	&	2	&	2	\\ \hline
p--value	&	0.000	&	0.000	&	0.000	&	0.000	&	0.000	&	0.000	&	0.000	&	0.016	\\ \hline
\end{tabular}
}
\resizebox{0.7\textwidth}{!}{
\begin{tabular}{|c||c|c|c|c|c|c|} 
\hline
\multicolumn{7}{|c|}{Agreement pooled according to polarity} \\
\hline 
\multirow{2}{*}{\shortstack[c]{Relations pooled \\according to strength}}  &	\multicolumn{3}{c|}{Total Sample}    	&\multicolumn{3}{c|}{Core Sample} \\			
\cline{2-7} 		
 &	Disbelieved	&	Believed	&	Neither	&	Disbelieved	&	Believed	&	Neither	\\ \hline
$\chi^2$	&	213.405	&	463.158	&	116.882	&	105.506	&	281.212	&	11.626	\\ \hline
DF	&	2	&	2	&	2	&	2	&	2	&	2	\\ \hline
p--value	&	0.000	&	0.000	&	0.000	&	0.000	&	0.000	&	0.003	\\ \hline
\hline 
\multirow{2}{*}{\shortstack[c]{Relations pooled \\according to polarity}} & \multicolumn{3}{c|}{Total Sample}    	&\multicolumn{3}{c|}{Core Sample} \\			
\cline{2-7} 				
&	Disbelieved	&	Believed	&	Neither	&	Disbelieved	&	Believed	&	Neither	\\ \hline
$\chi^2$	&	213.633	&	351.965	&	50.852	&	68.486	&	212.382	&	8.311	\\ \hline
DF	&	2	&	2	&	2	&	2	&	2	&	2	\\ \hline
p--value	&	0.000	&	0.000	&	0.000	&	0.000	&	0.000	&	0.016	\\ \hline
\end{tabular} 
}
\caption{Results of chi--squared goodness of fit test on relations carried out by arguments of a given acceptance level in Dialogue 2
on total and core samples according to a given pooling. DF stands for degrees of freedom. 
We have highlighted the fields with p--value greater than $0.05$, i.e. those for which we cannot reject the null hypothesis. 
These results were obtained using R.}
\label{tab:d2goodnessargpool}
\end{table}

\clearpage
\subsection{Dialogues 1\& 2: Changes in Beliefs}

\begin{table}[h]
\centering
\resizebox{\textwidth}{!}{
\begin{tabular}{|c|c||P{1.4cm}|c|c|c|c|c|c|c|c|c|c|c|c|c|c|} 
\cline{3-17} 					
\multicolumn{2}{c|}{}	 		&	Argument	&	\multicolumn{2}{c|}{A}			&	\multicolumn{2}{c|}{B}			&	\multicolumn{2}{c|}{C}			&	\multicolumn{2}{c|}{D}			&	\multicolumn{2}{c|}{E}			&	\multicolumn{2}{c|}{F}			&	\multicolumn{2}{c|}{G}			\\	\cline{3-17}
\multicolumn{2}{c|}{}	 		&	Between steps	&	Z	&	pvalue	&	Z	&	pvalue	&	Z	&	pvalue	&	Z	&	pvalue	&	Z	&	pvalue	&	Z	&	pvalue	&	Z	&	pvalue	\\	\hline
\multirow{8}{*}{\rot{Dialogue 1}}	&	\multirow{4}{*}{\rot{Total}}	&	1 and 2	&	-0.551	&	\cellg{0.716}	&	-5.997	&	0.000	&		&		&		&		&		&		&		&		&		&		\\	\cline{3-17} 
	&		&	2 and 3	&	1.890	&	\cellg{0.125}	&	5.846	&	0.000	&	1.783	&	\cellg{0.085}	&		&		&		&		&		&		&		&		\\	\cline{3-17} 
	&		&	3 and 4	&	-1.986	&	\cellg{0.063}	&	-5.826	&	0.000	&	0.780	&	\cellg{0.544}	&	-5.351	&	0.000	&		&		&		&		&		&		\\	\cline{3-17} 
	&		&	4 and 5	&	2.157	&	0.039			&	5.773	&	0.000	&	1.964	&	\cellg{0.051}	&	-1.134	&	\cellg{0.453}	&	2.875	&	0.005	&		&		&		&		\\	\cline{2-17} 																															
	&	\multirow{4}{*}{\rot{Core}}	&	1 and 2	&	0.069	&	\cellg{1.000}	&	-3.726	&	0.000	&		&		&		&		&		&		&		&		&		&		\\	\cline{3-17} 
	&		&	2 and 3	&	1.732	&	\cellg{0.250}	&	3.792	&	0.000	&	0.000	&	\cellg{1.000}	&		&		&		&		&		&		&		&		\\	\cline{3-17} 
	&		&	3 and 4	&	-0.577	&	\cellg{1.000}	&	-3.873	&	0.000	&	2.000	&	\cellg{0.125}	&	-0.069	&	\cellg{1.000}	&		&		&		&		&		&		\\ \cline{3-17} 
	&		&	4 and 5	&	1.000	&	\cellg{1.000}	&	3.873	&	0.000	&	0.046	&	\cellg{1.000}	&	-3.553	&	0.000	&	1.513	&	\cellg{0.188}	&		&		&		&		\\	\hline \hline

\multirow{8}{*}{\rot{Dialogue 2}}	&	\multirow{4}{*}{\rot{Total}}	&	1 and 2	&	-0.378	&	\cellg{1.000}	&	-4.126	&	0.000	&		&		&		&		&		&		&		&		&		&		\\	\cline{3-17} 
	&		&	2 and 3	&	-0.794	&	\cellg{0.656}	&	-1.102	&	\cellg{0.406}	&	5.449	&	0.000	&	0.622	&	\cellg{0.627}	&		&		&		&		&		&		\\	\cline{3-17} 
	&		&	3 and 4	&	-2.448	&	0.031			&	0.740	&	\cellg{0.688}	&	-4.781	&	0.000	&	-0.839	&	\cellg{0.440}	&	1.635	&	\cellg{0.156}	&	0.349	&	\cellg{0.699}	&		&		\\	\cline{3-17} 
	&		&	4 and 5	&	-0.013	&	\cellg{1.000}	&	0.764	&	\cellg{0.445}	&	1.297	&	\cellg{0.227}	&	1.154	&	\cellg{0.359}	&	-0.952	&	\cellg{0.434}	&	-1.393	&	\cellg{0.227}	&	0.421	&	\cellg{0.591}	\\	\cline{2-17} 		
		
	&	\multirow{4}{*}{\rot{Core}}	&	1 and 2	&	0.000	&	\cellg{1.000}	&	0.577	&	\cellg{1.000}	&		&		&		&		&		&		&		&		&		&		\\	\cline{3-17} 
	&		&	2 and 3	&	1.413	&	\cellg{0.500}	&	-3.011	&	0.003	&	3.316	&	0.000	&	1.413	&	\cellg{0.500}	&		&		&		&		&		&		\\	\cline{3-17} 
	&		&	3 and 4	&	-1.995	&	\cellg{0.125}	&	3.372	&	0.000	&	-1.717	&	\cellg{0.125}	&	-1.466	&	\cellg{0.234}	&	-2.653	&	0.003	&	-2.232	&	0.033	&		&		\\	\cline{3-17} 
	&		&	4 and 5	&	0.494	&	\cellg{1.000}	&	1.000	&	\cellg{1.000}	&	-0.647	&	\cellg{0.564}	&	1.996	&	\cellg{0.125}	&	-0.960	&	\cellg{0.500}	&	-0.576	&	\cellg{0.750}	&	1.308	&	\cellg{0.219}	\\	\hline
\end{tabular}
}
\caption{The results of the Wilcoxon signed rank test with Pratt adjustment between the acceptance levels declared by 
the participants at steps $i$ and $i+1$ for a given argument, in a given dialogue and sample. 
The results have been obtained using R library coin. 
We have highlighted the fields with p--value greater than $0.05$, i.e. those for which we cannot reject the null hypothesis. 
Due to the fact that arguments $F$ and $H$ appear only at the last stages of Dialogues 1 and 2 respectively, they possess only 
a single distribution and the change analysis cannot be performed for them. }
\label{tab:belchange}
\end{table}

\end{document}